\documentclass{article}

\usepackage{arxiv}
\usepackage[utf8]{inputenc} 
\usepackage[T1]{fontenc}    
\usepackage{hyperref}       
\usepackage{url}            
\usepackage{booktabs}       
\usepackage{amsfonts}       
\usepackage{nicefrac}       
\usepackage{microtype}      
\usepackage{lipsum}		
\usepackage{placeins} 
\usepackage{graphicx}
\usepackage{doi}
\usepackage{multirow}
\usepackage{colortbl}
\usepackage{dsfont}
\usepackage{amsmath,amssymb}
\usepackage{algorithm}
\usepackage{algorithmic}
\usepackage{subcaption}
\usepackage[acronym]{glossaries}

\title{Loss Functions and Metrics in Deep Learning}

\author{ \href{https://orcid.org/0000-0001-6662-0390}{\includegraphics[scale=0.06]{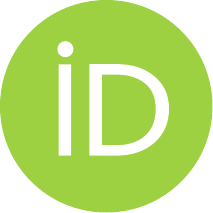}\hspace{1mm}Juan R.~Terven} \\
	CICATA-Qro\\
	Instituto Politecnico Nacional\\
	Mexico \\
	\texttt{jrtervens@ipn.mx} \\
    \And
    \href{https://orcid.org/0000-0002-5657-7752}{\includegraphics[scale=0.06]{orcid.pdf}\hspace{1mm}Diana M.~Cordova-Esparza} \\
    Facultad de Informática\\
    Universidad Autónoma de Querétaro\\
    Mexico \\
    \texttt{diana.cordova@uaq.mx} \\
    \And
    \href{https://orcid.org/0000-0003-0366-6249}{\includegraphics[scale=0.06]{orcid.pdf}\hspace{1mm}Alfonso Ramirez-Pedraza} \\
    Visión Robótica\\
    Centro de Investigaciones en Óptica A.C.\\
    Mexico \\
    \texttt{pedro.ramirez@cio.mx} \\
    \And
    \href{https://orcid.org/0000-0002-1938-8610}{\includegraphics[scale=0.06]{orcid.pdf}\hspace{1mm}Edgar A. Chavez-Urbiola} \\
    CICATA-Qro\\
    Instituto Politecnico Nacional\\
    Mexico \\
    \texttt{eachavezu@ipn.mx} \\
    \And
    \href{https://orcid.org/0000-0001-7257-7595}{\includegraphics[scale=0.06]{orcid.pdf}\hspace{1mm}Julio A. Romero-Gonzalez} \\
    Facultad de Informática\\
    Universidad Autónoma de Querétaro\\
    Mexico \\
    \texttt{julio.romero@uaq.mx} \\
}

\renewcommand{\undertitle}{Published in Springer Artificial Intelligence Review as: \\ A comprehensive survey of loss functions and metrics in deep learning \\\url{https://doi.org/10.1007/s10462-025-11198-7}\\
PLEASE CITE THE PUBLISHED VERSION}
\renewcommand{\shorttitle}{Published in Artificial Intelligence Review}

\hypersetup{
pdftitle={TITLE},
pdfsubject={cs, cs},
pdfauthor={Juan R.~Terven, Diana M.~Cordova-Esparza, Alfonso~Ramirez-Pedraza, Edgar~Chavez-Uurbiola, Julio~Romero-Gonzalez},
pdfkeywords={Deep learning, loss functions, performance metrics},
}

\begin{document}
\maketitle

\begin{abstract}
This paper presents a comprehensive review of loss functions and performance metrics in deep learning, highlighting key developments and practical insights across diverse application areas. We begin by outlining fundamental considerations in classic tasks such as regression and classification, then extend our analysis to specialized domains like computer vision and natural language processing including retrieval-augmented generation. In each setting, we systematically examine how different loss functions and evaluation metrics can be paired to address task-specific challenges such as class imbalance, outliers, and sequence-level optimization. Key contributions of this work include: (1) a unified framework for understanding how losses and metrics align with different learning objectives, (2) an in-depth discussion of multi-loss setups that balance competing goals, and (3) new insights into specialized metrics used to evaluate modern applications like retrieval-augmented generation, where faithfulness and context relevance are pivotal. Along the way, we highlight best practices for selecting or combining losses and metrics based on empirical behaviors and domain constraints. Finally, we identify open problems and promising directions, including the automation of loss-function search and the development of robust, interpretable evaluation measures for increasingly complex deep learning tasks. Our review aims to equip researchers and practitioners with clearer guidance in designing effective training pipelines and reliable model assessments for a wide spectrum of real-world applications.
\end{abstract}


\tableofcontents

\section{Introduction}
\label{intro}

Learning has become the dominant technology for solving problems involving unstructured data, such as images \cite{smith2018,jones2020,gonzalez2017,wang2019,chen2021,gatys2016,zhang2017,karras2017}, video, audio \cite{chen2019,liu2020,kim2018,zhang2021,wu2017}, and text \cite{smith2018-NLP,jones2020-NLP,gonzalez2017-NLP,wang2019-NPL,chen2021-NPL}. One of the critical components of deep learning is the selection of the loss function and performance metrics used for training and evaluating models. Loss functions measure how effectively a model can approximate the desired output, while performance metrics assess the model's ability to make accurate predictions on unseen data. Choosing the appropriate loss function and performance metric is essential for success in deep learning tasks. However, with many options to choose from, it can be challenging for practitioners to determine the most suitable method for their specific task.

In this paper, we present a detailed overview of the most commonly utilized loss functions and performance metrics in the field of deep learning. We analyze the strengths and weaknesses of each method and provide illustrative examples of its application in various deep learning tasks.

First, we delve into the prevalent regression and classification loss functions, such as mean squared error, cross-entropy, and hinge loss, delineating their respective advantages, limitations, and typical use cases. Subsequently, we explore standard tasks in computer vision, such as image classification, object detection, image segmentation, face recognition, and image generation. Finally, we conclude our review by outlining the prevalent loss functions and metrics employed in natural language processing and retrieval augmented generation (RAG).

This paper is structured as follows. Section \ref{sec:loss_vs_metrics} outlines the difference between loss functions and performance metrics. Sections \ref{subsec:Regression} and \ref{subsec:Classif} provide an overview of the most widely used losses and metrics in regression and classification. Sections \ref{subsec:img_class} to \ref{subsec:gen_models} delve into the prevalent computer vision tasks, detailing the associated loss functions and metrics. Section \ref{sec:NLP} concentrates on the use of losses and metrics in Natural Language Processing (NLP). Section \ref{sec:RAG} is dedicated to Retrieval Augmented Generation (RAG) losses and metrics. Section \ref{sec:combining_loss} discusses the common practice of combining multiple metrics, Section \ref{sec:challenges_and_trends} discusses challenges and trends, and Section \ref{sec:conclusion} rounds out the paper with a concluding statement. 

\subsection{Loss Functions vs. Performance Metrics}
\label{sec:loss_vs_metrics}

Loss functions and performance metrics are two distinct tools for evaluating the performance of a deep learning model and serve different purposes.

During training, a loss function measures the difference between the predicted and expected outputs of the model with the objective of optimizing the model parameters by minimizing this difference.

In contrast, a performance metric is used to evaluate the model after training. It helps to determine how well the model can generalize to new data and make accurate predictions. Performance metrics also aid in comparing different models or configurations to identify the best performing one.

The following list details the key differences between loss functions and performance metrics.

\begin{enumerate}
    \item During the training of a deep learning model, loss functions are used to optimize the model's parameters, whereas performance metrics are used to evaluate the model's performance after training.
    
    \item The choice of loss function typically depends on the model's architecture and the specific task at hand. In contrast, performance metrics are less dependent on the model's architecture and can be used to compare different models or configurations of a single model.
    
    \item The ultimate goal of training a deep learning model is to minimize the loss function, while evaluating a model aims to maximize the performance metric, with the exception of error performance metrics such as Mean Squared Error.
    
    \item Loss functions can be challenging to interpret as their values are often arbitrary and depend on the specific task and data. In contrast, performance metrics are often more interpretable and can be used across different tasks.
\end{enumerate}

\subsection{Properties of Loss Functions}
Loss functions have a series of properties that need to be considered when selected for a specific task:

\begin{enumerate}
    \item \textbf{Convexity}: A loss function is convex if any local minimum is also the global minimum. Convex loss functions are desirable because they can be easily optimized using gradient-based optimization methods.
    \item \textbf{Differentiability}: A loss function is differentiable if its derivative with respect to the model parameters exists and is continuous. Differentiability is essential because it allows the use of gradient-based optimization methods.
    \item \textbf{Robustness}: Loss functions should be able to handle outliers and not be affected by a small number of extreme values.
    \item \textbf{Smoothness}: A loss function should have a continuous gradient and no sharp transitions or spikes.
    \item \textbf{Sparsity}: A sparsity-promoting loss function should encourage the model to produce sparse output. This is useful when working with high-dimensional data and when the number of important features is small.
    \item \textbf{Monotonicity}: A loss function is monotonic if its value decreases as the predicted output approaches the true output. Monotonicity ensures that the optimization process is moving toward the correct solution.
\end{enumerate}

The tables below provide a summary of the loss functions and performance metrics that are discussed in this work. Table \ref{tab:general_tasks_losses_metrics} outlines the functions and metrics employed in general tasks such as regression, binary classification, and multiclass classification. Table \ref{tab:vision_tasks_losses_metrics} provides an overview of the loss functions and metrics related to computer vision, and Table \ref{tab:nlp_tasks_losses_metrics} summarizes the work on natural language processing.

\begin{table*}[htbp]
\caption{Loss functions and performance metrics for general tasks.}
\footnotesize
\centering
\begin{tabular}{p{2cm}p{5cm}p{5.5cm}}
\toprule
Deep Learning Task & Loss Functions & Performance Metrics\\
\midrule
Regression & Mean Squared Error (MSE) (\ref{loss:MSE}) \newline Mean Absolute Error (MAE) 
 (\ref{loss:MAE}) \newline Huber loss (\ref{loss:Huber}) \newline Log-Cosh (\ref{loss:logcos}) \newline Quantile loss (\ref{loss:Quantile}) \newline Poisson loss (\ref{loss:poisson}) & Mean Squared Error (MSE)(\ref{loss:MSE}) \newline Mean Absolute Error (MAE)(\ref{loss:MAE}) \newline Root Mean Squared Error (RMSE)(\ref{metric:RMSE})  \newline Mean Absolute Percentage Error (MAPE) (\ref{metric:MAPE}) \newline Symmetric MAPE (SMAPE) (\ref{metric:SMAPE}) \newline $R^2$ (\ref{metric:R2}) \newline Adjusted $R^2$ (\ref{metric:AdjR2}) \\
\midrule
Binary Classification & Binary Cross-Entropy (BCE) (\ref{loss:BCE}) \newline Weighted Cross-Entropy (\ref{loss:weighted_cross_entropy}) \newline Hinge loss (\ref{loss:hinge}) \newline Focal loss (\ref{loss:focal}) & Accuracy (\ref{metric:accuracy}) \newline Precision (\ref{metric:precision}) \newline Recall or True Positive Rate (TPR) (\ref{metric:recall_tpr}) \newline F1-Score (\ref{metric:f1}) \newline AUC-ROC (\ref{metric:AUC}) \newline Precision/Recall Curve (\ref{metric:prec-recall-curve}) \\
\midrule
Multi-Class Classification & Categorical Cross-Entropy (CCE) (\ref{loss:CCE}) \newline Weighted Cross-Entropy (\ref{loss:weighted_cross_entropy}) \newline Sparse Categorical Cross-Entropy (SCCE)(\ref{loss:SparseCCE}) \newline CCE w/label smoothing (\ref{loss:CE_loss_label_smoth})  \newline Focal loss (\ref{loss:focal}) \newline PolyLoss (\ref{loss:polyloss}) \newline Hinge loss (\ref{loss:hinge}) & Accuracy (\ref{metric:accuracy}) \newline Precision (\ref{metric:precision}) \newline Recall or True Positive Rate (TPR) (\ref{metric:recall_tpr}) \newline F1-Score (\ref{metric:f1}) \newline Precision/Recall Curve (\ref{metric:prec-recall-curve}) \\
\bottomrule
\end{tabular}
\label{tab:general_tasks_losses_metrics}
\end{table*}

\begin{table}[htbp]
\caption{Loss functions and performance metrics used in Computer Vision.}
\footnotesize
\centering
\begin{tabular}{p{3cm}p{5cm}p{5cm}}
\toprule
Computer Vision Task & Loss Functions & Performance Metrics\\
\midrule
Object Detection & Smooth L1 (\ref{loss:smooth_l1}) \newline Balanced L1 (\ref{loss:balanced_l1}) \newline Intersection over Union loss (\ref{loss:IoUloss_obj_det}) \newline GIoU (\ref{loss:GIoU}) \newline DIoU and CIoU (\ref{loss:DIoU_CIoU}) \newline Focal loss (\ref{loss:focal}) \newline YOLO loss (\ref{loss:yolo}) \newline Wing loss (\ref{loss:wing}) & Average Precision (\ref{metric:AP}) \newline Average Recall (\ref{metric:AR})\\
\midrule
Semantic Segmentation & Categorical Cross-Entropy (\ref{loss:CE_seg}) \newline Intersection over Union loss (IoU) (\ref{loss:IoUloss_obj_det}) \newline Dice Loss (\ref{loss:dice}) \newline Tversky loss (\ref{loss:tversky}) \newline Lovasz loss (\ref{loss:lovasz}) \newline Focal loss (\ref{loss:focal_seg}) & Intersection over Union (IoU) (\ref{loss:IoUloss_obj_det}) \newline Pixel Accuracy (\ref{metric:pixel_acc}) \newline Average Precision (AP) (\ref{metric:AP}) \newline Boundary F1 Score (\ref{metric:bf}) \newline Average Recall (AR) (\ref{metric:AR}) \\
\midrule
Instance Segmentation & Categorical Cross-Entropy (CCE) (\ref{loss:CE_seg}) \newline Intersection over Union loss (IoU) (\ref{loss:IoUloss_obj_det}) \newline Smooth L1 (\ref{metric:AP}) \newline Average Recall (AR) (\ref{metric:AR}) & Masked Average Precision (\ref{metric:mask_ap}) \\
\midrule
Panoptic Segmentation & Categorical Cross-Entropy (CCE) (\ref{loss:CE_seg}) \newline Dice Loss (\ref{loss:dice}) & Panoptic Quality (PQ) (\ref{metric:pq})  \\
\midrule
Face Recognition & A-Softmax (\ref{loss:face_asoftmax}) \newline Center loss (\ref{loss:center}) \newline CosFace (\ref{loss:cosface}) \newline ArcFace (\ref{loss:arcface}) \newline Triplet loss (\ref{loss:triplet}) \newline Contrastive loss (\ref{loss:contrastive}) \newline Circle loss (\ref{loss:circle}) \newline Barlow Twins (\ref{loss:barlow}) \newline SimSiam (\ref{loss:simsiam}) & Accuracy (\ref{metric:accuracy}) \newline Precision (\ref{metric:precision}) \newline Recall (\ref{metric:recall_tpr}) \newline F1-Score (\ref{metric:f1}) \\
\midrule
Monocular Depth Estimation & Point-wise Error (\ref{loss:point-wise_error}) \newline Scale Invariant Error (\ref{loss:scale_invariant_error}) \newline Structural Similarity Index (\ref{loss:SSIM}) \newline Photometric loss (\ref{loss:photometric}) \newline Disparity Smoothness (\ref{loss:disp_smoothness}) \newline Appearance Matching (\ref{loss:appearance_matching}) \newline Left-Right Consistency (\ref{loss:left-right-consistency}) \newline BerHu loss (\ref{loss:BerHu}) \newline Edge loss (\ref{loss:edge}) \newline Minimum Reprojection (\ref{loss:min_rep}) \newline Scale-and-shift-invariant (SSI) & Mean Absolure Relative Error (\ref{metric:absrel_mde}) \newline Root Mean Squared Error (\ref{metric:rmse_mde}) \newline Logarithmic RMSE (\ref{metric:rmse_log_mde}) \newline Threshold Accuracy (\ref{metric:threshold_acc_mde}) \newline Mean Log10 Error (\ref{metric:mean_log10_mde}) \newline \% of Pixels with High Error (\ref{metric:pixels_high_error_mde}) \newline Weighted Human Disagrement Rate (\ref{metric:whdr_mde}) \newline Scale-Invariant Error (\ref{metric:scale_invariant_mde}) \\
\midrule
Image Generation & Reconstruction loss (\ref{loss:reconstruction}) \newline KL Divergence (\ref{loss:kl-loss}) \newline Perceptual loss (\ref{loss:perceptual}) \newline Adversarial Loss (\ref{loss:adversarial}) \newline   Wasserstein Loss (\ref{loss:wasserstein}) \newline Negative Log-Likelihood (\ref{loss:nll_flows}) \newline Contrastive Divergence (\ref{loss:contr_div}) & Peak Signal to Noise Ratio (PSNR) (\ref{metric:psnr}) \newline Structural Similarity Index (SSIM) (\ref{metric:SSIM}) \newline Inception Score (IS) (\ref{metric:IS}), \newline Fréchet Inception Distance (FID) (\ref{metric:FID}) \\
\bottomrule
\end{tabular}
\label{tab:vision_tasks_losses_metrics}
\end{table}

\begin{table*}[htbp]
\caption{Loss functions and performance metrics used in Natural Language Processing.}
\label{tab:nlp_tasks_losses_metrics}
\centering
\footnotesize
\begin{tabular}{p{3cm}p{5cm}p{5cm}}
\toprule
NLP Task 
& Loss Functions 
& Performance Metrics\\
\midrule

Text Classification 
& Token-level Cross-Entropy (\ref{loss:t-cce}) \newline Hinge loss (\ref{loss:hinge_nlp})
& Accuracy (\ref{metric:accuracy_nlp}) \newline Precision/Recall/F1 (\ref{metric:precision_nlp}) \newline AUC-ROC (\ref{metric:auc_roc_nlp}) 
\\
\midrule

Language Modeling 
& Token-level Cross-Entropy (\ref{loss:t-cce})
& Perplexity (\ref{metric:perplexity}) \newline BLEU (\ref{metric:bleu_score}) \newline ROUGE (\ref{metric:rouge_score})
\\
\midrule

Machine Translation 
& Token-level Cross-Entropy (\ref{loss:t-cce}) \newline MRT (\ref{loss:mrt_nlp}) \newline REINFORCE (\ref{loss:reinforce}) 
& BLEU (\ref{metric:bleu_score}) \newline ROUGE (\ref{metric:rouge_score}) \newline Perplexity (\ref{metric:perplexity}) \newline Exact match (\ref{metric:exact_match_nlp})
\\
\midrule

Name Entity Recognition 
& Token-level Cross-Entropy (\ref{loss:t-cce})
& Accuracy (\ref{metric:accuracy_nlp}) \newline Precision/Recall/F1 (\ref{metric:precision_nlp}) \newline Exact match (\ref{metric:exact_match_nlp})
\\
\midrule

Part-of-Speech Tagging 
& Token-level Cross-Entropy (\ref{loss:t-cce})
& Accuracy (\ref{metric:accuracy_nlp}) \newline Precision/Recall/F1 (\ref{metric:precision_nlp})
\\
\midrule

Sentiment Analysis 
& Token-level Cross-Entropy (\ref{loss:t-cce}) \newline Hinge loss (\ref{loss:hinge_nlp}) \newline Cosine Similarity (\ref{loss:cosine_nlp}) \newline Adj. Circle (\ref{loss:circle})
& Precision/Recall/F1 (\ref{metric:precision_nlp}) \newline AUC-ROC (\ref{metric:auc_roc_nlp})
\\
\midrule

Text Summarization 
& Token-level Cross-Entropy (\ref{loss:t-cce}) \newline MRT (\ref{loss:mrt_nlp}) \newline REINFORCE (\ref{loss:reinforce})
& BLEU (\ref{metric:bleu_score}) \newline ROUGE (\ref{metric:rouge_score}) \newline Exact match (\ref{metric:exact_match_nlp})
\\
\midrule

Question Answering 
& Token-level Cross-Entropy (\ref{loss:t-cce}) \newline Hinge loss (\ref{loss:hinge_nlp}) \newline Cosine Similarity (\ref{loss:cosine_nlp})
& Precision/Recall/F1 (\ref{metric:precision_nlp}) \newline Exact match (\ref{metric:exact_match_nlp})
\\
\midrule

Language Detection 
& Token-level Cross-Entropy (\ref{loss:t-cce}) \newline Hinge loss (\ref{loss:hinge_nlp})
& Accuracy (\ref{metric:accuracy_nlp}) \newline Precision/Recall/F1 (\ref{metric:precision_nlp})
\\
\midrule

Retrieval Augmented Generation (RAG) 
& CE (\ref{loss:t-cce}) \newline Contrastive loss (\ref{loss:contrastive}) \newline Marginal Ranking (\ref{loss:marginal_ranking_nlp}) \newline NLL (\ref{loss:nll}) \newline KL Divergence (\ref{loss:kl-loss})
& Answer Semantic Similarity (\ref{metric:ragas_seman_simil}) \newline Answer correctness (\ref{metric:ragas_ans_correct}) \newline Answer Relevance (\ref{metric:ragas_ans_rel}) \newline Context Precision (\ref{metric:ragas_context_prec}) \newline Context Recall (\ref{metric:ragas_context_rec}) \newline Faithfulness (\ref{metric:ragas_faithfulness}) \newline Summarization score (\ref{metric:ragas_sum_score}) \newline Context Entitities Recall (\ref{metric:ragas_context_ent_recall}) \newline Aspect Critique (\ref{metric:ragas_aspect_critique}) \newline Context Relevance (\ref{metric:ares_context_rel}) \newline Answer Faithfulness (\ref{metric:ares_answer_faith}) \newline Answer Relevance (\ref{metric:ares_answ_rel})
\\
\midrule

Speech Recognition 
& CTC (\ref{loss:CTC})
& Word Error Rate (WER) (\ref{metric:wer}) \newline Character Error Rate (CER) (\ref{metric:cer}) \newline Perplexity (\ref{metric:perplexity})
\\
\bottomrule
\end{tabular}
\end{table*}

\subsection{Implementation Considerations and Software Libraries}
\label{subsec:software_libraries}

Practical deep learning development involves decisions about software frameworks and tools for implementing, training, and evaluating models. Popular frameworks such as PyTorch \cite{NEURIPS2019_bdbca288_pytorch}, TensorFlow/Keras \cite{abadi2016tensorflow,abadi2016tensorflow2}, and MATLAB \cite{MATLAB} provide core functionalities like computational graphs, automatic differentiation, and pre-implemented losses (e.g., MSE, cross-entropy) alongside standard metrics such as accuracy or precision-recall. These frameworks also facilitate customization, enabling users to define novel loss functions and evaluation protocols.

\paragraph{PyTorch}
Developed by Facebook's AI Research, PyTorch offers a dynamic computational graph that supports flexible model-building in Python. Its \textit{torch.nn} module provides common loss functions, including \texttt{nn.MSELoss} for regression and \texttt{nn.CrossEntropyLoss} for multi-class classification. For metrics, PyTorch-based libraries (e.g., \texttt{torchmetrics}) or the framework’s built-in tensor operations allow quick computation of accuracy, F1 scores, and even more complex measures like AUC-ROC. This design encourages straightforward experimentation when creating or modifying custom losses and metrics.

\paragraph{TensorFlow and Keras}
TensorFlow, from Google, and its high-level Keras API are pivotal in deep learning. Modules like \texttt{tf.keras.losses} and \texttt{tf.keras.metrics} contain canonical loss functions such as \texttt{MeanSquaredError} or \texttt{BinaryCrossentropy}, along with evaluation metrics like \texttt{Accuracy} or \texttt{MeanIoU} (for segmentation). Keras simplifies customization by allowing users to define new losses in Python, returning a scalar loss from \texttt{y\_true} and \texttt{y\_pred}. Similarly, extended metrics can be created by subclassing \texttt{tf.keras.metrics.Metric}, enabling tailored evaluation workflows for specialized tasks.

\paragraph{MATLAB}
MATLAB’s Deep Learning Toolbox provides built-in functions such as \texttt{crossentropy} and \texttt{mse} for loss calculations, plus object-oriented templates to code custom layers and losses. MATLAB’s emphasis on data analysis, visualization, and vectorized operations can streamline metric computation. Though less common in open-source research, MATLAB remains well regarded in many industrial and academic settings for debugging, simulation, and rapid prototyping.

\paragraph{scikit-learn Integration}
While PyTorch and TensorFlow focus primarily on neural network construction, the Python-based \texttt{scikit-learn} \cite{pedregosa2011scikit} library offers an extensive suite of evaluation metrics. Functions like \texttt{accuracy\_score}, \texttt{precision\_score}, \texttt{recall\_score}, and \texttt{f1\_score} can be readily applied to predictions from deep networks, bridging deep learning with standardized machine-learning ecosystems. This synergy ensures that model assessments align with established criteria, thereby permitting direct comparisons with more classical methods.

\paragraph{Example: Custom Losses and Metrics}
Researchers often create specialized losses to incorporate domain knowledge. Below is a PyTorch example that combines MSE and L1 penalties:
\begin{verbatim}
import torch
import torch.nn as nn

class CustomLoss(nn.Module):
    def forward(self, predictions, targets):
        mse_loss = torch.mean((predictions - targets)**2)
        l1_penalty = torch.mean(torch.abs(predictions))
        return mse_loss + 0.01 * l1_penalty
\end{verbatim}

The equivalent TensorFlow/Keras version:
\begin{verbatim}
import tensorflow as tf

def custom_loss(y_true, y_pred):
    mse_loss = tf.reduce_mean(tf.square(y_pred - y_true))
    l1_penalty = tf.reduce_mean(tf.abs(y_pred))
    return mse_loss + 0.01 * l1_penalty
\end{verbatim}

For evaluation, a trained model’s predictions can easily be passed to \texttt{scikit-learn} metrics:
\begin{verbatim}
from sklearn.metrics import accuracy_score, precision_score, recall_score

# Suppose y_true, y_pred are NumPy arrays from a trained model.
acc = accuracy_score(y_true, y_pred)
prec = precision_score(y_true, y_pred, average='macro')
rec = recall_score(y_true, y_pred, average='macro')
\end{verbatim}

These examples illustrate the convenience of mixing deep learning frameworks with machine-learning libraries, allowing for both flexible modeling and standardized performance assessment.

\vspace{0.5em}
Overall, modern deep learning software enables seamless definition and customization of loss functions and metrics. This flexibility aids experimental rigor and reproducibility, letting practitioners fine-tune objectives, combine multiple losses for multi-task learning, and leverage validated metrics to benchmark performance across different methodologies.

In the following sections, we dive into two essential tasks in machine learning: regression and classification. We examine the loss functions and performance metrics used and offer practical guidance on when to use each, as well as real-world applications.

\section{Regression}
\label{subsec:Regression}
Regression is a supervised learning problem in machine learning that aims to predict a continuous output value based on one or more input features. Regression is used in various domains, including finance, healthcare, social sciences, sports, and engineering. Some practical applications include house price prediction \cite{harrison1978hedonic}, energy consumption forecasting \cite{zhao2012review_energy_consumption}, healthcare and disease prediction \cite{harrell1984regression_prognostic}, stock price forecasting \cite{shen2012stock}, and customer lifetime value prediction \cite{malthouse2005can_lifetime_value}.

In the following subsections, we review the most common lost functions and performance metrics used for regression.

\subsection{Regression Loss Functions}
\label{subsec:Regression_loss_functions}


In this section, we explore various regression loss functions that play a crucial role in evaluating and optimizing predictive models. A regression loss function \( \mathcal{L} \) is typically defined over a dataset 
\(\mathcal{D} = \{(x_i, y_i)\}_{i=1}^n\)
of \(n\) samples, where \(x_i \in \mathbb{R}^d\) denotes the \(d\)-dimensional input feature vector and \(y_i \in \mathbb{R}\) is the corresponding continuous target value. Let \(\hat{y}_i = f_\theta(x_i)\) be the prediction of a model \(f_\theta\), parameterized by \(\theta\). The fundamental objective in regression is to find the parameter vector \(\theta\) that minimizes the overall loss function

\begin{equation}
\label{eq:general_regression_loss}
\min_{\theta} \; \mathcal{L}\big(\{(x_i,y_i)\}_{i=1}^n,\, \theta\big).
\end{equation}

Table \ref{tab:guidelines_regression-loss} summarizes common regression loss functions, highlighting their uses, data traits, benefits, and drawbacks. It aids in choosing a suitable loss function by considering data nature, outlier robustness, and the balance between interpretability and optimization smoothness.

The following subsections describe each of these loss functions in more detail.

\begin{table*}[ht]
\caption{Guidelines for selecting a regression loss function based on usage, data characteristics, advantages, and limitations.}
\label{tab:guidelines_regression-loss}
\centering
\footnotesize
\begin{tabular}{p{1.3cm}p{2.8cm}p{2.8cm}p{2.8cm}p{2.8cm}}
\toprule
\textbf{Function} 
& \textbf{Usage} 
& \textbf{Data Characteristics} 
& \textbf{Advantages} 
& \textbf{Limitations} \\
\midrule

MSE
& 
Standard tasks (e.g., linear regression) \newline
If you want to penalize large errors more severely
&
Data with few or no heavy outliers \newline
Errors approximately Gaussian
&
Differentiable everywhere (smooth) \newline
Convex in parameters \newline
Commonly used, well-understood
&
Highly sensitive to outliers \newline
Scale-dependent; not always comparable across problems \newline
Can prioritize reducing large errors over smaller ones
\\
MAE
&
When robustness to outliers is desired \newline
L1-regularized methods, median-based tasks \newline
&
Skewed error distributions \newline
Presence of moderate outliers
&
Less sensitive to outliers than MSE \newline
Convex in parameters \newline
Interpretable as median error 
&
Non-differentiable at zero error \newline
Scale-dependent 
\\

Huber
&
When a smooth, robust loss is needed \newline
Want a tunable transition between MSE and MAE
&
Data with occasional outliers \newline
Need partial robustness but still smooth optimization
&
Combines advantages of MSE (for small errors) and MAE (for large errors) \newline
Less sensitive to outliers than MSE \newline
Fully differentiable
&
Requires choice of threshold \(\delta\) \newline
Not scale-invariant
\\
Log-Cosh
&
Similar use-cases to Huber where smoothness is preferred \newline
If you desire a simpler, always differentiable robust alternative
&
Presence of outliers but not extremely large ones
&
Smooth and differentiable everywhere \newline
Robust to moderate outliers 
&
More sensitive to small errors than Huber \newline
Less intuitive thresholding than Huber
\\
Quantile
&
When predicting intervals or specific percentiles \newline
Tasks requiring asymmetric error penalties
&
Heteroscedastic or skewed data \newline
Need to estimate conditional quantiles (e.g., risk, demand)
&
Captures the distribution of possible outcomes \newline
Robust to outliers when \(q \ne 0.5\) \newline
Generalization of MAE (\(q=0.5\))
&
Must choose relevant quantile(s) \newline
Non-differentiable at zero error \newline
Interpretation can be less straightforward than MSE/MAE
\\
Poisson
&
Count data prediction \newline
Data assumed to follow (or roughly follow) a Poisson process
&
Target is non-negative, discrete counts \newline
Often low-to-moderate integer values
&
Aligned with count-data likelihood \newline
Ensures non-negative predictions (with exponential link)
&
Requires careful handling of zero/low counts \newline
Assumes Poisson-like distribution 
\\
\bottomrule
\end{tabular}
\end{table*}

\subsubsection{Mean Squared Error (MSE)}
\label{loss:MSE}

Mean Squared Error (MSE), also referred to as the \emph{L2 loss}, is one of the most widely used regression loss functions \cite{lehmann2006theory}. It is defined as the average of the squared differences between the predicted values and the true values:

\begin{equation}
\label{eq:mse_loss}
\text{MSE}(\theta) 
= \frac{1}{n} \sum_{i=1}^n (y_i - \hat{y}_i)^2
= \frac{1}{n} \sum_{i=1}^n \bigl(y_i - f_\theta(x_i)\bigr)^2.
\end{equation}

\paragraph{Key Properties:}
\begin{itemize}
    \item \textbf{Non-negativity:} Since squared errors are always non-negative, MSE returns a value \( \geq 0\), with \(0\) indicating a perfect fit.
    \item \textbf{Sensitivity to outliers:} By squaring the errors, large errors have a disproportionately higher impact on the loss, making MSE sensitive to outliers.
    \item \textbf{Differentiability:} MSE is smooth and differentiable with respect to both \(\hat{y}_i\) and \(\theta\). The gradient of MSE with respect to a prediction \(\hat{y}_i\) is 
    \begin{equation}
    \frac{\partial}{\partial \hat{y}_i}\,\bigl(y_i - \hat{y}_i\bigr)^2 
    = -2\,(y_i - \hat{y}_i).
    \end{equation}
    This property facilitates gradient-based optimization methods such as Stochastic Gradient Descent (SGD).
    \item \textbf{Convexity (in predictions):} MSE is convex in terms of the predictions \(\hat{y}_i\). For linear models \(\hat{y}_i = \theta^\top x_i\), MSE is also convex in \(\theta\). However, in deep neural networks with non-linear activation functions, the overall error surface can become non-convex.
    \item \textbf{Scale-dependence:} The magnitude of MSE depends on the scale of the target variable \(y_i\), making cross-comparison of models on different scales less straightforward. Consequently, measures like Root Mean Squared Error (RMSE) or Mean Squared Percentage Error (MSPE) are often preferred for interpretability.
\end{itemize}

\paragraph{Optimization:}
Minimizing MSE can be interpreted as minimizing the squared \(\ell_2\)-norm of the residuals:
\begin{equation}
\min_{\theta} \frac{1}{n}\sum_{i=1}^n (y_i - f_\theta(x_i))^2.
\end{equation}
In a simple linear regression scenario where \(f_\theta(x_i) = \theta^\top x_i\), an analytic closed-form solution exists using the Normal Equation \cite{lehmann2006theory}. However, for more complex models such as neural networks, numerical methods like gradient descent are commonly employed.

\paragraph{Practical Considerations:}
MSE is easy to implement and interpret, but its emphasis on larger errors may yield suboptimal performance in the presence of outliers. In such situations, more robust alternatives like Mean Absolute Error (MAE) or Huber Loss may be preferable.

\subsubsection{Mean Absolute Error (MAE)}
\label{loss:MAE}

Mean Absolute Error (MAE), also referred to as the \emph{L1 loss}, is another common choice for regression tasks \cite{willmott2005advantages}. It measures the average of the absolute differences between the predicted values and the true values:

\begin{equation}
\label{eq:mae_loss}
\text{MAE}(\theta)
= \frac{1}{n}\sum_{i=1}^n \bigl|\,y_i - \hat{y}_i\bigr|
= \frac{1}{n}\sum_{i=1}^n \bigl|\,y_i - f_\theta(x_i)\bigr|.
\end{equation}

\paragraph{Key Properties:}
\begin{itemize}
    \item \textbf{Non-negativity:} MAE is always \( \geq 0\). A value of \(0\) indicates a perfect fit.
    \item \textbf{Robustness to outliers:} By using the absolute value rather than squaring the error, MAE penalizes large errors linearly, making it less sensitive to outliers than MSE.
    \item \textbf{Non-differentiability at 0:} The absolute value function \(\lvert z \rvert\) is not differentiable at \(z=0\). Consequently, the loss function is not differentiable when \(y_i = \hat{y}_i\). However, subgradient methods \cite{auslender2004interior,bai2018subgradient,bianchi2021stochastic,xiao2009dual} can be employed in gradient-based optimization to address this issue. Formally, the subgradient w.r.t.\ \(\hat{y}_i\) is:
    \begin{equation}
    \partial_{\hat{y}_i}\,\bigl|\,y_i - \hat{y}_i\bigr| = 
    \begin{cases}
        -1, & \text{if } \hat{y}_i > y_i,\\
        +1, & \text{if } \hat{y}_i < y_i,\\
        [-1,\, +1], & \text{if } \hat{y}_i = y_i.
    \end{cases}
    \end{equation}
    \item \textbf{Convexity (in predictions):} MAE is convex in \(\hat{y}_i\), ensuring a single global minimum. As with MSE, in deep neural networks with non-linearities, the overall loss surface becomes non-convex in the model parameters.
\end{itemize}

\paragraph{Optimization:}
Minimizing MAE can be understood as minimizing the \(\ell_1\)-norm of the residuals:
\begin{equation}
\min_{\theta} \frac{1}{n}\sum_{i=1}^n \left|\,y_i - f_\theta(x_i)\right|.
\end{equation}
Unlike the MSE case, there is no simple closed-form solution for linear models. Algorithms such as coordinate descent or iterative reweighted least squares (IRLS) can be used \cite{auslender2004interior}, or gradient-based methods can approximate a solution by using subgradients.

\paragraph{Practical Considerations:}
MAE is more robust to outliers compared to MSE but lacks the smoothness that MSE provides, potentially making optimization slower in certain scenarios. As with MSE, the absolute scale of \(y_i\) affects MAE’s value, necessitating the use of scale-invariant metrics such as the Mean Absolute Percentage Error (MAPE) or Normalized Mean Absolute Error (NMAE) for inter-problem comparisons.

\medskip

\noindent
\textbf{Summary of MSE vs.\ MAE:} 
Both MSE and MAE are key loss functions for regression tasks. MSE emphasizes large errors more heavily and is differentiable everywhere, whereas MAE is more robust to outliers but non-differentiable at zero error. The choice between the two often depends on the application’s tolerance to outliers, interpretability needs, and computational considerations during optimization.

\subsubsection{Huber Loss}
\label{loss:Huber}

The \emph{Huber loss}, introduced by Huber in \cite{huber1992robust}, is a piecewise-defined function that combines the advantageous properties of both the Mean Squared Error (MSE) and Mean Absolute Error (MAE). Let \(y\) be the true value, \(\hat{y}\) be the predicted value, and \(\delta\) be a user-specified threshold. The Huber loss for a single sample is defined by

\begin{equation}
\label{eq:hubber_loss}
L_{\mathrm{Huber}}(y, \hat{y}) = 
\begin{cases}
    \tfrac{1}{2}(y - \hat{y})^2, & \text{if } \lvert y - \hat{y}\rvert \le \delta,\\[6pt]
    \delta \bigl(\lvert y - \hat{y}\rvert - \tfrac{1}{2}\,\delta\bigr), & \text{otherwise}.
\end{cases}
\end{equation}

\noindent
When the prediction error \(\lvert y - \hat{y}\rvert\) is smaller than \(\delta\), the function behaves like a quadratic (similar to MSE). For large errors, it transitions into a linear penalty (similar to MAE). This property provides robustness to outliers compared to standard MSE while maintaining smoothness for smaller residuals.

\paragraph{Properties:}
\begin{itemize}
    \item \textbf{Robustness to Outliers:} For errors larger than \(\delta\), the linear penalty avoids the rapid growth that occurs in MSE. Thus, outliers have a smaller effect on parameter updates.
    \item \textbf{Smoothness and Differentiability:} 
    Except for a point of non-differentiability at \(\lvert y - \hat{y}\rvert = \delta\), the function remains differentiable elsewhere, enabling gradient-based optimization. In practice, subgradient or piecewise derivative implementations address the corner case.
    \item \textbf{Piecewise Gradient:} The gradient of \(L_{\mathrm{Huber}}\) with respect to \(\hat{y}\) is:
    \begin{equation}
    \frac{\partial L_{\mathrm{Huber}}}{\partial \hat{y}} = 
    \begin{cases}
      -\, (y - \hat{y}), & \text{if } \lvert y - \hat{y}\rvert \le \delta, \\[6pt]
      -\,\delta\, \mathrm{sign}(y - \hat{y}), & \text{if } \lvert y - \hat{y}\rvert > \delta.
    \end{cases}
    \end{equation}
    Here, \(\mathrm{sign}(z)\) is \(+1\) if \(z > 0\) and \(-1\) otherwise.
    \item \textbf{Choice of \(\delta\):} The threshold \(\delta\) can be chosen empirically by cross-validation. A smaller \(\delta\) behaves more like MAE, whereas a larger \(\delta\) behaves closer to MSE. 
\end{itemize}

\paragraph{Practical Considerations:}
Huber loss is commonly employed in robust regression settings (e.g., linear regression and time series forecasting) where outliers and noise may be present. Its piecewise-defined structure makes it suitable for scenarios in which the transition from quadratic to linear penalization can be controlled. Nonetheless, the introduction of \(\delta\) adds a hyperparameter that requires tuning.

\subsubsection{Log-Cosh Loss}
\label{loss:logcos}

The \emph{Log-Cosh} loss \cite{saleh2022statistical} uses the hyperbolic cosine to measure errors. For \(n\) samples, it is given by

\begin{equation}
\label{eq:logcos_loss}
L_{\mathrm{LogCosh}}
= \frac{1}{n}\sum_{i=1}^{n} \log\!\bigl[\cosh\bigl(y_i - \hat{y}_i\bigr)\bigr],
\end{equation}

\noindent
where \(y_i\) and \(\hat{y}_i\) respectively denote the true and predicted values for the \(i\)-th sample.

\paragraph{Properties:}
\begin{itemize}
    \item \textbf{Smoothness and Differentiability:}
    The function \(\log(\cosh(z))\) is smooth and differentiable for all \(z \in \mathbb{R}\). The derivative of \(\log(\cosh(z))\) with respect to \(z\) is \(\tanh(z)\). Thus, the gradient of a single-sample loss \( \log\!\bigl[\cosh(y - \hat{y})\bigr] \) with respect to \(\hat{y}\) is:
    \begin{equation}
    \frac{\partial}{\partial \hat{y}} \log\!\bigl[\cosh(y - \hat{y})\bigr]
    = -\,\tanh\bigl(y - \hat{y}\bigr).
    \end{equation}
    \item \textbf{Less Sensitivity to Outliers:} 
    Unlike MSE, the \(\cosh\) function grows exponentially only for large \(\lvert y - \hat{y}\rvert\) but the \(\log(\cdot)\) moderates this growth. Consequently, extreme errors do not dominate as heavily as in the squared error case, providing a degree of robustness.
    \item \textbf{Emphasis on Small Errors:}
    For moderate errors, \(\log(\cosh(z)) \approx \frac{z^2}{2}\) (since \(\cosh(z) \approx 1 + \frac{z^2}{2}\) for small \(z\)). Thus, the loss remains smooth and lightly penalizes small deviations. However, it can be more sensitive to smaller deviations compared to Huber loss because it continuously grows with \(\lvert y - \hat{y}\rvert\), though less sharply than MSE.
\end{itemize}

\paragraph{Practical Considerations:}
Log-Cosh loss is particularly useful when one desires a fully smooth and continuous loss landscape without the piecewise nature of Huber. It offers a balance of robustness and differentiability, making it attractive in complex models (e.g., deep networks) where smooth gradient feedback is advantageous.

\subsubsection{Quantile Loss}
\label{loss:Quantile}

Also known as \emph{quantile regression loss}, this function is often used to produce predictive intervals or asymmetric error penalties \cite{koenker2001quantile}. For a quantile level \( 0 < q < 1\), let \(\hat{y}\) be the predicted value and \(y\) be the true value. The quantile loss for a single prediction is given by:

\begin{equation}
\label{eq:quantile_loss}
L_{\mathrm{quantile}}(y, \hat{y}) 
= q \, \max\bigl(y - \hat{y},\, 0\bigr)
  \;+\; (1 - q)\,\max\bigl(\hat{y} - y,\, 0\bigr).
\end{equation}

\noindent
Intuitively, if the model underestimates (\( \hat{y} < y \)), the error term \( (y - \hat{y}) \) is weighted by \(q\). Conversely, if the model overestimates (\( \hat{y} > y \)), then the error term \( (\hat{y} - y) \) is weighted by \((1-q)\). Hence, quantile loss generalizes Mean Absolute Error (MAE), since for \( q = 0.5 \), the two coincide.

\paragraph{Properties:}
\begin{itemize}
    \item \textbf{Asymmetric Penalties:}
    By choosing different values of \(q\), one can penalize overestimation or underestimation more heavily. For example, \(q>0.5\) penalizes underestimates more strongly, which can be useful in safety-critical applications where under-forecasting is risky.
    \item \textbf{Non-differentiability at Residual 0:} 
    Similar to MAE, the function \(\max(\cdot,\,0)\) is not differentiable at 0. One typically uses subgradient or piecewise derivatives to handle optimization. Specifically, the subgradient of \(\max\bigl(\hat{y}-y,0\bigr)\) w.r.t.\ \(\hat{y}\) is 1 if \(\hat{y}>y\), and 0 if \(\hat{y}<y\). Analogous logic applies to \(\max\bigl(y-\hat{y},0\bigr)\).
    \item \textbf{Relation to Median and Other Quantiles:} 
    When \( q = 0.5\), the solution to minimizing \(\sum L_{\mathrm{quantile}}\) yields the conditional median of \(y\). More generally, different \(q\) values yield the conditional \(q\)-th quantile of the response variable, facilitating interval predictions.
\end{itemize}

\paragraph{Practical Use Cases:}
Quantile regression is particularly helpful when a single-point estimate is insufficient for decision making. By estimating upper and lower quantiles, it provides \emph{interval predictions} or confidence bounds for a variety of applications:

\begin{itemize}
    \item \textbf{Financial Risk Management:}
    Estimating Value-at-Risk (VaR) and Conditional Value-at-Risk (CVaR) for extreme loss scenarios \cite{chen2002application}.
    \item \textbf{Supply Chain and Inventory Management:}
    Predicting demand ranges to balance stockouts and overstock situations \cite{bruzda2020multistep}.
    \item \textbf{Energy Production:}
    Forecasting power output ranges for grid stability \cite{bremnes2004probabilistic}.
    \item \textbf{Economic Forecasting:}
    Generating uncertainty bounds for economic indicators \cite{gaglianone2012constructing}.
    \item \textbf{Weather Forecasting:}
    Producing uncertainty intervals for temperature or rainfall \cite{massidda2018quantile,zarnani2019quantile}.
    \item \textbf{Real Estate Pricing:}
    Offering predictive price intervals rather than a single estimate \cite{zietz2008determinants}.
    \item \textbf{Healthcare:}
    Modeling a range of patient outcomes \cite{winkelmann2006reforming}.
\end{itemize}

\noindent
Thus, quantile loss provides a flexible approach to addressing asymmetric costs and quantifying predictive uncertainty.

\subsubsection{Poisson Loss}
\label{loss:poisson}

\emph{Poisson loss} is utilized when the target variable \(y_i\) represents count data assumed to follow a Poisson distribution. It is derived from the negative log-likelihood of the Poisson distribution:

\begin{equation}
P(y_i \mid \lambda_i) = \frac{\lambda_i^{y_i}\, e^{-\lambda_i}}{y_i!},
\quad
y_i \in \{0,1,2,\dots\},
\end{equation}

\noindent
where \(\lambda_i>0\) is the Poisson rate parameter (i.e., the expected count). The corresponding negative log-likelihood (omitting constant terms that do not depend on \(\lambda_i\)) is:

\begin{equation}
\ell_i(\lambda_i)
= \lambda_i \;-\; y_i\,\log(\lambda_i).
\end{equation}

\noindent
In practice, for a dataset of \(n\) samples, the \emph{Poisson loss} is often written as

\begin{equation}
\label{eq:poisson}
L_{\mathrm{Poisson}} 
= \frac{1}{n}\sum_{i=1}^{n}
   \bigl[\hat{y}_i - y_i \,\log(\hat{y}_i)\bigr],
\end{equation}
where \(\hat{y}_i\) is the model’s predicted (non-negative) rate for the \(i\)-th observation.

\paragraph{Non-negativity and Link Functions:}
Because count data cannot be negative, \(\hat{y}_i\) must be constrained to \(\hat{y}_i > 0\). In generalized linear models or neural networks, one often uses an exponential link function:

\begin{equation}
\label{eq:link_fnc}
\hat{y}_i = \exp\bigl(\mathbf{w}^\top \mathbf{x}_i + b\bigr),
\end{equation}
so that \(\hat{y}_i\) is guaranteed to be positive for all \(\mathbf{x}_i\). Substituting \eqref{eq:link_fnc} into \eqref{eq:poisson} yields

\begin{align}
\label{eq:poisson2}
L_{\mathrm{Poisson}}
&= \frac{1}{n}\sum_{i=1}^{n}\Bigl[
  \exp\bigl(\mathbf{w}^\top \mathbf{x}_i + b\bigr)
  \;-\;
  y_i\,\log\bigl(\exp(\mathbf{w}^\top \mathbf{x}_i + b)\bigr)
\Bigr] \nonumber \\
&= \frac{1}{n}\sum_{i=1}^{n}
  \Bigl[
    \exp\bigl(\mathbf{w}^\top \mathbf{x}_i + b\bigr)
    \;-\;
    y_i \bigl(\mathbf{w}^\top \mathbf{x}_i + b\bigr)
  \Bigr],
\end{align}
where the simplification \(\log(\exp(\cdot)) = \cdot\) has been applied.

\paragraph{Applications:}
The Poisson distribution is natural for modeling the number of times an event occurs in a fixed interval. Example use-cases include:

\begin{itemize}
    \item \textbf{Traffic Modeling:}
    Predicting vehicle counts through a toll booth under varying conditions \cite{yang2012effects}.
    \item \textbf{Healthcare and Epidemiology:}
    Forecasting disease cases by region to inform resource allocation \cite{viel1994poisson}.
    \item \textbf{Insurance:}
    Modeling insurance claim frequencies for rate-making \cite{mouatassim2012poisson}.
    \item \textbf{Customer Service:}
    Predicting incoming call volumes to allocate call-center staff \cite{shen2008forecasting}.
    \item \textbf{Internet Usage:}
    Modeling the number of website visits or ad-clicks in a time interval \cite{avila2016click}.
    \item \textbf{Manufacturing:}
    Estimating defects or failures in a production line for quality control \cite{lambert1992zero}.
    \item \textbf{Crime Analysis:}
    Assessing occurrences of crimes by area to guide resource deployment \cite{osgood2000poisson}.
\end{itemize}

\noindent
By directly modeling count-based observations under a Poisson assumption, Poisson loss allows both interpretability (expected count rates) and theoretical alignment with the discrete nature of the response variable.

\subsection{Regression Performance Metrics}
\label{subsec:Regression_metrics}

Selecting the right performance metrics is key to accurately evaluating regression models. Table \ref{tab:guidelines_regression-metrics} provides guidelines for choosing among five commonly used metrics based on data characteristics, robustness requirements, and task objectives. These metrics each offer distinct advantages and limitations, helping researchers account for factors such as error sensitivity, interpretability, and data scale. In the subsequent sections, we delve into the details of these metrics, omitting Mean Squared Error (MSE) and Mean Absolute Error (MAE) because they have been previously discussed as loss functions.

\begin{table*}[ht]
\caption{Guidelines for selecting a regression metric based on usage, data characteristics, advantages and limitations.}
\label{tab:guidelines_regression-metrics}
\centering
\footnotesize
\begin{tabular}{p{1.3cm}p{2.8cm}p{2.8cm}p{2.8cm}p{2.8cm}}
\toprule
\textbf{Metric}
& \textbf{Usage}
& \textbf{Data Characteristics}
& \textbf{Advantages}
& \textbf{Limitations} \\
\midrule
RMSE
& Measures average error magnitude
& Continuous data with moderate outliers
& Same units as target \newline Penalizes large errors
& Sensitive to outliers \newline Does not distinguish error types \\
\midrule

MAPE
& Forecasting where relative error is critical
& Positive, non-zero data \newline Varying scales
& Scale-independent \newline Intuitive in percentage terms
& Undefined if true value = 0 \newline Very sensitive to outliers \\
\midrule

SMAPE
& Time-series with balanced penalty on over/under-prediction
& Varied scales \newline Cases with zero or near-zero values
& Symmetric treatment of over/under error \newline Useful in cost-sensitive scenarios
& Undefined if both true and predicted = 0 \newline Still sensitive to outliers \\
\midrule

$R^2$
& Explains proportion of variance in target
& Data with primarily linear relationships
& Intuitive interpretation as variance explained \newline Allows model comparison
& Can be misleading for non-linear data \newline Increases with more predictors \\
\midrule

Adjusted $R^2$
& Model comparison in multi-predictor scenarios
& Larger feature sets \newline Similar contexts as $R^2$
& Penalizes irrelevant predictors \newline More reliable for model selection
& Same assumptions as $R^2$ \newline Negative if worse than mean-only model \\
\bottomrule
\end{tabular}
\end{table*}

\subsubsection{Root Mean Squared Error (RMSE)}
\label{metric:RMSE}

\emph{Root Mean Squared Error} (RMSE) is defined as the square root of the Mean Squared Error (MSE). For \( n \) data points \(\{(y_i, \hat{y}_i)\}_{i=1}^n\), where \(y_i\) is the true value and \(\hat{y}_i\) is the predicted value, RMSE is given by

\begin{equation}
\label{eq:rmse}
\text{RMSE} 
= \sqrt{\frac{1}{n}\sum_{i=1}^n \bigl(y_i - \hat{y}_i\bigr)^2}\,.
\end{equation}

\paragraph{Properties and Interpretation:}
\begin{itemize}
    \item \textbf{Unit Consistency:} 
    Since RMSE is the square root of the average squared error, it has the same unit of measurement as the target \(y\). This property makes RMSE highly interpretable in many practical scenarios (e.g., predicting prices in currency units, demand in units sold, etc.).
    
    \item \textbf{Geometric Perspective:}
    RMSE can be viewed as the \(\ell_2\)-norm distance between the vector of true values 
    \(\mathbf{y} = [y_1, y_2, \ldots, y_n]^\top\)
    and the vector of predicted values
    \(\hat{\mathbf{y}} = [\hat{y}_1, \hat{y}_2, \ldots, \hat{y}_n]^\top\). 
    Specifically, 
    \begin{equation}
      \text{RMSE}(\mathbf{y}, \hat{\mathbf{y}}) 
      = \frac{1}{\sqrt{n}}\|\mathbf{y} - \hat{\mathbf{y}}\|_2.
    \end{equation}
    This represents the average “Euclidean distance” between predictions and actuals, emphasizing larger residuals (errors).

    \item \textbf{Sensitivity to Outliers:}
    Squaring the errors disproportionately penalizes large errors. Hence, RMSE is sensitive to outliers; a few large residuals can inflate its value significantly.

    \item \textbf{Scale Dependence:}
    RMSE is directly influenced by the scale of \(y\). Larger or more variable targets can naturally produce higher RMSE values. This can hamper cross-comparison between datasets with different ranges.

    \item \textbf{Normalization and Comparison:} 
    To mitigate scale effects and facilitate model comparisons across different datasets or target scales, one may normalize the RMSE by the standard deviation \(\sigma_y\) of the true values:
    \begin{equation}
    \label{eq:normalized_rmse}
    \text{Normalized RMSE} 
    = \frac{\text{RMSE}}{\sigma_y}.
    \end{equation}
    An RMSE comparable to or below \(\sigma_y\) often indicates a model that predicts on par with—or better than—simply predicting the mean.

    \item \textbf{Relationship to Standard Deviation:}
    When the model predictions \(\hat{y}_i\) approximate the true values \(y_i\) well, the distribution of residuals \((y_i - \hat{y}_i)\) tends to have smaller variance, yielding a lower RMSE. Comparing RMSE to \(\sigma_y\) can reveal whether a model is capturing most of the variability in the data.
\end{itemize}

\paragraph{Use Cases and Limitations:}
RMSE is widely employed in fields such as financial forecasting, demand prediction, and various regression problems, due to its direct penalization of large errors. However:
\begin{itemize}
    \item \emph{Outlier Sensitivity:} Excessively large errors can overshadow overall performance.
    \item \emph{Lack of Error-Type Distinction:} RMSE combines both systematic (bias) and random (variance) errors into a single measure. One cannot directly discern error structure from RMSE alone.
    \item \emph{Scale Effects:} Care must be taken when comparing RMSE across datasets with very different units or ranges of \(y\).
\end{itemize}

\subsubsection{Mean Absolute Percentage Error (MAPE)}
\label{metric:MAPE}

\emph{Mean Absolute Percentage Error} (MAPE) gauges the average relative error of predictions, expressed as a percentage. For \( n \) samples,

\begin{equation}
\label{eq:MAPE}
\text{MAPE} 
= \frac{1}{n}\sum_{i=1}^n 
  \left(\frac{\lvert y_i - \hat{y}_i\rvert}{\,y_i\,}\times 100 \right),
\end{equation}
where \(y_i\) denotes the true value and \(\hat{y}_i\) the predicted value.

\paragraph{Properties:}
\begin{itemize}
    \item \textbf{Scale Independence:}
    Because each term in MAPE is a ratio \(\lvert y_i - \hat{y}_i\rvert / y_i\), it is invariant to scaling of \(y_i\). This property makes MAPE suitable for comparing performance across different problems or datasets that involve significantly different ranges of values.
    
    \item \textbf{Interpretability:}
    Being expressed in percentage terms, MAPE has an intuitive interpretation in many business, finance, or retail contexts: for instance, “the model is off by 12\% on average.”

    \item \textbf{Sensitivity to Zero or Near-Zero Values:}
    If any \(y_i\) is zero (or extremely small), the term \(\lvert y_i - \hat{y}_i\rvert / y_i\) can become undefined or extremely large. A common workaround is to add a small constant \(\epsilon\) to \(y_i\) in the denominator when necessary:
    \begin{equation}
    \text{MAPE}_\epsilon 
    = \frac{1}{n}\sum_{i=1}^n 
      \left(\frac{|\,y_i - \hat{y}_i\,|}{\,\max\{y_i, \epsilon\}\,}\times 100\right).
    \end{equation}
    
    \item \textbf{Outlier Sensitivity:}
    While MAPE is a relative measure, a single large relative error can still inflate the metric, especially if \(y_i\) is small. Hence, outliers can disproportionately affect MAPE—albeit in terms of \emph{relative} deviation rather than absolute magnitude.

    \item \textbf{Non-Symmetric Error Treatment:}
    Over- and under-predictions of the same magnitude do not result in symmetrical impact on MAPE, given the percentage basis. For example, predicting 50 instead of 100 is a 50\% error, but predicting 150 instead of 100 is also a 50\% error in raw terms—however, the absolute effect on MAPE can differ if there is variation in denominators across samples.
\end{itemize}

\paragraph{Use Cases and Limitations:}
\begin{itemize}
    \item \emph{Industries With Relative Error Focus:} 
    Finance, retail, and forecasting tasks where percentage deviations matter more than raw error magnitudes often favor MAPE.
    \item \emph{Comparative Model Assessments:} 
    MAPE is well-suited for comparing models trained on data with different scales, thanks to its dimensionless property.
    \item \emph{Handling of Zero or Near-Zero Targets:} 
    Special caution is required when any \(y_i\approx 0\). Alternative metrics like the Symmetric MAPE (sMAPE), or an \(\epsilon\)-adjusted MAPE, might be used to mitigate infinite or excessively large percentage errors.
\end{itemize}

\noindent
In summary, MAPE provides an intuitive percentage-based evaluation of model performance but can pose challenges in scenarios with zero or small true values and can be disproportionately influenced by large percentage errors.

\subsubsection{Symmetric Mean Absolute Percentage Error (SMAPE)}
\label{metric:SMAPE}

\emph{Symmetric Mean Absolute Percentage Error} (SMAPE) is a variant of the Mean Absolute Percentage Error (MAPE) often used to evaluate the accuracy of forecasts in time series settings \cite{goodwin1999asymmetry}. For \(n\) samples \(\{(y_i, \hat{y}_i)\}_{i=1}^n\), SMAPE is defined as

\begin{equation}
\label{eq:SMAPE}
\text{SMAPE}
= \frac{2}{n}\sum_{i=1}^n
  \frac{\bigl\lvert y_i - \hat{y}_i \bigr\rvert}{\lvert y_i\rvert + \lvert \hat{y}_i\rvert}
  \times 100\,.
\end{equation}

\paragraph{Properties:}
\begin{itemize}
    \item \textbf{Symmetry for Over-/Under-Estimation:}
    Unlike MAPE, SMAPE employs the denominator \(\lvert y_i\rvert + \lvert \hat{y}_i\rvert\), which balances the penalty between over-prediction and under-prediction. Thus, overestimations and underestimations of the same magnitude produce identical error contributions.
    \item \textbf{Scale Independence:}
    By normalizing each absolute error by the sum of absolute values, SMAPE partially mitigates issues associated with the scale of \(y_i\). Nevertheless, if both \(\lvert y_i\rvert\) and \(\lvert \hat{y}_i\rvert\) are very small, SMAPE terms can still be large or undefined in cases where both are zero.
    \item \textbf{Handling of Zeros:}
    If both \(y_i\) and \(\hat{y}_i\) are zero, the denominator \(\lvert y_i\rvert + \lvert \hat{y}_i\rvert = 0\). A common workaround is to omit (or treat specially) these terms or add a small \(\epsilon\) in the denominator. 
    \item \textbf{Outlier Sensitivity:}
    While SMAPE is more balanced than MAPE in treating over- and under-estimation, large relative errors (e.g., due to very small denominators) can still exert an outsized effect on the overall metric.
\end{itemize}

\paragraph{Practical Implications:}
SMAPE’s symmetry property is particularly valuable in applications where over- and under-predictions have equally costly (albeit different) consequences:

\begin{itemize}
    \item \textbf{Inventory Management:}
    Over-predicted demand can lead to excess stock and wasted resources, whereas under-predicted demand causes stockouts and lost sales \cite{tian2021forecasting}.
    \item \textbf{Energy Demand Forecasting:}
    Over-prediction wastes resources via unneeded power generation; under-prediction can cause blackouts or emergency generation \cite{ramos2021short}.
    \item \textbf{Financial Markets:}
    Overestimating prices risks financial losses via unwise investments, while underestimation can miss profitable opportunities \cite{kumar2021grey}.
    \item \textbf{Sales Forecasting:}
    Overestimates lead to unused inventory or labor; underestimates cause missed revenues and poor customer service \cite{lawrence2000field}.
    \item \textbf{Transportation and Logistics:}
    Overestimates waste resources on underfilled vehicles or routes; underestimates lead to congestion or unmet demand \cite{chu2019deep}.
\end{itemize}

\noindent
Overall, SMAPE’s symmetry makes it a fitting choice where the miscost of over-prediction is comparably important to that of under-prediction, though care must be taken with zeros and outliers.

\subsubsection{Coefficient of Determination \texorpdfstring{$R^2$}{R2}}
\label{metric:R2}

The \emph{Coefficient of Determination}, denoted \(R^2\), measures how well a model explains the total variation in the target variable \(y\). Let \(\overline{y} = \tfrac{1}{n}\sum_{i=1}^n y_i\) be the mean of the observed \(y\)-values. Then,

\begin{equation}
\label{eq:rsquared}
R^2
= 1 \;-\;
\frac{\sum_{i=1}^n (y_i - \hat{y}_i)^2}{
      \sum_{i=1}^n (y_i - \overline{y})^2}\,.
\end{equation}

\paragraph{Interpretation:}
\begin{itemize}
    \item \textbf{Explained Variance:}
    The denominator \(\sum (y_i - \overline{y})^2\) is the total variance of \(y\). The numerator \(\sum (y_i - \hat{y}_i)^2\) is the residual sum of squares (RSS), representing the unexplained portion of variability by the model. Thus,
    \begin{equation}
      R^2
      = 1 - \frac{\text{RSS}}{\text{Total Variance}}.
    \end{equation}
    An \(R^2\) close to 1 indicates a model that explains most of the variation in the data, whereas \(R^2\approx 0\) indicates the model is barely better than the mean predictor. Notably, \(R^2\) can be negative if the model is worse than simply predicting \(\overline{y}\).

    \item \textbf{Model Comparison:}
    Higher \(R^2\) values generally indicate better explanatory power. However, \(R^2\) must be interpreted in the context of the problem. For instance, a low \(R^2\) might still be acceptable if the data naturally exhibits high variance and partial predictability.

    \item \textbf{Linear vs. Nonlinear Models:}
    \(R^2\) was originally introduced for linear regression but is often used for more complex models. Its interpretation can become less straightforward in highly nonlinear settings. Still, it remains a commonly reported metric for uniform comparisons across model types.
\end{itemize}

\paragraph{Limitations:}
\begin{itemize}
    \item \textbf{Overfitting in Multiple Regression:}
    \(R^2\) typically increases (or remains the same) when additional predictors are introduced, even if they have little explanatory power. \emph{Adjusted \(R^2\)} (Section~\ref{metric:AdjR2}) partially addresses this.
    \item \textbf{Inability to Detect Bias:}
    An \(R^2\) value does not disclose if the model systematically overestimates or underestimates the target variable.
    \item \textbf{Sensitivity to Outliers:}
    A few extreme points can shift the overall variance structure, thus affecting \(R^2\). Careful examination of residual plots is advisable.
    \item \textbf{Potential Misleading Use in Small Samples:}
    In small datasets, \(R^2\) can be highly unstable or misleading.
\end{itemize}

\subsubsection{Adjusted \texorpdfstring{$R^{2}$}{R2}}
\label{metric:AdjR2}

When multiple predictors are included in a regression model, \emph{Adjusted \(R^2\)} offers a way to account for the fact that standard \(R^2\) naturally increases (or remains constant) as variables are added. Adjusted \(R^2\) is computed as

\begin{equation}
\label{eq:adjusted}
\text{Adjusted } R^2
= 1 - \Bigl(\frac{1 - R^2}{1}\Bigr)
  \cdot \frac{n - 1}{n - k - 1},
\end{equation}
where \(n\) is the number of observations, \(k\) is the number of predictors, and \(R^2\) is as in \eqref{eq:rsquared}.

\paragraph{Key Rationale:}
\begin{itemize}
    \item \textbf{Penalty for Additional Predictors:}
    The factor \(\frac{n-1}{n-k-1}\) imposes a penalty for each added predictor, ensuring that the adjusted \(R^2\) will only increase if the newly added predictor sufficiently reduces the residual sum of squares.
    \item \textbf{Model Comparison:}
    Adjusted \(R^2\) facilitates fair comparison of nested models (e.g., same dataset, differing number of predictors). A large jump in \(R^2\) but only a small (or negative) change in adjusted \(R^2\) suggests that the added predictors may not be truly informative.
\end{itemize}

\paragraph{Benefits and Limitations:}
\begin{enumerate}
    \item \textbf{Overfitting Mitigation:}
    By penalizing model complexity, adjusted \(R^2\) reduces the risk of overfitting that is sometimes masked by a raw \(R^2\).

    \item \textbf{Interpretation Parallels:}
    Like \(R^2\), adjusted \(R^2\) also ranges up to 1, indicating how much variation is explained after penalizing for predictor count. Negative values imply the model is worse than predicting \(\overline{y}\).

    \item \textbf{Still Limited in Nonlinear or Complex Models:}
    Adjusted \(R^2\) remains most naturally interpreted in the context of linear regression or generalized linear modeling. It may not fully capture complex relationships (e.g., deep neural networks), though it is still sometimes reported as a reference.

    \item \textbf{Sample Size Considerations:}
    For small \(n\), the penalty factor can be large, and minor changes in \(\sum (y_i - \hat{y}_i)^2\) can trigger large swings in adjusted \(R^2\).
\end{enumerate}

\paragraph{Practical Example:}
Consider a regression for housing prices based on factors like square footage, number of bedrooms, location, etc. Adding an irrelevant predictor (e.g., day of the week the listing was posted) might trivially increase \(R^2\), but adjusted \(R^2\) would likely decrease unless the new variable genuinely improves explanatory power. Thus, adjusted \(R^2\) provides a more robust way to evaluate whether a predictor is meaningfully contributing to the model.

\section{Classification}
\label{subsec:Classif}

Classification is a supervised machine learning task in which a model maps input features \(\mathbf{x}_i \in \mathbb{R}^d\) to discrete labels \(y_i \in \mathcal{Y}\). The goal is to learn a decision function 
\begin{equation}
f_\theta: \mathbb{R}^d \to \mathcal{Y},
\end{equation}
parameterized by \(\theta\), that assigns each input \(\mathbf{x}_i\) to a specific category in \(\mathcal{Y}\). In typical settings:
\begin{itemize}
  \item \textbf{Binary Classification:} \(\mathcal{Y} = \{0, 1\}\).
  \item \textbf{Multi-Class Classification:} \(\mathcal{Y} = \{1, 2, \ldots, C\}\) for \(C\) classes.
  \item \textbf{Multi-Label Classification:} \(\mathcal{Y} \subseteq \{0,1\}^C\), allowing multiple classes to be assigned to the same instance.
\end{itemize}

Classification algorithms span a wide range of methods, such as Decision Trees, Naïve Bayes, \(k\)-Nearest Neighbors, Support Vector Machines, Random Forests, Gradient Boosting, and Neural Networks. Regardless of the specific algorithm, an appropriate choice of \emph{loss function} is critical for guiding the training procedure. The following subsections present the most common classification loss functions and discuss their properties and typical use-cases.

\subsection{Classification Loss Functions}
\label{subsec_classi_loss_functions}

Several loss functions can be used for classification tasks, depending on the nature of the data, the number of classes, and the model architecture. 
Table \ref{tab:guidelines-class-losses} summarizes key classification loss functions, detailing their usage scenarios, data traits, advantages, and limitations. Below, we provide a more detailed discussion of each function.

\begin{table*}[ht]
\caption{Guidelines for selecting a classification loss function based on usage, data characteristics, advantages, and limitations.}
\label{tab:guidelines-class-losses}
\centering
\footnotesize
\begin{tabular}{p{1.2cm}p{2.3cm}p{2.3cm}p{2.3cm}p{2.3cm}}
\toprule
\textbf{Function} 
& \textbf{Usage} 
& \textbf{Data Characteristics} 
& \textbf{Advantages} 
& \textbf{Limitations} \\
\midrule

BCE 
& Binary classification \newline Common in logistic regression 
& Labels in \{0,1\} \newline Sensitive to imbalance 
& Probabilistic interpretation \newline Strong penalty for confident mistakes 
& Overconfidence if not regularized \newline Often needs weighting \\
\midrule

CCE 
& Multi-class classification \newline Softmax-based outputs 
& One-hot labels \newline Class imbalance is possible 
& Differentiable \newline Strong penalty for misclassifications 
& Requires one-hot (extra memory) \newline Sensitive to class imbalance \\
\midrule

Sparse CCE 
& Large-class problems \newline Integer-labeled data 
& No one-hot encoding \newline Useful with large vocabularies 
& Memory efficient \newline Same probabilistic basis as CCE 
& Only for integer labels \newline Imbalance remains an issue \\
\midrule

Weighted Cross-Entropy
& Binary or multi-class tasks \newline Class imbalance or cost sensitivity 
& Labels in \{0,1\} or one-hot \newline Per-class or per-label weights 
& Mitigates imbalance by up-weighting minority classes \newline Easily integrated into CE framework 
& Requires careful weight tuning \newline Large weights can destabilize training \\
\midrule

Label Smoothing 
& Large-scale multi-class \newline Reduces overconfidence 
& One-hot or near one-hot \newline Uses small \(\epsilon\) 
& Improves generalization \newline Penalizes extreme confidence 
& Needs \(\epsilon\) tuning \newline May slow convergence \\
\midrule

NLL 
& Equivalent to CE with one-hot \newline General classification 
& Typically one-hot labels \newline Probability matching 
& Interpretable as \(-\log(\text{true prob.})\) \newline Common in likelihood frameworks 
& Suffers from imbalance \newline Overconfidence = large gradient \\
\midrule

PolyLoss 
& General framework \newline Extensions of CE, focal loss 
& Binary or multi-class \newline Flexible via polynomial terms 
& Tunable sensitivity \newline Additional hyperparameter(s) 
& Requires careful coefficient choice \newline Increases complexity \\
\midrule

Hinge 
& Margin-based classification \newline SVM-style learning 
& Binary labels \{-1, 1\} \newline Decision boundary approach 
& Encourages margin maximization \newline Zero loss beyond margin 
& Not probabilistic \newline Less suited for probabilistic outputs \\
\bottomrule
\end{tabular}
\end{table*}


\subsubsection{Binary Cross-Entropy Loss (BCE)}
\label{loss:BCE}

\emph{Binary Cross-Entropy} (BCE), sometimes referred to as \emph{Log Loss}, is a standard loss function for binary classification \cite{goodman1991objective}. Consider a dataset \(\{(\mathbf{x}_i, y_i)\}_{i=1}^n\) where each label \(y_i \in \{0,1\}\). Let \(\hat{p}_i = f_\theta(\mathbf{x}_i)\) be the model’s predicted probability of \(y_i=1\). The BCE loss for a single instance is given by

\begin{equation}
\label{eq:lyp}
L\bigl(y_i, \hat{p}_i\bigr)
= -\Bigl[y_i \,\log(\hat{p}_i) + (1-y_i)\,\log\bigl(1-\hat{p}_i\bigr)\Bigr].
\end{equation}

\noindent
In practice, the average loss over the dataset is minimized:

\begin{equation}
\label{eq:bce_dataset}
\mathcal{L}_{\mathrm{BCE}}(\theta)
= \frac{1}{n}\sum_{i=1}^n
  L\bigl(y_i, \hat{p}_i\bigr).
\end{equation}

\paragraph{Properties:}
\begin{itemize}
    \item \textbf{Information-Theoretic Origin:} Cross-entropy measures the distance between two Bernoulli distributions: the true distribution \(P(y)\) and the predicted distribution \(Q(y)\). Minimizing it encourages \(\hat{p}_i\) to match the empirical probabilities of the labels.
    \item \textbf{Differentiability:}
    BCE is differentiable w.r.t.\ \(\hat{p}_i\in(0,1)\). When \(\hat{p}_i = \sigma(z_i)\) (the logistic sigmoid), gradient-based methods such as Stochastic Gradient Descent (SGD) or its variants (Adam, RMSProp, etc.) naturally follow.
    \item \textbf{Punishing Confident Mistakes:}
    If \(\hat{p}_i\) is close to 1 but \(y_i=0\), or vice versa, the logarithmic term \(\log(\hat{p}_i)\) or \(\log(1-\hat{p}_i)\) becomes large in magnitude, heavily penalizing over-confident errors.
    \item \textbf{Class Imbalance:}
    In datasets where one class significantly outnumbers the other, BCE can lead to biased models. A \emph{Weighted BCE} variant alleviates this:

    \begin{equation}
    \label{eq:weight}
    L_{\mathrm{weighted}}(y_i, \hat{p}_i)
    = -\Bigl[w_1\,y_i \log(\hat{p}_i) + w_0\,(1-y_i)\,\log\bigl(1-\hat{p}_i\bigr)\Bigr],
    \end{equation}
    where \(w_1\) and \(w_0\) are class-specific weights, often taken inversely proportional to class frequencies. Weighted Cross-Entropy is discussed with more details in Section \ref{loss:weighted_cross_entropy}.
\end{itemize}

\subsubsection{Categorical Cross-Entropy Loss (CCE)}
\label{loss:CCE}

\emph{Categorical Cross-Entropy} (CCE), also known as \emph{Multi-Class Log Loss}, extends BCE to handle \(\mathbf{y}_i \in \{0,1\}^C\) in a one-hot representation across \(C\) classes. Let \(\hat{\mathbf{p}}_i = [\hat{p}_{i,1}, \ldots, \hat{p}_{i,C}]\) be the predicted probability distribution for sample \(i\). Then, the per-sample loss is

\begin{equation}
\label{eq:L}
L\bigl(\mathbf{y}_i, \hat{\mathbf{p}}_i\bigr)
= - \sum_{j=1}^C
    y_{i,j}\,\log \bigl(\hat{p}_{i,j}\bigr),
\end{equation}
where \(y_{i,j}\in \{0,1\}\) and \(\sum_j y_{i,j}=1\). Averaging over \(n\) samples yields the dataset-level loss:

\begin{equation}
\label{eq:cce_dataset}
\mathcal{L}_{\mathrm{CCE}}(\theta)
= -\frac{1}{n}\sum_{i=1}^n \sum_{j=1}^C
   y_{i,j}\,\log \bigl(\hat{p}_{i,j}\bigr).
\end{equation}

\paragraph{Key Points:}
\begin{itemize}
    \item \textbf{Softmax Output:}
    In neural networks, the model typically outputs logits \(\mathbf{z}_i \in \mathbb{R}^C\). A softmax function
    \begin{equation}
      \hat{p}_{i,j} = \frac{\exp(z_{i,j})}{\sum_{k=1}^C \exp(z_{i,k})}
    \end{equation}
    ensures \(\sum_j \hat{p}_{i,j} = 1\).
    \item \textbf{Log-Likelihood Interpretation:}
    Minimizing \(\mathcal{L}_{\mathrm{CCE}}\) is equivalent to maximizing the log-likelihood of the correct class. 
    \item \textbf{Penalty for Confident Misclassifications:}
    As in BCE, CCE heavily penalizes predictions that assign large probability to an incorrect class.
    \item \textbf{One-Hot Encoding Requirements:}
    Standard CCE assumes one-hot (or probability-like) target vectors. For integer-encoded labels, \emph{Sparse CCE} (Section~\ref{loss:SparseCCE}) is often more efficient.
\end{itemize}

\subsubsection{Sparse Categorical Cross-Entropy Loss}
\label{loss:SparseCCE}

\emph{Sparse Categorical Cross-Entropy} (Sparse CCE) addresses scenarios where class labels are given as integers \(y_i \in \{1,2,\ldots,C\}\) rather than one-hot vectors. Let \(\hat{p}_{i,j}\) be the predicted probability of class \(j\) for sample \(i\). The individual sample loss is:

\begin{equation}
\label{eq:H}
L_{\mathrm{sparse}}\bigl(y_i, \hat{\mathbf{p}}_i\bigr)
= -\log\bigl(\hat{p}_{i,\,y_i}\bigr),
\end{equation}
where \(\hat{p}_{i,\,y_i}\) is the probability assigned to the correct class \(y_i\). The overall loss is:

\begin{equation}
\label{eq:sparse}
\mathcal{L}_{\mathrm{SparseCCE}}(\theta)
= -\frac{1}{n}\sum_{i=1}^n
   \log \bigl(\hat{p}_{i,\,y_i}\bigr).
\end{equation}

\paragraph{Usage}
\begin{itemize}
    \item \textbf{Efficiency:}
    In classification tasks with a large number of classes, one-hot encoding can be memory-intensive. Sparse CCE indexes the correct probability directly, avoiding the need for full one-hot vectors.
    \item \textbf{Similarity to Standard CCE:}
    From an optimization standpoint, Sparse CCE is equivalent to CCE but uses integer labels to look up \(\hat{p}_{i,\,y_i}\).
    \item \textbf{Gradient Computation:}
    As with other cross-entropy losses, it remains fully differentiable, allowing gradient-based training in neural networks.
\end{itemize}

\subsubsection{Weighted Cross-Entropy (WCE)}
\label{loss:weighted_cross_entropy}

\emph{Weighted Cross-Entropy} is a variant of the standard cross-entropy loss designed to address class imbalance by assigning greater importance to minority classes \cite{lin2017focal,king2001logistic}. This approach helps the model focus more on infrequent classes, where misclassification is often costlier or more critical. 

\paragraph{Binary Weighted Cross-Entropy:}
In binary classification with labels \( y_i \in \{0,1\} \), let \(\hat{p}_i \in [0,1]\) be the model’s predicted probability of \(y_i = 1\). Denote \(\alpha_1\) and \(\alpha_0\) as class-specific weights for the positive and negative classes, respectively. The \emph{per-instance} weighted cross-entropy loss is:

\begin{equation}
\label{eq:binary_wce}
L_{\mathrm{WCE}}\bigl(y_i, \hat{p}_i\bigr)
= -\Bigl[\alpha_1\, y_i \, \log\bigl(\hat{p}_i\bigr)
         + \alpha_0\, (1 - y_i)\,\log\bigl(1 - \hat{p}_i\bigr)\Bigr].
\end{equation}

\noindent
The overall loss is obtained by averaging over \(n\) samples:
\begin{equation}
\mathcal{L}_{\mathrm{WCE}} 
= \frac{1}{n}\sum_{i=1}^{n} 
    L_{\mathrm{WCE}}\bigl(y_i, \hat{p}_i\bigr).
\end{equation}

\paragraph{Multi-Class Weighted Cross-Entropy:}
For multi-class problems with \(C\) classes, let 
\(\hat{\mathbf{p}}_i = [\hat{p}_{i,1}, \ldots, \hat{p}_{i,C}]^\top\) 
be the predicted probability distribution over classes for sample \(i\), and let 
\(\mathbf{y}_i\) be the one-hot encoded true label. Denote \(\alpha_c\) as the weight for class \(c\). The \emph{per-instance} weighted cross-entropy loss then generalizes to:

\begin{equation}
\label{eq:multiclass_wce}
L_{\mathrm{WCE}}\bigl(\mathbf{y}_i, \hat{\mathbf{p}}_i\bigr)
= -\sum_{c=1}^{C}
   \alpha_c \, y_{i,c}\,\log\bigl(\hat{p}_{i,c}\bigr).
\end{equation}

\noindent
Again, the total loss is the mean over all samples:
\begin{equation}
\mathcal{L}_{\mathrm{WCE}} 
= \frac{1}{n}\sum_{i=1}^{n} 
    L_{\mathrm{WCE}}\bigl(\mathbf{y}_i, \hat{\mathbf{p}}_i\bigr).
\end{equation}

\paragraph{Choosing Class Weights:}
\begin{itemize}
    \item \textbf{Inverse Frequency:} 
    A common heuristic is to set \(\alpha_c \propto 1 / f_c\), where \(f_c\) is the (relative) frequency of class \(c\). Less frequent classes thus receive higher weights.
    \item \textbf{Proportional to Cost:}
    In cost-sensitive contexts, the weights may reflect the application’s error cost for each class. For instance, in medical imaging, an under-diagnosed disease class might receive a large weight.
    \item \textbf{Normalization:}
    Often, weights \(\{\alpha_c\}\) are normalized so that \(\sum_c \alpha_c = C\). This helps stabilize gradients if some \(\alpha_c\) is very large.
\end{itemize}

\paragraph{Interpretation and Practical Use:}
\begin{itemize}
    \item \textbf{Class Imbalance Mitigation:}
    Weighted cross-entropy helps ensure that minority classes have sufficient impact on the gradient during training, reducing model bias toward majority classes.
    \item \textbf{Extended to Multi-Label Scenarios:}
    In multi-label classification, each label can be assigned a weight based on its frequency or importance. Then, a binary weighted cross-entropy term is computed per label and summed.
    \item \textbf{Comparison with Other Losses:}
    While WCE often provides a straightforward fix for imbalance, more advanced approaches like Focal Loss (Section~\ref{loss:focal}) or Polyloss (Section~\ref
    {loss:polyloss}) may also be used, especially if the number of easy vs.\ hard examples significantly differs.
\end{itemize}

\paragraph{Limitations:}
\begin{itemize}
    \item \textbf{Hyperparameter Tuning:}
    Determining a suitable weighting scheme can require extensive experimentation; overly large weights might harm overall performance or cause unstable training.
    \item \textbf{Distribution Shifts:}
    If the class distribution changes during deployment (e.g., a new data source), fixed weights may become suboptimal.
\end{itemize}

Weighted Cross-Entropy thus extends the standard cross-entropy by introducing per-class or per-label weighting factors, allowing models to place greater emphasis on classes of higher importance or lower frequency. This often yields more balanced decision boundaries when class distributions are skewed, aligning better with real-world objectives.

\subsubsection{Cross-Entropy Loss with Label Smoothing}
\label{loss:CE_loss_label_smoth}

\emph{Label smoothing} is a regularization strategy that softens the target distributions for multi-class classification \cite{szegedy2016rethinking,muller2019does}. Instead of training the model to output probability 1 for the ground truth class and 0 for others, the targets are replaced by a smoothed distribution, reducing the model’s overconfidence.

\paragraph{Formulation:}
Let \(\epsilon \in [0,1]\) be the smoothing factor, \(C\) the number of classes, and \(\hat{\mathbf{p}}_i\) the predicted probability vector for sample \(i\). In a one-hot label scenario, the original target \(\mathbf{y}_i\) has \(y_{i,j}=1\) for the true class \(j\), and 0 otherwise. Label smoothing modifies \(\mathbf{y}_i\) to \(\tilde{\mathbf{y}}_i\):

\begin{equation}
\label{eq:smoothed_labels}
\tilde{y}_{i,j} 
= (1-\epsilon)\,y_{i,j}
  + \frac{\epsilon}{C}\,.
\end{equation}

\noindent
Then the usual cross-entropy is computed with \(\tilde{\mathbf{y}}_i\):

\begin{align}
\label{eq:ce_lab_smooth}
L_{\mathrm{labelSmooth}}
&= - \sum_{j=1}^C
     \tilde{y}_{i,j}\,\log \bigl(\hat{p}_{i,j}\bigr) \nonumber \\
&= - \sum_{j=1}^C
     \Bigl[
       (1-\epsilon)\,y_{i,j} + \frac{\epsilon}{C}
     \Bigr]
     \log \bigl(\hat{p}_{i,j}\bigr).
\end{align}

\paragraph{Advantages:}
\begin{itemize}
    \item \textbf{Overconfidence Mitigation:}
    By disallowing the model to assign a probability of exactly 1 to any single class, label smoothing prevents overly confident predictions.
    \item \textbf{Better Generalization:}
    Empirical evidence suggests label smoothing helps models avoid overfitting on noisy data or unrepresentative training samples.
\end{itemize}

Typical values for \(\epsilon\) range from \(0.05\) to \(0.2\). Tuning \(\epsilon\) can be done via validation performance.

The model learns to distribute some probability mass to non-ground-truth classes, effectively acting as a regularizer. Large \(\epsilon\) might excessively blur the distinction between classes, whereas very small \(\epsilon\) may not significantly affect overconfidence.

\subsubsection{Negative Log-Likelihood}
\label{loss:nll}

In many classification tasks, the ground-truth distribution \(\mathbf{p}\) for a given sample is represented by a one-hot encoded vector, meaning exactly one class has probability 1 and all others 0. Let \(\mathbf{q}\) be the model's predicted distribution over classes. The \emph{cross-entropy} between \(\mathbf{p}\) and \(\mathbf{q}\) for a single instance is given by

\begin{equation}
\label{eq:nll1}
H(\mathbf{p}, \mathbf{q})
= -\sum_{i=1}^C p_i \,\log(q_i),
\end{equation}

\noindent
where \(C\) is the total number of classes, \(p_i \in \{0,1\}\) and \(\sum_{i=1}^C p_i = 1\). Since \(\mathbf{p}\) is one-hot, exactly one \(p_i\) equals 1, and all others are 0. Hence, \eqref{eq:nll1} simplifies to

\begin{equation}
\label{eq:nll2}
H(\mathbf{p}, \mathbf{q})
= -\log\bigl(q_{y_t}\bigr),
\end{equation}

\noindent
where \(y_t\) denotes the true class index. This formulation is equivalent to the \emph{Negative Log-Likelihood (NLL)}, often written as

\begin{equation}
\label{eq:nll3}
\mathrm{NLL}(\mathbf{x}, y_t)
= -\log \bigl( p(y_t \mid \mathbf{x}) \bigr).
\end{equation}

\noindent
Thus, minimizing cross-entropy for a one-hot true distribution is equivalent to minimizing the negative log-likelihood. Both metrics incentivize the model to place high probability on the correct class \(y_t\). In modern deep learning frameworks, the softmax activation is typically used to produce \(\mathbf{q}\), ensuring \(\sum_i q_i = 1\) and \(q_i \in (0,1)\). By descending the gradient of \(\mathrm{NLL}\), the model learns to adjust its parameters so that \(q_{y_t} \to 1\) for the correct class label.

\subsubsection{Polyloss}
\label{loss:polyloss}

\emph{PolyLoss} \cite{leng2022polyloss} proposes a generalized loss function family that encompasses widely used losses like Cross-Entropy (CE) and Focal Loss as special cases. The key idea is to represent a loss function as a linear combination of polynomial expressions of the predicted probabilities, thus offering a flexible framework for adapting loss behavior based on task requirements.

\paragraph{General Formulation:}
Let \(p_t\) be the predicted probability of the true class \(t\) for a single instance. PolyLoss posits that one may write the loss function \(L\) as

\begin{equation}
\label{eq:poly}
L
= \sum_{n=0}^{N}
  \alpha_n \,\bigl(1 - p_t\bigr)^n,
\end{equation}

\noindent
where \(\alpha_n\) are polynomial coefficients controlling how heavily each \((1 - p_t)^n\) term contributes. Large values of \(n\) cause the loss to become more sensitive to small deviations in \(p_t\); thus, the \(\alpha_n\) sequence provides a knob to balance or emphasize certain error behaviors.

\paragraph{Poly-1 Loss Variant:}
A prominent example within this family is the \emph{Poly-1 loss}, which adds a single linear term on top of the standard Cross-Entropy:

\begin{equation}
\label{eq:poly1}
L_{\mathrm{Poly-1}}
= \text{CE}(\mathbf{y}, \hat{\mathbf{p}})
  \;+\;
  \epsilon \,\bigl(1 - p_t\bigr),
\end{equation}

\noindent
where \(\text{CE}(\cdot)\) is the usual cross-entropy, and \(\epsilon\) is a hyperparameter. Note that setting \(\epsilon = 0\) reduces Poly-1 to the standard cross-entropy loss. When \(\epsilon > 0\), the term \(\epsilon \,(1 - p_t)\) penalizes confidently wrong predictions more heavily and can mitigate overfitting—particularly in imbalanced classification scenarios or tasks requiring high precision.

\paragraph{Use Cases and Advantages:}
\begin{itemize}
  \item \textbf{Imbalanced Datasets:} By fine-tuning polynomial coefficients \(\alpha_n\), PolyLoss can heighten penalties on hard examples or minority classes, analogous to focal loss but with more flexible control.
  \item \textbf{Task-Specific Adaptation:} Custom polynomial expansions may help the model learn from domain-specific misclassifications (e.g., medical diagnoses where false negatives are costlier than false positives).
  \item \textbf{Minimal Hyperparameter Tuning:} Poly-1 introduces only one additional parameter \(\epsilon\). Compared to more complex losses (e.g., focal loss with multiple parameters), PolyLoss can simplify hyperparameter search.
\end{itemize}

\noindent
By providing a Taylor-series-like expansion for the loss, PolyLoss helps unify and extend existing classification loss functions within a single polynomial framework.

\subsubsection{Hinge Loss}
\label{loss:hinge}

\emph{Hinge Loss} is frequently employed in \emph{maximum-margin} classification tasks, particularly in Support Vector Machines (SVMs) \cite{rosasco2004loss}. For a binary classification task with labels \(y \in \{-1, +1\}\), the hinge loss for a single instance is

\begin{equation}
\label{eq:hinge_loss}
L\bigl(y, f(\mathbf{x})\bigr)
= \max\Bigl(0,\; 1 - y \, f(\mathbf{x})\Bigr),
\end{equation}

\noindent
where \(f(\mathbf{x})\) is the raw (unthresholded) model output or decision function (often the signed distance from the decision boundary). The \emph{margin} for the sample \(\mathbf{x}\) is given by the product \(y \, f(\mathbf{x})\). A correct classification with a sufficient margin yields a positive product above 1, making the loss zero.

\paragraph{Characteristics:}
\begin{itemize}
    \item \textbf{Margin Maximization:}
    Hinge loss enforces not only that the prediction is correct (\(\mathrm{sign}(f(\mathbf{x})) = y\)) but also that the margin \(\lvert f(\mathbf{x})\rvert\) is at least 1. Predictions on the correct side but with margin less than 1 still incur a positive loss.
    \item \textbf{Linear Penalty:}
    When \(y \, f(\mathbf{x}) < 1\), the hinge loss increases linearly with the distance \(1 - y \, f(\mathbf{x})\). This \emph{linear} penalty differs from cross-entropy’s \emph{logarithmic} penalty, often resulting in sparser gradients.
    \item \textbf{Extension to Multi-Class:}
    Though primarily associated with binary SVMs, hinge loss can be extended to multi-class settings (e.g., structured SVMs) \cite{tsochantaridis2005large}, where sophisticated loss functions depend on inter-class margins.
\end{itemize}

\paragraph{Variants:}
\begin{itemize}
    \item \textbf{Squared Hinge Loss:}
    \begin{equation}
      L_{\mathrm{sq}}(y, f(\mathbf{x}))
      = \Bigl(\max(0,\; 1 - y \, f(\mathbf{x}))\Bigr)^2,
    \end{equation}
    penalizes margin violations more severely and can lead to smoother gradient behavior near the margin boundary.
\end{itemize}

\noindent
Overall, hinge loss is valued for promoting large margins, which often correlates with better generalization in classification tasks that fit well into a maximum-margin framework.

\subsection{Classification Performance Metrics}
\label{subsec:Classif_perform_metrics}
This section covers metrics for evaluating classification models. Table \ref{tab:metrics_used_classification} summarizes these metrics, with details in subsequent subsections. Table \ref{tab:guidelines-class-metrics} offers guidelines for selecting the right metric considering usage, data traits, pros, and cons, providing support for choosing the best metric for varied data and performance needs.

\begin{table*}[ht]
\caption{Metrics used in classification task.}
\footnotesize
\label{tab:metrics_used_classification}
\centering
\begin{tabular}{p{2cm}p{2cm}p{1cm}p{2.5cm}p{4cm}}
\toprule
Common Name & Other Names & Abbr & Definitions & Interpretations\\ 
\midrule
True Positive & Hit & TP & True Sample labeled true & Correctly labeled True Sample \\
True Negative & Rejection & TN & False Sample labeled false & Correctly labeled False sample \\
False Positive & False alarm, Type I Error & FP & False sample labeled True &Incorrectly labeled False sample\\
False Negative & Miss, Type II Error& FN &True sample labeled false & Incorrectly label True sample \\
\midrule
Recall & True Positive Rate & TPR & TP/(TP+FN)& $\%$ of True samples correctly labeled \\
Specificity &  True Negative Rate & SPC, TNR  & TN/(TN+FP) & $\%$ of False samples correctly labeled \\
Precision & Positive Predictive Value & PPV & TP/(TP+FP) & $\%$ of samples labeled True that really are True \\
Negative Predictive Value & & NPV  & TN/(TN+FN) & $\%$ of samples labeled False that really are False \\
\midrule
False Negative Rate & & FNR & FN/(TP+FN)=1-TPR & $\%$ of True samples incorrectly labeled \\
False Positive Rate & Fall-out& FPR & FP/(FP+FN)=1-SPC & $\%$ of False samples incorrectly labeled \\
False Discovery Rate & & FDR & FP/(TP+FP)=1-PPV & $\%$ of samples labeled True that are really False \\
True Discovery Rate & & TDR & FN/(TN+FN)=1-NPV & $\%$ of samples labeled False that are really True \\
\midrule
Accuracy & & ACC & $\frac{(TP+TN)}{(TP+TN+FP+FN)}$ & Percent of samples correctly labeled \\
F1 Score & & F1 & $\frac{(2*TP)}{((2*TP)+FP+FN)}$ & Approaches 1 as errors decline \\ 
\bottomrule
\end{tabular}
\end{table*}

\begin{table*}[ht]
\caption{Guidelines for selecting a classification metric based on usage, data characteristics, advantages, and limitations.}
\label{tab:guidelines-class-metrics}
\centering
\footnotesize
\begin{tabular}{p{1.3cm}p{2.8cm}p{2.8cm}p{2.8cm}p{2.8cm}}
\toprule
\textbf{Metric}  
& \textbf{Usage} 
& \textbf{Data Characteristics} 
& \textbf{Advantages} 
& \textbf{Limitations} \\
\midrule

Confusion Matrix
& Summarizes model outputs (TP, FP, FN, TN) \newline Assists in error localization 
& Binary/multi-class \newline Larger matrices for many classes 
& Straightforward error analysis \newline Pinpoints specific misclassifications 
& Not a single-performance measure \newline Harder to compare multiple models \\
\midrule

Accuracy
& Overall fraction of correct predictions 
& Mostly reliable if classes are balanced 
& Easy to interpret and compute 
& Misleading under class imbalance \newline Masks type of errors \\
\midrule

Precision
& Reliability of positive predictions 
& Works in binary/multi-class \newline Affected by imbalanced positives 
& Reduces false positives \newline Critical where false alarms are costly 
& Ignores missed positives \newline Must be combined with recall \\
\midrule

Recall
& Proportion of actual positives detected 
& Binary/multi-class \newline Sensitive to class imbalance 
& Minimizes missed positives \newline Key in diagnostics/fraud 
& Can increase false positives \newline Requires balance with precision \\
\midrule

F1-Score
& Harmonic mean of precision and recall 
& Helpful in skewed data \newline Binary/multi-class 
& Balances false positives/negatives \newline More insightful than accuracy alone 
& Weighs errors equally \newline Less useful if cost of errors differs \\
\midrule

Specificity
& Proportion of actual negatives correctly identified 
& Common in binary scenarios \newline Medical or screening tests 
& Reduces false alarms \newline Complements recall 
& Neglects how positives are handled \newline Less intuitive for balanced data \\
\midrule

False Positive Rate (FPR)
& Fraction of negatives incorrectly labeled positive 
& Part of ROC analysis \newline Extensible to multi-class (OvA, OvO) 
& Shows Type I error \newline Useful with TPR (ROC curves) 
& Does not capture false negatives \newline Needs TPR to complete the picture \\
\midrule

Negative Predictive Value (NPV)
& Accuracy of predicted negatives 
& Effective in medical tests \newline Affected by prevalence of negatives 
& Ensures few missed positives \newline Valuable in screening 
& Highly sensitive to class ratio \newline Less common as a primary metric \\
\midrule

False Discovery Rate (FDR)
& Fraction of positives that are false 
& Relevant when cost of false positives is high 
& Indicates trustworthiness of positive results 
& Ignores negatives \newline Must pair with recall or precision \\
\midrule

Precision-Recall Curve
& Visualizes trade-off between precision, recall 
& Suited for imbalanced data \newline Can do one-vs-all for multi-class 
& Shows performance across thresholds \newline Highlights handling of positives 
& Complex in multi-class \newline Summaries often needed (e.g., AP) \\
\midrule

AUC-ROC
& Combines TPR vs. FPR across thresholds 
& Standard in binary classification \newline Extended via OvA or OvO 
& Threshold-independent \newline Easy model comparison 
& May be optimistic with skewed data \newline Lacks cost/context of errors \\

\bottomrule
\end{tabular}
\end{table*}

\subsubsection{Confusion Matrix}
\label{metric:conf_matrix}

The \emph{confusion matrix} is a fundamental tool for visualizing the performance of a binary classifier. It tallies the number of times each combination of predicted and actual labels occurs. For binary classification (with possible true labels \(\{\text{Positive}, \text{Negative}\}\)), the confusion matrix is a \(2 \times 2\) table, as shown in \autoref{tab:confusion_matrix}.

\begin{table}[ht]
\centering
\footnotesize
\caption{Confusion Matrix for a Binary Classification Task.}
\label{tab:confusion_matrix}
\begin{tabular}{lll}
\toprule
 & \textbf{Predicted Positive} & \textbf{Predicted Negative} \\ 
\midrule
\textbf{Actual Positive} & True Positive (TP) & False Negative (FN) \\ 
\textbf{Actual Negative} & False Positive (FP) & True Negative (TN) \\ 
\bottomrule
\end{tabular}
\end{table}

\paragraph{Definitions:}
\begin{itemize}
    \item \textbf{True Positive (TP):} The model correctly classifies a positive instance as positive.
    \item \textbf{True Negative (TN):} The model correctly classifies a negative instance as negative.
    \item \textbf{False Positive (FP):} Also known as \emph{Type I error}, the model incorrectly classifies a negative instance as positive (a ``false alarm'').
    \item \textbf{False Negative (FN):} Also known as \emph{Type II error}, the model incorrectly classifies a positive instance as negative (a ``miss'').
\end{itemize}

Many performance metrics (e.g., Accuracy, Precision, Recall, F1-score) directly derive from \((\text{TP}, \text{TN}, \text{FP}, \text{FN})\). Observing the confusion matrix highlights where the classifier errs most frequently, enabling targeted model improvements.

\paragraph{Derived Metrics:}
\begin{align}
\text{Accuracy} &= \frac{\text{TP} + \text{TN}}{\text{TP} + \text{TN} + \text{FP} + \text{FN}}, \\
\text{Precision} &= \frac{\text{TP}}{\text{TP} + \text{FP}}, \quad 
\text{Recall} = \frac{\text{TP}}{\text{TP} + \text{FN}}, \\
F_1\text{-score} &= 2 \cdot \frac{\text{Precision} \times \text{Recall}}{\text{Precision} + \text{Recall}}.
\end{align}

These metrics express different aspects of classification performance. For instance, \emph{Precision} measures how often predicted positives are correct, while \emph{Recall} indicates the fraction of positive instances the model successfully identifies.

\subsubsection{Confusion Matrix in Multi-Class Classification}
\label{metric:multiclass_conf_matrix}

For classification tasks involving \(\mathrm{N}\) classes, the confusion matrix generalizes to an \(\mathrm{N} \times \mathrm{N}\) table. Each row \(i\) corresponds to the true class \(i\), while each column \(j\) corresponds to the model’s predicted class \(j\). Let \(M[i,j]\) denote the number of instances whose ground-truth label is class \(i\) but which the model predicts as class \(j\). Thus, each entry of the matrix satisfies:

\begin{equation}
\sum_{j=1}^{N} M[i,j]
= \text{(total number of class }i\text{ instances)},
\end{equation}
\begin{equation}
\sum_{i=1}^{N} M[i,j]
= \text{(total number of instances predicted as class }j\text{)}.
\end{equation}

The diagonal entries \(M[i,i]\) represent correct classifications of class \(i\). Off-diagonal entries \(M[i,j]\) (\(i \neq j\)) reveal misclassifications from class \(i\) to class \(j\).

\begin{table}[ht]
\centering
\caption{Example of a Confusion Matrix for a Three-Class Classification Task. Diagonal entries \(M[i,i]\) indicate correct predictions for class \(i\). Off-diagonal entries illustrate misclassifications.}
\label{tab:multi-class-confusion_matrix}
\begin{tabular}{c|ccc}
\toprule
 & \textbf{Pred. A} & \textbf{Pred. B} & \textbf{Pred. C} \\
\midrule
\textbf{Actual: A} & $M[1,1]$ & $M[1,2]$ & $M[1,3]$ \\
\textbf{Actual: B} & $M[2,1]$ & $M[2,2]$ & $M[2,3]$ \\
\textbf{Actual: C} & $M[3,1]$ & $M[3,2]$ & $M[3,3]$ \\
\bottomrule
\end{tabular}
\end{table}

\paragraph{Interpretation in Multi-Class Problems:}
\begin{itemize}
    \item \textbf{Diagonal Entries:} 
    High values on the main diagonal (i.e., \(M[i,i]\)) imply strong class-specific accuracy. 
    \item \textbf{Row Summations:} 
    For row \(i\), \(\sum_{j=1}^N M[i,j]\) is the total number of class-\(i\) samples. Comparing \(M[i,i]\) with this row sum highlights the fraction of class-\(i\) samples misclassified.
    \item \textbf{Column Summations:} 
    For column \(j\), \(\sum_{i=1}^N M[i,j]\) shows how many instances the model labeled as class \(j\). When large row sums align with high misclassifications in a particular column, it suggests confusion between certain class pairs or subsets.
\end{itemize}

\begin{figure}[ht]
\centering
\includegraphics[width=\linewidth]{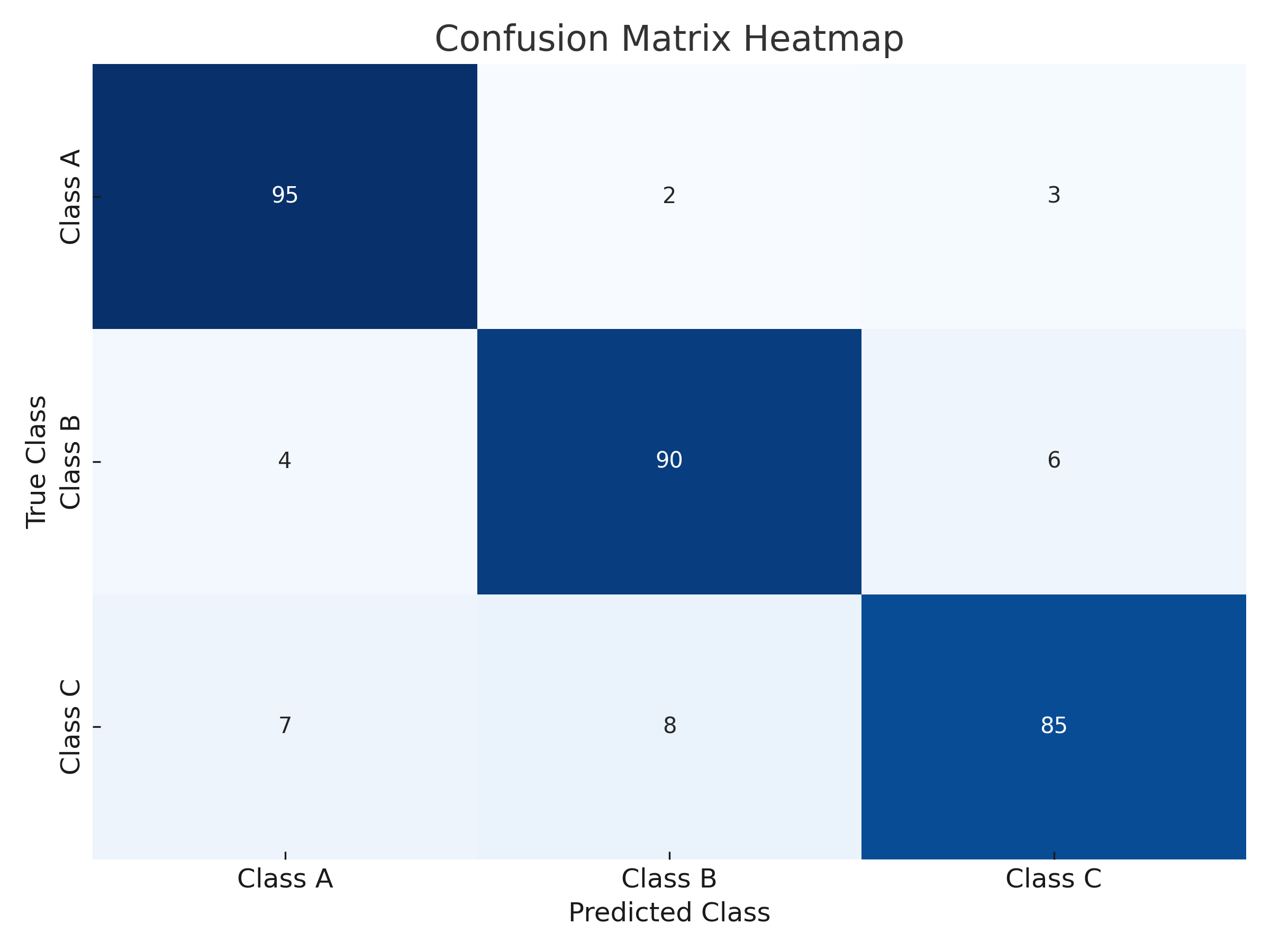}
\caption{Confusion Matrix Heatmap for a Three-Class Classification Task. 
Diagonal entries show correct predictions, off-diagonal entries represent misclassifications. Color intensity correlates with the number of samples. This format helps identify model strengths (e.g., class A has few misclassifications overall) and weaknesses (e.g., class C is misclassified as B relatively frequently).}
\label{fig:conf_matrix}
\end{figure}

\paragraph{Example Visualization:}
Figure \ref{fig:conf_matrix} illustrates a confusion matrix heatmap for three classes (A, B, C). Each cell’s color intensity is proportional to \(M[i,j]\). High diagonal intensities imply strong performance, whereas off-diagonal cells reveal specific misclassification patterns. For instance:
\begin{itemize}
    \item \(\mathbf{A \to A}\): 95 correct predictions for class A,
      \(\mathbf{A \to B}\): 2 misclassifications,
      \(\mathbf{A \to C}\): 3 misclassifications.
    \item \(\mathbf{B \to B}\): 90 correct predictions,
      \(\mathbf{B \to A}\): 4 errors,
      \(\mathbf{B \to C}\): 6 errors.
    \item \(\mathbf{C \to C}\): 85 correct identifications,
      \(\mathbf{C \to A}\): 7 errors,
      \(\mathbf{C \to B}\): 8 errors.
\end{itemize}
Observing that Class C is more frequently misclassified as Class B than A can guide adjustments in feature engineering or model capacity to reduce that confusion.

\subsubsection{Accuracy}
\label{metric:accuracy}

\emph{Accuracy} is one of the most widely used metrics for classification tasks and is defined as the ratio of correctly classified instances to the total number of instances \cite{taylor1997introduction}. Formally,

\begin{equation}
\label{eq:accuracy}
\text{Accuracy} 
= \frac{\text{Correctly Classified Samples}}{\text{Total Number of Samples}}.
\end{equation}

\noindent
When expressed in terms of the confusion matrix for binary classification, where the elements \((\text{TP}, \text{TN}, \text{FP}, \text{FN})\) denote true positives, true negatives, false positives, and false negatives respectively, Accuracy becomes

\begin{equation}
\label{eq:accuracy2}
\text{Accuracy} 
= \frac{\text{TP} + \text{TN}}{\text{TP} + \text{FP} + \text{TN} + \text{FN}}.
\end{equation}

\paragraph{Imbalance Consideration:}
Although Accuracy is intuitive, it can be misleading in cases of severe class imbalance. For example, a classifier that predicts \emph{negative} for all instances in a dataset where 99\% of samples truly are negative would achieve \(99\%\) accuracy but fail to detect the minority positive class entirely.

\paragraph{Multi-Class Accuracy:}
For a classification problem with \(C\) classes, Accuracy generalizes by summing the number of correct predictions across all classes and dividing by the total number of samples:

\begin{equation}
\label{eq:acc_multiclass}
\text{Accuracy}_{\mathrm{multi}}
= \frac{\sum_{i=1}^{C} \text{Correct}_i}{\text{Total Samples}},
\end{equation}

\noindent
where \(\text{Correct}_i\) is the number of correctly predicted samples for class \(i\). Although straightforward, multi-class Accuracy still suffers from the same imbalance issues if certain classes are underrepresented.

\paragraph{Balanced Accuracy:}
To mitigate imbalance effects, \emph{Balanced Accuracy} computes the mean of Recall (also known as Sensitivity) across classes:

\begin{equation}
\label{eq:balanced_acc}
\text{Balanced Accuracy}
= \frac{1}{C} \sum_{i=1}^{C} 
  \frac{\text{TP}_i}{\text{TP}_i + \text{FN}_i},
\end{equation}

\noindent
where \(\text{TP}_i\) and \(\text{FN}_i\) respectively denote true positives and false negatives for class \(i\). Balanced Accuracy ensures that each class’s performance contributes equally, making it particularly useful in imbalanced multi-class problems.

\subsubsection{Precision, PPV, or TDR}
\label{metric:precision}

\emph{Precision}, also referred to as \emph{Positive Predictive Value (PPV)} or \emph{True Discovery Rate (TDR)}, quantifies the fraction of predicted positives that are genuinely positive \cite{powers2020evaluation}. In binary classification:

\begin{equation}
\label{eq:prec}
\text{Precision}
= \frac{\text{TP}}{\text{TP} + \text{FP}},
\end{equation}

\noindent
where \(\text{TP}\) is the number of true positives and \(\text{FP}\) is the number of false positives. High Precision indicates that among all samples predicted as positive, a large proportion are correct—an essential property in domains where false alarms are costly (e.g., medical screenings, fraud detection).

\paragraph{Precision vs.\ Recall Trade-off:}
Maximizing Precision alone can be misleading if many actual positives are missed. Often, Precision is analyzed together with Recall (Section \ref{metric:recall_tpr}) to balance the cost of false positives against the cost of false negatives.

\paragraph{Precision in Multi-Class Classification:}
In multi-class settings with \(N\) classes, Precision is computed on a per-class basis and then averaged. Two common averaging schemes are:

\begin{itemize}
    \item \textbf{Macro-average Precision:}
    \begin{equation}
    \label{eq:macro_avg_prec}
    \text{Macro-Precision}
    = \frac{1}{N} \sum_{i=1}^{N} 
      \frac{\text{TP}_i}{\text{TP}_i + \text{FP}_i},
    \end{equation}
    where \(\text{TP}_i\) and \(\text{FP}_i\) refer to class \(i\). Macro-average treats all classes equally, regardless of class frequency.

    \item \textbf{Micro-average Precision:}
    \begin{equation}
    \label{eq:micro_avg_prec}
    \text{Micro-Precision}
    = \frac{\sum_{i=1}^{N} \text{TP}_i}{\sum_{i=1}^{N} (\text{TP}_i + \text{FP}_i)}.
    \end{equation}
    Micro-average aggregates global TP and FP counts across all classes, thus weighting the classes by their support (i.e., frequency).
\end{itemize}

Macro-average Precision is often preferred if one wants to measure performance equally across all classes, whereas Micro-average is influenced more by classes with higher sample counts.

\subsubsection{Recall, Sensitivity, or True Positive Rate (TPR)}
\label{metric:recall_tpr}

\emph{Recall}, also known as \emph{Sensitivity} or \emph{True Positive Rate (TPR)}, measures the proportion of actual positives correctly identified by the model \cite{powers2020evaluation}. In binary classification:

\begin{equation}
\label{eq:recall}
\text{Recall}
= \frac{\text{TP}}{\text{TP} + \text{FN}},
\end{equation}

\noindent
where \(\text{TP}\) denotes true positives and \(\text{FN}\) denotes false negatives. A high Recall means the model correctly identifies most of the positive cases, which is critical in scenarios where missing a positive instance (e.g., a disease diagnosis) is costly.

\paragraph{Recall in Multi-Class Classification:}
Similar to Precision, Recall must be evaluated per class and then averaged to gain an overall measure in multi-class settings:

\begin{itemize}
    \item \textbf{Macro-average Recall:}
    \begin{equation}
    \label{eq:macro_avg_recall}
    \text{Macro-Recall}
    = \frac{1}{N} \sum_{i=1}^{N} 
      \frac{\text{TP}_i}{\text{TP}_i + \text{FN}_i}.
    \end{equation}

    \item \textbf{Micro-average Recall:}
    \begin{equation}
    \label{eq:micro_avg_recall}
    \text{Micro-Recall}
    = \frac{\sum_{i=1}^{N} \text{TP}_i}{\sum_{i=1}^{N} (\text{TP}_i + \text{FN}_i)}.
    \end{equation}
\end{itemize}

\noindent
\emph{Macro-average} gives equal weight to each class, while \emph{Micro-average} aggregates the totals across classes. If class imbalance is severe, the choice of averaging method significantly affects the final reported Recall.

\paragraph{Precision–Recall Trade-off:}
Achieving high Recall often entails predicting more positives, which may inflate \(\text{FP}\) and thus reduce Precision. Conversely, focusing on high Precision can reduce Recall by being more conservative about labeling samples as positive. The \emph{F1-score} (or F-measure) is a common single metric that seeks to balance the two:

\begin{equation}
F_1 
= 2 \cdot \frac{\text{Precision} \times \text{Recall}}{\text{Precision} + \text{Recall}}.
\end{equation}

In practice, the choice to emphasize Precision or Recall (or a specific balance of both) depends on the cost of false positives vs.\ false negatives in the application domain.

\subsubsection{F1-Score}
\label{metric:f1}

The \emph{F1-score} is a statistical measure that combines \emph{Precision} and \emph{Recall} into a single indicator of a classification model’s overall performance \cite{powers2020evaluation}. Specifically, the F1-score is the \emph{harmonic mean} of Precision (\(\mathrm{Prec}\)) and Recall (\(\mathrm{Rec}\)), given by

\begin{equation}
\label{eq:f1}
F_1 
= 2 \cdot \frac{\mathrm{Prec} \;\times\; \mathrm{Rec}}{\mathrm{Prec} + \mathrm{Rec}}.
\end{equation}

\noindent
Because the harmonic mean heavily penalizes extreme disparities between Precision and Recall, the F1-score ensures neither metric is ignored. A higher F1-score indicates a more balanced classifier, capable of both correctly labeling the majority of positive instances (high Recall) and making relatively few false-positive predictions (high Precision).

\paragraph{Use Cases and Limitations:}
\begin{itemize}
    \item \textbf{Imbalanced Datasets:} F1 is especially meaningful when positive cases are rare (e.g., rare disease detection, fraud detection) and where a high Accuracy can be misleading. 
    \item \textbf{Lack of Cost Differentiation:} F1 does not differentiate between \(\mathrm{FP}\) (false positives) and \(\mathrm{FN}\) (false negatives) beyond their impact on Precision and Recall. If the cost of false positives greatly outweighs that of false negatives (or vice versa), a more specialized metric may be necessary.
\end{itemize}

\noindent
\textit{F-Beta Score:}  
An extension to F1 is the \(F_\beta\)-score, which provides a mechanism for weighting Precision and Recall differently. It is defined as

\begin{equation}
\label{eq:f-beta}
F_{\beta}
= \bigl(1 + \beta^2\bigr)
  \cdot
  \frac{\mathrm{Prec}\,\times\,\mathrm{Rec}}{
         \beta^2\,(\mathrm{Prec}) + \mathrm{Rec}},
\end{equation}

\noindent
where \(\beta \in \mathbb{R}^+\). When \(\beta > 1\), Recall receives more emphasis; when \(\beta < 1\), Precision is emphasized. This flexibility allows the metric to be tuned to application-specific costs of false positives and false negatives, making \(F_\beta\) a powerful tool in domains with asymmetrical error concerns.

\subsubsection{Specificity or True Negative Rate (TNR)}
\label{metric:specificity}

\emph{Specificity}, sometimes called the \emph{True Negative Rate (TNR)}, measures the proportion of actual negatives correctly identified as negative \cite{parikh2008understanding}. In binary classification, it is computed as:

\begin{equation}
\label{eq:specif}
\mathrm{Specificity}
= \frac{\mathrm{TN}}{\mathrm{TN} + \mathrm{FP}},
\end{equation}

\noindent
where \(\mathrm{TN}\) (true negatives) is the count of correctly identified negative instances, and \(\mathrm{FP}\) (false positives) is the count of negative instances incorrectly labeled as positive.

\paragraph{Interpretation and Relevance:}
\begin{itemize}
    \item \textbf{Medical Diagnostics:} High Specificity is vital when the cost of a false positive is high (e.g., unnecessary anxiety, invasive follow-up tests). 
    \item \textbf{Financial Fraud Detection:} Overzealous blocking of legitimate transactions (\(\mathrm{FP}\)) can harm customers, making Specificity important.
    \item \textbf{Complementary Metric to Recall:} Whereas Recall (or Sensitivity) concentrates on capturing actual positives, Specificity focuses on \emph{rejecting} actual negatives correctly. In many cases, both metrics are examined together to capture a classifier’s trade-off between false negatives and false positives.
\end{itemize}

\paragraph{Example Trade-offs:}
\begin{itemize}
    \item A test with high Recall but low Specificity may flag most positives but incur many false alarms (\(\mathrm{FP}\)). 
    \item A test with high Specificity but low Recall will likely miss numerous actual positives, but it will rarely mislabel negatives as positives.
\end{itemize}

\noindent
When combined with Recall, Specificity paints a more complete picture of a classifier’s performance, aiding in balancing error types in high-stakes domains. 

\subsubsection{False Positive Rate (FPR)}
\label{metric:FPR}

\emph{False Positive Rate} (\(\mathrm{FPR}\)) measures the proportion of actual negatives that are incorrectly predicted as positive. Also known as the \emph{Type I Error rate}, it is an important complement to \emph{Specificity} (also called the True Negative Rate, \(\mathrm{TNR}\)). Formally, from the confusion matrix elements \(\mathrm{TN}\) (true negatives) and \(\mathrm{FP}\) (false positives), \(\mathrm{FPR}\) is defined as

\begin{equation}
\label{eq:fpr}
\mathrm{FPR}
= \frac{\mathrm{FP}}{\mathrm{TN} + \mathrm{FP}}.
\end{equation}

\noindent
The relationship between FPR and Specificity is given by
\begin{equation}
\mathrm{FPR} 
= 1 - \mathrm{Specificity},
\end{equation}
since
\begin{equation}
\mathrm{Specificity}
= \frac{\mathrm{TN}}{\mathrm{TN} + \mathrm{FP}}.
\end{equation}

\paragraph{Threshold Dependence:}
FPR is heavily influenced by the decision threshold used to classify instances as positive vs.\ negative. In probabilistic models:
\begin{itemize}
    \item \textbf{Lower Threshold:} More instances are labeled as positive, potentially increasing \(\mathrm{FP}\), hence increasing \(\mathrm{FPR}\).
    \item \textbf{Higher Threshold:} Fewer instances are labeled as positive, likely reducing \(\mathrm{FP}\) and hence \(\mathrm{FPR}\).
\end{itemize}

\paragraph{Practical Importance:}
FPR is critical in applications where the cost of a false positive can be high:
\begin{itemize}
    \item \textbf{Financial Fraud Detection:}
    A high FPR means legitimate transactions are flagged as fraudulent, harming customer satisfaction and possibly incurring fees.
    \item \textbf{Medical Tests:}
    A high FPR implies many healthy individuals are flagged as diseased, leading to unnecessary, invasive follow-up tests and anxiety.
\end{itemize}

\paragraph{Usage in ROC Analysis:}
FPR is commonly plotted on the \(x\)-axis of a \emph{Receiver Operating Characteristic} (ROC) curve against the \emph{True Positive Rate} (TPR) on the \(y\)-axis. Varying the classification threshold traces out the ROC curve. An ideal model aims for high TPR and low FPR. \emph{Area Under the ROC Curve} (AUC) then provides a single, threshold-independent measure of how well the model balances \(\mathrm{TPR}\) and \(\mathrm{FPR}\) (see Section~\ref{metric:AUC}).

\subsubsection{Negative Predictive Value (NPV)}
\label{metric:npv}

\emph{Negative Predictive Value} (\(\mathrm{NPV}\)) measures the proportion of negative predictions that are correct. In other words, among all instances labeled negative by the model, \(\mathrm{NPV}\) indicates the fraction that truly are negative. From the confusion matrix elements \(\mathrm{TN}\) (true negatives) and \(\mathrm{FN}\) (false negatives), \(\mathrm{NPV}\) is defined as

\begin{equation}
\label{eq:npv}
\mathrm{NPV}
= \frac{\mathrm{TN}}{\mathrm{TN} + \mathrm{FN}}.
\end{equation}

\paragraph{Interpretation and Relevance:}
\begin{itemize}
    \item \textbf{High-Impact Misses (False Negatives):} NPV is central in contexts where missing a positive case (\(\mathrm{FN}\)) is particularly detrimental (e.g., undiagnosed disease). A high NPV means the model rarely overlooks actual positives when it predicts negative.
    \item \textbf{Prevalence Sensitivity:} Similar to Positive Predictive Value (PPV), \(\mathrm{NPV}\) is sensitive to the prevalence of the target condition. If negatives are abundant, a moderate \(\mathrm{TN}\) plus a small \(\mathrm{FN}\) can yield a high NPV.
    \item \textbf{Comparison with PPV:} PPV measures the correctness of positives, whereas NPV measures the correctness of negatives. Together, they provide a holistic view of a model’s predictive reliability for both classes.
\end{itemize}

\paragraph{Use Cases:}
\begin{itemize}
    \item \textbf{Medical Diagnostics (Rare Diseases):} In screening for a rare disease, \(\mathrm{FN}\) can be fatal if a patient goes untreated. Hence, high NPV (few missed positives when labeled negative) is crucial.
    \item \textbf{Quality Control:} In manufacturing, incorrectly labeling defective items as “non-defective” (\(\mathrm{FN}\)) can allow faulty products to pass inspection, incurring recalls or safety issues.
\end{itemize}

\paragraph{Prevalence Effects:}
Because \(\mathrm{NPV}\) depends on the ratio of actual negatives, changes in the underlying data distribution (e.g., disease prevalence in a population) can affect \(\mathrm{NPV}\) significantly. In designing classification systems or screening tools for low-prevalence conditions, ensuring a sufficiently high NPV often necessitates strategies to reduce \(\mathrm{FN}\), such as lowering the threshold for a positive test or employing further confirmatory tests.

\subsubsection{False Discovery Rate (FDR)}
\label{metric:fdr}

\emph{False Discovery Rate} (\(\mathrm{FDR}\)) quantifies the proportion of predicted positives that are actually negative. It is defined in terms of \(\mathrm{FP}\) (false positives) and \(\mathrm{TP}\) (true positives) as:

\begin{equation}
\label{eq:fdr}
\mathrm{FDR}
= \frac{\mathrm{FP}}{\mathrm{TP} + \mathrm{FP}}.
\end{equation}

\noindent
Thus, \(\mathrm{FDR} = 1 - \mathrm{Precision}\), highlighting its relationship to Precision (\(\mathrm{PPV}\)).

\paragraph{Interpretation:}
\begin{itemize}
    \item \textbf{Positive Class Reliability:} A low FDR indicates that among all instances predicted positive, only a small fraction are false positives. This is essential in domains that cannot afford erroneous positive alerts.
    \item \textbf{Threshold Dependence:} Like other metrics (e.g., FPR), FDR may shift as one adjusts the classification threshold. A more lenient threshold often inflates \(\mathrm{FP}\), driving FDR higher.
\end{itemize}

\paragraph{Domain-Specific Applications:}
\begin{itemize}
    \item \textbf{Scientific Research / Multiple Hypothesis Testing:} FDR control is critical to limit the number of false discoveries—untrue hypotheses declared significant. Techniques like the Benjamini–Hochberg procedure specifically target a controlled FDR in statistical inference.
    \item \textbf{Medical Diagnostics \& Fraud Detection:} Minimizing FDR helps ensure that most of the flagged positives (\(\mathrm{TP} + \mathrm{FP}\)) are actually correct positives (\(\mathrm{TP}\)). Excessive false discoveries can lead to wasted resources, unnecessary treatments, or alarm fatigue.
\end{itemize}

\paragraph{FDR vs.\ FPR:}
While \(\mathrm{FPR}\) is the proportion of false positives among all actual negatives, \(\mathrm{FDR}\) is the proportion of false positives among predicted positives. The choice between focusing on FDR or FPR depends on whether one cares more about:
\begin{itemize}
    \item \(\mathrm{FPR}\): The effect of mislabeling actual negatives.
    \item \(\mathrm{FDR}\): The trustworthiness of declared positives.
\end{itemize}

\noindent
In essence, FDR is invaluable for maintaining credibility in positive predictions. Where false alarms (false positives) are costly or resource-intensive, controlling FDR is a priority.

\subsubsection{Precision-Recall Curve}
\label{metric:prec-recall-curve}

The \emph{Precision-Recall (PR) curve} visualizes the trade-off between \emph{Precision} and \emph{Recall} at varying decision thresholds for a binary classifier \cite{powers2020evaluation}. Recall that:

\begin{equation}
\mathrm{Precision} = \frac{\mathrm{TP}}{\mathrm{TP} + \mathrm{FP}}, 
\qquad
\mathrm{Recall} = \frac{\mathrm{TP}}{\mathrm{TP} + \mathrm{FN}},
\end{equation}
where \(\mathrm{TP}, \mathrm{FP}, \mathrm{FN}\) denote true positives, false positives, and false negatives, respectively.

\paragraph{Computing the PR Curve:}
\begin{enumerate}
    \item Have a binary classifier that outputs a probability or \emph{score} \(s(x)\) for the positive class.
    \item For thresholds \(\tau\) in \(\{0, 0.01, 0.02, \ldots, 1.0\}\), predict “positive” if \(s(x) \geq \tau\), otherwise “negative.”
    \item At each threshold, compute \(\mathrm{Precision}(\tau)\) and \(\mathrm{Recall}(\tau)\). 
    \item Plot \(\mathrm{Recall}(\tau)\) on the \(x\)-axis vs.\ \(\mathrm{Precision}(\tau)\) on the \(y\)-axis.
\end{enumerate}

A PR curve is thus a parametric plot of \(\mathrm{(Recall}, \mathrm{Precision})\) pairs as \(\tau\) varies.

\paragraph{Interpretation of the PR Curve:}
\begin{itemize}
    \item \textbf{Top-Right Corner \((1,1)\)}: Corresponds to a hypothetical “perfect” classifier with no false positives and no false negatives. Approaching this corner indicates jointly high Precision and Recall across many thresholds.
    \item \textbf{Area Under the PR Curve (AUC-PR)}: Summarizes the model’s ability to maintain high Precision and high Recall simultaneously. Unlike the ROC curve, the baseline for AUC-PR depends on the fraction of positive instances in the data (class prevalence).
    \item \textbf{Shape of the Curve}: 
        - A steep initial slope often implies the model can achieve significantly higher Recall before Precision drops sharply.  
        - A more gradual slope suggests the model must sacrifice Precision in order to capture more positives (higher Recall).
    \item \textbf{Model Comparison}: Models whose PR curves lie consistently above another’s generally exhibit better performance at a wide range of thresholds.
\end{itemize}

\begin{figure}[ht]
\centering
\includegraphics[width=\linewidth]{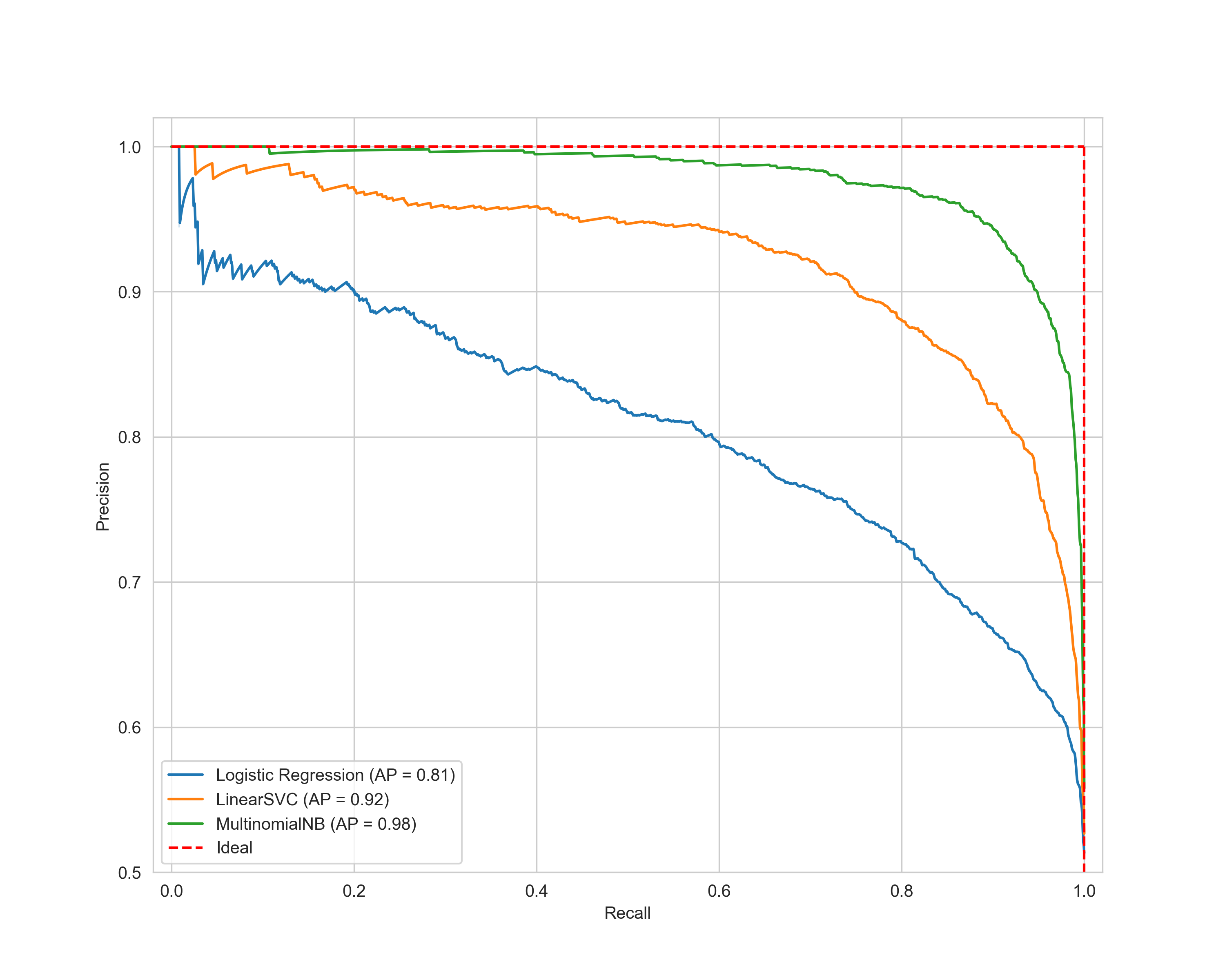}
\caption{Precision-Recall curves for three models trained on the same dataset. The dashed line indicates ideal performance. Each curve’s Average Precision (AP) provides a numerical summary akin to AUC-PR.\label{fig:pr_curve}}
\end{figure}

\subsubsection{Precision-Recall Curve in Multi-Class Classification}
\label{metric:prec-recall-curve-multiclass}

Although PR curves are traditionally formulated for binary classification, they can be extended to multi-class problems using a \emph{One-vs-All (OvA)} approach for each class, treating that class as “positive” and merging all other classes as “negative.” 

\paragraph{Computing Multi-Class PR Curves:}
\begin{enumerate}
    \item For each class \(c \in \{1,2,\ldots,C\}\), obtain probability scores \(s_c(x)\) indicating the model’s confidence that \(x\) belongs to class \(c\).
    \item For each class \(c\), treat it as the positive class and compute \(\mathrm{Precision}_c(\tau)\) and \(\mathrm{Recall}_c(\tau)\) across thresholds \(\tau \in [0,1]\).
    \item Plot a PR curve for each class with \(\mathrm{Recall}_c(\tau)\) on the \(x\)-axis and \(\mathrm{Precision}_c(\tau)\) on the \(y\)-axis.
\end{enumerate}

Each curve shows how well the model distinguishes class \(c\) from all other classes combined.

\paragraph{Interpretation and Use of Multi-Class PR Curves:}
\begin{itemize}
    \item \textbf{Per-Class Analysis}: A curve approaching \((1,1)\) suggests strong class-specific discrimination.  
    \item \textbf{AUC-PR per Class}: One may compute the area under each class’s PR curve, providing a single-value measure of how well the model isolates that class from the rest.
    \item \textbf{Macro vs.\ Micro Averages}: 
        - \emph{Macro-average} treats each class equally, averaging \(\mathrm{Precision}_c\) and \(\mathrm{Recall}_c\) across classes.  
        - \emph{Micro-average} aggregates across all classes, thus weighting classes by their frequencies.
\end{itemize}

\paragraph{Advantages and Limitations:}
Multi-class PR curves offer detailed, class-by-class insights—especially vital in imbalanced datasets or where certain classes are more critical. However, visualizing many classes can become cumbersome. Summary statistics (e.g., macro/micro averaged AUC-PR) or advanced visualization methods may be necessary when \(C\) is large.

\subsubsection{Area Under the Receiver Operating Characteristic Curve (AUC-ROC)}
\label{metric:AUC}

The \emph{Receiver Operating Characteristic} (ROC) curve plots the \emph{True Positive Rate} (TPR) vs.\ the \emph{False Positive Rate} (FPR) at various thresholds. Recall that:

\begin{equation}
\mathrm{TPR} = \frac{\mathrm{TP}}{\mathrm{TP} + \mathrm{FN}},
\qquad
\mathrm{FPR} = \frac{\mathrm{FP}}{\mathrm{TN} + \mathrm{FP}}.
\end{equation}

\paragraph{Definition and Computation:}
\begin{enumerate}
    \item For each threshold \(\tau\) in \([0,1]\), classify an instance as positive if its score \(s(x) \geq \tau\), otherwise negative.
    \item Compute \(\mathrm{TPR}(\tau)\) and \(\mathrm{FPR}(\tau)\).
    \item Plot \(\mathrm{FPR}\) on the \(x\)-axis vs.\ \(\mathrm{TPR}\) on the \(y\)-axis. The resulting curve is the ROC curve.
\end{enumerate}

\paragraph{Interpretation of the ROC Curve:}
\begin{itemize}
    \item \textbf{Diagonal Line}: A purely random classifier. Points above the diagonal indicate better-than-random performance; points below indicate worse than random.
    \item \textbf{Area Under the Curve (AUC)}: Provides a single-value summary of performance. An AUC of 1.0 indicates perfect separability; 0.5 indicates random guessing.
    \item \textbf{Top-Left Corner}: Ideally, a model aims for high TPR (capturing most positives) and low FPR (minimizing false positives).
\end{itemize}

\paragraph{Limitations in Imbalanced Datasets:}
When the positive class is rare, the ROC curve can be overly optimistic, since \(\mathrm{FPR}\) can remain low simply because \(\mathrm{TN}\) is large. In such cases, the Precision-Recall curve might be more informative, highlighting performance on the minority class more sensitively.

\begin{figure}[ht]
\centering
\includegraphics[width=\linewidth]{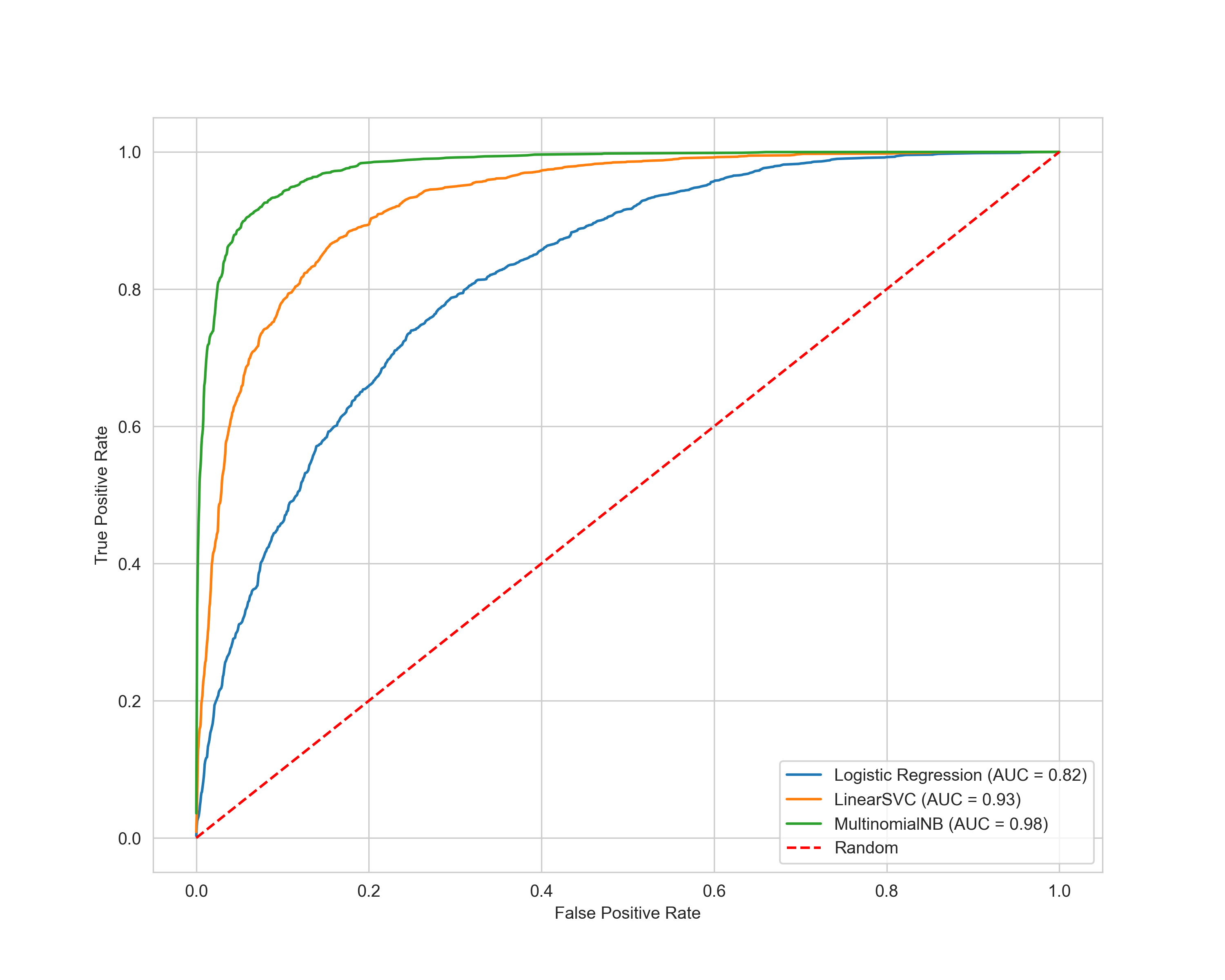}
\caption{ROC curves for three models on the same dataset. The dashed line indicates random performance. Each model’s AUC (area under the curve) is noted in the legend.\label{fig:roc_curve}}
\end{figure}

\subsubsection{Multi-Class Receiver Operating Characteristic Curve (AUC-ROC)}
\label{metric:multi-roc}

Though ROC curves and AUC values were originally devised for binary classification, they can extend to multi-class tasks via multiple strategies.

\paragraph{One-vs-All (OvA) Approach:}
\begin{enumerate}
    \item For each class \(c\), treat it as the positive class, merging all other classes as negative.
    \item Construct a binary ROC curve for class \(c\). Compute its AUC, \(\mathrm{AUC}_c\).
    \item Repeat for each class, yielding \(C\) ROC curves and \(C\) AUC scores.
\end{enumerate}
One can then take the \emph{macro-average} or \emph{weighted-average} of these per-class AUC scores to obtain a single measure of multi-class performance.

\paragraph{One-vs-One (OvO) Approach:}
In problems with \(C\) classes, an OvO approach involves constructing a binary ROC curve for each pair of classes \(\{i, j\}\), treating class \(i\) as positive and \(j\) as negative (and vice versa). This yields \(\frac{C(C-1)}{2}\) curves. A mean AUC across all pairs can then summarize the classifier’s multi-class capability, although this can become computationally expensive for large \(C\).

\paragraph{Macro and Micro-Averaging in Multi-Class ROC:}
\begin{itemize}
    \item \textbf{Macro-Averaged AUC}: Calculates AUC for each OvA (or OvO) ROC curve and then averages them. This treats each class (or class pair) equally, regardless of frequency.
    \item \textbf{Micro-Averaged AUC}: Aggregates the confusion matrix counts over all classes into a single \(\mathrm{TPR}\) and \(\mathrm{FPR}\) curve, effectively weighting classes by their prevalence.
\end{itemize}

\paragraph{Visualization and Interpretation:}
Visualizing multi-class ROC curves can be done by overlaying each OvA curve on a single plot or employing a grid of subplots. The final choice depends on the number of classes and the intended level of detail. Metrics such as macro- and micro-averaged AUCs facilitate simpler quantitative comparisons between models.

\bigskip
\noindent
In the preceding sections, we have covered a wide variety of loss functions and metrics for both regression and classification scenarios. Next, we delve into more specialized applications in computer vision, such as image classification (Section~\ref{subsec:img_class}), object detection (Section~\ref{subsec:obj_det}), image segmentation (Section~\ref{subsec:segmentation}), face recognition (Section~\ref{subsec:face_recog}), monocular depth estimation (Section~\ref{subsec:depth_est} and image generation (Section~\ref{subsec:gen_models}). These tasks introduce additional challenges and specialized metrics tailored to spatial, structural, and interpretability considerations.

\section{Image Classification}
\label{subsec:img_class}

Image classification involves categorizing an image as a whole into a specific label~\cite{paperswithcode_image_classification}.

Image classification was the first task where deep learning demonstrated significant success, marking a critical moment in the advancement of artificial intelligence. This breakthrough occurred primarily with the development and application of Convolutional Neural Networks (CNNs). In 2012, the AlexNet architecture~\cite{krizhevsky2012}, presented by Alex Krizhevsky, Ilya Sutskever, and Geoffrey Hinton, achieved state-of-the-art performance substantially better than the previous best results. This achievement established the potential of deep learning models to outperform traditional machine learning methods that relied heavily on hand-engineered features.

AlexNet's success led to a surge in research and development in deep learning. This resulted in significant improvements in deep learning algorithms and their applications in various domains. Image classification played a key role in this revolution, enabling advances that expanded the scope of deep learning into numerous other applications. This breakthrough has laid a strong foundation for further innovations in the field of AI.

Image classification is used in a diverse range of sectors, demonstrating its broad applicability. In medical imaging, it is used to improve diagnostic accuracy by automatically detecting and classifying abnormalities such as tumors, fractures, and pathological lesions in various types of scans, including X-rays, MRIs, and CT images \cite{Varshney2023ResnetTL,AsharaniKagolanu2022MultiClassMI,Saleh2023MedicalIC,Harika2023ClassificationOC}. In agriculture, it helps to monitor crop health, predict yields, and detect plant diseases through images captured by drones or satellites \cite{Shah2023ApplicationOD,Yuan2023ComputerVA,Nidhis2019ClusterBP,Tendolkar2021ModifiedCH}. The retail industry uses this technology for automatic product categorization, inventory management, and online shopping enhanced with visual search capabilities \cite{peng2021rp2klargescaleretailproduct,shihab2022vistavisiontransformerenhanced,Boriya2019ViSeRAV,Alghamdi2024EmpoweringRT}. Security and surveillance systems use image classification to analyze footage for unauthorized activities, manage crowds, and identify individuals or objects of interest \cite{M2023UnusualCA,Ayad2023ConvolutionalNN}. In environmental monitoring, it is instrumental in tracking deforestation, monitoring changes in the natural environment, and assessing water quality using satellite imagery \cite{Ayad2023ConvolutionalNN,Ramadasan2024ForestRE,Ahmed2022AMA,Farahnakian2023CNNbasedBP,Ilteralp2021ADM}. In addition, in the manufacturing sector, image classification automates quality control processes by identifying defects and verifying assembly, ensuring that products meet stringent standards with minimal human intervention \cite{Abdullah2024LeatherIQ,Mamun2024DetectionAC,Pranoto2023BurrsAS,Yang2020AHD,brown2021deeplabb}.

\subsection{Image Classification Loss Functions}
\label{subsec:img_class_loss}

The core loss functions used for image classification are the same as those described in Section~\ref{subsec_classi_loss_functions}, such as Binary Cross-Entropy, Categorical Cross-Entropy, Weighted Cross-Entropy, Focal Loss, and Hinge Loss (see Table~\ref{tab:guidelines-class-losses}). Despite sharing a common foundation with standard classification tasks, image classification often presents additional considerations:

\begin{itemize}
    \item \textbf{Large-Scale Datasets:} 
    Datasets like ImageNet~\cite{deng2009imagenet}, with millions of labeled images across hundreds or thousands of classes, frequently employ the same cross-entropy-based losses. However, they may require more sophisticated regularization techniques (e.g., label smoothing~\cite{szegedy2016rethinking} or focal loss~\cite{lin2017focal}) to handle a large number of classes and to mitigate overfitting.

    \item \textbf{Class Imbalance in Visual Data:}
    Many image datasets, especially in medical imaging (e.g., rare pathology detection) or specialized industrial inspection, can exhibit severe class imbalance. Weighted Cross-Entropy or Focal Loss becomes essential for emphasizing minority classes, ensuring that rare but crucial categories receive sufficient attention during training.

    \item \textbf{Architectural Considerations:}
    Deep convolutional neural networks (CNNs)~\cite{krizhevsky2012} or Vision Transformers~\cite{dosovitskiy2020image} typically serve as the backbone for image classification. Although these architectures affect the feature extraction process, the fundamental loss computation remains unchanged.

    \item \textbf{Label Smoothing in Vision Tasks:}
    Label Smoothing (Section~\ref{loss:CE_loss_label_smoth}) is particularly common in large-scale vision classification. By replacing one-hot targets with a softly distributed label vector, models avoid becoming overconfident and can better generalize.

    \item \textbf{Multi-Label Image Classification:}
    In some applications, a single image can contain multiple labels (e.g., an image with both ``cat'' and ``outdoor''). Binary cross-entropy losses are often used in multi-label classification setups, with each label treated as an independent positive/negative prediction.
\end{itemize}

Overall, while the mathematical definitions of these loss functions remain consistent across various domains, image classification tasks can require fine-tuning hyperparameters or incorporating additional techniques (like data augmentation or specialized regularizers) to optimize performance in large-scale or highly imbalanced image data.

\subsection{Image Classification Metrics}
\label{subsec:img_class_metrics}

Likewise, the performance metrics for image classification are the same as those described in Section~\ref{subsec:Classif_perform_metrics} (see Table~\ref{tab:metrics_used_classification}). Nonetheless, several image-specific factors and additional measures can be pertinent:

\begin{itemize}
    \item \textbf{Top-k Accuracy:}
    A frequently reported metric in large-scale image classification (e.g., ImageNet benchmarks) is \emph{Top-5 Accuracy}, where a prediction is considered correct if the true class is among the top-5 highest-probability predictions. This metric better reflects performance in settings with many similar classes.

    \item \textbf{Per-Class Accuracy and Balanced Metrics:}
    In highly imbalanced datasets, per-class accuracy or Balanced Accuracy (\S\ref{metric:accuracy}) can offer a more equitable view of model performance across rare and frequent categories. This is crucial in medical imaging, where detecting a rare disease can be more critical than accurately labeling a common condition.

    \item \textbf{Precision-Recall Curves for Rare Classes:}
    For multi-class or multi-label vision tasks where a particular class is rare yet high-impact (e.g., detection of a rare defect in industrial inspection), analyzing Precision-Recall curves (\S\ref{metric:prec-recall-curve} and \S\ref{metric:prec-recall-curve-multiclass}) can give deeper insight into model performance, especially when focusing on minority classes.

    \item \textbf{Confusion Matrices for Visual Inspection:}
    In multi-class image classification, confusion matrices (\S\ref{metric:conf_matrix} and \S\ref{metric:multiclass_conf_matrix}) are often visualized as heatmaps, allowing quick identification of classes that the model confuses (e.g., visually similar species of animals).

    \item \textbf{Robustness and Interpretability:}
    Beyond the standard metrics, recent research in image classification sometimes involves robustness evaluations (e.g., under adversarial attacks~\cite{goodfellow2014explaining} or noise~\cite{carlini2017towards}) and interpretability metrics (e.g., saliency map consistency~\cite{daroya2023cose}). While these are not strictly performance metrics in the conventional sense, they reflect additional real-world considerations.

    \item \textbf{Multi-Label Precision, Recall, and F1:}
    In tasks where images can have multiple labels simultaneously (e.g., a single image could be ``dog,'' ``outdoor,'' and ``playful''), metrics like Micro-/Macro-Precision, Recall, and F1 (\S\ref{metric:precision}, \S\ref{metric:recall_tpr}, and \S\ref{metric:f1}) need to be computed across all relevant labels. This ensures a complete view of performance when each image can belong to multiple categories.
\end{itemize}

While the core classification metrics (Accuracy, Precision, Recall, F1-score, ROC-AUC, Precision-Recall AUC, etc.) remain fundamentally the same, image classification scenarios often require additional approaches (e.g., top-k accuracy, multi-label evaluation) and visual methods (e.g., confusion matrices as heatmaps) to handle the unique challenges posed by large-scale or imbalanced image data.

\section{Object Detection}
\label{subsec:obj_det}

Object detection in deep learning is a computer vision technique that involves localizing and recognizing objects in images or videos. It is common in various applications such as autonomous driving \cite{bojarski2016,codevilla2018,sadeghian2019,chen2020}, surveillance \cite{chalapathy2017,li2018,zhang2016}, human-computer interaction \cite{kim2018-CI,lotte2018,sutskever2014,li2017}, and robotics \cite{levine2018,zhang2019,gualtieri2018,hussein2017}. Object detection involves identifying the presence of an object, determining its location in an image, and recognizing the object's class.

\subsection{Object Detection Loss Functions}
\label{subsubsec:obj_det_loss}

Object detection requires both localization---the accurate regression of bounding box coordinates around objects of interest---and classification---the correct assignment of class labels. Consequently, models typically employ composite loss functions comprising a classification component (e.g., cross-entropy-based) and a regression component (e.g., L1/L2 variants). By jointly penalizing misclassifications and bounding-box inaccuracies, these loss functions guide the training process toward accurate and robust detection across a variety of data distributions.

Table~\ref{tab:guidelines-objdet-losses} provides an overview of commonly used loss functions in object detection, highlighting their applications, advantages, and limitations across different tasks and datasets.

\noindent
Formally, for an image \(\mathbf{I}\) containing \(N\) ground-truth objects, a typical detection model outputs:
\begin{equation}
\Bigl\{\bigl(\hat{c}_i, \hat{\mathbf{b}}_i\bigr)\Bigr\}_{i=1}^{M},
\end{equation}
where \(\hat{c}_i\) denotes predicted class probabilities (or logits) for the \(i\)-th detected region, \(\hat{\mathbf{b}}_i \in \mathbb{R}^4\) are the predicted bounding-box coordinates (e.g., \(\{x_\mathrm{min}, y_\mathrm{min}, x_\mathrm{max}, y_\mathrm{max}\}\) or center/width/height parameterization), and \(M\) is the number of proposals or anchors. The overall loss \(\mathcal{L}\) is often a sum:
\begin{equation}
\label{eq:objdet_loss_generic}
\mathcal{L} 
= \mathcal{L}_{\mathrm{class}}\bigl(\{c_i\}, \{\hat{c}_i\}\bigr)
+ \lambda\,\mathcal{L}_{\mathrm{reg}}\bigl(\{\mathbf{b}_i\}, \{\hat{\mathbf{b}}_i\}\bigr),
\end{equation}
where \(\lambda\) is a weighting factor, and \(\{c_i,\mathbf{b}_i\}\) denote ground-truth class and bounding-box parameters. Below, we focus on the main bounding-box regression losses \(\mathcal{L}_{\mathrm{reg}}\).

\begin{table*}[ht!]
\caption{Guidelines for selecting an object detection loss function based on usage, data characteristics, advantages, and limitations.}
\label{tab:guidelines-objdet-losses}
\centering
\footnotesize
\begin{tabular}{p{1.3cm}p{2.8cm}p{2.8cm}p{2.8cm}p{2.8cm}}
\toprule
\textbf{Function} 
& \textbf{Usage} 
& \textbf{Data Characteristics} 
& \textbf{Advantages} 
& \textbf{Limitations} \\
\midrule

Smooth L1 Loss 
& Bounding box regression (e.g., RPN) 
& Handles moderate outliers; requires threshold $\beta$ 
& Robust to outliers; less sensitive than MSE 
& Requires $\beta$ tuning; can degrade to L1 or near-0 loss \\

\midrule
Balanced L1 Loss 
& Enhanced L1 for varying error magnitudes 
& Helpful for large or small bounding-box errors 
& Dynamic penalty; improves convergence, outlier handling 
& Extra hyperparameters; slightly higher complexity \\

\midrule
IoU Loss 
& Directly optimizes overlap in object detection 
& Suitable when bounding-box overlap is primary metric 
& Aligns with common evaluation metric; intuitive 
& Zero gradient if no overlap; limited partial-overlap sensitivity \\

\midrule
GIoU Loss 
& Extension of IoU for non/partially overlapping boxes 
& Useful when frequent non-overlaps occur 
& Overcomes IoU’s zero-gradient issue; better feedback 
& Does not handle centroid distance or aspect ratio \\

\midrule
DIoU, CIoU 
& Improved bounding-box regression (centroid, aspect ratio) 
& Applicable to precise alignment needs 
& DIoU handles spatial misalignment; CIoU adds aspect-ratio term 
& More parameters ($\alpha$, etc.); higher tuning complexity \\

\midrule
Focal Loss 
& Addresses severe class imbalance 
& Effective for rare-class detection in large datasets 
& Down-weights easy negatives; focuses on hard examples 
& Extra parameters ($\gamma$, $\alpha$); can be tricky to tune \\

\midrule
YOLO Loss 
& Composite loss for YOLO detectors (localization + classification + objectness) 
& Designed for grid-based real-time detection 
& Single-pass optimization; efficient in practice 
& Requires balancing components; may struggle with small objects \\

\midrule
Wing Loss 
& Robust regression for facial landmarks and similar tasks 
& Suitable for data with heavy outliers 
& Less sensitive to outliers; emphasizes medium errors 
& Requires careful tuning of $\omega$ and $\epsilon$; instability risk \\

\bottomrule
\end{tabular}
\end{table*}

\subsubsection{Smooth L1 Loss}
\label{loss:smooth_l1}

\emph{Smooth L1} (sometimes called \emph{L1-Smooth} or Huber-like) is a robust loss function commonly used in bounding-box regression. Originally introduced in the Fast R-CNN framework \cite{girshick2015fast}, Smooth L1 mitigates outliers more effectively than Mean Squared Error (MSE), while retaining differentiability at zero. For a single coordinate, it is defined as:

\begin{equation}
\label{eq:smoothl1}
L_{\mathrm{smoothL1}}(y, \hat{y})
=
\begin{cases}
\tfrac{1}{2}\,(y - \hat{y})^2, & \text{if } \lvert y - \hat{y}\rvert < \beta, \\[6pt]
\lvert y - \hat{y}\rvert - \tfrac{1}{2}\,\beta, & \text{otherwise},
\end{cases}
\end{equation}
where \(y\) is the true bounding-box coordinate (e.g., an offset or width/height), \(\hat{y}\) is the predicted value, and \(\beta > 0\) is a threshold that controls the transition between the L2-like (quadratic) region and the L1-like (linear) region.

\paragraph{Relation to Huber Loss:}
Smooth L1 is closely related to Huber Loss (Section~\ref{loss:Huber}). In fact, one can consider Huber Loss scaled by \(1/\beta\), with subtle differences in slope and transitions \cite{pytorch_smoothl1loss}:
\begin{itemize}
  \item \textbf{As} \(\beta \to 0\), Smooth L1 \(\to\) absolute L1 loss.  
  \item \textbf{As} \(\beta \to +\infty\), Smooth L1 \(\to\) constant zero in the wide region, while Huber transitions to MSE behavior.
  \item \textbf{Slope Differences:} In Smooth L1, the slope in the L1 region is always 1, whereas in Huber Loss, the slope in the linear region is \(\beta\).
\end{itemize}

\paragraph{Usage:}
Smooth L1 is employed extensively in two-stage object detectors (e.g., Fast R-CNN, Faster R-CNN \cite{ren2015faster}). The Region Proposal Network (RPN) uses Smooth L1 for refining anchor boxes into region proposals, and the second stage uses it again for final bounding-box adjustments. Empirically, it offers stable gradients and is less sensitive to large bounding-box errors than MSE.

\paragraph{Practical Considerations:}
\begin{itemize}
  \item \textbf{Anchor/Proposal Mappings:} Bounding-box regression often predicts offsets relative to anchor or proposal boxes, making the typical \((x, y, w, h)\) transformation more robust when combined with Smooth L1.
  \item \textbf{Hyperparameter \(\beta\):} Typical values range from \(0.5\) to \(1.0\). Tuning \(\beta\) can yield small but noticeable improvements in detection accuracy for particular datasets.
  \item \textbf{Gradient Behavior:} The piecewise structure ensures that minor deviations are penalized quadratically (promoting stable convergence), whereas large deviations shift to a linear penalty, reducing the effect of outliers.
\end{itemize}

\subsubsection{Balanced L1 Loss}
\label{loss:balanced_l1}

\emph{Balanced L1} \cite{pang2019libra} is an enhancement over standard L1 for bounding-box regression, designed to improve robustness when object sizes vary widely or when large vs.\ small errors must be balanced differently. It adaptively modulates the penalty based on the magnitude of the error \(x = y - \hat{y}\), applying a logarithmic form for large errors and a quadratic-like form for small errors. The loss is defined as:

\begin{equation}
\label{eq:balanced_l1}
L_{\mathrm{balanced}}(x)
=
\begin{cases} 
\beta \,\ln\Bigl(\dfrac{|x|}{\beta} + 1\Bigr), & \text{if } |x|\ge \beta, \\[6pt]
\dfrac{x^2 + \beta^2}{2\beta}, & \text{otherwise},
\end{cases}
\end{equation}
where \(x\) is the bounding-box error and \(\beta\) is a threshold controlling the transition between these regions.

\paragraph{Properties:}
\begin{itemize}
    \item \textbf{Logarithmic Penalty for Large Errors:} When \(\lvert x\rvert \gg \beta\), the penalty grows more slowly (logarithmically), mitigating gradient explosions that can occur with pure L1 in the presence of outliers or misaligned anchors.
    \item \textbf{Quadratic Penalty for Small Errors:} Near \(\lvert x\rvert < \beta\), the squared term \(\frac{x^2}{2\beta}\) fosters fine-grained accuracy, akin to a smoothed L1 approach.
    \item \textbf{Stabilizing Training:} By balancing the response to large vs.\ small errors, the Balanced L1 Loss can yield more stable gradient updates and faster convergence.
\end{itemize}

\paragraph{Benefits for Object Detection:}
\begin{itemize}
    \item \textbf{Varying Object Scales:} Datasets like MS COCO or Pascal VOC contain objects of vastly different sizes. Balanced L1 helps ensure large errors (e.g., on very big bounding boxes) do not overly dominate training while still providing precise adjustment for smaller bounding boxes.
    \item \textbf{Outlier Resistance:} Sudden large bounding-box offsets, often arising in early training stages or challenging examples, are handled gracefully via the log function, preventing gradient spikes.
    \item \textbf{Integration with Classification Loss:} As with Smooth L1, Balanced L1 is commonly added to a classification loss (e.g., Focal Loss or cross-entropy) for the overall detection loss \eqref{eq:objdet_loss_generic}.
\end{itemize}

\paragraph{Hyperparameter Tuning:}
\(\beta\) generally lies in a range (\(0.5\) to \(2\)), and in some implementations, additional coefficients modulate the log or quadratic segments. Empirical tuning or cross-validation can improve performance on specific datasets (indoor vs.\ outdoor images, small objects vs.\ large objects, etc.).

\paragraph{When to Use Balanced L1:}
While standard Smooth L1 suffices for many detection tasks, Balanced L1 can be advantageous when:
\begin{itemize}
  \item The dataset exhibits a broad distribution of object scales or unusual bounding-box aspect ratios.
  \item One observes slow convergence or instability due to extreme bounding-box outliers.
  \item Precisely localizing small objects while also handling large ones is critical.
\end{itemize}

Balanced L1 loss offers a flexible, adaptive mechanism for bounding-box regression, addressing limitations of purely linear or purely quadratic losses in heterogeneous detection scenarios.

\subsubsection{Intersection Over Union (IoU) Loss}
\label{loss:IoUloss_obj_det}

\emph{Intersection over Union} (IoU) measures the overlap between two bounding boxes, typically denoted as the predicted box \(B_p\) and the ground-truth box \(B_t\). Mathematically, the IoU is defined as:

\begin{equation}
\label{eq:iou_loss}
\mathrm{IoU}
= \frac{\mathrm{Area}\bigl(B_p \cap B_t\bigr)}{\mathrm{Area}\bigl(B_p \cup B_t\bigr)},
\end{equation}

\noindent
where \(\mathrm{Area}(\cdot)\) calculates the 2D area of a bounding box, and \(B_p \cap B_t\) denotes their intersection region. The corresponding \emph{IoU Loss} is then:

\begin{equation}
\label{eq:iou2}
L_{\mathrm{IoU}}(B_p, B_t) = 1 - \mathrm{IoU}(B_p, B_t).
\end{equation}

\begin{figure}[ht]
  \centering
  \includegraphics[width=\linewidth]{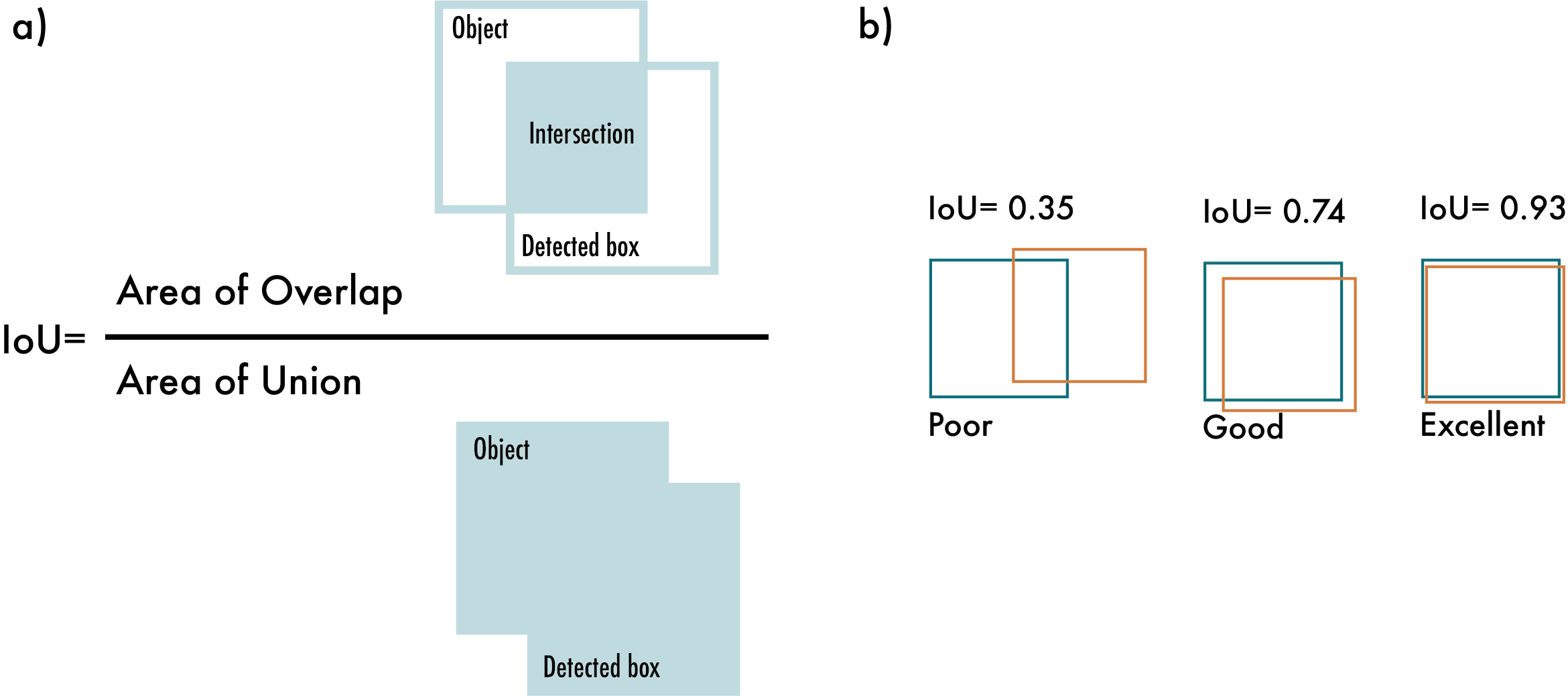}
  \caption{Intersection Over Union (IoU). (a) The IoU is computed as the ratio of the intersection area of two boxes over their union area. (b) Examples of three different IoU values for varied box positions.}
  \label{fig:iou_det}
\end{figure}

\paragraph{Usage in Object Detection:}
One-stage detectors (e.g., SSD \cite{liu2016ssd} and YOLO \cite{redmon2016you}) commonly include IoU Loss as a localization term in a multi-task objective alongside a classification loss. Minimizing \(1 - \mathrm{IoU}\) encourages a high overlap between \(B_p\) and \(B_t\), thus improving bounding-box alignment. Refer to Figure~\ref{fig:iou_det}.

\paragraph{Limitations:}
\begin{itemize}
  \item \textbf{Zero Gradient for Non-Overlapping Boxes:} If \(B_p\cap B_t = \emptyset\), the intersection area is zero, yielding \(\mathrm{IoU} = 0\). This offers no gradient signal to adjust \(B_p\), making optimization difficult when boxes do not overlap initially.
  \item \textbf{Sensitivity to Partial Overlaps:} IoU may change slowly when boxes partially overlap but have varying aspect ratios or positions, limiting its ability to distinguish subtly different bounding-box configurations.
\end{itemize}

These issues motivate generalizations such as GIoU, DIoU, and CIoU, which enhance gradient signals for bounding-box regression under non-overlapping or partially overlapping conditions.

\subsubsection{GIoU Loss}
\label{loss:GIoU}

\emph{Generalized IoU (GIoU) Loss} \cite{rezatofighi2019generalized} extends IoU Loss by accounting for cases where bounding boxes do not overlap or only partially overlap. It incorporates the area outside both bounding boxes but within their smallest enclosing box. Formally:

\begin{equation}
\label{eq:giou}
L_{\mathrm{GIoU}}\bigl(B_p, B_t\bigr)
= 1 - \mathrm{IoU}\bigl(B_p, B_t\bigr)
  + \frac{\mathrm{Area}\bigl(\mathcal{C}\setminus (B_p \cup B_t)\bigr)}{\mathrm{Area}(\mathcal{C})},
\end{equation}

\noindent
where \(\mathcal{C}\) is the smallest bounding box that encloses both \(B_p\) and \(B_t\). Compared to the standard IoU loss, GIoU offers:

\begin{itemize}
    \item \textbf{Non-Overlap Sensitivity:} By including the region \(\mathcal{C}\setminus (B_p \cup B_t)\) in the loss, GIoU penalizes disjoint boxes and guides them to move closer, even if they do not initially intersect.
    \item \textbf{More Stable Gradients:} GIoU provides a continuous gradient signal through all stages of bounding-box refinement, improving convergence.
\end{itemize}

\paragraph{Limitations of GIoU:}
\begin{enumerate}
    \item \textbf{Spatial Misalignment:}
    GIoU does not directly address centroid distance between boxes, which can still result in suboptimal positioning even if GIoU is high.
    \item \textbf{Aspect Ratio Differences:}
    GIoU does not incorporate width-height ratio mismatches, potentially leading to inaccurate regressions when object shapes vary significantly.
\end{enumerate}

\subsubsection{DIoU and CIoU Losses}
\label{loss:DIoU_CIoU}

\emph{Distance-IoU (DIoU)} and \emph{Complete-IoU (CIoU)} \cite{zheng2020distance} are further refinements of IoU-based losses, introduced to handle spatial misalignment and aspect ratio differences more explicitly.

\paragraph{DIoU Loss:}
\begin{equation}
\label{eq:diou}
L_{\mathrm{DIoU}}\bigl(B_p, B_t\bigr)
= 1 - \mathrm{IoU}\bigl(B_p, B_t\bigr)
  + \frac{d^2\bigl(B_p, B_t\bigr)}{c^2},
\end{equation}
where \(d\bigl(B_p, B_t\bigr)\) is the Euclidean distance between the centroids of the two boxes, and \(c\) is the diagonal length of their smallest enclosing box. Hence, DIoU penalizes large centroid offsets even if the boxes overlap reasonably well.

\paragraph{CIoU Loss:}
\begin{equation}
\label{eq:ciou2}
L_{\mathrm{CIoU}}\bigl(B_p, B_t\bigr)
= L_{\mathrm{DIoU}}\bigl(B_p, B_t\bigr)
  + \alpha \cdot v,
\end{equation}
where \(v\) measures the consistency in aspect ratios:

\begin{equation}
v = \frac{4}{\pi^2} 
    \bigl(\arctan(\tfrac{w_t}{h_t}) - \arctan(\tfrac{w_p}{h_p})\bigr)^2,
\end{equation}
and \(\alpha\) is a weighting factor defined as:

\begin{equation}
\label{eq:alpha_iou}
\alpha = \frac{v}{(1 - \mathrm{IoU}) + v}.
\end{equation}

\noindent
Here, \((w_t, h_t)\) and \((w_p, h_p)\) are the ground-truth and predicted bounding-box widths and heights, respectively. By combining \(\mathrm{DIoU}\) with an aspect-ratio alignment term, CIoU yields more comprehensive bounding-box alignment.

\paragraph{Advantages of DIoU and CIoU:}
\begin{itemize}
    \item \textbf{Centroid Alignment:} DIoU addresses scenarios where boxes overlap in area but remain spatially misaligned, improving positioning accuracy.
    \item \textbf{Aspect Ratio Correction:} CIoU adjusts for mismatches in width-height ratio, reducing the chance that two boxes have significant overlap but very different shapes.
    \item \textbf{Stable, Continuous Gradients:} Both losses provide a gradient signal even when boxes do not fully overlap, aiding the early stages of training.
\end{itemize}

\paragraph{Practical Considerations:}
\begin{itemize}
    \item \textbf{Computational Complexity:}
    Calculating DIoU and CIoU involves centroid distance and aspect ratio terms but adds minimal overhead compared to simpler IoU computations.
    \item \textbf{Hyperparameters:}
    The weighting factor \(\alpha\) in CIoU is dynamically computed \eqref{eq:alpha_iou}; thus, no extra hyperparameter is usually needed beyond standard bounding-box regression settings.
    \item \textbf{Integration in Detection Pipelines:}
    Both DIoU and CIoU replace the standard bounding-box loss term in multi-task objectives. They can be seamlessly plugged into frameworks such as YOLO or Faster R-CNN to enhance localization quality.
\end{itemize}

By addressing both centroid offset and aspect-ratio mismatch, DIoU and CIoU further refine IoU-based losses, leading to improved localization accuracy and more robust bounding-box regression in modern object detection pipelines.

\subsubsection{Focal Loss}
\label{loss:focal}

\emph{Focal Loss}, introduced by Lin~\emph{et~al.} \cite{lin2017focal}, is an enhancement of the standard cross-entropy loss aimed at tackling severe class imbalance. In many detection tasks, background (negative) samples vastly outnumber foreground (positive) samples, causing traditional loss functions to bias the model toward the majority class. Focal Loss modifies cross-entropy by down-weighting well-classified examples, thus shifting focus onto “hard,” misclassified samples.

\paragraph{Definition:}
Let \(p \in [0,1]\) be the predicted probability for the ground-truth class \(y\in\{0,1\}\). Define
\begin{equation}
p_t 
= \begin{cases}
    p,   & \text{if } y=1,\\
    1-p, & \text{if } y=0.
\end{cases}
\end{equation}
The \emph{Focal Loss} is then:

\begin{equation}
\label{eq:focal}
FL(p_t)
= -\alpha_t 
    \,(1 - p_t)^{\gamma} 
    \,\log\bigl(p_t\bigr),
\end{equation}
where \(\alpha_t \in [0,1]\) is a class-dependent weighting factor (often set inversely proportional to class frequency), and \(\gamma \ge 0\) is the \emph{focusing parameter} that reduces the relative loss for well-classified examples.

\paragraph{Interpretation:}
\begin{itemize}
    \item \textbf{Focus on Hard Examples:} When \(p_t\) is large (the example is predicted correctly with high confidence), \((1-p_t)^\gamma\approx 0\), thus diminishing the contribution of that “easy” example to the loss. Conversely, misclassified or uncertain examples (\(p_t\) small) incur larger losses, guiding the model to improve minority or hard-to-classify instances.
    \item \textbf{Imbalanced Data:} By setting \(\alpha_t\) based on class frequencies, Focal Loss helps the model pay sufficient attention to underrepresented classes.
    \item \textbf{Typical Hyperparameters:} Common values for \(\gamma\) range from 1 to 5. Larger \(\gamma\) emphasizes misclassified examples more strongly, but if too large, it may destabilize training.
\end{itemize}

\paragraph{Applications:}
Focal Loss has been extensively adopted in:
\begin{itemize}
    \item \textbf{One-stage Detectors:} RetinaNet \cite{lin2017focal} uses Focal Loss to mitigate class imbalance in dense object detection.
    \item \textbf{Medical Imaging:} Enhances detection of rare pathologies by strongly penalizing misclassifications of minority disease classes.
    \item \textbf{Segmentation and Pose Estimation:} Similar imbalance scenarios (background vs.\ foreground pixels, etc.) can benefit from the focusing property.
\end{itemize}

\subsubsection{YOLO Loss}
\label{loss:yolo}

\emph{YOLO Loss} \cite{redmon2016you} is a composite loss function specifically crafted for the YOLO family of one-stage object detectors. YOLO processes an image in a single forward pass, dividing the input into grid cells, each predicting bounding-box coordinates, an “objectness” score, and class probabilities.

The standard YOLO loss combines three terms:

\begin{enumerate}
    \item \textbf{Localization Loss:}
    \begin{equation}
      L_{\mathrm{loc}}=\sum_{\text{pred boxes}}
          \lambda_{\mathrm{coord}}
          \Bigl[
            (x_i - \hat{x}_i)^2 
          + (y_i - \hat{y}_i)^2
          + (w_i - \hat{w}_i)^2 
          + (h_i - \hat{h}_i)^2
          \Bigr],
    \end{equation}
    where \(\bigl(x_i, y_i, w_i, h_i\bigr)\) and \(\bigl(\hat{x}_i,\hat{y}_i,\hat{w}_i,\hat{h}_i\bigr)\) are ground-truth and predicted box parameters, respectively, and \(\lambda_{\mathrm{coord}}\) controls the relative importance of localization accuracy.

    \item \textbf{Objectness (Confidence) Loss:}
    \begin{equation}
    \label{eq:yolo_obj_loss}
    L_{\mathrm{obj}}
    = \sum_{\text{pred boxes}}
      \bigl(C_i - \hat{C}_i\bigr)^2,
    \end{equation}
    where \(C_i\in\{0,1\}\) indicates whether an object is present, and \(\hat{C}_i\) is the predicted confidence score (sometimes interpreted as the IoU of the predicted box with any ground truth).

    \item \textbf{Classification Loss:}
    \begin{equation}
      L_{\mathrm{class}}
      = \sum_{\text{pred boxes}}
        -\sum_{c=1}^{C}
           p_i(c)\,\log\bigl(\hat{p}_i(c)\bigr),
    \end{equation}
    where \(p_i(c)\) is the one-hot vector of the true class and \(\hat{p}_i(c)\) is the predicted probability for class \(c\). Typically, standard cross-entropy is used here.
\end{enumerate}

\paragraph{YOLO Loss:}
\begin{equation}
\mathcal{L}_{\mathrm{YOLO}} 
= L_{\mathrm{loc}}
 + L_{\mathrm{obj}}
 + L_{\mathrm{class}}.
\end{equation}
Adjusting the weighting factors in each term (e.g., \(\lambda_{\mathrm{coord}}\)) balances the trade-off among bounding-box precision, object confidence, and classification accuracy.

YOLO’s grid-based approach relies heavily on these combined losses to perform localization and classification simultaneously. Early YOLO versions used sum-of-squares for bounding-box terms, which can be sensitive to object scale. Subsequent YOLO variants introduced modifications like IoU-based or GIoU-based losses for more robust box regression.

\subsubsection{Wing Loss}
\label{loss:wing}

\emph{Wing Loss} \cite{feng2018wing} was proposed to address the limitations of pure \(L_1\) or \(L_2\) (MSE) losses in tasks sensitive to both small and large localization errors. In bounding-box regression (and related tasks such as facial landmark localization), mild errors should be penalized differently from outliers.

\paragraph{Definition:}
Let \(x\) be the prediction error (e.g., the difference between predicted and ground-truth coordinates), and let \(\omega\) and \(\epsilon\) be hyperparameters. Wing Loss is piecewise-defined:

\begin{equation}
\label{eq:wing_obj_loss}
\mathrm{WingLoss}(x)
= 
\begin{cases}
\omega \,\ln\!\Bigl(1 + \frac{|x|}{\epsilon}\Bigr),
 & \text{if } |x| < \omega,\\[6pt]
|x| - C,
 & \text{otherwise},
\end{cases}
\end{equation}
where 
\begin{equation}
C = \omega 
    - \omega\,\ln\Bigl(1 + \frac{\omega}{\epsilon}\Bigr).
\end{equation}
The constant \(C\) ensures continuity of the function at \(|x| = \omega\).

\paragraph{Properties:}
\begin{itemize}
    \item \textbf{Logarithmic Region:} For errors \(|x|<\omega\), the penalty grows \emph{logarithmically} with \(\lvert x\rvert\), increasing sensitivity to small and moderate errors compared to L1’s linear slope.
    \item \textbf{Linear Region:} For larger errors (\(|x|\ge\omega\)), the loss reverts to a linear form \(\lvert x\rvert -C\), preventing extreme outliers from dominating gradient updates.
    \item \textbf{Hyperparameters \(\omega\) and \(\epsilon\):}
          - \(\omega\) sets the threshold where the function transitions from the logarithmic regime to the linear regime.  
          - \(\epsilon\) controls the steepness of the log curve. Large \(\epsilon\) increases sensitivity to moderate errors, but may require careful training to avoid instability.
\end{itemize}

\paragraph{Usage:}
\begin{itemize}
    \item \textbf{Facial Landmarks:} Wing Loss was originally proposed for face alignment tasks, ensuring small errors are penalized more precisely without overly punishing large outliers.
    \item \textbf{Potential for Bounding-Box Regression:} The same principle can be extended to object bounding-box localization, especially in tasks where small positional deviations can significantly impact performance (e.g., medical imaging).
\end{itemize}

Similar to Smooth L1 (Section~\ref{loss:smooth_l1}), Wing Loss dampens large errors but handles small errors differently via a logarithmic term. Users may prefer Wing Loss when the distribution of regression errors is such that capturing “near-accurate” predictions is crucial, and absolute outliers must not dominate training.

\subsection{Object Detection Metrics}
\label{subsec:obj_det_metrics}

In object detection, a model must correctly localize objects (bounding box regression) and classify them (object category). During evaluation, we typically rely on True Positives (TP), False Positives (FP), False Negatives (FN), and (in some definitions) True Negatives (TN) to quantify how well the detector performs. However, unlike in image-level classification, the notion of positive vs.\ negative predictions depends on spatial overlap between boxes, usually measured by the \emph{Intersection over Union} (IoU) (see Figure~\ref{fig:iou_det} and Section~\ref{loss:IoUloss_obj_det}).

\paragraph{True Positive (TP):}
A predicted bounding box is deemed a true positive if:
\begin{equation}
\mathrm{IoU}\bigl(B_p, B_t\bigr) \;\;\geq\;\; \tau,
\end{equation}
where \(B_p\) is the predicted box, \(B_t\) is the ground-truth box, and \(\tau\in[0,1]\) is an IoU threshold. Common thresholds (\(0.25, 0.5, 0.75\)) reflect different stringency levels. For instance, \(\tau=0.5\) is typical in many benchmarks, whereas higher thresholds (e.g., \(0.75\)) demand more precise localization.

\paragraph{False Positive (FP):}
A predicted box is labeled as a false positive if it does not meet the IoU threshold for \emph{any} ground-truth box (i.e., no match), or if it is deemed redundant after a match has already been assigned to another prediction. High FP counts degrade \emph{Precision}.

\paragraph{False Negative (FN):}
A ground-truth box not matched by any prediction above threshold \(\tau\) is a false negative, indicating the model missed a true object in the image. High FN counts degrade \emph{Recall}.

\paragraph{True Negative (TN):}
In many detection settings, TN is less frequently reported. Conceptually, TN indicates that the model correctly concludes no object is present in a negative region. However, since potential bounding-box search space is huge, TN is not always explicitly counted.

\subsubsection*{Common IoU Thresholds in Object Detection}

\begin{itemize}
    \item \textbf{IoU\,\(\ge0.5\):} Widely used in benchmarks like Pascal VOC~\cite{everingham2010pascal}, establishing a balance between moderate localization precision and practical acceptance. 
    \item \textbf{IoU\,\(\ge0.75\):} Used in stricter scenarios (e.g., autonomous driving) where precise localization is critical and false positives (poorly localized boxes) can have severe consequences.
    \item \textbf{IoU\,\(\ge0.25\):} Occasionally used in tasks requiring high recall (e.g., medical imaging), ensuring that even loosely fitting bounding boxes can count as detections to avoid missing important findings.
\end{itemize}

\noindent
Typical object detection metrics include:
\begin{itemize}
    \item \emph{Average Precision (AP)} and \emph{Mean Average Precision (mAP)}
    \item \emph{Intersection over Union (IoU)} (Section~\ref{loss:IoUloss_obj_det})
    \item \emph{Precision-Recall Curve} (Section~\ref{metric:prec-recall-curve})
\end{itemize}

\subsubsection{Average Precision (AP)}
\label{metric:AP}

In multi-class object detection, a detector must correctly localize and classify objects across multiple categories. \emph{Average Precision (AP)} quantifies per-category performance and then averages over categories (often termed \emph{mean Average Precision} or mAP). By doing so, each class’s performance is individually assessed, preventing categories with abundant instances from overshadowing others.

AP calculation typically incorporates an IoU threshold \(\tau\), such that a predicted box is considered a correct detection if \(\mathrm{IoU} \ge \tau\). Datasets like COCO \cite{lin2014microsoft} examine multiple IoU thresholds (\(\tau=0.5,0.55,\ldots,0.95\)) to measure performance from lenient to strict localization criteria.

\subsubsection*{Pascal VOC AP Computation}

The Pascal VOC dataset \cite{everingham2010pascal} includes 20 object categories. The classical procedure for computing AP in VOC is:

\begin{enumerate}
    \item \emph{Compute IoU:} For each detection, compute IoU with ground-truth objects. A detection is matched to the ground-truth box with the highest IoU above \(\tau\).
    \item \emph{Precision and Recall Curve:} For each category, sort predictions by confidence score (descending). Moving through these detections from highest to lowest confidence defines a sequence of precision-recall pairs \(\bigl(\mathrm{Rec}(\theta),\,\mathrm{Prec}(\theta)\bigr)\) as the decision threshold \(\theta\) changes.
    \item \emph{11-Point Interpolation:} VOC uses recall levels \(\{0,0.1,0.2,\dots,1.0\}\). For each recall level \(r\), the \emph{interpolated precision} is the maximum precision for any recall \(\ge r\). This yields a piecewise constant precision-recall curve.
    \item \emph{Area Under Curve:} The AP is the area under this interpolated curve, computed as a sum over the discrete recall bins:
    \begin{equation}
    \mathrm{AP}
    = \sum_{r\in\{0,0.1,\dots,1.0\}}
      \frac{\mathrm{Prec}_{\mathrm{interp}}(r)}{11}.
    \end{equation}
\end{enumerate}

\subsubsection*{Microsoft COCO AP Computation}

The COCO dataset \cite{lin2014microsoft} spans 80 categories and employs a more extensive AP definition:

\begin{enumerate}
    \item \emph{Intersection over Union:} As before, compute the IoU between each detection and ground truth. Match each detection to at most one ground truth if \(\mathrm{IoU}\ge\tau\).
    \item \emph{Precision-Recall Pairs:} For a wide range of model confidence thresholds, measure the proportion of TPs among all predicted positives (Precision) and the fraction of GT objects detected (Recall).
    \item \emph{101-Point Interpolation:} COCO samples 101 recall levels \(\{0,0.01,0.02,\ldots,1.0\}\). For each recall level \(r\), compute the maximum precision for recall \(\ge r\).
    \item \emph{Compute AP for Each IoU:} Unlike VOC, COCO averages over multiple IoU thresholds: \(0.50,0.55,\ldots,0.95\). 
    \item \emph{Mean AP over Categories:} Average the per-category AP to obtain \(\mathrm{mAP}\). Optionally, separate calculations by object size (small, medium, large).
\end{enumerate}

Common COCO metrics include \(\mathrm{AP}_{50}\) (AP at IoU\(\>=0.5\)) and \(\mathrm{AP}_{50:95}\) (average AP across \(\mathrm{IoU}\in\{0.50,\dots,0.95\}\)), summarizing performance across varying localization strictness.  
Table~\ref{tab:coco_eval_metrics} lists all common COCO evaluation metrics.

\subsubsection{Average Recall (AR)}
\label{metric:AR}

\emph{Average Recall (AR)} summarizes how well the model detects all objects over multiple IoU thresholds and possible per-image detection limits. It complements AP by focusing on the \emph{recall} dimension—i.e., the proportion of ground-truth objects found—averaged across different scenarios.

\paragraph{General Steps to Compute AR:}
\begin{enumerate}
    \item \textbf{Compute IoU:} For each detection, compute \(\mathrm{IoU}\) with each ground-truth box in the same image.
    \item \textbf{Match Detections and Ground Truths:} A ground truth is \emph{matched} if at least one detection has \(\mathrm{IoU}\ge\tau\). Each ground truth can only be matched once.
    \item \textbf{Measure Recall:} For a given IoU threshold, the recall is \(\frac{\mathrm{TP}}{\mathrm{GT}}\), where GT is the number of ground-truth boxes.
    \item \textbf{Average Over IoU Thresholds:} Vary \(\tau\) in a set (e.g., \(\{0.5,0.55,\dots,0.95\}\)) and average the resulting recall values.
    \item \textbf{Average Over Max Detections:} Models often limit the maximum number of detections (e.g., 100) per image. Repeat the recall calculation with different detection limits (e.g., 1, 10, 100) and average.
    \item \textbf{Aggregate Over the Dataset:} Finally, average across all images in the dataset to obtain a single AR value.
\end{enumerate}

\paragraph{AR in COCO:}
COCO commonly reports \(\mathrm{AR}\) at different detection cutoffs (AR@1, AR@10, AR@100) and object sizes (small, medium, large). This helps assess the trade-off between detection quantity and detection quality, revealing how the detector performs when restricted to fewer or more bounding-box predictions.

\bigskip
\noindent
In general, these object detection metrics: AP (including variations like mAP, AP\(_{50}\), AP\(_{50:95}\)) and AR, together with IoU-based definitions of TP, FP, and FN, provide a robust framework to evaluate the precision and robustness of detection models in tasks ranging from generic object localization to specialized domains (e.g. autonomous driving or medical imaging). 

\begin{table*}[ht]
\caption{This table summarizes key metrics from the COCO dataset for evaluating object detection models, including Average Precision (AP) and Average Recall (AR) across various IoU thresholds and by object size. AP at different IoU thresholds measures model accuracy with varying overlap levels, while AP and AR across scales evaluate performance on different object sizes.}
\label{tab:coco_eval_metrics}
\centering
\begin{tabular}{l p{8cm}}
\toprule
Average Precision (AP) &  \\ 
\midrule
AP & $\%$ AP at IoU=.50:.05:.95 (primary challenge metric)\\
$AP^{IoU=.50}$ & $\%$ AP at IoU=.50 (PASCAL VOC metric) \\
$AP^{IoU=.75}$ & $\%$ AP at IoU=.75 (strict metric) \\
\midrule
AP Across Scales: &\\
\midrule
$AP^{small}$ & $\%$ AP for small objects: area $< 32^2$ \\
$AP^{medium}$ & $\%$ AP for medium objects: $32^2 <$ area $< 96^2$ \\
$AP^{large}$ & $\%$ AP for large objects: area $> 96^2$ \\
\midrule
Average Recall (AR): & \\ 
\midrule
$AR^{max=1}$ & $\%$ AR given 1 detection per image \\ 
$AR^{max=10}$ & $\%$ AR given 10 detections per image \\ 
$AR^{IoU=100}$  & $\%$ AR given 100 detection per image \\ 
\midrule
AR Across Scales: &  \\ 
\midrule
$AR^{small}$ & $\%$ AR for small objects: area $< 32^2$ \\ 
$AR^{medium}$ & $\%$ AR for medium objects: $32^2 <$ area $< 96^2$ \\ 
$AR^{large}$ & $\%$ AR for large objects: area $> 96^2$ \\
\bottomrule
\end{tabular}
\end{table*}

\section{Image Segmentation}
\label{subsec:segmentation}

\emph{Image segmentation} is the process of assigning a label or category to each pixel in an image, effectively partitioning the scene into meaningful regions. Unlike image-level classification or object detection, segmentation operates at the pixel level, requiring a detailed understanding of both local features and global context. Deep segmentation networks typically learn to predict a pixel’s class label by considering its surrounding region or the entire image context.

Common segmentation approaches are usually grouped into:
\begin{itemize}
    \item \textbf{Semantic Segmentation:} Focuses on partitioning the image into “stuff” categories (e.g., road, sky, grass), where all instances of the same class share one label \cite{long2015fully,badrinarayanan2017segnet,chen2018-SS,chen2017-SS,ronneberger2015,oktay2018,he2014,zheng2015}.
    \item \textbf{Instance Segmentation:} Aims to detect and segment each occurrence of a “thing” class (e.g., people, cars) individually \cite{he2017mask,liu2018path,kirillov2019,kirillov2020,wang2019-IS}. Different instances of the same class receive distinct masks.
    \item \textbf{Panoptic Segmentation:} Unifies semantic and instance segmentation \cite{kirillov2019panoptic,xiong2019,liu2019,cheng2020,wang2021}, assigning each pixel either a “stuff” label or a unique “thing” identifier.
\end{itemize}

Segmentation has broad applications, including but not limited to:
\begin{itemize}
    \item \emph{Scene Understanding:} In robotics or autonomous driving, detailed environmental parsing enhances navigation and decision-making \cite{li2009towards,hua2016,zhou2017,song2015,yu2015}.
    \item \emph{Medical Imaging:} Accurate pixel-level delineation of anatomy or pathology is critical for diagnosis and treatment planning \cite{pham2000current,yan2018,liu2017-MIA,havaei2017,milletari2016,kamnitsas2017,chen2019-MIA}.
    \item \emph{Robotic Perception:} Pixel-wise classification facilitates grasping and manipulation by identifying object boundaries \cite{pan2020cross,qi2017,toshev2014,jegou2017,iandola2016}.
    \item \emph{Autonomous Vehicles:} Road-lane detection, pedestrian segmentation, and obstacle identification demand precise pixel-level predictions \cite{ha2017mfnet,mnih2015,chen2020-AV1,gal2016,alahi2016,chen2021-AV1}.
    \item \emph{Video Surveillance:} Segmentation aids in crowd analysis, anomaly detection, and activity recognition \cite{ammar2020deep,redmon2017,yeung2016,tran2015,hasan2016}.
    \item \emph{Augmented Reality:} Seamlessly overlays virtual elements by segmenting foreground objects or the user’s environment \cite{ko2020novel,cao2018,kim2018-AR,tekin2018,bessesen2018}.
\end{itemize}

\subsection{Segmentation Loss Functions}
\label{subsec:seg_loss_func}

Numerous loss functions have been devised to handle the unique challenges of segmentation, including dealing with heavy class imbalance and enforcing coherent shapes in the output masks. Table~\ref{tab:guidelines-seg-loss} summarizes the most common segmentation loss functions, along with their typical usage scenarios, data properties for which they work best, and practical advantages/limitations. Although many segmentation tasks can be approached using standard classification losses (e.g., pixel-wise cross-entropy), specialized losses often incorporate region-based considerations or adapt to heavily imbalanced pixel distributions.

The sections that follow present deeper explanations for each loss function and how they apply to different segmentation contexts. Depending on the nature of the application—medical imaging, large-scale street-scene parsing, or instance-level segmentation tasks—practitioners might favor certain losses for their robustness to imbalance or their direct optimization of region overlap metrics.

\begin{table*}[ht!]
\caption{Guidelines for selecting image segmentation loss functions based on usage, data characteristics, advantages, and limitations.}
\label{tab:guidelines-seg-loss}
\centering
\footnotesize
\begin{tabular}{p{1.3cm}p{2.8cm}p{2.8cm}p{2.8cm}p{2.8cm}}
\toprule
\textbf{Function} 
& \textbf{Usage} 
& \textbf{Data Characteristics} 
& \textbf{Advantages} 
& \textbf{Limitations} \\
\midrule

Cross-Entropy 
& Multi-class or binary segmentation \newline Pixel-wise probability comparison 
& Balanced or large-scale datasets \newline Works in general segmentation tasks 
& Straightforward, stable \newline Well-established 
& Sensitive to imbalance \newline May require weighting or complementary losses \\
\midrule

Dice Loss 
& Overlap-based segmentation \newline Common in medical imaging 
& Highly imbalanced data \newline Different class sizes 
& Directly optimizes overlap \newline Robust to imbalance 
& Instability if overlap is very low \newline May need combination with CE \\
\midrule

IoU (Jaccard) Loss 
& Maximizing Intersection over Union \newline Used in semantic segmentation 
& Data where precision and recall both matter \newline Multiple object classes 
& Correlates directly with IoU metric \newline Penalizes FPs/FNs 
& Implementation complexity \newline May need surrogate for better gradients \\
\midrule

Tversky Loss
& Extension of Dice \newline Adjusts penalties for FPs/FNs 
& Severe class imbalance \newline Medical/rare-class data 
& Hyperparameters ($\alpha,\beta$) for fine control \newline Focus on crucial errors 
& Requires careful tuning \newline Risk of bias toward certain errors \\
\midrule

Lovász Loss 
& Direct IoU optimization \newline Smooth approximation 
& Imbalanced data \newline IoU as primary metric 
& Aligns well with final IoU \newline Differentiable surrogate 
& More complex implementation \newline Fewer available examples \\
\midrule

Focal Loss 
& Class-imbalanced segmentation \newline Emphasizes hard examples 
& Highly skewed data \newline Medical, aerial imagery 
& Down-weights easy cases \newline $\alpha,\gamma$ fine-tuning 
& Hyperparameter sensitivity \newline Over-focus on hard samples \\

\bottomrule
\end{tabular}
\end{table*}

\subsubsection{Cross-Entropy Loss for Segmentation}
\label{loss:CE_seg}

\emph{Cross-Entropy (CE) loss} is a widely used measure in segmentation tasks to quantify the divergence between the model’s predicted class probabilities and the true (one-hot or binary) labels on a \emph{per-pixel} basis. In \emph{multi-class segmentation}, one assumes each pixel \(i\) belongs to exactly one of \(C\) classes. Let \(p_{i,c}\in [0,1]\) be the predicted probability that pixel \(i\) is class \(c\), and let \(y_{i,c}\in\{0,1\}\) be a one-hot indicator denoting the actual class label for pixel \(i\). Then the CE loss is

\begin{equation}
\label{eq:CE_seg}
L_{\mathrm{CE}}
= -\frac{1}{N}\sum_{i=1}^{N} 
  \sum_{c=1}^{C}
  y_{i,c}\,\log\bigl(p_{i,c}\bigr),
\end{equation}

\noindent
where \(N\) is the total number of pixels in the image. Minimizing \(L_{\mathrm{CE}}\) encourages \(p_{i,c}\) to approach 1 for the correct class \(c\), and to remain near 0 for all other classes.

\paragraph{Binary Segmentation Case:}
For \emph{binary} foreground/background segmentation, the cross-entropy loss simplifies to the \emph{Binary Cross-Entropy} (BCE) form:

\begin{equation}
\label{eq:CE_bin_seg}
L_{\mathrm{BCE}}
= -\frac{1}{N}\sum_{i=1}^{N}
   \Bigl[
    y_i\,\log\bigl(p_i\bigr)
    + (1 - y_i)\,\log\bigl(1 - p_i\bigr)
   \Bigr],
\end{equation}

\noindent
where \(y_i\in\{0,1\}\) indicates whether pixel \(i\) belongs to the object of interest (1) or the background (0), and \(p_i\) is the predicted probability of pixel \(i\) being foreground. As with multi-class CE, minimizing BCE aligns predicted probabilities with the true binary labels at each pixel.

\paragraph{Discussion and Advantages:}
\begin{itemize}
    \item \textbf{Pixel-Wise Classification}: 
    CE-based losses treat each pixel classification independently, making them straightforward and efficient for large-scale segmentation tasks.
    \item \textbf{Probabilistic Interpretation}: 
    By modeling segmentation as per-pixel Bernoulli or multinomial distributions, one can leverage well-established gradient-based optimization and network architectures.
    \item \textbf{Limitations with Imbalance}: 
    In cases where the foreground is much smaller than the background (e.g., medical imaging), standard CE may underweight crucial classes. Solutions include weighting schemes, Focal Loss (\S\ref{loss:focal}), or region-based losses (e.g., Dice Loss).
\end{itemize}

\subsubsection{Dice Loss}
\label{loss:dice}

\emph{Dice Loss}, derived from the \emph{Dice Similarity Coefficient (DSC)} or Sørensen–Dice index \cite{sorensen1948method}, offers an overlap-based metric for segmentation. In tasks where the foreground class is relatively rare—such as in lesion detection or other small-object scenarios—Dice-based losses can outperform pixel-wise classification losses by directly optimizing the spatial overlap between predicted and ground-truth regions.

\paragraph{Definition:}
Consider \(\hat{y}_i\) as the set of pixels in the predicted segmentation mask and \(y_i\) as the set of pixels in the ground-truth mask. The DSC measures the overlap:

\begin{equation}
\label{eq:dice_loss}
\text{DSC}(\hat{y}_i, y_i) 
= \frac{2\,|\hat{y}_i \cap y_i|}{|\hat{y}_i| + |y_i|}.
\end{equation}

\noindent
The \emph{Dice Loss} is then formulated as

\begin{equation}
L_{\mathrm{Dice}} 
= 1 - \text{DSC}(\hat{y}_i, y_i),
\end{equation}
which yields values from 0 (no overlap) to 1 (perfect overlap) in the coefficient domain, and from 1 to 0 in the loss domain. 

\paragraph{Pixel-Level Implementation:}
In practice, \(\hat{y}_i\) and \(y_i\) can be represented as continuous maps (e.g., predicted probabilities and binary ground-truth labels). Summations or integrals over these maps approximate the set cardinalities and intersections. This approach is particularly effective for imbalanced datasets because it amplifies small-object region overlap and de-emphasizes large background areas.

\paragraph{Advantages and Considerations:}
\begin{itemize}
    \item \textbf{Imbalance Robustness}: 
    Dice Loss intrinsically focuses on matching the pixels in the smaller target region, improving sensitivity to rare classes or fine structures.
    \item \textbf{Direct Overlap Optimization}: 
    Unlike pixel-wise classification losses, Dice Loss directly penalizes inadequate region overlap, often leading to better boundary adherence and shape consistency.
    \item \textbf{Hyperparameter-Free}: 
    The classic Dice formulation does not introduce extra hyperparameters, though variants like Tversky Loss (\S\ref{loss:tversky}) do provide weighting mechanisms for FN vs.\ FP.
\end{itemize}

\noindent
In general, dice loss is a natural choice in segmentation tasks, particularly in medical imaging, where overlapping predicted and ground-truth masks are more relevant than merely classifying each pixel independently.

\subsubsection{Intersection Over Union (IoU) Loss for Segmentation}
\label{loss:iou_seg}

\emph{Intersection over Union} (\(\mathrm{IoU}\)), also called the \emph{Jaccard Index}, measures the degree of overlap between the predicted segmentation mask \(\hat{y}_i\) and the ground-truth mask \(y_i\) at the pixel level. Formally, the IoU for each pixel \(i\) reflects whether that pixel belongs to both masks (intersection) or to at least one of the two masks (union). Let \(\lvert y_i \cap \hat{y}_i\rvert\) and \(\lvert y_i \cup \hat{y}_i\rvert\) respectively denote the pixel-level intersection and union; the IoU-based loss is often defined as:

\begin{equation}
\label{eq:iou_seg}
L_{\mathrm{IoU}}
= 1 
  - \frac{1}{N}\sum_{i=1}^N 
    \frac{\lvert y_i \cap \hat{y}_i\rvert}{
          \lvert y_i \cup \hat{y}_i\rvert},
\end{equation}

\noindent
where \(N\) is the number of pixels (or potentially superpixels/regions) in an image, \(y_i\) indicates the ground-truth label of pixel \(i\), and \(\hat{y}_i\) is the predicted label. In practice, these sets are often implemented as continuous probability maps (for \(\hat{y}_i\)) and binary indicators (for \(y_i\)).

\paragraph{Interpretation:}
\begin{itemize}
    \item \textbf{IoU Score}: Ranges from 0 (no overlap) to 1 (perfect overlap).  
    \item \textbf{Loss}: \(L_{\mathrm{IoU}}=0\) when the predicted mask exactly matches the ground truth, and approaches 1 when there is minimal or no overlap.
\end{itemize}

\paragraph{Advantages:}
\begin{itemize}
    \item \textbf{Balance of Precision and Recall}: IoU intrinsically captures how many true positives (correctly overlapped pixels) exist relative to the sum of all positives (both predicted and actual).
    \item \textbf{Direct Alignment with Evaluation}: IoU is a popular performance metric in many segmentation benchmarks \cite{rahman2016optimizing,van2019deep,long2015fully,luc2016semantic}; using it as a loss can align training with the final evaluation measure.
\end{itemize}

\paragraph{Challenges:}
\begin{itemize}
    \item \textbf{Gradient Near Zero Overlap}: 
    Similar to other region-based losses, if the model’s prediction initially fails to overlap with the ground truth, the IoU gradient may provide weaker feedback than pixel-wise losses (cf.\ Section~\ref{loss:CE_seg}).
    \item \textbf{Multiple Classes or Large Images}: 
    Extending IoU to multi-class setups requires summing or averaging per-class IoU, which can be computationally more involved if done naively for large images.  
\end{itemize}

In many modern segmentation frameworks, IoU loss can be combined with other pixel-wise losses or with Dice-based metrics (\S\ref{loss:dice}) to balance local pixel accuracy and global region-overlap optimization.

\subsubsection{Tversky Loss}
\label{loss:tversky}

\emph{Tversky Loss}, introduced by Salehi~\emph{et~al.} \cite{salehi2017tversky}, generalizes Dice-based losses to better address class imbalances and varying error penalties in segmentation tasks. It introduces two hyperparameters, \(\alpha\) and \(\beta\), controlling how false positives and false negatives are weighted. Consider the sets \(\hat{y}_i\) (predicted mask) and \(y_i\) (ground-truth mask):

\begin{equation}
\label{eq:tv_loss}
\mathrm{Tversky}(\hat{y}_i, y_i) 
= 1 
  - \frac{\lvert \hat{y}_i \cap y_i\rvert}{
          \lvert \hat{y}_i \cap y_i\rvert 
        + \alpha\,\lvert \hat{y}_i \backslash y_i\rvert 
        + \beta\,\lvert y_i \backslash \hat{y}_i\rvert}.
\end{equation}

\noindent
Here, \(\lvert \hat{y}_i \cap y_i\rvert\) is the intersection (true positives), \(\lvert \hat{y}_i \backslash y_i\rvert\) is the set difference (false positives), and \(\lvert y_i \backslash \hat{y}_i\rvert\) denotes missed pixels (false negatives). By adjusting \(\alpha,\beta\in[0,1]\), practitioners emphasize or de-emphasize certain types of errors.

\paragraph{Key Motivations:}
\begin{itemize}
    \item \textbf{Class Imbalance}: 
    If one class (e.g., lesions in medical images) is significantly smaller, Tversky Loss can be tuned (\(\alpha < \beta\) or vice versa) to penalize false negatives more heavily.
    \item \textbf{Enhanced Customization}: 
    Different domains may require prioritizing false positives vs. false negatives. Tversky Loss provides a direct mechanism for these domain-specific trade-offs.
\end{itemize}

\paragraph{Typical Settings:}
\begin{itemize}
    \item \(\alpha=\beta=0.5\): Recovers Dice Loss (Section~\ref{loss:dice}).  
    \item \(\alpha + \beta=1\): Preserves a balancing pattern; e.g., \(\alpha=0.7,\beta=0.3\) puts more emphasis on penalizing missed positives. 
\end{itemize}

\paragraph{Use Cases:}
\begin{itemize}
    \item \textbf{Medical Imaging}: 
    Pathologies can be tiny and critical to detect. Weighting false negatives more heavily can reduce missed diagnoses.
    \item \textbf{Multi-Class Extensions}: 
    Tversky Loss can be applied per class and then summed or averaged. This approach helps if multiple classes are imbalanced in different ways.
\end{itemize}

\noindent
In general, Tversky Loss provides a customizable tool for balancing segmentation errors, especially in high-stakes or highly imbalanced tasks. By choosing appropriate \(\alpha\) and \(\beta\), one can tailor the loss configuration to a desired error profile.

\subsubsection{Lovász Loss}
\label{loss:lovasz}

\emph{Lovász Loss}, proposed by Berman~\emph{et~al.} \cite{berman2018lovasz}, is designed to directly optimize Intersection over Union (IoU), offering a smooth, differentiable surrogate to the Jaccard index. Rather than aggregating pixel-wise errors, the loss treats the IoU as a submodular set function and applies a convex Lovász extension that approximates the discrete IoU in a manner suitable for gradient-based optimization.

\paragraph{Definition:}
Given predicted segmentation probabilities \(\{p_i\}_{i=1}^N\) and binary ground truth labels \(\{y_i\}_{i=1}^N\) (0 or 1), Lovász Loss reorders pixels by \emph{confidence errors} and sums a piecewise linear function derived from the Lovász extension. Conceptually:

\begin{equation}
L_{\mathrm{Lovasz}} 
= \sum_{i \,\in\, \mathrm{errors}} 
  \phi\bigl(m_i\bigr),
\end{equation}
where \(m_i\) is the margin-based error term for pixel \(i\) (e.g., \(m_i = 1 - p_i\) if \(y_i = 1\) or \(m_i = p_i\) if \(y_i=0\)), and \(\phi(\cdot)\) is the Lovász extension that approximates the IoU gap.

\paragraph{Advantages:}
\begin{itemize}
    \item \textbf{Direct IoU Optimization:}
    Since IoU is a set-based metric, standard pixel-wise losses (e.g., cross-entropy) may not fully align with maximizing IoU. Lovász Loss bridges this gap by operating on IoU-related errors directly.
    \item \textbf{Imbalance Handling:}
    By focusing on misclassified pixels (i.e., those with high margin errors), Lovász Loss naturally emphasizes classes or regions that are under-segmented.
    \item \textbf{Smooth Differentiability:}
    The Lovász extension yields a continuous, piecewise linear approximation to the discrete Jaccard index, enabling backpropagation without discontinuities.
\end{itemize}

\paragraph{Usage:}
Lovász Loss is valuable in tasks where IoU is the primary performance metric (e.g., large-scale semantic segmentation benchmarks). It often outperforms purely pixel-level losses in cases of class imbalance or when small object segmentation is crucial, thanks to its alignment with overlap-based evaluation criteria.

\subsubsection{Focal Loss for Segmentation}
\label{loss:focal_seg}

\emph{Focal Loss}, introduced by Lin~\emph{et~al.} \cite{lin2017focal} to handle class imbalance in object detection, has also been adopted for semantic segmentation—particularly when one or more foreground classes are rare or difficult to detect.

\paragraph{Definition:}
For a binary segmentation scenario, let \(p\in[0,1]\) denote the predicted probability that a pixel belongs to the foreground class. Denote the ground-truth label by \(y\in\{0,1\}\). Then the Focal Loss is

\begin{equation}
\label{eq:focal_seg}
FL(p_t)
= - \alpha_t
      \,(1 - p_t)^{\gamma}
      \,\log\bigl(p_t\bigr),
\end{equation}

\noindent
where
\begin{equation}
p_t 
= \begin{cases}
     p,   & \text{if } y=1,\\
     1-p, & \text{if } y=0,
  \end{cases}
\end{equation}
\(\alpha_t\in[0,1]\) is a class weighting factor (often used to weight foreground vs.\ background), and \(\gamma\ge0\) is the focusing parameter. In multi-class segmentation, Focal Loss typically applies to each class channel separately and sums or averages them.

\paragraph{Mechanism:}
\begin{itemize}
    \item \textbf{Down-Weight Easy Examples:}
    When a pixel is correctly classified (\(p_t\approx1\)), \((1 - p_t)^\gamma \approx 0\). This diminishes the contribution of “easy” pixels to the loss, redirecting the model’s attention to harder or misclassified pixels.
    \item \textbf{Adjustable Class Importance:}
    The factor \(\alpha_t\) can emphasize minority classes, ensuring that performance on these classes is not overshadowed by more frequent classes or background pixels.
\end{itemize}

\paragraph{Applications:}
\begin{itemize}
    \item \textbf{Highly Imbalanced Segmentation:}
    Focal Loss shines in contexts where large portions of the image belong to background or dominant classes, while small, important regions (e.g., tumors in medical imaging) risk neglect.
    \item \textbf{Focus on Ambiguous Boundaries:}
    By heavily penalizing misclassifications, Focal Loss encourages the network to refine boundaries and subtle regions that standard cross-entropy might overlook.
\end{itemize}

\paragraph{Hyperparameter Tuning:}
\begin{itemize}
    \item \(\alpha\): Balances importance between positive (foreground) and negative (background).  
    \item \(\gamma\): Governs how much the loss “focuses” on hard examples. Typical values range from \(1\) to \(5\), though dataset-specific tuning is often required.
\end{itemize}

\noindent
In general, Focal Loss for segmentation provides a strong alternative to plain cross-entropy in scenarios demanding fine-grained attention to rare or challenging classes, leading to improved coverage of small or ambiguous structures.

\subsection{Segmentation Metrics}
\label{subsec:seg_metrics}

Various metrics exist to assess the performance of segmentation models, taking into account both per-pixel classification accuracy and the consistency of predicted object boundaries or instances. Table~\ref{tab:guidelines-seg-metrics} presents a concise overview of commonly used segmentation metrics, highlighting their typical usage scenarios, data characteristics, primary advantages, and key limitations.

\begin{table*}[ht!]
\caption{Guidelines for selecting a metric for segmentation based on usage, data characteristics, advantages, and limitations.}
\label{tab:guidelines-seg-metrics}
\centering
\footnotesize
\begin{tabular}{p{1.3cm}p{2.6cm}p{2.6cm}p{2.6cm}p{2.6cm}}
\toprule
Metric 
& Usage 
& Data Characteristics 
& Advantages 
& Limitations \\
\midrule

Pixel accuracy 
& Semantic segmentation tasks where each pixel belongs to a class \newline Works if class frequencies are not too skewed 
& Images labeled at pixel level \newline Suitable when classes are relatively balanced 
& Very simple \newline Fast to compute \newline Intuitive interpretation 
& Dominated by majority classes \newline Ignores subtle region errors \newline No sense of boundary quality \\
\midrule

Boundary F1 
& Scenarios emphasizing precise object contours \newline Evaluates boundary alignment 
& Images where boundary delineation is crucial \newline Uses a distance threshold for matches 
& Focuses on boundary correctness \newline Captures alignment of object edges 
& Interior segmentation ignored \newline Choice of distance threshold affects scores \newline Can be less robust to noise near edges \\
\midrule

Mask AP 
& Instance segmentation \newline Requires accurate pixel-level masks for each object 
& Images with multiple instances \newline Each instance has a ground-truth mask \newline Typically evaluated at various IoU thresholds 
& Captures both localization and mask quality \newline Standard benchmark in instance seg (e.g., COCO) 
& Sensitive to IoU thresholds \newline Complex matching logic for multiple predictions \newline Calibration of confidence scores is critical \\
\midrule

Panoptic Quality 
& Panoptic segmentation \newline Unifies stuff + things 
& Images with pixel-level labels for both \newline Combines semantic + instance segments 
& Single metric for instance detection and area coverage \newline Balances region accuracy + object recognition 
& Requires matching instances + evaluating stuff classes \newline Overlaps can complicate matching \newline Not as common outside panoptic tasks \\
\bottomrule
\end{tabular}
\end{table*}

\subsubsection{Pixel Accuracy}
\label{metric:pixel_acc}

\emph{Pixel Accuracy} is a simple ratio describing how many pixels are classified correctly. For an image of size \(N\) pixels and \(C\) classes, let \(\hat{y}_i\in\{1,\ldots,C\}\) be the model’s predicted label for pixel \(i\), and \(y_i\) the ground-truth label. Then,

\begin{equation}
\label{eq:pix_acc}
\text{Pixel Accuracy}
= \frac{1}{N}
  \sum_{i=1}^{N}
  \mathbb{I}\bigl[\hat{y}_i = y_i\bigr],
\end{equation}
where \(\mathbb{I}[\cdot]\) is an indicator function equal to 1 if the expression is true, and 0 otherwise.

\paragraph{Interpretation and Limitations:}
\begin{itemize}
    \item \textbf{Balanced vs.\ Imbalanced Classes:} 
    Pixel accuracy may be dominated by large or frequently occurring classes. In tasks with severe foreground-background imbalance, high pixel accuracy might conceal poor performance on smaller classes.
    \item \textbf{Simplicity:} 
    Despite its limitations, pixel accuracy remains a fast and intuitive measure for many semantic segmentation scenarios, especially when classes appear in roughly equal proportions.
    \item \textbf{Complementary Metrics:} 
    Metrics like IoU (\S\ref{loss:iou_seg}) or Dice are often reported alongside pixel accuracy for a more robust assessment of segmentation quality, especially in unbalanced contexts.
\end{itemize}

\subsubsection{Boundary F1 Score (BF)}
\label{metric:bf}

The \emph{Boundary F1 Score (BF)} \cite{csurka2013good}, sometimes denoted \(\mathrm{BF}\), evaluates segmentation quality by focusing on \textit{boundary delineation}. Instead of measuring per-pixel overlaps, it checks how accurately predicted boundaries align with ground-truth boundaries.

The steps to compute the BF score are the following:
\begin{enumerate}
    \item \textbf{Match Predicted and Ground-Truth Boundaries:}
    For each ground-truth boundary, find the closest predicted boundary within a distance threshold \(\delta\). Similarly, each predicted boundary is matched to any ground-truth boundary within the same threshold.
    \item \textbf{Compute Precision (P):}
    \begin{equation}
    P 
      = \frac{TP}{TP + FP},
    \end{equation}
    where \(TP\) (true positives) counts predicted boundary points that are matched to a ground-truth boundary, and \(FP\) (false positives) counts predicted boundary points that have no ground-truth match.
    \item \textbf{Compute Recall (R):}
    \begin{equation}
    R 
      = \frac{TP}{TP + FN},
    \end{equation}
    where \(FN\) (false negatives) is the number of ground-truth boundary points not matched by any prediction.
    \item \textbf{Boundary F1 Score:}
    \begin{equation}
    \label{eq:bf}
    \mathrm{BF} 
    = \frac{2 \cdot P \cdot R}{P + R}.
    \end{equation}
\end{enumerate}

\paragraph{Interpretation:}
\begin{itemize}
    \item \textbf{Distance Threshold \(\delta\):} 
    Defines how close a predicted boundary must be to the ground-truth boundary to count as a match. Smaller \(\delta\) demands higher precision of boundary alignment; larger \(\delta\) is more lenient.
    \item \textbf{Focused on Boundary Quality:} 
    Suited for tasks where accurate object outlines are paramount, e.g., medical image analysis of organs or lesions with critical boundary definitions.
\end{itemize}

Although BF Score emphasizes boundary precision, it may not reflect how well large interior regions are segmented. Thus, it is commonly used alongside region-based metrics (IoU, Dice) for a complete segmentation evaluation.

\subsubsection{Masked Average Precision (Mask AP)}
\label{metric:mask_ap}

\emph{Masked Average Precision (Mask AP)} is the standard evaluation metric for \emph{instance segmentation}, extending the concept of bounding-box \textbf{AP} to pixel-level segmentation masks. Instead of checking whether a predicted bounding box overlaps sufficiently with a ground-truth box, \emph{Mask AP} requires the predicted segmentation mask to achieve a certain \emph{Intersection over Union} (IoU) threshold with the corresponding ground-truth mask.

\paragraph{Definition:}
Let \(\mathcal{G}\) be the set of ground-truth masks, and \(\mathcal{P}\) be the set of predicted masks for a given dataset of images. Each predicted mask \(\mathbf{p}_i\in\mathcal{P}\) is accompanied by a confidence score \(s_i\). For each ground-truth mask \(\mathbf{g}_j\in\mathcal{G}\) associated with a category \(c_j\), define the mask IoU as:

\begin{equation}
\mathrm{IoU}(\mathbf{p}_i,\mathbf{g}_j)
= \frac{\mathbf{p}_i \cap \mathbf{g}_j}{\mathbf{p}_i \cup \mathbf{g}_j}.
\end{equation}

A predicted mask \(\mathbf{p}_i\) is considered a \emph{true positive} if (1) its predicted category matches the category of a ground-truth mask, (2) it has the highest IoU with that ground-truth mask compared to other predictions of the same category, and (3) the IoU surpasses a given threshold \(\tau\) (e.g., 0.50, 0.75). Otherwise, it is counted as a \emph{false positive}.

After sorting predictions by decreasing confidence score \(s_i\), define:
\begin{equation}
\mathrm{Precision}(k)
= \frac{\text{\# of true positives among top-}k\ \text{predictions}}
       {k},
\end{equation}
\begin{equation}
\mathrm{Recall}(k)
= \frac{\text{\# of true positives among top-}k\ \text{predictions}}
       {\text{total \# of ground-truth masks in that category}}.
\end{equation}
By varying the decision threshold on \(s_i\) (from high to low), a precision–recall curve is traced.

\paragraph{Average Precision (AP) at a Fixed IoU Threshold:}
For each category (or across categories), the \(\mathrm{AP}\) at IoU threshold \(\tau\) is the area under the precision–recall curve. Formally, if one samples recall at discrete points \(\{r_j\}\), the average precision can be approximated as:

\begin{equation}
\mathrm{AP}_\tau
= \int_{0}^{1}
   \mathrm{Precision}(r)
   \,d(r),
\end{equation}
where \(\mathrm{Precision}(r)\) is interpolated at each recall level \(r\). In practice, the MS-COCO dataset uses 101 recall points (from 0 to 1 in increments of 0.01) and averages the maximum precision observed at or above each recall.

\paragraph{Mean Mask AP Over Multiple IoU Thresholds:}
To account for varying segmentation qualities, one typically computes AP at multiple IoU thresholds \(\{\tau_k\}\), such as \(\tau\in \{0.50, 0.55,\ldots,0.95\}\). The final \emph{mean Mask AP} is the average of \(\mathrm{AP}_\tau\) across these thresholds, thereby providing a more comprehensive evaluation:

\begin{equation}
\mathrm{mAP}_\text{mask}
= \frac{1}{|\{\tau_k\}|}
  \sum_{\tau_k}
    \mathrm{AP}_{\tau_k}.
\end{equation}

\paragraph{Usage:}
\begin{itemize}
    \item \textbf{Instance Segmentation Benchmarks:}
      Datasets like MS-COCO, Cityscapes, or LVIS employ \emph{mask IoU} and average precision across multiple IoU thresholds. This helps gauge the model’s ability to accurately localize instances and produce precise boundaries.

    \item \textbf{Category-level or Overall AP:}
      One can compute Mask AP separately for each object category or average across categories to yield an overall performance measure. Typically reported are \(\mathrm{AP}_{50}\) (IoU=0.50), \(\mathrm{AP}_{75}\) (IoU=0.75), and the \(\mathrm{AP}\) over \(\tau\in[0.50:0.05:0.95]\).

    \item \textbf{Object Size Variations:}
      Additional breakdowns—\(\mathrm{AP}_\text{S}, \mathrm{AP}_\text{M}, \mathrm{AP}_\text{L}\)—reflect performance on small, medium, and large objects, respectively, illustrating how the model scales across object sizes.
\end{itemize}

\paragraph{Limitations:}
\begin{itemize}
    \item \textbf{Strict IoU Thresholds:}
      AP relies on discrete IoU thresholds. Small changes in mask boundaries can affect the IoU, potentially flipping a prediction between true and false positive if near the threshold.
    \item \textbf{Confidence Calibration:}
      Precise rank ordering of predicted masks by confidence is crucial. Poor calibration may degrade the precision–recall curve and thus the final AP.
    \item \textbf{Multiple Instances and Overlapping:}
      In crowded scenes, partial overlap or occlusion among objects can complicate matching predictions to ground truth, affecting the measured IoU and scoring.
\end{itemize}

\emph{Masked Average Precision (Mask AP)} generalizes object detection’s bounding-box AP to pixel-level segmentation tasks. By computing how well predicted masks match ground-truth masks across various IoU thresholds and confidence levels, \textbf{Mask AP} serves as the principal benchmark for instance segmentation performance. This measure ensures both correct object localization \emph{and} accurate mask delineation.

\subsubsection{Panoptic Quality (PQ)}
\label{metric:pq}

\emph{Panoptic Quality (PQ)} \cite{kirillov2019panoptic} is a unified metric for \emph{panoptic segmentation}, which merges semantic segmentation (“stuff” classes) and instance segmentation (“thing” classes). PQ assesses both the accuracy of instance segmentation and the correct recognition of classes and object identities.

\paragraph{Definition:}
Let \(\mathrm{TP}\) be the set of matched (predicted, ground-truth) segments with an Intersection-over-Union \(\mathrm{IoU}(p,g)\) above a threshold (commonly 0.5). Let \(\mathrm{FP}\) be the set of unmatched predicted segments, and \(\mathrm{FN}\) be the set of ground-truth segments not matched by any prediction. Then,

\begin{align}
\label{eq:pq}
PQ 
&= \underbrace{\frac{\sum_{(p,g)\in \mathrm{TP}} \mathrm{IoU}(p,g)}{\lvert \mathrm{TP}\rvert}}_{\text{Segmentation Quality (SQ)}} 
   \;\times\; 
   \underbrace{\frac{\lvert \mathrm{TP}\rvert}{\lvert \mathrm{TP}\rvert + \tfrac12\,\lvert \mathrm{FP}\rvert + \tfrac12\,\lvert \mathrm{FN}\rvert}}_{\text{Recognition Quality (RQ)}}.
\end{align}

\paragraph{Components:}
\begin{itemize}
    \item \textbf{Segmentation Quality (SQ):} 
    Averages IoU over all correctly matched segments (\(\mathrm{TP}\)), measuring how well each object’s shape overlaps with its ground truth.
    \item \textbf{Recognition Quality (RQ):} 
    Reflects detection performance. By counting \(\mathrm{TP}\) over \(\mathrm{TP} + \frac12\,(\mathrm{FP}+\mathrm{FN})\), RQ penalizes both spurious predictions and missed instances.
\end{itemize}

\paragraph{Interpretation:}
\begin{itemize}
    \item \textbf{Range and Meaning:} 
    \(0 \le PQ \le 1\). Higher PQ implies better per-instance segmentation (\(\mathrm{SQ}\) high) and accurate instance recognition (\(\mathrm{RQ}\) high). 
    \item \textbf{Comprehensive vs.\ Other Metrics:} 
    PQ is more holistic than mIoU or average precision alone, as it requires both correct pixel-level segmentation and correct instance labeling. 
    \item \textbf{Application to “Stuff” vs.\ “Things”:} 
    Panoptic segmentation divides classes into “stuff” (e.g., roads, sky) and “things” (e.g., cars, people). Many benchmarks provide PQ metrics separately or in aggregate for each category group.
\end{itemize}

PQ is considered a robust choice in real-world scenarios with multiple overlapping objects and background classes. It balances region accuracy and instance identification, making it invaluable for tasks such as autonomous driving and complex scene understanding.

\section{Face Recognition}
\label{subsec:face_recog}

\emph{Face recognition} is the task of matching an individual's face in an image or video to a corresponding identity in a database of faces. Deep learning models, typically Convolutional Neural Networks (CNNs) or Transformers \cite{vaswani2017attention}, are trained on large datasets of face images to extract discriminative features. These features enable accurate matching between input faces and stored identities. Face recognition finds applications in areas such as \emph{security} \cite{owayjan2015face,sayem2018integrating}, \emph{social media} \cite{indrawan2013face}, and \emph{biometric identification systems} \cite{chang2003multimodal,senior2002face}.

\subsection{Face Recognition Loss Functions and Metrics}
\label{subsec:face_recog_loss}

Loss functions for face recognition often focus on preserving fine-grained relational structure among face embeddings (i.e., feature vectors). Broadly, they can be categorized as:
\begin{itemize}
    \item \textbf{Classification-based losses:} 
    \(\mathrm{Softmax}\), \(\mathrm{A\text{-}Softmax}\), \emph{Center Loss}, \emph{Large-Margin Cosine Loss}, and \emph{Additive Angular Margin Loss}.
    \item \textbf{Representation-based losses:} 
    \emph{Triplet Loss}, \emph{Contrastive Loss}, \emph{Circle Loss}, and \emph{Barlow Twins Loss}.
\end{itemize}

Common evaluation metrics for face recognition mirror general classification metrics—accuracy, precision, recall, F1-score, ROC curves—but often specialized protocols measure \emph{verification} (is this the same person or not?) and \emph{identification} (which person is this?). Table~\ref{tab:guidelines-face-losses} summarizes frequently used face-recognition loss functions, including their typical data properties, advantages, and limitations. Below, we discuss the details of each.

\begin{table*}[ht!]
\caption{Guidelines for selecting a suitable loss function for face recognition based on usage, data characteristics, advantages, and limitations.}
\label{tab:guidelines-face-losses}
\centering
\footnotesize
\begin{tabular}{p{1.3cm}p{2.8cm}p{2.8cm}p{2.8cm}p{2.8cm}}
\toprule
\textbf{Function} 
& \textbf{Usage} 
& \textbf{Data Characteristics} 
& \textbf{Advantages} 
& \textbf{Limitations} \\
\midrule

Softmax
& Multi-class classification \newline Common for image recognition 
& Large labeled datasets \newline Clear class boundaries 
& Simple and widely used \newline Stable training 
& Lacks explicit margin \newline Limited discriminative power \\

A-Softmax 
& Face recognition \newline Tasks needing angular margin 
& Intraclass similarity \newline Need for angular separability 
& Improves class separability \newline Enforces angular constraints 
& Sensitive to margin hyperparams \newline Can be unstable if not tuned \\

Center Loss
& Combined with softmax \newline Encourages intra-class clustering 
& Data with well-defined classes \newline Useful for identification tasks 
& Tightens within-class features \newline Improves clustering 
& Requires balancing two losses \newline Needs careful center updates \\

CosFace 
& Large-margin face recognition \newline Cosine-based separation 
& High intraclass variability \newline Need robust boundaries 
& Direct cosine margin \newline Usually stable training 
& Requires margin tuning \newline May not generalize to all tasks \\

ArcFace 
& Face/ID recognition \newline Angular margin in feature space 
& Subtle interclass differences \newline High-precision scenarios 
& Additive margin improves separation \newline Strong geometric intuition 
& Sensitive to margin and scale \newline Higher compute cost \\

Triplet Loss 
& Verification and retrieval \newline Enforces relative distances 
& Needs anchor-positive-negative \newline Good for similarity tasks 
& Flexible for unseen classes \newline Learns a metric space 
& Triplet mining is complex \newline Slower convergence \\

Contrastive loss 
& Siamese networks \newline Pairwise similarity/dissimilarity 
& Requires similar/dissimilar labels \newline Potential imbalanced pairs 
& Straightforward metric learning \newline Extensible to new classes 
& Single margin for all negatives \newline Sensitive to pair sampling \\

Circle Loss 
& Tasks with imbalanced data \newline Simultaneous pos/neg optimization 
& Need pairwise similarity scores \newline Complex data distributions 
& Optimizes both pos and neg pairs \newline Clear similarity-space target 
& Tuning margins can be tricky \newline Sensitive to hyperparams \\

Barlow Twins 
& Self-supervised \newline Learns decorrelated representations 
& Large unlabeled data \newline High-dimensional embeddings 
& Removes redundancy \newline No explicit negatives needed 
& Requires large batch size \newline Sensitive to augmentations \\

SimSiam 
& Self-supervised \newline No negative pairs 
& Large unlabeled data \newline Relies on augmentations 
& Simple approach \newline Stop-gradient avoids collapse 
& Sensitive to hyperparams \newline May learn low-level features \\

\bottomrule
\end{tabular}
\end{table*}

\subsubsection{Softmax Loss}
\label{loss:face_softmax}

\emph{Softmax Loss} (cross-entropy) is a canonical classification objective. In face recognition, the network typically outputs a \emph{logit} vector \(\mathbf{f}(x)\) for each identity class, which is mapped to probabilities via the softmax function. For a sample \((x, y)\) (face image \(x\) of true identity \(y\)), let \(\mathbf{W} \in \mathbb{R}^{d\times n}\) be the learnable weights (one column per class) and \(\mathbf{b}\in\mathbb{R}^n\) the bias. Then each logit is
\begin{equation}
f_j(x)
= \mathbf{w}_j^\top \mathbf{x} + b_j,
\end{equation}
where \(\mathbf{x}\in \mathbb{R}^d\) is the deep feature vector extracted by the network. The probability of class \(j\) is

\begin{equation}
\label{eq:prob_classi}
P(y=j \mid x;\mathbf{W})
= \frac{\exp\!\bigl(f_j(x)\bigr)}{\sum_{k=1}^{n} \exp\!\bigl(f_k(x)\bigr)}.
\end{equation}

The cross-entropy loss for a batch of \(N\) samples is

\begin{equation}
\label{eq:softmax_logl}
\mathcal{L}_{\mathrm{softmax}}
= -\frac{1}{N}\sum_{i=1}^{N}
   \log \bigl( P\bigl(y=y_i \mid x_i;\mathbf{W}\bigr)\bigr).
\end{equation}

\paragraph{Limitations for Face Recognition:}
Although softmax can learn to classify multiple identities, it does not explicitly enforce large margins between different classes or compress intra-class variability. This may hinder the network’s ability to discriminate among visually similar faces. Consequently, more sophisticated margin-based losses or representation-learning objectives are often employed to enhance separation in the embedding space.

\subsubsection{A-Softmax (SphereFace) Loss}
\label{loss:face_asoftmax}

\emph{A-Softmax} \cite{liu2017sphereface}, also referred to as \emph{SphereFace Loss}, improves upon standard softmax by introducing an \emph{angular margin}, thus enhancing the embedding discriminativeness. The key idea is to \emph{normalize} both the weights and feature vectors so that classification depends on the angle between \(\mathbf{x}\) and each class vector \(\mathbf{w}_j\).

\paragraph{Derivation:}
\begin{enumerate}
    \item \textbf{Weight Normalization:}
    Let \(\|\mathbf{w}_j\|=1\) for each class \(j\). In practice, each column of \(\mathbf{W}\) is normalized, ensuring \(\|\mathbf{w}_j\|=1\). 

    \item \textbf{Feature Normalization:}
    Consider the feature vector \(\mathbf{x}\) with norm \(\|\mathbf{x}\|\). The logit for class \(j\) effectively becomes \(\|\mathbf{x}\|\;\|\mathbf{w}_j\|\cos(\theta_j)=\|\mathbf{x}\|\cos(\theta_j)\), where \(\theta_j\) is the angle between \(\mathbf{x}\) and \(\mathbf{w}_j\).

    \item \textbf{Angular Margin $m$:}
    A-Softmax modifies the logit for the true class \(y\) by replacing \(\cos(\theta_y)\) with \(\cos(m\theta_y)\). This imposes a stricter requirement for correct classification, forcing \(\theta_y\) to be smaller (or $\cos(\theta_y)$ bigger) by a factor of \(m\). Formally,
    \begin{equation}
    \label{eq:a-softmax-angular}
    \cos(m\,\theta_y) 
    = \cos\!\bigl(m \arccos(\cos(\theta_y))\bigr).
    \end{equation}

    \item \textbf{Scaled Logits:}
    The final logits become
    \begin{equation}
    \hat{y}_y 
    = \|\mathbf{x}\|\cos(m\,\theta_y),
    \quad 
    \hat{y}_j
    = \|\mathbf{x}\|\cos(\theta_j),\;\; j \neq y.
    \end{equation}
    The cross-entropy loss uses these adjusted logits:
    \begin{equation}
    \mathcal{L}_{\mathrm{A\text{-}softmax}}
    = -\frac{1}{N}\sum_{i=1}^N 
      \log
      \frac{\exp\!\Bigl(\|\mathbf{x}_i\|\cos\bigl(m\,\theta_{y_i}\bigr)\Bigr)}{
            \exp\!\Bigl(\|\mathbf{x}_i\|\cos\bigl(m\,\theta_{y_i}\bigr)\Bigr)
          + \sum_{j\neq y_i}
             \exp\!\Bigl(\|\mathbf{x}_i\|\cos(\theta_j)\Bigr)}.
    \end{equation}

\end{enumerate}

\paragraph{Benefits for Face Recognition:}
\begin{itemize}
    \item \textbf{Margin-Based Separation:}
    By enforcing \(\cos(m\theta)\) for the true class, the network encourages tighter intra-class clustering and larger inter-class separation in the angular space.
    \item \textbf{Better Discriminability:}
    Compared to plain softmax, A-Softmax reduces misclassification among visually similar faces, thus improving verification and identification metrics.
\end{itemize}

\paragraph{Practical Considerations:}
\begin{itemize}
    \item \textbf{Margin $m$ Tuning:}
    Larger $m$ leads to stronger separation but can be harder to optimize if $m$ is too large.
    \item \textbf{Scaling Factor:}
    Some implementations multiply the norms by an additional scaling factor $s$ to stabilize training.
\end{itemize}

A-Softmax generally exemplifies how incorporating an \emph{angular margin} contributes to more resilient face embeddings. This method is favored in numerous face recognition systems that require enhanced inter-class distinction and tightly-knit intra-class clusters.

\subsubsection{Center Loss}
\label{loss:center}

\emph{Center Loss} \cite{wen2016discriminative} is designed to reduce intra-class variation in feature representations by learning a \emph{class center} for each identity and penalizing the distance between each sample’s feature vector and its corresponding class center. When combined with a primary classification loss (e.g., softmax), this approach yields embeddings that are not only correctly classified but also tightly clustered by identity.

\paragraph{Formulation:}
Let \(\{\mathbf{x}_i\}_{i=1}^n\) be the feature vectors extracted by the network for a mini-batch of \(n\) samples, and let \(y_i\in\{1,\ldots,K\}\) denote the class (identity) label of sample \(i\). For each class \(k\), we maintain a center vector \(\mathbf{c}_k \in \mathbb{R}^d\). The \emph{Center Loss} is

\begin{equation}
\label{eq:center_loss}
L_{\mathrm{center}}
= \frac{1}{2}\sum_{i=1}^{n}
  \bigl\|\mathbf{x}_i - \mathbf{c}_{y_i}\bigr\|_2^2.
\end{equation}

During backpropagation, \(\{\mathbf{c}_k\}\) are also learnable parameters. To avoid abrupt fluctuations, the class centers are updated via an exponential moving average. For a sample \((\mathbf{x}_i, y_i)\), the update rule is

\begin{equation}
\label{eq:centers_update}
\mathbf{c}_{y_i}^{(t+1)}
= \mathbf{c}_{y_i}^{(t)}
  + \alpha \,\bigl(\mathbf{x}_i - \mathbf{c}_{y_i}^{(t)}\bigr),
\end{equation}

\noindent
where \(\alpha\in(0,1)\) is a small learning rate for the center vectors, ensuring smoother shifts.

Center Loss is typically added to the primary classification loss (e.g., softmax cross-entropy):

\begin{equation}
\label{eq:center_loss2}
L
= L_{\mathrm{softmax}} + \lambda\,L_{\mathrm{center}},
\end{equation}

\noindent
where \(\lambda>0\) balances the classification term and the clustering term. As a result, the network learns to position each sample close to its class center, improving intra-class compactness and discriminative power—critical qualities in face recognition tasks, where multiple faces must be distinguished with high accuracy.

\subsubsection{CosFace: Large-Margin Cosine Loss}
\label{loss:cosface}

\emph{CosFace} \cite{wang2018cosface}, also known as the \emph{Large-Margin Cosine Loss}, enhances face recognition by introducing a margin in \emph{cosine space} rather than in the angular domain (as in SphereFace). The key idea is to directly subtract a margin \(m\) from the cosine similarity of the correct class, thereby enforcing a stricter separation between classes.

Let \(\mathbf{x}_i\in \mathbb{R}^d\) be the feature for a sample \(i\), and let \(\mathbf{w}_j\in \mathbb{R}^d\) be the weight vector representing class \(j\). CosFace normalizes both \(\mathbf{x}_i\) and \(\mathbf{w}_j\) so that \(\|\mathbf{x}_i\|=1\) and \(\|\mathbf{w}_j\|=1\). The cosine of the angle \(\theta_j\) between \(\mathbf{x}_i\) and \(\mathbf{w}_j\) is:

\begin{equation}
\label{eq:cosface_cosine}
\cos\theta_j
= \mathbf{w}_j^\top \mathbf{x}_i.
\end{equation}

If \(y_i\) is the correct class for sample \(i\), the logit for this class is adjusted to \(\cos\theta_{y_i} - m\). All other classes keep \(\cos\theta_j\) unchanged. Let \(s>0\) be a scale factor. The \emph{CosFace Loss} for sample \(\mathbf{x}_i\) is:

\begin{equation}
\label{eq:cosface1}
L_i
= -\log \frac{\exp\bigl(s(\cos\theta_{y_i} - m)\bigr)}
           {\exp\bigl(s(\cos\theta_{y_i} - m)\bigr)
            + \sum_{j\neq y_i}
               \exp\bigl(s\,\cos\theta_j\bigr)}.
\end{equation}

\paragraph{Interpretation:}
\begin{itemize}
    \item \textbf{Margin in Cosine Space:}
    By subtracting \(m\) from the correct class’s cosine similarity, CosFace forces a larger gap between different classes in the normalized feature space.  
    \item \textbf{Stable Optimization:}
    Unlike SphereFace, which manipulates angles using \(\cos(m\theta)\), CosFace’s direct subtraction of \(m\) is simpler and avoids potential non-monotonic behaviors.  
    \item \textbf{Normalization:}
    Both \(\mathbf{x}_i\) and \(\mathbf{w}_j\) are normalized, preventing the network from trivially increasing norms to minimize the loss. Hence, the model truly learns to discriminate via angular separation.  
\end{itemize}

CosFace often outperforms standard softmax or SphereFace in challenging face benchmarks by providing a direct margin in the cosine domain, leading to better generalization and discriminability.

\subsubsection{ArcFace: Additive Angular Margin Loss}
\label{loss:arcface}

\emph{ArcFace} \cite{deng2019arcface}, also called \emph{Additive Angular Margin Loss}, refines margin-based approaches by placing a margin \emph{additively} in the \emph{angle} itself. This ensures a more geometrically interpretable boundary on the hypersphere.

\paragraph{Angular Margin:}
After normalizing \(\mathbf{x}_i\) and \(\mathbf{w}_{y_i}\), let \(\theta_{y_i}\) be the angle between \(\mathbf{x}_i\) and the correct class weight \(\mathbf{w}_{y_i}\). ArcFace modifies \(\theta_{y_i}\) to \(\theta_{y_i} + m\), where \(m>0\) is a constant margin. Consequently, \(\cos(\theta_{y_i}+m)\) is used in place of \(\cos(\theta_{y_i})\). With scale factor \(s\), the logit for the correct class becomes:

\begin{equation}
s\,\cos\bigl(\theta_{y_i}+m\bigr).
\end{equation}

\paragraph{Loss Definition:}
For a sample \((\mathbf{x}_i,y_i)\), the \emph{ArcFace Loss} is:

\begin{equation}
\label{eq:arcface_loss1}
L_i
= -\log
   \frac{\exp\bigl(s \,\cos(\theta_{y_i} + m)\bigr)}
        {\exp\bigl(s \,\cos(\theta_{y_i} + m)\bigr)
         + \sum_{j \neq y_i}
           \exp\bigl(s\,\cos(\theta_j)\bigr)}.
\end{equation}

\paragraph{Geometric Interpretation:}
\begin{itemize}
    \item \textbf{Additive \emph{Angle} Margin:}
    By shifting the angle by \(m\), ArcFace enforces that the model must produce a higher cosine for the correct class to compensate for this extra margin. This effectively increases inter-class separations on the hypersphere.
    \item \textbf{Stable and Interpretable:}
    The additive nature in the angle domain yields a monotonic boundary, making it simpler to optimize than purely multiplicative angle margins.
    \item \textbf{State-of-the-Art Accuracy:}
    ArcFace typically outperforms earlier margin-based losses (SphereFace, CosFace) on tasks like LFW or MegaFace, providing a robust face embedding with high intra-class compactness and inter-class separability \cite{manwatkar2023}.
\end{itemize}

In general, ArcFace advances the margin-based paradigm by adding a direct angular offset, leading to improved face verification and identification performance under challenging conditions of high identity similarity or limited training data.

\subsubsection{Triplet Loss}
\label{loss:triplet}

\emph{Triplet Loss} \cite{schultz2003learning,schroff2015facenet} is a classical metric-learning approach in face recognition. It aims to ensure that images of the same person (positive) lie closer in the learned embedding space to a reference image (anchor) than do images of different persons (negative). By enforcing a margin separation, the model learns discriminative features, reducing intra-class distances while increasing inter-class distances.

\paragraph{Definition:}
A \emph{triplet} consists of an \emph{anchor} \(A\), a \emph{positive} \(P\) (same identity as \(A\)), and a \emph{negative} \(N\) (different identity). Let \(\mathbf{f}_A, \mathbf{f}_P, \mathbf{f}_N \in \mathbb{R}^d\) be their respective embeddings. \emph{Triplet Loss} imposes:

\begin{equation}
\label{eq:triplet_loss}
L_{\mathrm{triplet}}
= \max \Bigl\{0,\,
    \|\mathbf{f}_A - \mathbf{f}_P\|_2^2
    - \|\mathbf{f}_A - \mathbf{f}_N\|_2^2
    + \alpha
  \Bigr\},
\end{equation}

\noindent
where \(\|\cdot\|_2\) denotes the Euclidean (L2) distance and \(\alpha>0\) is the margin parameter. Minimizing \(L_{\mathrm{triplet}}\) enforces:

\begin{equation}
\|\mathbf{f}_A - \mathbf{f}_P\|_2^2 
+ \alpha
< \|\mathbf{f}_A - \mathbf{f}_N\|_2^2,
\end{equation}
so that the anchor–positive distance is smaller than the anchor–negative distance by at least \(\alpha\).

\paragraph{Interpretation and Key Points:}
\begin{itemize}
    \item \textbf{Margin $\alpha$:}
    A sufficiently large margin fosters clearer separation but may slow training if set too high. Tuning $\alpha$ is critical for balanced learning.
    \item \textbf{Sampling Strategy:}
    Effectiveness depends heavily on how triplets are chosen. Hard or semi-hard negative mining can accelerate convergence by providing informative examples \cite{wu2017sampling,hoffer2015deep}.
    \item \textbf{Embedding Consistency:}
    Because the loss compares distances among triplets, the network learns a \emph{metric space} where same-identity samples cluster tightly, and different-identity samples are pushed apart.
\end{itemize}

Triplet Loss has proven successful in applications like \emph{FaceNet} \cite{schroff2015facenet}, where a single embedding space suffices for tasks like face verification or recognition without needing explicit class-based training for each new identity.

\subsubsection{Contrastive Loss}
\label{loss:contrastive}

\emph{Contrastive Loss} \cite{chopra2005learning} encourages pairs of images deemed \emph{similar} to lie close together in the embedding space and pairs deemed \emph{dissimilar} to lie further apart. This approach is often implemented via a Siamese network architecture, which processes pairs \((x_i^a, x_i^p)\) or \((x_i^a, x_i^n)\) to learn meaningful relative distances.

\paragraph{Definition:}
Given a labeled pair \((x_i^a, x_i^p)\) indicating \emph{similarity} (same identity) or \((x_i^a, x_i^n)\) indicating \emph{dissimilarity} (different identities), let the network’s embeddings be \(f(x_i^a)\) and \(f(x_i^p)\) or \(f(x_i^n)\). A binary label \(y_i\in\{0,1\}\) signifies whether the pair is similar (\(y_i=1\)) or dissimilar (\(y_i=0\)). The \emph{Contrastive Loss} is often written as:

\begin{equation}
\label{eq:contrastive}
L
= \frac{1}{N} \sum_{i=1}^{N} 
  \Bigl[
    y_i\,\|f(x_i^a) - f(x_i^p)\|_2^2
  + (1 - y_i)\,
    \max\!\Bigl(0,\,
        m - \|f(x_i^a) - f(x_i^n)\|_2^2
      \Bigr)
  \Bigr],
\end{equation}

\noindent
where \(m>0\) is a margin hyperparameter, \(N\) is the number of pairs in a batch, and \(\|\cdot\|_2\) denotes the L2 distance.

\paragraph{Terms Explanation:}
\begin{itemize}
    \item \textbf{Similar Pairs ($y_i=1$):}
    The term \(\|f(x_i^a) - f(x_i^p)\|_2^2\) is minimized, bringing matched samples closer.
    \item \textbf{Dissimilar Pairs ($y_i=0$):}
    The loss encourages \(\|f(x_i^a) - f(x_i^n)\|_2^2\ge m\). If the distance is already beyond \(m\), the contribution is 0, i.e., no further penalty.
\end{itemize}

\paragraph{Key Properties:}
\begin{itemize}
    \item \textbf{Margin $m$:}
    Controls how separate dissimilar pairs must be. Too small a margin may under-separate classes, while too large may impede convergence.
    \item \textbf{Metric Generalization:}
    Since the model learns a shared embedding space, \emph{new} classes (identities) can be recognized without retraining a classifier—only the distance function is used.
    \item \textbf{Possible Limitation:}
    All negative pairs are treated equally, assuming a uniform notion of dissimilarity. This might be simplistic if some negative pairs are “less dissimilar” than others \cite{manwatkar2023,wu2017sampling,hoffer2015deep}.
\end{itemize}

Contrastive Loss underlies many face verification systems, especially those requiring flexible addition of new individuals. By pushing same-identity pairs together and different-identity pairs beyond a margin, the learned feature space naturally supports verifying whether two faces belong to the same person.

\subsubsection{Circle Loss}
\label{loss:circle}

\emph{Circle Loss} \cite{sun2020circle} aims to enhance the discriminative power of learned embeddings by simultaneously maximizing positive-pair similarity and minimizing negative-pair similarity, all while maintaining a \emph{circle-like} decision boundary in the similarity space. This approach handles challenges like \emph{imbalance} and \emph{complex distributions}, which are prevalent in tasks such as face recognition.

\paragraph{Definition:}
Let \(\{s_{pos_i}\}\) and \(\{s_{neg_j}\}\) represent similarity scores for positive and negative pairs, respectively. Circle Loss introduces two margins \(O_{pos}<O_{neg}\), plus slack variables \(\alpha_{pos_i}, \alpha_{neg_j}\) to account for whether each pair already meets its margin requirement. A scaling factor \(\gamma>0\) amplifies the effects of violating these margins. Formally, each batch contributes:

\begin{equation}
\label{eq:circle_loss}
\begin{aligned}
\alpha_{pos_i} &= \max\bigl(O_{pos} - s_{pos_i},\, 0\bigr),\\
\alpha_{neg_j} &= \max\bigl(s_{neg_j} - O_{neg},\, 0\bigr),\\
\mathrm{sum}_{pos} &= \sum_i \exp\!\Bigl[-\gamma\, \alpha_{pos_i}\,(s_{pos_i} - O_{pos})\Bigr],\\
\mathrm{sum}_{neg} &= \sum_j \exp\!\Bigl[\;\gamma\, \alpha_{neg_j}\,(s_{neg_j} - O_{neg})\Bigr],\\
L_{\mathrm{Circle}} &= \ln\!\Bigl(1 + \mathrm{sum}_{pos}\,\times \mathrm{sum}_{neg}\Bigr).
\end{aligned}
\end{equation}

\paragraph{Interpretation:}
\begin{itemize}
    \item \textbf{Margins $(O_{pos}, O_{neg})$:} 
    Encourages positive similarities \(s_{pos_i}\) to exceed \(O_{pos}\) and negative similarities \(s_{neg_j}\) to remain below \(O_{neg}\). Typically \(O_{pos} < O_{neg}\).  
    \item \textbf{Slack Variables $\alpha_{pos_i}, \alpha_{neg_j}$:} 
    If a pair already satisfies its margin, its slack variable is 0, contributing minimally to the exponential terms.  
    \item \textbf{Scaling $\gamma$:}
    Magnifies the penalty for pairs failing their margin constraints. Larger $\gamma$ imposes harsher penalties but may make training more sensitive.
\end{itemize}

\paragraph{Practical Considerations:}
\begin{itemize}
    \item \textbf{Margin Tuning:}
    Choosing $O_{pos}$ and $O_{neg}$ (or a gap $m$ with $O_{neg}=O_{pos}+m$) is crucial. Inadequate margins may yield insufficient separation; overly large margins risk training instability.
    \item \textbf{Similarity Space Convergence:}
    Circle Loss directs optimization toward a “circle-like” boundary in the $(s_{neg}, s_{pos})$ space, focusing the network on achieving consistent separation for both positive and negative pairs.
\end{itemize}

Circle Loss often excels in scenarios with complex inter-class relationships or skewed distributions, driving robust separation in the similarity space without requiring explicit angular or cosine-based margins.

\subsubsection{Barlow Twins Loss}
\label{loss:barlow}

\emph{Barlow Twins} \cite{zbontar2021barlow} is a self-supervised approach that learns invariant and decorrelated feature representations. By processing two different augmentations of the same input, it aligns them in embedding space while reducing redundancy across feature dimensions. Though not specific to face recognition, its emphasis on robust, non-redundant features can benefit large-scale facial representation learning.

\paragraph{Notation:}
Consider embeddings $\mathbf{z}_a, \mathbf{z}_b\in\mathbb{R}^{D}$ produced by the network for two augmentations of the same image. Each embedding is normalized across a batch of size $N$, yielding $\mathbf{z}_{a_{\mathrm{norm}}}$ and $\mathbf{z}_{b_{\mathrm{norm}}}$ with zero mean and unit variance over the batch dimension.

\paragraph{Cross-Correlation Computation:}
Construct the $D\times D$ cross-correlation matrix

\begin{equation}
\label{eq:barlow_cc}
\mathbf{C}
= \frac{1}{N} \;\mathbf{z}_{a_{\mathrm{norm}}}^\top 
   \,\mathbf{z}_{b_{\mathrm{norm}}}.
\end{equation}

\paragraph{Difference and Scaling:}
Subtract the identity matrix $\mathbf{I}$ and square the result. Then optionally scale off-diagonal entries by a factor $\lambda$:

\begin{align}
\label{eq:barlow_diff}
\mathbf{C}_{\mathrm{diff}} 
&= (\mathbf{C} - \mathbf{I})^2,
\\
\label{eq:barlow_scale_diag}
\mathbf{C}_{\mathrm{diff}_{ij}}
&=
\begin{cases}
\mathbf{C}_{\mathrm{diff}_{ij}}, & i=j,\\
\lambda \,\mathbf{C}_{\mathrm{diff}_{ij}}, & i\neq j.
\end{cases}
\end{align}

\paragraph{Final Barlow Twins Loss:}
Sum up all elements of the adjusted matrix:

\begin{equation}
\label{eq:barlow1}
L
= \sum_{i=1}^D \sum_{j=1}^D 
   \mathbf{C}_{\mathrm{diff}_{ij}}.
\end{equation}

\paragraph{Goals:}
\begin{itemize}
    \item \textbf{High Diagonal Correlation:}
    $(C - I)$ drives diagonal elements of $\mathbf{C}$ to be 1, forcing $\mathbf{z}_a$ and $\mathbf{z}_b$ to match for each dimension.
    \item \textbf{Low Off-Diagonal Correlation:}
    Off-diagonal entries are penalized to be near 0, reducing redundancy across different dimensions of the learned representation.
\end{itemize}

Barlow Twins does not rely on negative examples, thus reducing data expansion overhead. However, it typically requires large embedding dimensionality and carefully chosen augmentations to perform effectively. In face recognition contexts, it can produce high-level features that remain robust under various transformations (occlusion, lighting changes), though additional fine-tuning or specialized augmentation may be needed.

\subsubsection{SimSiam Loss}
\label{loss:simsiam}

\emph{SimSiam} \cite{chen2021exploring} is another self-supervised method that learns invariant representations by comparing two augmented views of the same image. Although not specifically tailored for face recognition, it can be applied for unsupervised facial feature learning, leveraging data augmentations to achieve robust embeddings.

\paragraph{Architecture and Loss:}
Given two augmentations $x_1$ and $x_2$ of the same image, the network produces features $\mathbf{f}_1, \mathbf{f}_2$ and predictions $\mathbf{p}_1, \mathbf{p}_2$. The \emph{SimSiam Loss} is defined via negative cosine similarity between $\mathbf{p}_i$ on one view and $\mathbf{f}_j$ on the other:

\begin{equation}
\label{eq:sim_siam}
L
= -\tfrac12
  \left[
    \frac{\mathbf{p}_1^\top \mathbf{f}_2}{
          \|\mathbf{p}_1\|\;\|\mathbf{f}_2\|}
  + \frac{\mathbf{p}_2^\top \mathbf{f}_1}{
          \|\mathbf{p}_2\|\;\|\mathbf{f}_1\|}
  \right].
\end{equation}

The \emph{stop-gradient} operation is applied to $\mathbf{f}_2$ in the first term and $\mathbf{f}_1$ in the second, preventing trivial collapse (identical outputs for all inputs).

\paragraph{Advantages:}
\begin{enumerate}
    \item \textbf{No Negative Pairs:}
    Unlike contrastive methods requiring pairs of dissimilar samples, SimSiam relies purely on positive pairs (two augmentations of the same image), simplifying data structures.
    \item \textbf{Stop-Gradient Mechanism:}
    Crucial for avoiding degenerate solutions where all embeddings converge to a single vector.
    \item \textbf{Simple Architecture:}
    SimSiam uses a straightforward predictor head and a symmetric design, making it less complex than some other self-supervised methods.
\end{enumerate}

\paragraph{Drawbacks:}
\begin{enumerate}
    \item \textbf{Hyperparameter Tuning:}
    Performance can be sensitive to learning rate, weight decay, or batch size, requiring careful tuning.
    \item \textbf{Dependence on Augmentations:}
    Like other self-supervised methods, the choice and strength of image transformations significantly impact the learned features.
    \item \textbf{Potentially Non-semantic Features:}
    There is no absolute guarantee the model captures class-level semantics. The learned features might be robust to certain perturbations but not necessarily contain identity-relevant details.
\end{enumerate}

In face recognition, SimSiam can serve as a pretraining approach to obtain general facial embeddings, later refined with supervised objectives (e.g., ArcFace or CosFace) for improved identity separation. Its reliance on strong augmentations and careful parameter settings underscores the importance of domain knowledge in designing unsupervised facial representation workflows.

\section{Monocular Depth Estimation (MDE)}
\label{subsec:depth_est}

Monocular depth estimation in computer vision infers depth from a single 2D image, unlike stereo methods using multiple images. It has key applications in areas like autonomous driving and robotics, where multi-camera setups are often impractical. The main challenge is inferring 3D depth from a 2D projection, using cues like object size and scene context. Traditional methods with hand-crafted features had limited generalization. However, deep learning, especially CNNs and attention mechanisms, has dramatically improved depth estimation, leveraging large datasets for better accuracy. Supervised, self-supervised, and unsupervised methods using stereo pairs or video sequences are popular. Despite progress, challenges remain, such as handling varied lighting and occlusions and ensuring scalability for real-world scenarios. Current research aims to improve model efficiency, reduce computational needs, and integrate geometric and physics-based constraints for better predictions.

The following sections explain the loss functions and metrics utilized for depth estimation.

\subsection{Depth Estimation Loss Functions}
\label{subsec:depth_est_losses}

The evolution of loss functions in monocular depth estimation (MDE) reflects a shift from simple global measures, like scale-invariant or point-wise errors, to composite losses that balance global accuracy and fine-grained detail. Modern approaches integrate structural and semantic priors through edge-aware and smoothness losses, while photometric consistency losses have become standard in self-supervised methods. Task-specific enhancements, such as scale-consistent and boundary-aware losses, address challenges like scale ambiguity and boundary delineation. The integration of adversarial, perceptual, and multi-domain training losses enables robustness to noisy data and improves generalization across diverse datasets. In general, MDE loss functions have advanced to prioritize depth accuracy, structural consistency, and computational efficiency.

Table \ref{tab:guidelines-mde-losses} provides a detailed overview of commonly used loss functions in depth estimation, emphasizing their uses, characteristics of the data, advantages, and disadvantages.

\begin{table*}[ht!]
\caption{Guidelines for selecting a depth estimation loss function based on usage, data characteristics, advantages, and limitations.}
\label{tab:guidelines-mde-losses}
\centering
\footnotesize
\begin{tabular}{p{1.2cm}p{2.6cm}p{2.6cm}p{2.6cm}p{2.6cm}}
\toprule
\textbf{Function} 
& \textbf{Usage} 
& \textbf{Data Characteristics} 
& \textbf{Advantages} 
& \textbf{Limitations} \\
\midrule

Point-wise \newline MAE 
& Minimizes absolute difference \newline Useful for moderate depth ranges 
& Moderate variance \newline Less prone to outliers 
& Easy to interpret \newline Robust to large errors 
& Under-penalizes big residuals \newline Slower convergence sometimes \\
\midrule

Point-wise \newline MSE
& Minimizes squared difference \newline Common in regression 
& Works when outliers are rare \newline Prefers smaller variance 
& Penalizes large errors heavily \newline Well-known, simple 
& Overemphasizes big errors \newline Can oversmooth predictions \\
\midrule

Scale Invariant 
& Emphasizes relative depths \newline Common for monocular 
& Uncertain or inconsistent scale \newline Single-view depth 
& Robust to global scale shifts \newline Focuses on relative structure 
& May lose absolute scale \newline Requires careful parameter tuning \\
\midrule

SSIM 
& Preserves structural details \newline Perceptual measure 
& Depth maps with strong spatial cues \newline Local contrast vital 
& Encourages texture/edge fidelity \newline Improves visual quality 
& Less focus on pixel-level accuracy \newline Can be sensitive to noise \\
\midrule

Photometric
& Self-supervised from image recon. \newline Minimizes appearance diffs 
& Multi-view data \newline No explicit depth labels 
& Uses geometry consistency \newline Good for unsupervised setups 
& Assumes consistent illumination/visibility \newline Struggles with occlusions/dynamics \\
\midrule

Disparity Smoothness
& Regularizes depth in flat regions \newline Weighted by image gradients 
& Clear edges help \newline Texture-less areas benefit 
& Reduces random noise \newline Preserves important contours 
& Can oversmooth actual edges \newline Needs balance with edge terms \\
\midrule

Appearance Matching
& Reconstructs one view from another \newline Uses L1 + SSIM 
& Stereo/multi-view data \newline Lacks direct absolute supervision 
& Balances structure and pixel accuracy \newline Effective in stereo setups 
& Depends on photometric constancy \newline Sensitive to big viewpoint or light changes \\
\midrule

L-R Consistency 
& Matches left/right disparities \newline Penalizes inconsistent maps 
& Stereo pairs with known calibration \newline Synchronized captures 
& Improves stereo alignment \newline Reduces mismatched disparity 
& Needs two views \newline Not for single-view data \\
\midrule

BerHu
& Combines $L_1$ and $L_2$ \newline Handles wide error range 
& Large dynamic depth range \newline Adaptive threshold 
& Balances large/small residuals \newline Robust to big deviations 
& Threshold tuning is critical \newline Varies across datasets \\
\midrule

Edge Loss 
& Aligns depth edges with image edges \newline Focuses on boundaries 
& Useful for preserving sharp contours \newline Edge-based gradient check 
& Sharp transitions are maintained \newline Reduces boundary blur 
& Sensitive to image noise \newline Requires good edge detection \\
\midrule

Min Reproj.
& Picks min photometric error across views \newline Good for occlusion handling 
& Multi-view with dynamic objects \newline Avoids invalid projections 
& Improves self-supervised stability \newline Reduces occlusion artifacts 
& Can oversmooth in multi-valid areas \newline May need extra masking \\
\midrule

SSI
& Scale/shift alignment for diverse data \newline Robust cross-dataset 
& Varied scales across datasets \newline Ambiguous absolute depth 
& Better generalization \newline Less scale-dependent 
& Extra overhead for scale/shift \newline Sensitive to severe outliers \\
\bottomrule
\end{tabular}
\end{table*}

\subsubsection{Point-wise Error}
\label{loss:point-wise_error}

\emph{Point-wise errors} measure per-pixel discrepancies between the predicted depth map \(\hat{D}\) and the ground truth \(D\). Let \(\{\hat{d}_i\}_{i=1}^N\) and \(\{d_i\}_{i=1}^N\) be the predicted and ground-truth depths for \(N\) valid pixels. Two common forms are \emph{Mean Absolute Error (MAE)} and \emph{Mean Squared Error (MSE)}.

\paragraph{Mean Absolute Error (MAE):}
MAE penalizes the average absolute deviation between prediction and ground truth:
\begin{equation}
\label{eq:pointwise-mae}
\mathcal{L}_{\text{MAE}}
= \frac{1}{N}
  \sum_{i=1}^N
  \lvert \hat{d}_i - d_i \rvert.
\end{equation}
MAE treats all residuals equally, making it \emph{robust to moderate outliers} since large errors are weighted linearly.

\paragraph{Mean Squared Error (MSE):}
MSE penalizes large errors more heavily due to squaring:
\begin{equation}
\label{eq:pointwise-mse}
\mathcal{L}_{\text{MSE}}
= \frac{1}{N}
  \sum_{i=1}^N
  (\hat{d}_i - d_i)^2.
\end{equation}
While MSE is sensitive to outliers, it can drive the network to correct major depth misestimates quickly in early training stages.

\paragraph{Usage and Limitations:}
Point-wise errors are straightforward and have been employed in \cite{eigen2014depth} and subsequent MDE works to minimize overall pixel-level depth residuals. However, they do not capture structural relationships or relative depth cues among neighboring pixels. Consequently, point-wise errors alone may yield suboptimal boundary delineation or fail in texture-less regions. To mitigate these shortcomings, many modern MDE frameworks combine point-wise errors with structural metrics (e.g., SSIM \S\ref{loss:SSIM}) or smoothness constraints (e.g., edge-aware regularization).

\subsubsection{Scale Invariant Error}
\label{loss:scale_invariant_error}

\emph{Scale-invariant error} addresses the inherent scale ambiguity in monocular depth estimation. Proposed by Eigen~\emph{et~al.} \cite{eigen2014depth}, it shifts emphasis from absolute depth values to \emph{relative} depth consistency.

\paragraph{Definition:}
Let \(\{\hat{d}_i\}_{i=1}^N\) and \(\{d_i\}_{i=1}^N\) be predicted and ground-truth depths for \(N\) pixels. The scale-invariant loss uses logarithms of these depths:

\begin{equation}
\label{eq:scale-inv-error}
\mathcal{L}_{\mathrm{scale\text{-}inv}}
= \frac{1}{2} 
  \sum_{i=1}^N
  \bigl(\log \hat{d}_i - \log d_i\bigr)^2
  \;-\;
  \frac{\lambda}{N^2}
  \biggl(\sum_{i=1}^N
    \bigl(\log \hat{d}_i - \log d_i\bigr)
  \biggr)^2.
\end{equation}

\noindent
The first term is an \(\ell_2\) measure on \(\log\)-transformed depths, prioritizing multiplicative (relative) errors over additive. The second term counteracts mean bias in the log space, aligning the overall scale of \(\hat{D}\) to that of \(D\). The hyperparameter \(\lambda\) controls how strongly the model penalizes that global log offset; commonly, \(\lambda=1\) reproduces the original scale-invariant formulation \cite{eigen2014depth}.

\paragraph{Interpretation and Advantages:}
\begin{itemize}
    \item \textbf{Relative Depth Focus:}
    By operating in log space, the model can match the \emph{shape} of depth surfaces even if the absolute scale is off.  
    \item \textbf{Scale Ambiguity Mitigation:}
    This error metric tolerates global scaling errors, crucial in monocular setups lacking absolute distance references.  
    \item \textbf{Combination with Other Losses:}
    Often supplemented by smoothness (to encourage local regularity) or photometric alignment (in self-supervised settings), ensuring that relative depth structure remains consistent across the image.
\end{itemize}

Scale-invariant error is valuable in tasks where only relative depth matters. Nonetheless, if absolute metric scale is required (e.g., robotics tasks with real-world distances), additional constraints or multi-view geometry may be necessary to resolve scale fully. 

\subsubsection{Structural Similarity Index Measure (SSIM)}
\label{loss:SSIM}

\emph{Structural Similarity Index Measure (SSIM)} \cite{wang2004image,wang2017structural} compares two images by factoring in luminance, contrast, and local structure. In Monocular Depth Estimation (MDE), SSIM is utilized to preserve perceptual fidelity in predicted depth maps by encouraging local structural consistency with respect to the ground truth.

\paragraph{Formulation:}
Let \(\hat{D}\) and \(D\) be the predicted and ground-truth depth maps, respectively, and denote \(\mu_{\hat{D}}, \mu_D\) as local means, \(\sigma_{\hat{D}}, \sigma_D\) as local standard deviations, and \(\sigma_{\hat{D}D}\) as the cross-covariance of \(\hat{D}\) and \(D\) over a small spatial window. The SSIM index for that window is:

\begin{equation}
\label{eq:ssim-mde}
\text{SSIM}(\hat{D}, D)
= \frac{\bigl(2\mu_{\hat{D}}\mu_D + C_1\bigr)\,\bigl(2\,\sigma_{\hat{D}D} + C_2\bigr)}{\bigl(\mu_{\hat{D}}^2 + \mu_D^2 + C_1\bigr)\,\bigl(\sigma_{\hat{D}}^2 + \sigma_D^2 + C_2\bigr)},
\end{equation}

\noindent
where \(C_1\) and \(C_2\) are small constants to avoid division by zero. SSIM ranges from \(-1\) to \(1\), with \(1\) indicating perfect structural alignment.

To use SSIM in a loss objective, we convert similarity to dissimilarity:

\begin{equation}
\label{eq:ssim-mde-loss}
\mathcal{L}_{\text{SSIM}}
= 1 \;-\; \text{SSIM}(\hat{D}, D).
\end{equation}

Since SSIM is computed in local windows, a typical practice is to average over all windows:

\begin{equation}
\label{eq:ssim-mde-window}
\mathcal{L}_{\text{SSIM}}
= \frac{1}{N}
  \sum_{i=1}^N
  \Bigl(
    1 - \text{SSIM}\bigl(\hat{D}_i, D_i\bigr)
  \Bigr),
\end{equation}

\noindent
where \(N\) is the total number of local patches. This formulation encourages \(\hat{D}\) to exhibit structurally consistent edges and gradients relative to \(D\). SSIM loss is often combined with numeric errors (e.g., \(\ell_1\) or MSE) or other structural priors (e.g., smoothness) to balance overall depth accuracy with localized structural fidelity.

\subsubsection{Photometric Loss}
\label{loss:photometric}

\emph{Photometric Loss} \cite{zhou2017unsupervised} is commonly used in \emph{self-supervised} monocular depth estimation. It leverages photometric consistency between different viewpoints rather than requiring ground-truth depth. The idea is to warp a source image \(I_t\) into a reference view \(I_r\) using the predicted depth \(\hat{D}\) and the camera pose \(T\). Then the difference between \(I_r\) and the warped image \(I_s\) (synthesized from \(I_t\)) forms the loss.

\paragraph{Warping Procedure:}
Given a depth map \(\hat{D}\) and relative pose \(T\), a pixel in the reference view is projected into 3D using \(\hat{D}\), then reprojected into the source frame to sample a color from \(I_t\). Denote the warped image as:

\begin{equation}
\label{eq:photometric-warping}
I_s
= \mathcal{W}(I_t, \hat{D}, T).
\end{equation}

\paragraph{Photometric Consistency Loss:}
A combined measure of absolute intensity difference and structural similarity \cite{zhou2017unsupervised}:

\begin{equation}
\label{eq:photometric-mde}
\mathcal{L}_{\mathrm{photometric}}
= \frac{1}{N}
  \sum_{i=1}^N
  \Bigl[
    \alpha \,\frac{1 - \text{SSIM}\bigl(I_r^i, I_s^i\bigr)}{2}
  + (1-\alpha)\,\lvert I_r^i - I_s^i\rvert
  \Bigr],
\end{equation}

\noindent
where \(N\) indexes valid pixels, \(\alpha\in[0,1]\) weights the SSIM term relative to the absolute difference term, and \(I_r^i, I_s^i\) are pixel intensities in the reference and warped images respectively.

\paragraph{Usage and Limitations:}
\begin{itemize}
    \item \textbf{Self-Supervised Depth:}
    Photometric loss is crucial when ground-truth depth is unavailable. The network infers \(\hat{D}\) purely from multi-view photometric consistency.
    \item \textbf{Occlusions and Dynamic Scenes:}
    Real-world scenarios with movement or reflectance changes violate the brightness constancy assumption. Solutions include masking invalid regions, minimum reprojection losses, and auto-masking of dynamic objects.
    \item \textbf{Combination with Smoothness/Edge-Aware:}
    Due to the possibility of local minima or incorrect warps, photometric loss is typically combined with smoothness (\S\ref{loss:disp_smoothness}) and, optionally, geometric constraints for more robust depth estimates.
\end{itemize}

\subsubsection{Disparity Smoothness Loss}
\label{loss:disp_smoothness}

\emph{Disparity Smoothness Loss} \cite{Godard_2017_CVPR} acts as a spatial regularizer, penalizing abrupt or noisy depth variations except where image edges imply real depth discontinuities. By weighting predicted depth gradients with image gradients, it promotes piecewise-smooth depth except near intensity edges.

\paragraph{Formulation:}
Let \(\hat{D}_{i,j}\) be the predicted depth (or disparity) at pixel \((i,j)\), and let \(I_{i,j}\) be the corresponding intensity in a reference image. The smoothness loss is typically:

\begin{equation}
\label{eq:disp-smooth-mde}
\mathcal{L}_{\mathrm{smooth}}
= \sum_{i,j}
  \Bigl[
    \bigl|\partial_x \hat{D}_{i,j}\bigr|\,
    e^{-\lvert\partial_x I_{i,j}\rvert}
  + \bigl|\partial_y \hat{D}_{i,j}\bigr|\,
    e^{-\lvert\partial_y I_{i,j}\rvert}
  \Bigr],
\end{equation}

\noindent
where \(\partial_x\) and \(\partial_y\) denote horizontal and vertical gradients. Multiplying by \(\exp\!\bigl(-\lvert\partial_x I_{i,j}\rvert\bigr)\) or \(\exp\!\bigl(-\lvert\partial_y I_{i,j}\rvert\bigr)\) reduces the penalty near image edges.

\paragraph{Interpretation and Extensions:}
\begin{itemize}
    \item \textbf{Spatial Coherence:}
    In flat or texture-less regions, depth should remain relatively smooth. This term suppresses spurious gradient noise.
    \item \textbf{Edge Preservation:}
    Penalizing depth gradients less near strong intensity edges prevents over-smoothing objects’ boundaries.
    \item \textbf{Log-Scaled and Multi-Scale Variants:}
    Some works apply smoothing to \(\log(\hat{D})\) to handle wide depth ranges uniformly; multi-scale smoothing addresses coherence across multiple image resolutions.
\end{itemize}

\paragraph{Implementation in MDE Pipelines:}
Smoothness loss is often combined with (i) \emph{photometric reconstruction} (\S\ref{loss:photometric}), (ii) \emph{point-wise errors} (\S\ref{loss:point-wise_error}), or (iii) \emph{scale-invariant terms} (\S\ref{loss:scale_invariant_error}). This synergy helps produce depth maps that are not only globally accurate but also consistent at object boundaries, reflecting the real scene structure.

\subsubsection{Appearance Matching Loss}
\label{loss:appearance_matching}

\emph{Appearance Matching Loss} \cite{Godard_2017_CVPR} enforces that a warped or reconstructed image should closely resemble its corresponding original view. By relying on photometric consistency, it obviates the need for ground-truth depth, making it particularly suited for self-supervised or semi-supervised depth estimation approaches. 

\paragraph{Definition:}
Let \(I\) be the original (reference) image and \(\tilde{I}\) be the reconstructed (warped) image, derived from predicted disparity/depth. The \emph{Appearance Matching Loss} is typically defined as a weighted sum of an \(\ell_1\) intensity difference and a Structural Similarity (SSIM) term:

\begin{equation}
\label{eq:app_matching-mde}
\mathcal{L}_{\mathrm{ap}}
= \frac{1}{N} 
  \sum_{i,j}
  \Bigl[
    \alpha\,\frac{1 - \mathrm{SSIM}\bigl(I_{ij}, \tilde{I}_{ij}\bigr)}{2}
    + (1-\alpha)\,\lvert I_{ij} - \tilde{I}_{ij}\rvert
  \Bigr],
\end{equation}

\noindent
where \(N\) denotes the total number of pixels, \(\mathrm{SSIM}\) is computed over a local window (often \(3\times 3\)), and \(\alpha\in[0,1]\) balances the SSIM term against the absolute difference. A common choice is \(\alpha=0.85\).

\paragraph{Interpretation and Usage:}
\begin{itemize}
    \item \textbf{Photometric Consistency:}
    Encourages the warped image \(\tilde{I}\) to align with the reference \(I\), leveraging stereo or temporal viewpoint shifts to infer geometry.
    \item \textbf{SSIM Emphasis vs.\ Pixel-level Accuracy:}
    The SSIM portion preserves local structure (edges, textures), while the \(\ell_1\) term maintains per-pixel intensity fidelity.
    \item \textbf{Self-Supervised Depth:}
    The network learns depth by minimizing reconstruction error, sidestepping the need for ground-truth disparities.
\end{itemize}

\paragraph{Practical Variations:}
\begin{itemize}
    \item \textbf{SSIM Computation:}
    Often simplified to a \(3 \times 3\) box filter for computational efficiency.
    \item \textbf{Loss Combination:}
    Typically paired with smoothness losses (\S\ref{loss:disp_smoothness}) or left-right consistency (\S\ref{loss:left-right-consistency}) to ensure globally coherent depth.
\end{itemize}

\subsubsection{Left-Right Consistency Loss}
\label{loss:left-right-consistency}

\emph{Left-Right Consistency Loss} \cite{Godard_2017_CVPR} is commonly employed in stereo-based or self-supervised depth estimation, reinforcing consistency between left-view and right-view disparity (or depth) maps \emph{without requiring ground-truth labels}. It leverages two views, \(I^l\) and \(I^r\), and their learned disparity fields \(\mathbf{d}^l\) and \(\mathbf{d}^r\).

\paragraph{Disparity Warping:}
The network warps \(\mathbf{d}^r\) into the left coordinate frame via \(\mathbf{d}^l\). For pixel \((i,j)\), the location \((j + d^l_{ij})\) in the right disparity map is used. The \emph{Left-Right Consistency Loss} penalizes divergence between these:

\begin{equation}
\label{eq:left-right-consistency-left}
\mathcal{L}_{\mathrm{lr}}
= \frac{1}{N}
  \sum_{i,j}
  \Bigl\lvert 
    d^l_{i,j}
    - d^r_{\,i,\,(j + d^l_{i,j})}
  \Bigr\rvert,
\end{equation}

\noindent
where \(N\) is the number of valid pixels. The argument \(j + d^l_{i,j}\) indicates the horizontal shift due to the left-view disparity.

In many depth estimation tasks, especially when ground-truth data is scarce or expensive to obtain, stereo pairs serve as a form of weak supervision. By coupling \(\mathcal{L}_{\mathrm{lr}}\) with a photometric reconstruction loss (see \S\ref{loss:appearance_matching}), the network learns to produce consistent disparities on both views and minimize appearance differences when one view is warped into the other. This setup removes the need for labeled depth maps, making the training process \emph{unsupervised} (or self-supervised). Specifically, disparities are learned by enforcing internal geometric and photometric consistency, eliminating dependence on actual depth annotations. Enforcing left-right consistency complements the photometric loss, improving overall depth quality by penalizing mismatched disparities across views. Furthermore, large stereo datasets can be collected without manual labeling, making this approach appealing for real-world scenarios where labeled depth is limited.

\paragraph{Total Pipeline Loss:}
As introduced by \cite{Godard_2017_CVPR}, \(\mathcal{L}_{\mathrm{lr}}\) is typically combined with appearance matching (\S\ref{loss:appearance_matching}) and smoothness (\S\ref{loss:disp_smoothness}) terms:

\begin{equation}
\label{eq:left-right-consistency-mde}
\mathcal{L}
= \alpha_{\mathrm{ap}}\,\mathcal{L}_{\mathrm{ap}}
 + \alpha_{\mathrm{ds}}\,\mathcal{L}_{\mathrm{ds}}
 + \alpha_{\mathrm{lr}}\,\mathcal{L}_{\mathrm{lr}},
\end{equation}

\noindent
where \(\alpha_{\mathrm{ap}}, \alpha_{\mathrm{ds}}, \alpha_{\mathrm{lr}}\) are hyperparameters controlling the relative importance of each term.

\paragraph{Advantages and Considerations:}
\begin{itemize}
    \item \textbf{Stereo Regularization:}
    Ensures that left-view and right-view disparities remain consistent, reducing artifacts or mismatches.
    \item \textbf{Training with Single-View Inputs:}
    Although only one view might feed the main encoder, the other disparity map is still predicted, ensuring synergy between the two. This is especially useful when extending stereo-based methods to monocular sequences in a self-supervised manner.
    \item \textbf{Occlusion Handling:}
    Real stereo pairs have occluded regions visible only in one view. Usually, masks or confidence measures are introduced so that $\mathcal{L}_{\mathrm{lr}}$ excludes invalid pixels.
    \item \textbf{Generalizability:}
    By removing the need for explicit labels, left-right consistency methods can be adapted to various domains where collecting ground-truth depth is infeasible.
\end{itemize}

\subsubsection{BerHu Loss (Reverse Huber)}
\label{loss:BerHu}

\emph{BerHu Loss} \cite{zwald2012berhu,laina2016deeper} is a “Reverse Huber” formulation that switches between $\ell_1$ and $\ell_2$ norms based on a threshold $\delta$. It aims to robustly handle both small and large depth errors across wide depth ranges.

\paragraph{Definition:}
For a predicted depth map \(\{\hat{d}_i\}\) and ground-truth \(\{d_i\}\), BerHu defines:

\begin{equation}
\label{eq:berhu-mde}
\mathcal{L}_{\mathrm{BerHu}}(\hat{d}_i, d_i)
= 
\begin{cases}
\lvert \hat{d}_i - d_i\rvert, 
  & \text{if } \lvert \hat{d}_i - d_i\rvert \le \delta,\\[6pt]
\frac{(\hat{d}_i - d_i)^2 + \delta^2}{2\,\delta}, 
  & \text{otherwise},
\end{cases}
\end{equation}

\noindent
where \(\delta\) is a threshold controlling where the loss transitions from $\ell_1$ to $\ell_2$. If \(\lvert \hat{d}_i - d_i\rvert \le \delta\), the penalty is linear; if it exceeds \(\delta\), the penalty grows quadratically.

\paragraph{Adaptive Threshold:}
One strategy sets \(\delta\) as a fraction of the maximum error in a batch \(\{\max_i |\hat{d}_i - d_i|\}\), ensuring the threshold scales with current training dynamics. This can improve stability across varying depth ranges.

\paragraph{Usage and Benefits:}
\begin{itemize}
    \item \textbf{Small Errors vs.\ Large Errors:}
    $\ell_1$ handles finer deviations well, while the $\ell_2$ region strongly penalizes large outliers.
    \item \textbf{Wide Depth Ranges:}
    Particularly effective where the disparity spans multiple orders of magnitude (e.g., indoor vs.\ outdoor scenes).
\end{itemize}

\paragraph{Limitations:}
\begin{itemize}
    \item \textbf{Threshold Tuning:}
    Incorrect \(\delta\) can hamper performance or slow convergence, motivating adaptive or data-driven choices.
    \item \textbf{Structural Awareness:}
    Like point-wise errors, BerHu does not inherently encode geometric or structural priors. Many MDE pipelines combine BerHu with edge-aware or scale-invariant terms to capture robust scene structure.
\end{itemize}

Thus, BerHu Loss offers a middle ground that linearly penalizes moderate errors but escalates penalties for large errors, making it suitable for real-world depth estimation tasks with varying depth scales.

\subsubsection{Edge Loss}
\label{loss:edge}

\emph{Edge Loss} \cite{paul2022edge} aims to preserve sharp transitions in the predicted depth map near object boundaries or edges. While smoothness terms (\S\ref{loss:disp_smoothness}) encourage local coherence, they can inadvertently blur true discontinuities. Edge Loss mitigates this by aligning depth gradients with corresponding image gradients, ensuring that depth discontinuities match actual intensity edges in the reference image.

\paragraph{Formulation:}
Let \(\hat{D}\) be the predicted depth map and \(I\) be a reference intensity image. Denote \(\nabla_x\) and \(\nabla_y\) as gradient operators in the horizontal and vertical directions. The Edge Loss typically penalizes mismatches between the depth gradients \(\nabla_x \hat{D}, \nabla_y \hat{D}\) and the image gradients \(\nabla_x I, \nabla_y I\). A common form is:

\begin{equation}
\label{eq:edgeloss-mde}
\mathcal{L}_{\mathrm{edge}}
= \sum_{i,j} 
  \Bigl(
    \bigl\lvert 
      \nabla_x \hat{d}_{i,j}
      - \nabla_x I_{i,j}
    \bigr\rvert 
  + \bigl\lvert
      \nabla_y \hat{d}_{i,j}
      - \nabla_y I_{i,j}
    \bigr\rvert
  \Bigr),
\end{equation}

\noindent
where \(\hat{d}_{i,j}\) is the predicted depth at pixel \((i,j)\), and \(I_{i,j}\) is the intensity in the reference image. By enforcing \(\nabla_x \hat{D}\approx \nabla_x I\) and \(\nabla_y \hat{D}\approx \nabla_y I\), the network learns to place depth discontinuities only where genuine intensity edges exist.

\paragraph{Usage and Practical Aspects:}
\begin{itemize}
    \item \textbf{Edge-Aware Regularization}: 
    Unlike pure smoothness terms, Edge Loss prioritizes preserving structural detail, preventing over-smoothing around object boundaries.
    \item \textbf{Sensitivity to Noise}: 
    Since it directly uses image gradients, noise or artifacts in \(I\) may produce spurious edges. Real implementations often incorporate edge-aware weighting or denoising to reduce false gradients.
    \item \textbf{Multi-Scale or Combined Approach}: 
    Some approaches apply Edge Loss across multiple scales or combine it with point-wise or photometric objectives (e.g., \S\ref{loss:photometric}) to refine both global and boundary-level accuracy.
\end{itemize}

Edge Loss thus establishes a balance between enforcing smooth depth in homogeneous regions and preserving crisp depth transitions at real scene edges, leading to more perceptually convincing and structurally consistent depth predictions in MDE.

\subsubsection{Minimum Reprojection Loss}
\label{loss:min_rep}

\emph{Minimum Reprojection Loss} \cite{godard2019digging} is a self-supervised strategy to handle occlusions and dynamic objects in multi-view (or stereo) data. Instead of averaging photometric errors across all source views, it \emph{selects the minimal} reprojection error per pixel, thus ignoring view-specific occlusions or erroneous projections.

\paragraph{Motivation:}
When warping multiple source images \(\{I_t^k\}\) into the reference view \(I_r\) using a predicted depth \(\hat{D}\) and known camera transformations, some pixel reprojections may be invalid due to occlusion or dynamic scene elements. By taking the minimum photometric error among all source images, the network automatically emphasizes valid, well-aligned viewpoints while disregarding noisy reprojected pixels.

\paragraph{Definition:}
For each pixel \(i\), we compute a photometric error \(\mathcal{E}_{\mathrm{photo}}\) w.r.t.\ each warped image \(\hat{I}_t^{k}\). Commonly, \(\mathcal{E}_{\mathrm{photo}}\) is a blend of SSIM and \(\ell_1\)-intensity differences \cite{zhou2017unsupervised}, e.g.,

\begin{equation}
\label{eq:photometric-min-rep-loss}
\mathcal{E}_{\mathrm{photo}}\bigl(I_r^i, \hat{I}_t^{k,i}\bigr)
= \alpha\,
  \frac{1 - \mathrm{SSIM}\bigl(I_r^i, \hat{I}_t^{k,i}\bigr)}{2}
  \;+\;
  (1-\alpha)\,
  \lvert I_r^i - \hat{I}_t^{k,i}\rvert,
\end{equation}

\noindent
where \(k\) indexes source images, \(\alpha\in[0,1]\) balances the SSIM and absolute difference terms, and \(i\) indexes pixels. The \emph{Minimum Reprojection Loss} is then:

\begin{equation}
\label{eq:min-rep-loss-mde}
\mathcal{L}_{\mathrm{min\text{-}reproj}}
= \frac{1}{N}
  \sum_{i=1}^N
    \min_{k}
    \mathcal{E}_{\mathrm{photo}}\bigl(I_r^i, \hat{I}_t^{k,i}\bigr),
\end{equation}

\noindent
where \(N\) is the total number of valid pixels. Each pixel \emph{chooses} the source view \(k\) yielding the least reprojection error.

\paragraph{Advantages and Caveats:}
\begin{itemize}
    \item \textbf{Occlusion Handling}: 
    If a pixel is occluded in one source view, it might be visible in another. Taking the minimum reduces penalizing erroneous warping of occluded regions.
    \item \textbf{Dynamic Scenes}: 
    Minimizes artifacts from moving objects or inconsistent image patches across views.
    \item \textbf{Potential Oversmoothing}: 
    In certain regions with multiple valid reprojections, the model can over-rely on the minimum, ignoring fine details. Automasking or advanced heuristics can mitigate this phenomenon \cite{godard2019digging}.
\end{itemize}

Minimum Reprojection Loss is typically paired with smoothness (\S\ref{loss:disp_smoothness}), edge-aware constraints (\S\ref{loss:edge}), or geometric consistency to build a robust self-supervised depth estimation framework. This synergy ensures both local coherence and the ability to disregard invalid or occluded pixels in multi-view training data, resulting in sharper, more accurate depth maps.

\subsubsection{Scale-and-shift-invariant Loss (SSI)}
\label{loss:ssi}

The \emph{Scale-and-Shift-Invariant Loss (SSI Loss)}~\cite{ranftl2020towards} was designed to address the challenges associated with depth (or disparity) prediction tasks, particularly where ground-truth depth values may vary across datasets due to differences in scale and shift. By making the loss function invariant to scale and shift ambiguities, it enables robust training and facilitates better generalization across diverse datasets with varying depth or disparity annotations.

Let \(\hat{d} \in \mathbb{R}^M\) denote the \emph{predicted} disparities (or depths) and \(d \in \mathbb{R}^M\) represent the \emph{ground-truth} disparities, where \(M\) is the number of valid pixels. The SSI Loss is defined as in~\eqref{eq:scale-shift-inv1}:

\begin{equation}
\label{eq:scale-shift-inv1}
L_{\text{ssi}}(\hat{d}, d) 
= \frac{1}{2M} \sum_{i=1}^M r\bigl(\hat{d}^{*}_i - d^{*}_i\bigr),
\end{equation}

where \(\hat{d}^*\) and \(d^*\) are the \emph{scale- and shift-aligned} predicted and ground-truth disparities, respectively, and \(r(\cdot)\) is a chosen loss function (e.g., mean squared error, absolute error, or trimmed absolute error).

\paragraph{Scale-and-shift alignment.}
The alignment of predicted and ground-truth disparities involves solving the least-squares problem:

\begin{equation}
\label{eq:scale-shift-alignment}
s, t \;=\; \arg \min_{s,\,t} 
\sum_{i=1}^M \Bigl( s\,\hat{d}_i \;+\; t \;-\; d_i \Bigr)^2,
\end{equation}
where \(s\) and \(t\) are the \emph{scale} and \emph{shift} parameters. The resulting aligned predictions and ground truths are given by

\begin{equation}
\label{eq:scale-shift-aligned}
\hat{d}^* \;=\; s\,\hat{d} \;+\; t,
\quad
d^* \;=\; d.
\end{equation}

Alternatively, robust estimators for \(\hat{d}\) may be used for scale and shift::

\begin{equation}
\label{eq:scale-shift-robust}
t(\hat{d}) \;=\; \mathrm{median}(\hat{d}),
\quad
s(\hat{d}) \;=\; \frac{1}{M}\sum_{i=1}^M \bigl|\hat{d}_i - t(\hat{d})\bigr|.
\end{equation}

\paragraph{Advantages and trade-offs.}
This loss offers several benefits, including robustness to large variations across datasets, improved generalization, and numerical stability. By explicitly handling scale and shift ambiguities, \(L_{\text{ssi}}\) can integrate heterogeneous data sources, each possibly having different disparity ranges. Numerical stability is also enhanced by working in disparity space (rather than log-space), avoiding issues that can arise from taking the logarithm of small or zero depth values.

A potential downside is the computational overhead of estimating \(\{s, t\}\) for each training sample, especially if robust estimation is used. While robust estimators mitigate the influence of outliers, extreme ground-truth errors can still negatively affect training. Additionally, the SSI Loss’s effectiveness hinges upon accurate alignment; if scale/shift estimates are poor, performance may degrade.

\paragraph{Implementation details.}
In practice, the SSI Loss is implemented via the following steps:
\begin{enumerate}
    \item Compute the optimal \(s\) and \(t\) (using least-squares~\eqref{eq:scale-shift-alignment} or robust approaches~\eqref{eq:scale-shift-robust}).
    \item Align the predicted and ground-truth disparities \(\hat{d}^*, d^*\) using \(s\) and \(t\) as in~\eqref{eq:scale-shift-aligned}.
    \item Apply a robust loss function \(r(\cdot)\) (e.g., trimmed absolute error) to the aligned values:
    \begin{equation}
      L_{\text{ssi}}(\hat{d}, d) 
      \;=\; \frac{1}{2M}\sum_{i=1}^M r\bigl(\hat{d}^*_i - d^*_i\bigr).
    \end{equation}
    \item Optionally include regularization terms such as a gradient-matching loss~\eqref{eq:scale-shift-regul}:
    \begin{equation}
    \label{eq:scale-shift-regul}
      L_{\text{reg}} 
      = \frac{1}{M} 
        \sum_{k=1}^K 
        \sum_{i=1}^M 
        \bigl|\nabla R^k_i\bigr|,
    \end{equation}
    where \(R = \hat{d}^* - d^*\) is the residual, and \(K\) is the number of scales (for multi-scale gradient consistency).
\end{enumerate}

\paragraph{Combining with gradient matching.}
In~\cite{ranftl2020towards}, the authors add a gradient-matching regularization term to sharpen discontinuities and refine local details. The total loss function is given in~\eqref{eq:scale-shift-final-mde}:

\begin{equation}
\label{eq:scale-shift-final-mde}
L \;=\; L_{\text{ssi}} \;+\; \alpha\,L_{\text{reg}},
\end{equation}
where \(L_{\text{ssi}}\) is the scale-and-shift-invariant term described above, and \(L_{\text{reg}}\) is a multi-scale gradient-matching loss. Specifically, \eqref{eq:scale-shift-gradient-matching-mde} details \(L_{\text{reg}}\):

\begin{equation}
\label{eq:scale-shift-gradient-matching-mde}
L_{\text{reg}}
= \frac{1}{M} \sum_{k=1}^K \sum_{i=1}^M 
  \Bigl(\bigl|\nabla_x R^k_i\bigr| \;+\; \bigl|\nabla_y R^k_i\bigr|\Bigr),
\end{equation}
where again \(R = \hat{d}^* - d^*\), and \(\nabla_x\) and \(\nabla_y\) denote horizontal and vertical derivatives, respectively. By tuning \(\alpha\), one can adjust the strength of the gradient-based penalty relative to the global alignment term, offering flexibility across different data domains or tasks.

In general, combining \(L_{\text{ssi}}\) with gradient matching has proven advantageous, yielding:
\begin{itemize}
    \item \emph{Complementary strengths}: \(L_{\text{ssi}}\) ensures global alignment while \(L_{\text{reg}}\) sharpens object boundaries and local details.
    \item \emph{Improved cross-dataset generalization}: The combined loss often demonstrates superior zero-shot performance on unseen datasets.
    \item \emph{Controllability}: By adjusting the hyperparameter \(\alpha\), practitioners can balance global alignment against high-frequency refinement depending on task requirements.
\end{itemize}

\subsection{Depth Estimation Metrics}
\label{subsec:depth_est_metrics}

Assessing the performance of depth estimation models requires the use of several metrics that reflect diverse dimensions of accuracy and quality. These metrics assess absolute and relative error rates, pixel-wise accuracy, and the alignment of predicted depth with the actual data. Listed below are the most commonly employed metrics for depth estimation; although some have been previously introduced, our focus here will be on their application in depth estimation.

The choice of metrics depends on the specific application and dataset. For example:
\begin{itemize}
    \item \(\text{AbsRel}\), \(\text{RMSE}\), and \(\delta\) are commonly used for absolute depth datasets (e.g., KITTI \cite{geiger2012wekitti}, NYU \cite{silberman2011indoornyu,silberman2012indoornyu2}).
    \item WHDR is specifically suited for datasets with ordinal depth annotations (e.g., DIW \cite{chen2016single_diw}).
    \item Scale-Invariant Error and RMSE(log) are ideal for tasks with varying scales or relative depth datasets.
\end{itemize}

Every single metric provides unique information, and combining multiple metrics leads to a comprehensive evaluation of the effectiveness of depth estimation.

Table \ref{tab:guidelines-mde-metrics} provides a summary of commonly used metrics in depth estimation, highlighting their uses, characteristics of the data, advantages, and disadvantages.

\begin{table*}[ht!]
\caption{Guidelines for selecting depth estimation metrics based on usage, data characteristics, advantages and limitations.}
\label{tab:guidelines-mde-metrics}
\centering
\footnotesize
\begin{tabular}{p{1.3cm}p{2.8cm}p{2.8cm}p{2.8cm}p{2.8cm}}
\toprule
\textbf{Function} 
& \textbf{Usage} 
& \textbf{Data Characteristics} 
& \textbf{Advantages} 
& \textbf{Limitations} \\
\midrule
AbsRel
& Evaluates average relative deviation\newline between predicted and ground truth
& Emphasizes errors in smaller depths,\newline useful for moderate ranges
& Highlights relative discrepancies\newline good for tasks needing fine-grained accuracy
& Can undervalue errors at large depths\newline sensitive to scale variations \\
\midrule
RMSE
& Measures overall deviation,\newline penalizes large errors heavily
& Suitable for datasets without extremely\newline high depth ranges
& Common and intuitive,\newline highlights outlier errors
& Large errors can dominate,\newline not scale-invariant \\
\midrule
RMSE(log)
& Focuses on proportional differences\newline in log space
& Effective across wide depth scales,\newline mitigates scale mismatches
& Balances errors for both small\newline and large depths
& Sensitive near zero-depth values,\newline can be less intuitive \\
\midrule
Threshold \newline Accuracy \(\delta\)
& Fraction of pixels within a\newline multiplicative error bound
& Useful for quick checks\newline in diverse depth ranges
& Simple to interpret,\newline highlights near-accurate predictions
& Discontinuous metric,\newline choice of threshold is arbitrary \\
\midrule
Mean Log10
& Assesses log-scale deviations,\newline robust for varied depth distributions
& Suitable for indoor-outdoor mixes,\newline handles large-scale diversity
& Scale invariant,\newline balances small and large depth errors
& Amplifies near-zero errors,\newline can be harder to interpret than RMSE \\
\midrule
Percentage of \newline High Error
& Identifies proportion of pixels\newline exceeding a chosen relative error
& Useful in safety-critical settings,\newline focuses on large deviations
& Emphasizes worst-case performance,\newline reveals catastrophic failures
& Threshold selection is subjective,\newline ignores moderate errors \\
\midrule
WHDR
& Evaluates ordinal consistency\newline with human-labeled pairs
& Applicable to sparse or relative data,\newline no absolute ground truth
& Handles subjective depth ordering,\newline good for datasets with pairwise labels
& Cannot assess absolute depth,\newline depends on annotation quality \\
\midrule
Scale-\newline Invariant
& Focuses on structural accuracy\newline regardless of global scale
& Helpful when true scale is unknown\newline or for cross-dataset tasks
& Robust to scale mismatch,\newline leverages log-space corrections
& Less useful when exact scale\newline is crucial, can mask global errors \\
\bottomrule
\end{tabular}
\end{table*}

\subsubsection{Mean Absolute Relative Error (AbsRel)}
\label{metric:absrel_mde}

\emph{Mean Absolute Relative Error (AbsRel)} weights errors by the ground-truth depth, highlighting inaccuracies at smaller depths more strongly \cite{eigen2014depth}. Let \(\{\hat{d}_i\}\) and \(\{d_i\}\) be predicted and true depths for \(M\) valid pixels:

\begin{equation}
\label{eq:absrel-mde}
\mathrm{AbsRel}
= \frac{1}{M}
  \sum_{i=1}^M
  \frac{\lvert\hat{d}_i - d_i\rvert}{d_i}.
\end{equation}

\paragraph{Interpretation:}
\begin{itemize}
    \item \textbf{Relative Emphasis:}
    Large errors at shallow depths can heavily impact the metric, reflecting the significance of small-scale geometry in many applications.
    \item \textbf{Limitations:}
    If any \(d_i\approx 0\) or the dataset spans wide depth ranges, AbsRel can be unstable or less intuitive, motivating complementary metrics like RMSE.
\end{itemize}

\subsubsection{Root Mean Squared Error (RMSE)}
\label{metric:rmse_mde}

\emph{RMSE} \cite{eigen2014depth} quantifies the overall deviation between predicted \(\hat{d}_i\) and ground-truth \(d_i\), placing stronger penalization on large residuals:

\begin{equation}
\label{eq:rmse-mde}
\mathrm{RMSE}
= \sqrt{\frac{1}{M}
  \sum_{i=1}^M
  \bigl(\hat{d}_i - d_i\bigr)^2}.
\end{equation}

\paragraph{Usage:}
\begin{itemize}
    \item \textbf{Global Error Sensitivity:}
    RMSE magnifies large outliers, beneficial in catching significant depth mispredictions but can overshadow small, systematic errors.
    \item \textbf{Interpretation:}
    RMSE is dimensionally in units of depth, making it straightforward but sometimes less robust when scale ambiguities exist.
\end{itemize}

\subsubsection{Logarithmic RMSE (RMSE(log))}
\label{metric:rmse_log_mde}

\emph{RMSE(log)} \cite{eigen2014depth} measures errors in log space, emphasizing relative over absolute deviations. Formally,

\begin{equation}
\label{eq:rmselog-mde}
\mathrm{RMSE(log)}
= \sqrt{\frac{1}{M}
  \sum_{i=1}^M
  \bigl[\log(\hat{d}_i) - \log(d_i)\bigr]^2}.
\end{equation}

\paragraph{Advantages:}
\begin{itemize}
    \item \textbf{Large Depth Range Handling:}
    Suitable for datasets mixing near-field and far-field distances (e.g., indoor vs.\ outdoor).
    \item \textbf{Limitations:}
    Depth values near zero cause numerical issues, and the log transform can complicate direct interpretability compared to linear measures.
\end{itemize}

\subsubsection{Threshold Accuracy (\texorpdfstring{\(\boldsymbol{\delta}\)}{delta})}
\label{metric:threshold_acc_mde}

\emph{Threshold Accuracy} \cite{eigen2014depth} measures the fraction of pixels whose predicted depth falls within a factor of the ground truth:

\begin{equation}
\delta
= \max\!\Bigl(
    \frac{\hat{d}_i}{d_i},\,
    \frac{d_i}{\hat{d}_i}
  \Bigr).
\end{equation}
One reports the percentage of pixels satisfying \(\delta < \delta_{\mathrm{threshold}}\). Common thresholds are \(\delta_1=1.25\), \(\delta_2=1.25^2\), \(\delta_3=1.25^3\).

\paragraph{Interpretation:}
\begin{itemize}
    \item \textbf{Relative Criterion:}
    Captures how many pixels are \emph{relatively} accurate within a multiplicative bound.
    \item \textbf{Multiple Thresholds:}
    \(\delta_1, \delta_2, \delta_3\) reflect progressively looser accuracy constraints, giving a multi-level view of performance.
\end{itemize}

\subsubsection{Mean Log10 Error}
\label{metric:mean_log10_mde}

\emph{Mean Log10 Error} \cite{eigen2014depth} measures the absolute difference in base-10 logarithmic space:

\begin{equation}
\label{eq:meanlog10-mde}
\mathrm{Mean\ Log10}
= \frac{1}{M}
  \sum_{i=1}^M
  \bigl\lvert
    \log_{10}(\hat{d}_i)
    - \log_{10}(d_i)
  \bigr\rvert.
\end{equation}

\paragraph{Advantages and Limitations:}
\begin{itemize}
    \item \textbf{Logarithmic Emphasis:}
    Similar to RMSE(log), it focuses on relative discrepancies, beneficial in multi-scale or cross-dataset applications.
    \item \textbf{Interpretation Challenges:}
    Depth near zero or extremely large can skew the metric, and base-10 logs may be less intuitive than linear errors for some tasks.
\end{itemize}

\subsubsection{Percentage of Pixels with High Error}
\label{metric:pixels_high_error_mde}

This metric \cite{geiger2012wekitti} evaluates the fraction of pixels whose relative error exceeds a threshold \(\tau\). Formally,

\begin{equation}
\label{eq:percent-pix-high-error}
\mathrm{Percent\ Error}
= \frac{1}{M}
  \sum_{i=1}^M
  \mathbb{1}\!\Bigl(
    \frac{\lvert \hat{d}_i - d_i\rvert}{d_i}
    > \tau
  \Bigr),
\end{equation}
where \(\mathbb{1}(\cdot)\) is the indicator function. Threshold \(\tau\) might be \(0.1\), \(0.2\), or \(0.5\), depending on desired strictness.

\paragraph{Interpretation:}
\begin{itemize}
    \item \textbf{Focus on Worst-Case Pixels:}
    Identifies how often the model severely mispredicts depth, which can be critical for safety-related or robotics tasks.
    \item \textbf{Threshold Choice:}
    Lower \(\tau\) yields a stricter standard, highlighting small permissible errors, whereas higher \(\tau\) is more lenient but less informative about mild deviations.
\end{itemize}

\subsubsection{Weighted Human Disagreement Rate (WHDR)}
\label{metric:whdr_mde}

\emph{Weighted Human Disagreement Rate (WHDR)} \cite{chen2016single_diw} evaluates how well predicted depths match human-labeled ordinal relations (closer, same, or farther) in pairs of image pixels. It is especially suitable when absolute ground-truth depths are unavailable, and only pairwise judgments exist:

\begin{equation}
\label{eq:whdr-mde}
\mathrm{WHDR}
= \frac{
    \sum_{(i,j)} 
      w_{ij}\,
      \mathbb{1}\Bigl(
        l_{ij}\,\bigl(\hat{d}_i - \hat{d}_j\bigr) < 0
      \Bigr)
    }{
    \sum_{(i,j)} w_{ij}
  },
\end{equation}

\noindent
where \(l_{ij}\in\{-1,0,1\}\) is the ordinal relationship for pair \((i,j)\) (e.g., \(l_{ij}=-1\) if \(i\) is nearer than \(j\)), and \(w_{ij}\) weights that pair’s importance.

\paragraph{Usage:}
\begin{itemize}
    \item \textbf{Ordinal-Only Datasets:}
    WHDR is crucial in large-scale data lacking metric depths but providing relative ranks or “closer/farther” judgments.
    \item \textbf{Limitations:}
    Metric scale is not assessed, and it depends on the quality of human annotations. Also, ties or equal distances can complicate interpretation.
\end{itemize}

\subsubsection{Scale-Invariant Error (Metric Form)}
\label{metric:scale_invariant_mde}

\emph{Scale-Invariant Error} \cite{eigen2014depth} can also serve as a metric, complementing or replacing standard \(\ell_2\) measures by focusing on relative depth consistency. For predicted \(\{\hat{d}_i\}\) and true \(\{d_i\}\), the scale-invariant metric is:

\begin{equation}
\label{eq:scale-inv-error-metric}
\mathrm{Scale\text{-}Invariant\ Error}
= \frac{1}{M}
  \sum_{i=1}^M
  \Bigl[
    \log(\hat{d}_i)
    - \log(d_i)
    + \frac{1}{M}
      \sum_{j=1}^M
      \bigl(
        \log(d_j)
        - \log(\hat{d}_j)
      \bigr)
  \Bigr]^2.
\end{equation}

\paragraph{Interpretation:}
\begin{itemize}
    \item \textbf{Relative Depth Emphasis:}
    Penalizes shape mismatches rather than absolute scale. Especially apt for cross-dataset or relative depth tasks.
    \item \textbf{Drawback:}
    Omits large global scale errors if uniform scaling is off, so some applications needing exact metric consistency prefer absolute-based metrics.
\end{itemize}

\paragraph{Conclusion on MDE Metrics:}
Each metric captures different aspects: \(\mathrm{AbsRel}\) or \(\mathrm{RMSE}\) stress absolute residuals, \(\mathrm{RMSE(log)}\) and \(\mathrm{MeanLog10}\) focus on multiplicative or log-space errors, \(\delta\) thresholds reveal how many pixels lie within a given factor, \(\mathrm{WHDR}\) handles purely ordinal data, and \(\mathrm{Scale\text{-}Invariant Error}\) emphasizes relative shape. In practice, researchers often report multiple metrics to provide a comprehensive assessment of the depth estimation performance of a model.

\section{Image Generation}
\label{subsec:gen_models}

Image generation in deep learning involves using artificial neural networks to generate new images. This task has progressed significantly with models such as Variational Autoencoders (VAEs) \cite{kingma2013auto,bowman2016,an2015,zhao2017-VA}, Generative Adversarial Networks (GANs) \cite{goodfellow2020generative,goodfellow2014,yu2017,isola2017,yang2017}, Normalized Flow models (NFs) \cite{dinh2016density,kingma2018glow,grathwohl2018ffjord,rezende2015,rezende2015b}, Energy-Based Models (EBMs) \cite{lecun2006tutorial,du2019implicit,hinton2006,zhao2015,belanger2016,song2019}, and Diffusion Models \cite{sohl2015deep,ho2020denoising,grathwohl2020,sohl2015-DM,song2021}. These models can generate high-quality images that can be used in various applications such as image super-resolution \cite{park2003super,dong2014,ledig2017,zhang2018-ISR}, denoising \cite{tian2020deep,xie2012,lehtinen2018}, inpainting \cite{bertalmio2000image,yu2018,pathak2016,liu2018-II}, and style transfer \cite{gatys2015neural,johnson2016,huang2017-ST,chen2018-ST}.

\emph{Variational Autoencoders (VAEs)} are generative models that use deep learning techniques to create new data and learn latent representations of the input data. They consist of an encoder and a decoder. The encoder compresses input data into a lower-dimensional latent space, while the decoder reconstructs the original data from points in the latent space. The process is trained to minimize the difference between original and reconstructed data and ensure the latent space approximates a standard Gaussian distribution. VAEs can generate new data by feeding the decoder points sampled from the latent space.

\emph{Generate Adversarial Networks (GANs)} involve two neural networks, a Generator, and a Discriminator, playing a game against each other. The Generator tries to create data similar to the training data, while the Discriminator tries to distinguish between the real and generated data. Through this process, both networks improve: the Generator learns to produce increasingly realistic data, while the Discriminator becomes better at distinguishing between real and artificial data. This adversarial process continues until an equilibrium is reached, at which point the Generator is producing realistic data and the Discriminator is, at best, randomly guessing whether the data is real or generated. This equilibrium is conceptually referred to as a Nash equilibrium in game theory \cite{nash1996non}. 

\emph{Normalizing Flows} use invertible transformations to generate diverse outputs and provide an exact and tractable likelihood for a given sample, enabling efficient sampling and density estimation. However, they can be computationally intensive to train.

\emph{Energy-Based Models (EBMs)} learn a scalar energy function to distinguish real data points from unlikely ones. A neural network often parameterizes this function and learns from the data. Sampling in EBMs is typically done via Markov Chain Monte Carlo (MCMC) methods. EBMs can represent a wide variety of data distributions but can be challenging to train due to the intractability of the partition function and the computational expense of MCMC sampling.

\emph{Diffusion models} use a random process to transform simple data distribution, like Gaussian noise, into the desired target distribution. This is controlled by a trained neural network, allowing the generation of high-quality data, like images, through a smooth transition from noise to the desired data.

The following sections will review the common lost functions and performance metrics used for image generation.

\subsection{Image Generation Loss Functions}
\label{subsec:img_gen_losses}

The loss function in a VAE consists of the reconstruction loss and the Kullback-Leibler Divergence Loss (KL). The reconstruction loss measures how well the decoded data matches the original input data. The KL divergence measures how much the learned distribution in the latent space deviates from a target distribution, usually a standard normal distribution. KL-divergence is used as a regularization term to ensure that the distributions produced by the encoder remain close to a unit Gaussian, penalizing the model if the learned distributions depart from it. 

The most common loss function used in GANs is the adversarial loss, which is the sum of the cross-entropy loss between the generator's predictions and the real or fake labels. Later, WGAN \cite{arjovsky2017wasserstein} applied the Wasserstein distance as an alternative to training GANs to improve stability and avoid mode collapse that occurs when the generator network stops learning the underlying data distribution and begins to produce a limited variety of outputs, rather than a diverse range of outputs that accurately represent the true data distribution.

Normalizing Flows are typically trained using maximum likelihood estimation. Given a dataset, the aim is to maximize the log-likelihood of the data under the model by minimizing the negative log-likelihood of the data. 

During training, Energy-based models (EBMs) minimize a loss function that encourages the energy function to assign lower energy values to data points from the training data and higher energy values to other points. Different types of EBMs use different loss functions, such as Contrastive Divergence (CD)\cite{carreira2005contrastive}, Maximum Likelihood Estimation (MLE)\cite{myung2003tutorial}, and Noise-Contrastive Estimation (NCE) \cite{gutmann2010noise}.  

Diffusion models use a denoising loss function based on the Mean Absolute Error (MAE) or the Mean Squared Error (MSE) between the original and reconstructed data.

Table \ref{tab:guidelines-imggen-losses} offers an in-depth examination of popular loss functions in image generation, highlighting their applications, data attributes, benefits, and drawbacks.

In the following sections, we describe in detail each of these losses.

\begin{table*}[ht!]
\caption{Guidelines for selecting a loss function for image generation, based on usage, data characteristics, advantages and limitations.}
\label{tab:guidelines-imggen-losses}
\centering
\footnotesize
\begin{tabular}{p{1.3cm}p{2.8cm}p{2.8cm}p{2.8cm}p{2.8cm}}
\toprule
\textbf{Function} 
& \textbf{Usage} 
& \textbf{Data Characteristics} 
& \textbf{Advantages} 
& \textbf{Limitations} \\
\midrule

MSE 
& Autoencoders \newline Continuous image reconstruction \newline Image restoration 
& Real-valued intensities \newline Typically unimodal 
& Stable training \newline Easy to implement 
& Can produce overly smooth outputs \newline Sensitive to outliers \\
\midrule

BCE 
& Binary image generation \newline Reconstruction tasks with binary data 
& Binary or probabilistic pixel outputs \newline Pixel-wise classification 
& Direct probability interpretation \newline Common in classification-based setups 
& May saturate gradients \newline Slow convergence if predictions are poor \\
\midrule

KL Divergence 
& Variational autoencoders \newline Distribution matching in latent space 
& Known or approximated target distribution \newline Typically continuous latent variables 
& Aligns predicted and target distributions \newline Popular in generative modeling 
& Highly sensitive to zero probabilities \newline Can be numerically unstable \\
\midrule

Perceptual
& Super-resolution, style transfer, \newline advanced image generation tasks
& Real-valued or color images \newline Requires pretrained feature extractor (e.g., VGG)
& Preserves semantic details \newline Less blurring than pixel-based losses \newline Aligned with human perception
& Depends on pretrained models \newline Domain mismatch can degrade features \newline Ignores exact pixel-level fidelity \\
\midrule

Adversarial Loss 
& GAN-based image generation \newline Domain adaptation for realism 
& Unlabeled data \newline Real vs. synthetic distributions 
& Generates high-fidelity images \newline Encourages realistic outputs 
& Prone to mode collapse \newline Training can be unstable \\
\midrule

Wasserstein 
& WGANs \newline Stable image generation 
& Continuous or discrete data \newline Flexible for various domains 
& More stable training \newline Less mode collapse than standard GANs 
& Requires Lipschitz constraints \newline Involves extra gradient penalty steps \\
\midrule

Negative Log-likelihood (NFs)
& Normalizing flows \newline Exact likelihood-based generation 
& Invertible transformations \newline Tractable Jacobian determinant 
& Provides exact density estimates \newline Flexible architectures 
& High computational cost \newline Memory-intensive for large images \\
\midrule

Contrastive Divergence 
& Energy-based image models \newline Restricted Boltzmann Machines 
& High-dimensional image data \newline Continuous or discrete inputs 
& Approximate training of intractable likelihood \newline Generally more efficient than full MCMC 
& Biased gradient estimates \newline May need multiple sampling steps \\
\bottomrule
\end{tabular}
\end{table*}

\subsubsection{Reconstruction Loss}
\label{loss:reconstruction}

\emph{Reconstruction loss} measures how faithfully a model reproduces the input image (or data) in tasks such as Variational Autoencoders (VAEs), image restoration, and generative image synthesis. Its primary objective is to penalize discrepancies between the original input \(\mathbf{x}\) and the reconstructed output \(\hat{\mathbf{x}}\), driving the model to learn representations that preserve key image features.

\paragraph{Mean Squared Error (MSE):}
For continuous-valued image data, a common form of reconstruction loss is \emph{Mean Squared Error} (MSE). Let \(\{\mathbf{x}_i\}_{i=1}^N\) be the original images and \(\{\hat{\mathbf{x}}_i\}_{i=1}^N\) be their reconstructions. The MSE loss is:

\begin{equation}
\label{eq:rec_loss}
\mathcal{L}_{\mathrm{recon}}
= \frac{1}{N} \sum_{i=1}^N 
  \|\mathbf{x}_i - \hat{\mathbf{x}}_i\|_2^2,
\end{equation}

\noindent
where \(\|\cdot\|_2\) denotes the Euclidean norm. MSE directly measures pixel-level squared differences, making it suitable for continuous image intensities. However, it may over-penalize small local shifts (e.g., slight misalignments in edges or textures).

\paragraph{Binary Cross-Entropy (BCE):}
For binary or probabilistic image outputs, \emph{Binary Cross-Entropy} (BCE) is often preferred. Let \(\mathbf{y}_i\in\{0,1\}^D\) be a binary ground-truth vector and \(\hat{\mathbf{y}}_i\in[0,1]^D\) the model’s predicted probabilities. Then the BCE loss is:

\begin{equation}
\label{eq:bce_recloss}
\mathcal{L}_{\mathrm{BCE}}
= - \frac{1}{N}\sum_{i=1}^{N}
    \sum_{j=1}^D
    \Bigl[
      y_{i,j}\,\log(\hat{y}_{i,j})
      + (1 - y_{i,j})\,\log\bigl(1 - \hat{y}_{i,j}\bigr)
    \Bigr],
\end{equation}

\noindent
where \(D\) is the dimensionality (e.g., number of pixels). BCE measures the divergence between the true binary distribution and the predicted Bernoulli outputs, making it a natural choice for tasks involving binarized images or probabilistic pixel maps.

\paragraph{Usage in Image Generation:}
\begin{enumerate}
    \item \textbf{Variational Autoencoders (VAEs):}
    Reconstruction loss is combined with a Kullback–Leibler (KL) divergence term to ensure the latent code follows a chosen prior distribution. Minimizing MSE or BCE encourages faithful image reconstructions while the KL term regularizes the latent space.
    \item \textbf{Image Restoration:}
    In denoising, super-resolution, and inpainting, reconstruction loss enforces that the model output remains close to the pristine target image, recovering lost detail or removing artifacts.
    \item \textbf{Generative Models:}
    Models like \(\beta\)-VAEs or certain GAN variations use reconstruction losses to keep outputs close to real data distributions. This includes reconstructing inputs or partially corrupted images.
    \item \textbf{Anomaly Detection:}
    Reconstruction loss can highlight anomalous patterns if a model trained on “normal” images produces large residuals for out-of-distribution samples.
\end{enumerate}

\paragraph{Limitations and Extensions:}
\begin{itemize}
    \item \textbf{Overly Smooth Outputs:}
    MSE can induce blurring or loss of sharp details. Perceptual losses or adversarial terms (\S\ref{loss:adversarial}) may be added to enhance realism.
    \item \textbf{Limited Structural Awareness:}
    Pixel-wise metrics do not explicitly account for high-level image features or semantics. SSIM-based terms (\S\ref{loss:SSIM}) or perceptual losses (\S\ref{loss:perceptual}) can improve structural fidelity.
    \item \textbf{Adaptive Weighting:}
    Some tasks require balancing reconstruction fidelity with constraints on latent space size, style consistency, or domain adaptation. Hence, $\lambda_{\mathrm{recon}}$ may be tuned in multi-objective loss setups.
\end{itemize}

In general, reconstruction loss remains an essential element of image generation and restoration tasks, providing a direct measure of fidelity between the original and the reconstructed images.

\subsubsection{Kullback--Leibler Divergence Loss}
\label{loss:kl-loss}

\emph{Kullback--Leibler (KL) Divergence} \cite{Csiszar_KL} measures how one probability distribution \(q\) diverges from a target distribution \(p\). In a supervised setting, \(p(\mathbf{x})\) often represents the true or “target” probabilities over classes, while \(q(\mathbf{x})\) is the model’s predicted distribution. Let \(x_1, x_2, \dots, x_n\) be the outcomes or class labels. The KL divergence is defined as:

\begin{equation}
\label{eq:kl}
\mathrm{KL}\bigl(p\,\|\,q\bigr)
= \sum_{i=1}^n 
  p(x_i)\,
  \log
    \!\Bigl(
      \frac{p(x_i)}{\,q(x_i)\,}
    \Bigr).
\end{equation}

\paragraph{Interpretation:}
\begin{itemize}
    \item \textbf{Distribution Matching:} 
    KL divergence quantifies how many extra bits are needed to encode samples from \(p\) using \(q\). Minimizing KL enforces that \(q(x_i)\approx p(x_i)\) for all \(i\).
    \item \textbf{Asymmetric Measure:} 
    \(\mathrm{KL}(p\|q)\) is not the same as \(\mathrm{KL}(q\|p)\). The choice of which distribution is in the numerator vs.\ denominator can greatly affect optimization behavior, especially near zero probabilities.
\end{itemize}

\paragraph{Usage in Generative Models:}
\begin{itemize}
    \item \textbf{Variational Autoencoders (VAEs):} 
    VAEs minimize \(\mathrm{KL}(q_\phi(\mathbf{z}|\mathbf{x})\,\|\,p_\theta(\mathbf{z}|\mathbf{x}))\) or \(\mathrm{KL}(q_\phi(\mathbf{z}|\mathbf{x})\,\|\,p(\mathbf{z}))\) in the ELBO objective, guiding the learned latent distribution \(q_\phi(\mathbf{z}|\mathbf{x})\) to match a prior \(p(\mathbf{z})\).
    \item \textbf{Reinforcement Learning:} 
    KL divergence is used to keep an agent’s updated policy close to a reference or previous policy (e.g., TRPO, PPO). Minimizing \(\mathrm{KL}(\pi_{\mathrm{new}}\|\pi_{\mathrm{old}})\) or vice versa ensures stable policy updates.
\end{itemize}

\paragraph{Sensitivity to Zero Probabilities:}
\begin{itemize}
    \item \textbf{Numerical Instability:}
    When \(p(x_i)=0\) but \(q(x_i)>0\) (or vice versa), the KL term can be undefined or infinite. This demands care in initial probability assignments or smoothing.
    \item \textbf{Practical Remedy:}
    A small constant \(\varepsilon\) (e.g., \(10^{-7}\)) is often added to both \(p\) and \(q\) to avoid taking the log of zero.
\end{itemize}

\paragraph{Comparison with Other Divergences:}
\begin{itemize}
    \item \textbf{Forward vs.\ Reverse KL:}
    Minimizing \(\mathrm{KL}(p\|q)\) differs from \(\mathrm{KL}(q\|p)\). For instance, \(\mathrm{KL}(p\|q)\) is more sensitive if \(p(x_i)\) is significantly non-zero but \(q(x_i)\approx0\).
    \item \textbf{Alternative Metrics:}
    In some scenarios, Jensen–Shannon divergence (symmetric version) or Wasserstein distance (\S\ref{loss:wasserstein}) can be more stable or yield different optimization trajectories, especially in GAN training.
\end{itemize}

By matching the model’s predicted distribution \(q(x_i)\) to the true (or target) distribution \(p(x_i)\), KL divergence loss fosters distribution-level correctness, which is especially critical in generative modeling contexts (e.g., VAEs) and policy optimization in reinforcement learning. However, one must handle zero-probability events carefully to avoid numerical instability or poor gradient signals.

\subsubsection{Perceptual Loss}
\label{loss:perceptual}

\emph{Perceptual Loss}, sometimes referred to as \emph{feature reconstruction loss} or \emph{VGG-based loss}, measures the discrepancy between high-level feature representations of the original and generated images rather than their raw pixel-level differences \cite{johnson2016perceptual,bruna2015super,simonyan2014very}. By comparing deep network activations (e.g., from a pretrained classification network), Perceptual Loss aims to ensure that the reconstructed image \(\hat{\mathbf{x}}\) not only matches \(\mathbf{x}\) in low-level details but also captures semantically important features and textures.

\paragraph{Motivation:}
Pixel-wise metrics such as MSE (\S\ref{loss:reconstruction}) often produce blurred or less realistic outputs, as they overemphasize exact local correspondences. Perceptual Loss mitigates this issue by forcing the generated image to have similar \emph{feature activations} in a deep CNN’s intermediate layers, leading to sharper and more visually coherent results.

\paragraph{Formulation:}
Let \(\phi(\cdot)\) be a pretrained network (often a classification CNN such as VGG-16 \cite{simonyan2014very}), and \(\phi_\ell(\mathbf{x})\in \mathbb{R}^{C\times H\times W}\) be its \(\ell\)-th layer activations when input \(\mathbf{x}\) is forwarded. Then the \emph{Perceptual Loss} between an original image \(\mathbf{x}\) and a reconstructed image \(\hat{\mathbf{x}}\) is computed as the \(\ell_2\)-distance between their feature activations:

\begin{equation}
\label{eq:perceptual}
\mathcal{L}_{\mathrm{perc}}(\mathbf{x}, \hat{\mathbf{x}})
= \sum_{\ell \in \mathcal{L}}
  \frac{1}{C_\ell\,H_\ell\,W_\ell}
  \bigl\|\,
    \phi_\ell(\mathbf{x})
    \;-\;
    \phi_\ell(\hat{\mathbf{x}})
  \bigr\|_2^2,
\end{equation}

\noindent
where \(\mathcal{L}\) is the set of chosen layers, and \(C_\ell, H_\ell, W_\ell\) are the channel, height, and width dimensions of layer \(\ell\). The normalization factor \(\tfrac{1}{C_\ell\,H_\ell\,W_\ell}\) ensures consistent scale across layers.

\paragraph{Usage in Image Generation:}
\begin{itemize}
    \item \textbf{Super-Resolution and Deblurring}: 
    Perceptual Loss helps recover sharp edges and fine structures that pixel-wise errors tend to oversmooth \cite{ledig2017photo}.
    \item \textbf{Style Transfer}: 
    By matching feature activations of style and content images, models achieve visually pleasing stylization \cite{gatys2015neural,johnson2016perceptual}.
    \item \textbf{GANs and VAEs}: 
    Combining Perceptual Loss with adversarial or reconstruction objectives yields images with globally and semantically coherent details \cite{dosovitskiy2016generating,isola2017image}.
\end{itemize}

\paragraph{Advantages and Limitations:}
\begin{itemize}
    \item \textbf{Preserving High-Level Semantics}: 
    Focuses on how the generated image appears to a deep network, correlating better with human perception of sharpness and structure.
    \item \textbf{Decreased Local Fidelity}: 
    It does not guarantee pixel-perfect alignment and might overlook small color mismatches if they do not significantly alter high-level features.
    \item \textbf{Dependence on Pretrained Networks}: 
    Typically, \(\phi\) is a CNN trained on large image datasets (e.g., ImageNet), introducing a domain bias. If the target domain differs from the CNN’s training distribution, perceived “feature relevance” might be suboptimal.
\end{itemize}

\paragraph{Extensions:}
\begin{itemize}
    \item \textbf{Layer Selection}: 
    Perceptual Loss can combine activations from multiple layers to capture both coarse and fine-level attributes.
    \item \textbf{Style Loss (Gram Matrices)}: 
    In style transfer or artistic generation, one may also penalize discrepancies in correlation statistics (Gram matrices) of layer features, capturing “texture” alignment \cite{gatys2015neural}.
    \item \textbf{Multiscale Variants}: 
    Computing Perceptual Loss at different spatial resolutions can enhance stability and capture multi-level semantics.
\end{itemize}

Perceptual Loss thus balances high-level structural resemblance with a more “human-like” notion of image quality, making it a staple in advanced image generation, super-resolution, and style transfer pipelines. While it lacks pixel-accuracy guarantees, its synergy with low-level reconstruction terms or adversarial objectives often yields outputs that are both visually appealing and domain-faithful.

\subsubsection{Adversarial Loss}
\label{loss:adversarial}

\emph{Adversarial Loss} is foundational to Generative Adversarial Networks (GANs), introduced by Goodfellow~\emph{et~al.} \cite{goodfellow2020generative}. It arises from a minimax game between two networks: a \emph{generator} \(G\) (which maps noise vectors to synthesized data) and a \emph{discriminator} \(D\) (which classifies inputs as real or generated). By training \(G\) to produce samples indistinguishable from real data, and simultaneously training \(D\) to distinguish real from generated, this adversarial setup propels the generator’s outputs toward realism.

\paragraph{Basic Formulation:}
Assume \(\mathbf{z}\sim p_z(\mathbf{z})\) is noise drawn from a latent distribution, and \(\mathbf{x}\sim p_{\mathrm{data}}(\mathbf{x})\) is a real-data sample from the target distribution. The \emph{Adversarial Loss} (in its non-saturated version) can be stated as:

\begin{equation}
\label{eq:adver_loss}
\begin{split}
\mathcal{L}_{\mathrm{adv}}(G,D)
&= -\,\mathbb{E}_{\mathbf{x}\sim p_{\mathrm{data}}} \bigl[\log D(\mathbf{x})\bigr] \\
&\quad -\,\mathbb{E}_{\mathbf{z}\sim p_z} \bigl[\log\!\bigl(1 - D\bigl(G(\mathbf{z})\bigr)\bigr)\bigr].
\end{split}
\end{equation}

Here,
\begin{itemize}
    \item \(D(\mathbf{x})\) outputs the probability that \(\mathbf{x}\) is a real sample.
    \item \(G(\mathbf{z})\) generates a synthetic sample from noise \(\mathbf{z}\).
\end{itemize}

GAN training is often viewed as a two-player minimax game. The discriminator \(D\) maximizes:

\begin{equation}
\label{eq:discrim}
\max_{D}
\Bigl\{
  \mathbb{E}_{\mathbf{x}\sim p_{\mathrm{data}}}[\log D(\mathbf{x})]
  + \mathbb{E}_{\mathbf{z}\sim p_z}[\log\bigl(1 - D(G(\mathbf{z}))\bigr)]
\Bigr\}.
\end{equation}

Meanwhile, the generator \(G\) minimizes:

\begin{equation}
\label{eq:generator}
\min_{G}
\mathbb{E}_{\mathbf{z}\sim p_z}
  \bigl[
    \log \bigl(1 - D(G(\mathbf{z}))\bigr)
  \bigr].
\end{equation}

Alternatively, one may train the generator by \(\max_{G}\,\mathbb{E}_{\mathbf{z}\sim p_z}[\log D(G(\mathbf{z}))]\) to help mitigate gradient saturation issues when \(D\) is strong.

The training procedure alternates between:
\begin{enumerate}
    \item \textbf{Discriminator Update:}  
    Holding \(G\) fixed, optimize \(D\) to distinguish real \(\mathbf{x}\) vs.\ fake \(G(\mathbf{z})\).
    \item \textbf{Generator Update:}  
    Holding \(D\) fixed, optimize \(G\) to produce samples that fool \(D\).
\end{enumerate}
Over time, \(G\) improves at creating realistic samples, while \(D\) refines its discrimination boundary. Under ideal conditions, this converges to a \emph{Nash equilibrium} where \(G\)'s outputs are indistinguishable from real data.

\paragraph{Advantages and Challenges:}
\begin{itemize}
    \item \textbf{High-Fidelity Generation:}
    GANs can learn complex, high-dimensional data distributions, yielding photorealistic images in tasks like image synthesis, inpainting, or super-resolution.
    \item \textbf{Mode Collapse:}
    The generator may collapse to a limited subset of modes, producing less diverse outputs. Techniques like Wasserstein distance (\S\ref{loss:wasserstein}) or multi-objective strategies mitigate this.
    \item \textbf{Training Instability:}
    Adversarial objectives can be sensitive to hyperparameters and require careful balancing of \(G\) and \(D\) capacities. Convergence is not guaranteed but can be stabilized via gradient penalties or spectral normalization.
\end{itemize}

\paragraph{Extensions:}
\begin{itemize}
    \item \textbf{Wasserstein GANs (WGANs):}
    Replace KL/JS divergences with Earth Mover’s distance for more stable training (\S\ref{loss:wasserstein}).
    \item \textbf{Conditional GANs:}
    Incorporate class labels or other conditions into \(G\) and \(D\), enabling class-specific image synthesis or domain adaptation.
    \item \textbf{CycleGAN, StyleGAN, etc.:}
    Specialized architectures leverage adversarial loss for unpaired image-to-image translation, style generation, and more.
\end{itemize}

Adversarial Loss underpins GAN-based approaches by framing generation as a competition between generating realistic samples and detecting fakes. Despite challenges in training stability, it remains one of the most powerful tools in image generation, pushing the state of the art in visual realism and diversity.

\subsubsection{Wasserstein Loss}
\label{loss:wasserstein}

\emph{Wasserstein Loss}, introduced by Arjovsky~\emph{et~al.} \cite{arjovsky2017wasserstein}, provides an alternative objective to standard adversarial loss in Generative Adversarial Networks (GANs). Rather than directly matching distributions through JS or KL divergences, Wasserstein Loss employs the \emph{Earth Mover's Distance} (EMD), offering smoother, more stable gradients when learning to transform a generator's distribution \(p_g\) toward the real data distribution \(p_d\).

\paragraph{Definition:}
Let \(p_d(\mathbf{x})\) be the distribution of real data and \(p_g(\mathbf{x})\) be the distribution of generated data. The \emph{Wasserstein-1} distance (\(\mathrm{W}\)) between these two distributions can be expressed as an optimal transport problem:

\begin{equation}
\label{eq:wasser}
\mathrm{W}\bigl(p_d,\,p_g\bigr)
= \inf_{\gamma \in \Gamma(p_d, p_g)} 
  \int_{\mathbf{x},\mathbf{y}}
  \lVert \mathbf{x} - \mathbf{y}\rVert
  \;d\gamma(\mathbf{x},\mathbf{y}),
\end{equation}

\noindent
where
\begin{itemize}
    \item \(\Gamma(p_d, p_g)\) is the set of all joint distributions \(\gamma(\mathbf{x},\mathbf{y})\) whose marginals are \(p_d\) and \(p_g\).
    \item \(\lVert \cdot \rVert\) typically denotes the Euclidean distance.
    \item \(\gamma(\mathbf{x},\mathbf{y})\) indicates how “mass” is transported from \(\mathbf{y}\) in distribution \(p_g\) to \(\mathbf{x}\) in distribution \(p_d\).
\end{itemize}

The integral represents the minimum “cost” to morph one distribution into the other, where cost is the distance \(\|\mathbf{x} - \mathbf{y}\|\) weighted by \(\gamma(\mathbf{x},\mathbf{y})\).

In practice, one implements \(\mathrm{W}(p_d, p_g)\) in GAN training by defining a critic (akin to a discriminator) that approximates the Wasserstein distance through its Lipschitz-constrained outputs \cite{arjovsky2017wasserstein}. Formally, for a function \(f_\theta\) constrained to be 1-Lipschitz,

\begin{equation}
\mathrm{W}\bigl(p_d, p_g\bigr)
\approx \max_{\theta,\,\|f_\theta\|\leq 1}
\bigl(
   \mathbb{E}_{\mathbf{x}\sim p_d} [f_\theta(\mathbf{x})]
  -\mathbb{E}_{\mathbf{z}\sim p_z} [f_\theta(G(\mathbf{z}))]
\bigr).
\end{equation}

Hence the \emph{Wasserstein Loss} for the generator and critic can be written similarly to:

\begin{equation}
\mathcal{L}_{\mathrm{critic}}
= \mathbb{E}_{\mathbf{x}\sim p_d}[f_\theta(\mathbf{x})]
 - \mathbb{E}_{\mathbf{z}\sim p_z}[f_\theta(G(\mathbf{z}))],
\end{equation}
\begin{equation}
\mathcal{L}_{\mathrm{gen}}
= -\mathbb{E}_{\mathbf{z}\sim p_z}[f_\theta(G(\mathbf{z}))].
\end{equation}

\paragraph{Advantages Over Classic GAN Losses:}
\begin{itemize}
    \item \textbf{Smooth Gradient Signals:}
    Wasserstein distance yields non-saturating gradients even when distributions have disjoint supports, reducing training instability.
    \item \textbf{Mitigates Mode Collapse:}
    The generator is penalized proportionally to how far $p_g$ is from covering $p_d$, rather than simply whether samples look real or fake.
    \item \textbf{Stable Convergence:}
    Empirically, WGAN training can exhibit reduced mode collapse and more robust progression, though hyperparameters and Lipschitz constraints remain critical.
\end{itemize}

\paragraph{Lipschitz Constraint and WGAN-GP:}
To ensure \(\|f_\theta\|\le 1\), Arjovsky~\emph{et~al.} proposed weight clipping, but Gulrajani~\emph{et~al.} \cite{gulrajani2017improved} introduced a \emph{gradient penalty} (GP) approach, often referred to as WGAN-GP. This enforces the 1-Lipschitz condition by penalizing deviations from \(\|\nabla f_\theta\|\!=1\), improving training stability and removing the discrete clamp on weights.

\paragraph{Limitations:}
\begin{itemize}
    \item \textbf{Lipschitz Enforcement Overhead:}
    Gradient penalties or clipping can add complexity and slow training.
    \item \textbf{Hyperparameter Sensitivity:}
    WGAN still requires careful tuning (penalty coefficients, learning rates) to fully reap its benefits of stable training.
\end{itemize}

\subsubsection{Negative Log-Likelihood in Normalizing Flows}
\label{loss:nll_flows}

\emph{Normalizing Flows} form a class of generative models where a bijective and differentiable transformation \(f_{\theta}\) maps data \(\mathbf{x}\) from a complex distribution \(p_{\theta}(\mathbf{x})\) to a simpler, tractable base distribution \(p_{z}(\mathbf{z})\) (often an isotropic Gaussian). Let \(\mathbf{z} = f_{\theta}(\mathbf{x})\) denote the forward transformation, implying \(\mathbf{x} = f_{\theta}^{-1}(\mathbf{z})\). Because \(f_{\theta}\) is invertible, the model can compute an \emph{exact} likelihood for data \(\mathbf{x}\) via the change of variables formula.

\paragraph{Log-likelihood Computation:}
For a single data point \(\mathbf{x}\), the \emph{log-likelihood} under the flow model \(p_{\theta}(\mathbf{x})\) is given by

\begin{equation}
\label{eq:nll_norm_flows}
\log p_{\theta}(\mathbf{x})
= \log p_{z}\bigl(f_{\theta}(\mathbf{x})\bigr)
  + \log\!\Bigl\lvert 
      \det 
        \Bigl(
          \frac{\partial f_{\theta}(\mathbf{x})}{\partial \mathbf{x}}
        \Bigr)
    \Bigr\rvert,
\end{equation}

\noindent
where
\begin{itemize}
    \item \(\mathbf{z} = f_{\theta}(\mathbf{x}) \in \mathbb{R}^d\) is the latent representation, distributed according to a simple prior \(p_z(\mathbf{z})\) (e.g., \(\mathcal{N}(\mathbf{0},\mathbf{I})\)).
    \item \(\det \bigl(\tfrac{\partial f_{\theta}(\mathbf{x})}{\partial \mathbf{x}}\bigr)\) is the Jacobian determinant of \(f_{\theta}\) evaluated at \(\mathbf{x}\). This term accounts for the volume change induced by the mapping.
\end{itemize}

\paragraph{Training via Negative Log-likelihood:}
To learn the parameters \(\theta\) of the flow, one maximizes the likelihood of a dataset \(\{\mathbf{x}_i\}_{i=1}^N\) or equivalently minimizes its negative log-likelihood. Thus, the loss function is:

\begin{equation}
\label{eq:nll_norm_flows2}
\begin{split}
\mathcal{L}(\theta)
&= -\frac{1}{N}\sum_{i=1}^{N}
    \log p_{\theta}(\mathbf{x}_i) \\
&= -\frac{1}{N}\sum_{i=1}^{N}
    \Bigl[
      \log p_{z}\bigl(f_{\theta}(\mathbf{x}_i)\bigr)
      + \log\!\bigl\lvert
         \det \frac{\partial f_{\theta}(\mathbf{x}_i)}{\partial \mathbf{x}_i}
        \bigr\rvert
    \Bigr].
\end{split}
\end{equation}

\noindent
Minimizing \(\mathcal{L}(\theta)\) via stochastic gradient methods encourages \(f_{\theta}\) to transform real samples \(\mathbf{x}_i\) into plausible latent codes \(\mathbf{z}\) that align with the base distribution, while accurately capturing the data distribution’s structure.

A key challenge is computing and differentiating through the Jacobian determinant \(\det(\partial f_{\theta}/\partial \mathbf{x})\). To keep this tractable, Normalizing Flow architectures, such as RealNVP \cite{dinh2016density} and Glow \cite{kingma2018glow}, design invertible layers (e.g., \emph{affine coupling}, \emph{invertible \(1\times 1\) convolutions}) that yield either a block triangular Jacobian or otherwise simplify the determinant calculation. Thus, the log-determinant term remains efficiently computable and backprop-friendly.

\paragraph{Advantages and Limitations:}
\begin{itemize}
    \item \textbf{Exact Likelihood}: 
    Unlike typical GANs (\S\ref{loss:wasserstein}) or VAEs that approximate likelihood, Normalizing Flows give an exact $\log p_{\theta}(\mathbf{x})$.
    \item \textbf{Flexible Distributions}: 
    Coupling-based transformations allow for complex multimodal distributions if stacked sufficiently.
    \item \textbf{Computational Cost}: 
    The necessity of computing and differentiating \(\log|\det(\cdot)|\) can be memory-intensive. Flow-based models are often larger and slower to train than latent-variable methods with approximate posteriors (e.g., VAEs).
\end{itemize}

Hence, \emph{Negative Log-Likelihood} in Normalizing Flows integrates invertible transformations with tractable determinant structures, offering exact density estimation and generative sampling. Despite heavier computational demands, this approach is highly appealing for tasks demanding precise likelihood evaluations.

\subsubsection{Contrastive Divergence}
\label{loss:contr_div}

\emph{Contrastive Divergence} (CD) \cite{hinton2002training} is a method for approximately maximizing the log-likelihood of an Energy-Based Model (EBM), such as a Restricted Boltzmann Machine (RBM), when direct computation of the gradient is intractable. Rather than sampling from the model distribution until equilibrium, CD relies on short Markov Chain Monte Carlo (MCMC) runs, starting from real data samples, to estimate the negative phase gradient.

\paragraph{Energy-Based Models and Log-likelihood:}
Let \(\{x_i\}_{i=1}^N\) be a dataset of \(N\) points. An EBM parameterized by \(\theta\) defines an energy function \(E(x;\theta)\), so that the model’s density is
\begin{equation}
p(x;\theta)
= \frac{\exp\bigl(-E(x;\theta)\bigr)}{Z(\theta)},
\end{equation}
where 
\begin{equation}
Z(\theta)
= \int \exp\bigl(-E(x;\theta)\bigr)\,dx
\end{equation}
is the partition function. The \emph{log-likelihood} of the data is

\begin{equation}
\label{eq:cont_div}
L(\theta)
= \frac{1}{N}\sum_{i=1}^N
  \log p\bigl(x_i;\theta\bigr).
\end{equation}

\noindent
Maximizing \(L(\theta)\) is equivalent to minimizing \(-L(\theta)\). However, computing \(\nabla_\theta L(\theta)\) exactly involves an expectation under \(p(x;\theta)\), which is typically intractable.

\paragraph{Positive and Negative Phases:}
\begin{itemize}
    \item \textbf{Positive Phase:}
    Computed from the data distribution directly. For data samples \(\{x_i\}\), the gradient term is:

    \begin{equation}
    \label{eq:cont_div3}
    \frac{1}{N}\sum_{i=1}^{N}
      \frac{\partial E(x_{i};\theta)}{\partial \theta}.
    \end{equation}

    \item \textbf{Negative Phase:}
    Involves an expectation under the model’s own distribution, 
    \begin{equation}
      -\,\Bigl\langle
         \frac{\partial E(x;\theta)}{\partial \theta}
        \Bigr\rangle_{p(x;\theta)}.
    \end{equation}
    Since exact sampling from \(p(x;\theta)\) is expensive, Contrastive Divergence approximates this expectation via a short MCMC chain (e.g., \(k\) steps of Gibbs sampling) starting from the data point \(x_i\). Let \(x'_i\) be the sample drawn after these \(k\) steps. Then the negative phase gradient is estimated by:

    \begin{equation}
    \label{eq:cont_div4}
    -\,\Bigl\langle
       \frac{\partial E(x';\theta)}{\partial \theta}
      \Bigr\rangle_{\mathrm{CD}},
    \end{equation}

    where \(\mathrm{CD}\) indicates the distribution induced by a short chain from data points.
\end{itemize}

\paragraph{Parameter Update (CD-$k$):}
Putting the two phases together, the update rule for \(\theta\) after observing \(\{x_i\}\) is:

\begin{equation}
\label{eq:cont_div5}
\Delta \theta
= \eta
  \Bigl[
    \frac{1}{N}\sum_{i=1}^N
      \frac{\partial E(x_{i};\theta)}{\partial \theta}
    \;-\;
    \Bigl\langle
      \frac{\partial E(x';\theta)}{\partial \theta}
    \Bigr\rangle_{\mathrm{CD}}
  \Bigr],
\end{equation}

\noindent
where \(\eta\) is the learning rate, and \(x'_i\) are samples obtained from a short chain (e.g., \(k\) steps of Gibbs sampling) starting at the real data \(x_i\).

\paragraph{Advantages and Limitations:}
\begin{itemize}
    \item \textbf{Computational Efficiency:}
    CD-$k$ requires fewer MCMC steps than full equilibrium sampling, enabling faster training than exact gradient methods.
    \item \textbf{Bias:}
    Because the negative phase is only partially explored with a short chain, the gradient is a biased approximation of the true log-likelihood gradient. Increasing \(k\) can reduce this bias but raises computational cost.
    \item \textbf{Scope of Use:}
    Especially prominent in RBMs, DBMs, or other EBMs with local or pairwise interactions. More advanced approaches include \emph{Persistent Contrastive Divergence} (PCD) \cite{desjardins2010tempered}, where the chain is not reinitialized from data each update, and \emph{Mean-Field} methods \cite{opper2001advanced} for approximate inference.
\end{itemize}

\emph{Contrastive Divergence} trades off exactness for computational feasibility when training energy-based models, often yielding good empirical performance despite the inherent approximation in the negative phase. This makes it a practical cornerstone for learning parametric EBMs in high-dimensional image-generation tasks.

\subsection{Image Generation Metrics}
\label{subsec:img_gen_metrics}

Evaluating the quality of generated images requires metrics that capture both low-level fidelity (e.g., pixel-wise accuracy) and high-level perceptual or distributional realism. Researchers have developed a range of quantitative measures, each highlighting different aspects of the generative model’s output. This section introduces four commonly used metrics:

\begin{itemize}
    \item \textbf{Peak Signal-to-Noise Ratio (PSNR):} 
        Assesses image fidelity by comparing signal power to noise power on a pixel-wise basis.
    \item \textbf{Structural Similarity Index (SSIM):} 
        Evaluates perceptual similarity by modeling luminance, contrast, and structural components.
    \item \textbf{Inception Score (IS) \cite{salimans2016improved}:} 
        Estimates both the diversity of generated samples and the alignment of each sample with a class-conditional label distribution.
    \item \textbf{Fréchet Inception Distance (FID) \cite{heusel2017gans}:} 
        Measures the distance between real and generated image distributions by comparing feature-space statistics.
\end{itemize}

In the following subsections, we present each metric in detail, discussing its mathematical formulation, practical application, and associated strengths and weaknesses for various image generation tasks.

\subsubsection{Peak Signal-to-Noise Ratio (PSNR)}
\label{metric:psnr}

\emph{Peak Signal-to-Noise Ratio (PSNR)} is a traditional metric that quantifies the fidelity between an original image \(\mathbf{x}\) and a reconstructed (or generated) image \(\hat{\mathbf{x}}\). Widely employed in image/video compression and restoration, PSNR measures pixel-wise accuracy in a logarithmic decibel (dB) scale.

Let the Mean Squared Error (MSE) between \(\mathbf{x}\) and \(\hat{\mathbf{x}}\) be

\begin{equation}
\mathrm{MSE}
= \frac{1}{HW}
  \sum_{i=1}^{H}
  \sum_{j=1}^{W}
    \bigl(x_{i,j} - \hat{x}_{i,j}\bigr)^2,
\end{equation}
where \(H\) and \(W\) are the image height and width (assuming a single channel for simplicity). If \( \mathrm{MAX}_I\) denotes the maximum possible pixel intensity (e.g., 255 in 8-bit grayscale), then \emph{PSNR} is given by

\begin{equation}
\label{eq:psnr}
\mathrm{PSNR}
= 10 \,\log_{10}\!\Bigl(
     \frac{(\mathrm{MAX}_I)^2}{\mathrm{MSE}}
   \Bigr).
\end{equation}

\paragraph{Interpretation:}
\begin{itemize}
    \item \textbf{Higher is Better:}
    Since PSNR measures the ratio of signal power (i.e., squared maximum intensity) to error power, a larger PSNR (in dB) indicates fewer pixel-wise deviations, implying stronger fidelity.
    \item \textbf{Reference-based:}
    PSNR requires a reference “ground-truth” image, making it suited for tasks like super-resolution, denoising, or inpainting where the objective is to match a high-quality target.
\end{itemize}

\paragraph{Strengths and Limitations:}
\begin{itemize}
    \item \textbf{Advantages:}
    \begin{itemize}
       \item \emph{Simplicity}: PSNR is straightforward to compute and interpret.
       \item \emph{Historical Precedence}: Widely used in compression/restoration literature, providing easy comparisons across studies.
    \end{itemize}
    \item \textbf{Limitations:}
    \begin{itemize}
       \item \emph{Poor Perceptual Correlation}: PSNR may not reflect human visual quality well; two images can have similar PSNR yet differ significantly in perceived detail.
       \item \emph{Unsuitable for Generative Tasks}: For generative models aiming to create novel images (rather than exact reconstructions), PSNR’s pixel-wise emphasis fails to capture realism or diversity.
    \end{itemize}
\end{itemize}

\paragraph{Usage:}
While PSNR remains a standard for evaluating tasks with a known ground truth (e.g., super-resolution vs.\ a high-resolution reference, denoising vs.\ an uncorrupted image), it is less informative in adversarial or large-scale generative scenarios where the model produces new, unconstrained samples. In those cases, metrics like Inception Score (IS) or Fréchet Inception Distance (FID) (\S\ref{metric:FID}) often provide more perceptually meaningful assessments \cite{mukherkjee2022brain}.

\subsubsection{Structural Similarity Index (SSIM)}
\label{metric:SSIM}

\emph{Structural Similarity Index (SSIM)} \cite{wang2004image} is a perceptual metric for comparing two images \(\mathbf{x}\) and \(\mathbf{y}\). It serves as an improvement over purely pixel-wise measures (e.g., MSE or PSNR) by capturing local patterns of luminance, contrast, and structural details, thus correlating better with human visual perception.

\paragraph{Motivation and Components:}
SSIM decomposes similarity into three core components:
\begin{enumerate}
    \item \textbf{Luminance} — The mean intensity (\(\mu_\mathbf{x}, \mu_\mathbf{y}\)) of each image region.
    \item \textbf{Contrast} — The standard deviation (\(\sigma_\mathbf{x}, \sigma_\mathbf{y}\)) or intensity variation.
    \item \textbf{Structure} — The normalized covariance (\(\sigma_{\mathbf{x}\mathbf{y}}\)) capturing how pixel intensities co-vary between \(\mathbf{x}\) and \(\mathbf{y}\).
\end{enumerate}
By combining these in a single index, SSIM yields a measure ranging from \(-1\) to \(1\), where \(1\) indicates perfect similarity, \(0\) indicates no correlation, and \(-1\) indicates complete dissimilarity.

\paragraph{Definition:}
The SSIM for two image patches \(\mathbf{x}\) and \(\mathbf{y}\) of equal size is defined as

\begin{equation}
\label{eq:ssim1}
\mathrm{SSIM}(\mathbf{x}, \mathbf{y})
= \frac{
    \bigl(2\,\mu_\mathbf{x}\,\mu_\mathbf{y} + C_1\bigr)
    \bigl(2\,\sigma_{\mathbf{x}\mathbf{y}} + C_2\bigr)
  }{
    \bigl(\mu_\mathbf{x}^2 + \mu_\mathbf{y}^2 + C_1\bigr)
    \bigl(\sigma_\mathbf{x}^2 + \sigma_\mathbf{y}^2 + C_2\bigr)
  },
\end{equation}

\noindent
where:
\begin{itemize}
    \item \(\mu_\mathbf{x}, \mu_\mathbf{y}\) are mean intensities (luminance) of \(\mathbf{x}\) and \(\mathbf{y}\).
    \item \(\sigma_\mathbf{x}, \sigma_\mathbf{y}\) are standard deviations (contrast) of \(\mathbf{x}\) and \(\mathbf{y}\).
    \item \(\sigma_{\mathbf{x}\mathbf{y}}\) is the covariance capturing structural similarity.
    \item \(C_1, C_2\) are small stabilizing constants, typically set to \((K_1 L)^2\) and \((K_2 L)^2\) where \(L\) is the dynamic range (e.g., 255 for 8-bit grayscale) and \(K_1, K_2\ll 1\).
\end{itemize}

\paragraph{Interpretation:}
\begin{itemize}
    \item \textbf{Robust to Brightness/Contrast Shifts:}
    Because SSIM separately compares luminance and contrast, small changes in overall brightness or global intensity do not overly penalize the metric.
    \item \textbf{Local Evaluation:}
    SSIM is often computed over local windows, then averaged across the image, providing a spatially aware measure of perceptual differences.
\end{itemize}

\paragraph{Relevance to Image Generation:}
\begin{itemize}
    \item \textbf{Objective Quality Assessment:}
    In tasks like super-resolution, denoising, or inpainting, SSIM compares the generated image to a ground-truth reference, reflecting perceptual fidelity more accurately than MSE or PSNR alone.
    \item \textbf{Limits for Generative Tasks:}
    SSIM requires a reference target. In purely generative applications (e.g., unconditional GANs) where no exact ground truth exists, SSIM cannot measure diversity or distribution coverage. 
\end{itemize}

While SSIM is valuable for assessing local perceptual quality, it does not account for global distributional properties or sample diversity. Hence, it is often used alongside \emph{Inception Score} or \emph{Fréchet Inception Distance} (\S\ref{metric:FID}) for a more holistic evaluation of generative models \cite{mukherkjee2022brain}.

In general, \emph{SSIM} provides a more perceptually aligned alternative to pixel-based error metrics, allowing image generation tasks that rely on fidelity to a reference image to benefit from a measure that better aligns with the human visual system.

\subsubsection{Inception Score (IS)}
\label{metric:IS}

\emph{Inception Score (IS)} \cite{salimans2016improved} provides a single numeric measure of both the \emph{quality} (or recognizability) and the diversity of images generated by a model. The metric leverages a pretrained Inception-v3 classifier, originally trained on ImageNet, to evaluate how confidently generated samples can be classified and whether the overall set of generated images spans multiple classes.

\paragraph{Definition:}
Let \(p_g(\mathbf{x})\) be the distribution of generated images \(\mathbf{x}\). When fed through the Inception model, each image \(\mathbf{x}\) yields a conditional label distribution \(p(y \mid \mathbf{x})\), where \(y\) indexes ImageNet classes. Define the marginal distribution of classes as

\begin{equation}
p(y)
= \int
   p(y \mid \mathbf{x})\,
   p_g(\mathbf{x})\, d\mathbf{x}
\quad\text{(in practice, approximated by sampling)}.
\end{equation}

The Inception Score is computed as

\begin{equation}
\label{eq:is}
\mathrm{IS}
= \exp\Bigl(
    \mathbb{E}_{\mathbf{x}\sim p_g}
    \bigl[
      \mathrm{KL}\bigl(p(y\mid\mathbf{x})\,\bigl\|\,p(y)\bigr)
    \bigr]
  \Bigr),
\end{equation}

\noindent
where 
\(\mathrm{KL}\bigl(p(y\mid\mathbf{x})\|p(y)\bigr)\) is the Kullback--Leibler divergence of the conditional distribution \(p(y\mid\mathbf{x})\) from the marginal \(p(y)\).

\paragraph{Interpretation:}
\begin{itemize}
    \item \textbf{Quality (Confidence):}
    The term \(p(y\mid \mathbf{x})\) is expected to have low entropy (i.e., be peaked around a single class), implying that \(\mathbf{x}\) is recognizable as a valid object.
    \item \textbf{Diversity:}
    The marginal \(p(y)\) should have high entropy—i.e., images are distributed among many classes. A high KL divergence indicates that \(\mathbf{x}\) strongly activates a distinct class label relative to the overall distribution, reflecting class diversity.
    \item \textbf{Exponential Form:}
    Taking \(\exp\) of the average KL divergence yields a positive scalar, commonly larger than 1. Higher IS values suggest better generative performance.
\end{itemize}

\paragraph{Practical Computation:}
\begin{enumerate}
    \item \textbf{Sampling Generated Images:} 
    Draw a sufficiently large set \(\{\mathbf{x}_i\}\) from your generative model.
    \item \textbf{Inception Predictions:}
    For each \(\mathbf{x}_i\), compute \(p(y\mid \mathbf{x}_i)\) using Inception-v3’s softmax output. 
    \item \textbf{Approximate $p(y)$:} 
    Estimate the marginal \(p(y)\) by averaging conditional predictions:
    \begin{equation}
      p(y)
      \approx \frac{1}{N}\sum_{i=1}^N p(y\mid \mathbf{x}_i).
    \end{equation}
    \item \textbf{Compute $\mathrm{IS}$:}
    Evaluate 
    \(\mathbb{E}_{\mathbf{x}\sim p_g}[\mathrm{KL}(p(y\mid \mathbf{x})\|\, p(y))]\)
    and exponentiate the result.
\end{enumerate}

\paragraph{Limitations:}
\begin{enumerate}
    \item \emph{Dependence on Inception-v3:}
    The metric heavily relies on ImageNet-trained features, making it less reliable in domains or label distributions far from ImageNet classes.
    \item \emph{Mismatch with Human Perception:}
    A high IS might not align with subjective visual realism; images can be “classifiable” yet lack photorealism or contain subtle artifacts.
    \item \emph{Potential Blindness to Mode Collapse:}
    If a few modes already suffice for diversity among recognized classes, the model may still yield repeated patterns within each class that IS fails to penalize.
    \item \emph{Insensitivity to Class Balance:}
    IS does not penalize generating certain classes more often than others if coverage of classes is still broad.
    \item \emph{Sample Efficiency:}
    A large number of generated samples is needed for a stable estimate, and the metric can exhibit high variance for smaller sets.
\end{enumerate}

\paragraph{Usage Context:}
Despite these limitations, \emph{Inception Score} remains a popular and straightforward heuristic for evaluating image quality and diversity in generative tasks—particularly when real-world labels are in line with ImageNet concepts. However, for thorough assessment, practitioners often complement IS with measures like \emph{Fréchet Inception Distance} (\S\ref{metric:FID}) or user studies to better capture perceptual fidelity and distribution coverage.

\subsubsection{Fréchet Inception Distance (FID)}
\label{metric:FID}

\emph{Fréchet Inception Distance (FID)} \cite{heusel2017gans} evaluates the realism and diversity of generated images by measuring how closely their feature distribution matches that of real images. Unlike the Inception Score (\S\ref{metric:IS}), which focuses on per-image class distributions, FID compares the global statistics of high-level feature activations between real and generated datasets.

\paragraph{Feature Extraction:}
Let \(\mathbf{x} \in X\) be samples from the real dataset and \(\hat{\mathbf{x}} \in \hat{X}\) be samples from the model’s generated dataset. Both sets are fed through a fixed Inception network, and feature activations (from a chosen layer, often a pool or hidden representation) are extracted:

\begin{equation}
\mathbf{x}' = \mathrm{Inception}(\mathbf{x}), 
\quad 
\hat{\mathbf{x}}' = \mathrm{Inception}(\hat{\mathbf{x}}).
\end{equation}
Denote \(\{\mathbf{x}'\}\) and \(\{\hat{\mathbf{x}}'\}\) as sets of feature vectors for real and generated data, respectively.

\paragraph{Modeling with Multivariate Gaussians:}
Assume that these feature activations follow a multivariate Gaussian distribution in \(\mathbb{R}^d\). Let
\begin{equation}
\mu_x, \Sigma_x 
\quad\text{and}\quad 
\mu_{\hat{x}}, \Sigma_{\hat{x}}
\end{equation}
be the mean and covariance of real and generated activations. The FID then defines the \emph{Fréchet distance} between these Gaussians as:

\begin{align}
\label{eq:fid}
\mathrm{FID}(\mathbf{x}', \hat{\mathbf{x}}')
&=
\|\mu_{x} - \mu_{\hat{x}}\|^{2}
+ \mathrm{Tr}
   \bigl(
     \Sigma_{x}
     + \Sigma_{\hat{x}}
     - 2\,(\Sigma_{x}\,\Sigma_{\hat{x}})^{1/2}
   \bigr),
\end{align}

\noindent
where:
\begin{itemize}
    \item \(\|\mu_{x} - \mu_{\hat{x}}\|\) is the Euclidean distance of means, capturing shifts in the distribution’s center.
    \item \(\mathrm{Tr}(\cdot)\) denotes the trace operator (sum of diagonal elements).
    \item \((\Sigma_{x}\,\Sigma_{\hat{x}})^{1/2}\) is the matrix square root of the product of the two covariance matrices, well-defined since \(\Sigma_{x}\) and \(\Sigma_{\hat{x}}\) are positive semi-definite.
\end{itemize}

\paragraph{Advantages:}
\begin{itemize}
    \item \textbf{Realism and Diversity:}
    By comparing mean and covariance of feature embeddings, FID evaluates both the average “location” (quality) and the spread (diversity) of generated samples.
    \item \textbf{Less Sensitive to Noise than IS:}
    FID can capture subtle distribution shifts even if images appear recognizable to Inception but differ in style or texture.
\end{itemize}

\paragraph{Limitations:}
\begin{enumerate}
    \item \textbf{Domain Dependence on Inception:}
    Similar to Inception Score, the features are extracted from an ImageNet-trained network, which may be suboptimal for domains far removed from natural images.
    \item \textbf{Gaussian Assumption:}
    FID implicitly treats feature activations as Gaussian, which might not hold perfectly. Deviations from normality can impact the metric’s accuracy.
    \item \textbf{Requires Large Sample Sets:}
    Reliable estimation of means and covariances often demands a sufficiently large dataset for both real and generated samples.
\end{enumerate}

\paragraph{Usage Context:}
Despite these limitations, \emph{FID} has become a standard metric for evaluating generative models in vision tasks. By simultaneously penalizing discrepancies in feature-space location and spread, it provides a more robust assessment of how faithfully a model reproduces the underlying data distribution than single-sample or classification-based metrics.

\section{Natural Language Processing (NLP)}
\label{sec:NLP}

Natural Language Processing (NLP) is a subfield of artificial intelligence that focuses on the interaction between computers and human languages. It involves the development of algorithms and models that allow computers to understand, interpret, and generate human language in a way that is both meaningful and useful. NLP encompasses a wide range of tasks, from basic text processing and translation \cite{vaswani2017attention} to complex language generation \cite{brown2020language}.

The main goal of NLP is to bridge the gap between human communication and machine understanding, allowing more natural and efficient interactions with technology. This includes enabling machines to comprehend the subtleties of human language, such as syntax, semantics, context, and sentiment.

Tha main tasks in NLP are the following:
\begin{enumerate}
    \item Text Classification: Involves categorizing text into predefined classes or categories. Examples include sentiment analysis \cite{pang2008opinion}, spam detection \cite{metsis2006spam}, and topic classification \cite{blei2003latent}.
    \item Language Modeling: Predicts the probability of a sequence of words. It is a foundational task in NLP, crucial for applications like text generation and autocomplete \cite{brown2020language}.
    \item Machine Translation: Converts text from one language to another. Popular applications include translating documents, websites, and real-time speech translation \cite{vaswani2017attention}.
    \item Named Entity Recognition (NER): Identifies and classifies proper names, such as people, organizations, and locations, within text \cite{chieu2003named}.
    \item Part-of-Speech Tagging: Involves labeling each word in a sentence with its corresponding part of speech, such as noun, verb, adjective, etc \cite{church1989stochastic}.
    \item Text Summarization: Automatically generates a concise summary of a larger body of text. It can be extractive (selecting key sentences) or abstractive (generating new sentences) \cite{shi2021neural}.
    \item Question Answering: Involves building systems that can automatically answer questions posed in natural language \cite{devlin2018bert}.
\end{enumerate}

These tasks need specialized techniques, including different loss functions and performance metrics for model optimization and evaluation. NLP evolves rapidly due to machine learning advances and large dataset availability. The next sections will discuss key loss functions and metrics, explaining their applications and relevance.

\subsection{Loss Functions Used in NLP}
\label{subsec:Losses_in_NLP}

Loss functions are central to training robust and accurate Natural Language Processing (NLP) models. From classifying text and predicting next-word tokens to generating coherent sentences in machine translation or summarization, carefully chosen loss functions guide each model’s parameters toward meaningful language representations and better performance on downstream tasks.

While general-purpose losses such as \emph{Cross-Entropy} or \emph{Hinge Loss} appear across multiple domains (including computer vision), they also feature prominently in NLP settings because many language tasks, such as text classification and sequence labeling, boil down to discriminative objectives. Moreover, certain loss formulations specifically address the intricacies of language modeling and sequence generation, acknowledging the sequential, probabilistic nature of text.

In this section, we provide an overview of commonly used loss functions in NLP, each one exhibits different strengths and trade-offs, from straightforward implementations for single-label classification to more elaborate designs that handle sequential dependencies, large vocabularies, or exposure bias in generated text. By selecting the most appropriate loss function, NLP practitioners can balance training efficiency, model interpretability, and the desired linguistic or functional outcome.

Table~\ref{tab:guidelines-nlp-losses} summarizes these widely used losses for NLP tasks, detailing their typical application scenarios, data assumptions, advantages, and potential pitfalls. Understanding when and how to leverage each loss function is critical for achieving state-of-the-art results in modern NLP pipelines.

\begin{table*}[ht!]
\caption{Guidelines for selecting a loss function in NLP based on usage, data characteristics, advantages and limitations.}
\label{tab:guidelines-nlp-losses}
\centering
\footnotesize
\begin{tabular}{p{1.3cm}p{2.8cm}p{2.8cm}p{2.8cm}p{2.8cm}}
\toprule
\textbf{Function}
& \textbf{Usage} 
& \textbf{Data Characteristics} 
& \textbf{Advantages} 
& \textbf{Limitations} \\
\midrule

Cross-Entropy Loss 
& Language modeling\newline Machine translation\newline Summarization\newline Token-level classification 
& Typically discrete tokens\newline Large vocabularies\newline One-hot or sparse labels 
& Direct optimization of token accuracy\newline Well-established and easy to implement 
& May not capture global sequence quality\newline Can lead to exposure bias \\
\midrule

Hinge Loss 
& Binary text classification\newline Sentiment analysis\newline Margin-based sequence labeling 
& Mostly binary labels\newline Structured or sparse features 
& Promotes large margin separation\newline Robust to outliers 
& Primarily for binary tasks\newline Needs adaptation for multi-class\newline Less common in modern NLP \\
\midrule

Cosine Similarity Loss 
& Semantic similarity\newline Paraphrase identification\newline Information retrieval 
& Dense embeddings\newline Continuous vector spaces 
& Aligns well with embedding-based tasks\newline Captures semantic closeness 
& Zero similarity can be ambiguous\newline Not ideal for sparse inputs \\
\midrule

Marginal Ranking Loss 
& Ranking documents\newline Question-answer retrieval\newline Recommendation systems 
& Paired relevance data\newline Positive vs.\ negative samples 
& Improves ordering of relevant vs.\ irrelevant\newline Flexible margin-based design 
& Pairwise comparisons can be expensive\newline Fine-grained ranking errors may go unpenalized \\
\midrule

CTC Loss 
& Speech recognition\newline Sequence alignment-free tasks\newline OCR-like problems 
& Input longer than output\newline Unaligned sequence data\newline Blank symbol handling 
& Learns alignment automatically\newline Handles variable-length outputs 
& More complex to implement\newline Can struggle with large vocabularies \\
\midrule

Minimum Risk Training 
& Machine translation\newline Summarization\newline Dialog systems 
& Full-sequence metrics\newline BLEU or ROUGE-based evaluation 
& Directly optimizes final metric\newline Better global coherence 
& Computationally expensive sampling\newline Requires careful tuning of scaling factors \\
\midrule

REINFORCE 
& Non-differentiable metric optimization\newline Generative tasks (MT, summarization) 
& Sequence-level rewards\newline Discrete token actions 
& Flexible for custom rewards\newline Improves performance on metrics like BLEU 
& High variance in gradient estimates\newline Slow convergence without variance reduction \\

\bottomrule
\end{tabular}
\end{table*}

\subsubsection{Cross-Entropy Loss (Token-Level)}
\label{loss:t-cce}

\emph{Cross-Entropy} (CE) loss is fundamental in NLP tasks involving sequence prediction, as it quantifies the divergence between the model’s predicted probability distribution and the true distribution for each token. In practice, the ground-truth token is encoded as a one-hot vector, and the model’s output is typically a softmax distribution over the vocabulary.

\paragraph{Definition:}
Let \(\mathbf{y} = (y_1, y_2, \dots, y_T)\) be the target token sequence of length \(T\), and let \(\hat{\mathbf{y}} = (\hat{y}_1, \hat{y}_2, \dots, \hat{y}_T)\) be the model’s predicted tokens. During training, the model actually produces a probability distribution \(\mathbf{p}_t = p(\cdot\mid \mathbf{y}_{<t})\in\mathbb{R}^V\) at each timestep \(t\) over a vocabulary of size \(V\). The token-level cross-entropy loss is then:

\begin{equation}
\label{eq:tce}
\mathcal{L}_{\mathrm{CE}}
= -\frac{1}{T}
   \sum_{t=1}^{T}
     \log p\bigl(y_t \mid y_{<t}\bigr),
\end{equation}

\noindent
where \(p\bigl(y_t \mid y_{<t}\bigr)\) denotes the predicted probability (from the model’s softmax layer at time \(t\)) of the correct token \(y_t\). Formally, if \(\mathbf{p}_t\) is the predicted distribution over all possible tokens at step \(t\), then \(p\bigl(y_t\mid y_{<t}\bigr) = \mathbf{p}_{t}[y_t]\). 

Minimizing \(\mathcal{L}_{\mathrm{CE}}\) pushes the model to assign higher probability mass to the correct tokens at each timestep, effectively learning the conditional distributions needed for sequence generation or token-level classification.

\paragraph{Usage in NLP Tasks:}
\begin{enumerate}
    \item \textbf{Language Modeling \cite{radford2018improving}}:
    Predicting the next word given previous context. Cross-entropy penalizes the model when it assigns low probability to the correct next word, thus learning coherent text generation.

    \item \textbf{Machine Translation \cite{vaswani2017attention}}:
    Mapping a source-language sentence into a target-language sentence. CE is computed at each target token to align the generated translation with reference translations, guiding the model to produce contextually accurate sequences.

    \item \textbf{Summarization \cite{torres2014automatic}}:
    Generating a condensed summary of a longer text. The cross-entropy loss encourages the system to reproduce relevant tokens from reference summaries, ensuring fidelity and coverage of essential information.

    \item \textbf{Token Labeling Tasks}:
    In part-of-speech tagging \cite{schmid1994part} or named entity recognition \cite{etzioni2005unsupervised}, each token’s label is predicted. CE at each timestep penalizes errors in assigning the correct label, effectively treating each token as a classification instance.
\end{enumerate}

\paragraph{Strengths and Limitations:}
\begin{itemize}
    \item \textbf{Strengths}:
    \begin{itemize}
      \item \emph{Direct Token-Level Supervision}:
      The model receives a clear gradient signal for each token misclassification, supporting stable training in sequence tasks.
      \item \emph{Broad Applicability}:
      Widely used across classification-based NLP tasks, from language modeling to sequence labeling.
    \end{itemize}
    \item \textbf{Limitations}:
    \begin{itemize}
      \item \emph{Local Focus}:
      Cross-entropy penalizes token mismatches independently, ignoring potential correlation or reordering effects in the final output (e.g., important for machine translation beyond lexical matches).
      \item \emph{Discrepancy With End Metrics}:
      For tasks evaluated by BLEU, ROUGE, or other sequence-level metrics, token-level CE may not capture global sequence properties or semantic coherence.
      \item \emph{Exposure Bias}:
      Models trained with “teacher forcing” see ground-truth history at each step, causing potential discrepancies at inference if previous tokens are mispredicted.
    \end{itemize}
\end{itemize}

Token-level \emph{Cross-Entropy Loss} remains the primary objective in many supervised and sequence-to-sequence frameworks, reliably guiding the model to align its per-token distributions with the ground truth. However, advanced or specialized tasks might incorporate additional objectives (e.g., sequence-level RL or minimum risk training) to address holistic output fidelity and potential mismatch with final evaluation metrics.

\subsubsection{Hinge Loss}
\label{loss:hinge_nlp}

\emph{Hinge Loss} is a margin-based objective often employed in binary classification tasks, where labels are encoded as \(\{-1, +1\}\). Although hinge loss is more traditionally linked to Support Vector Machines (SVMs) in computer vision or general classification, it also features in various NLP contexts such as sentiment analysis, spam detection, or structured prediction under \emph{Structured SVMs} \cite{taskar2003max,tsochantaridis2005large,martins2012structured,wang2017adversarial}.

\paragraph{Definition:}
For an input \(\mathbf{x}\in\mathcal{X}\), a model (e.g., linear or kernel-based) produces a real-valued score \(f(\mathbf{x})\). The true label \(y\in\{-1,+1\}\). The \emph{Hinge Loss} is defined by:

\begin{equation}
\label{eq:hinge_nlp}
\mathcal{L}_{\mathrm{hinge}}(\mathbf{x},y) 
= \max\bigl(0,\;1 - y\,f(\mathbf{x})\bigr).
\end{equation}

\noindent
This implies:
\begin{itemize}
    \item If \(y\,f(\mathbf{x}) \ge 1\), the loss is \(0\), indicating that \(\mathbf{x}\) is not only classified correctly but also lies outside or on the margin boundary (i.e., with sufficient confidence).
    \item If \(y\,f(\mathbf{x}) < 1\), the model incurs a penalty proportional to \(\bigl(1 - y\,f(\mathbf{x})\bigr)\), pushing parameters to adjust until \(\mathbf{x}\) crosses the margin or is better separated.
\end{itemize}

\paragraph{Usage in NLP:}
\begin{enumerate}
    \item \textbf{Binary Text Classification:}  
    Tasks like \emph{spam detection} or \emph{sentiment analysis} often reduce to a \(\{-1,+1\}\)-label scenario. A linear or kernel-based SVM with hinge loss defines a max-margin boundary separating positive from negative classes.

    \item \textbf{Structured SVMs for Sequence Labeling:}  
    In part-of-speech tagging \cite{schmid1994part} or named entity recognition \cite{etzioni2005unsupervised}, hinge loss can be adapted to \emph{structured prediction} (structured hinge loss). The model’s output is an entire label sequence, and the margin constraints incorporate sequence dependencies or feature-based interactions \cite{taskar2003max,tsochantaridis2005large}.

    \item \textbf{Sparse Linear Models in NLP:}  
    L1- or L2-regularized hinge loss is employed to promote sparse solutions or certain generalization behaviors \cite{moore2011l1}. This can be beneficial when the input feature space is large (e.g., high-dimensional lexical features).

    \item \textbf{Adaptive Losses and Adversarial Frameworks:}  
    Recent methods adapt hinge-based objectives for balancing precision and recall (e.g., in document-level relation extraction \cite{wang2023adaptive}) or for adversarially aligning model predictions with performance metrics in structured tasks \cite{wang2017adversarial}.
\end{enumerate}

\paragraph{Advantages:}
\begin{itemize}
    \item \textbf{Max-Margin Principle:}  
      By ensuring \(\max(0,1 - y\,f(\mathbf{x}))\), the hinge loss explicitly drives the decision boundary away from training samples, encouraging robust classification.
    \item \textbf{Suitability for Sparse and Structured Models:}  
      Hinge loss, paired with regularization (L1/L2), can yield sparser parameter vectors in large-scale NLP tasks with vast lexical or n-gram features.
\end{itemize}

\paragraph{Limitations:}
\begin{itemize}
    \item \textbf{Binary Nature:}  
      Extending hinge loss to multi-class problems requires one-vs.-all, one-vs.-one, or structured multi-class formulations, which can complicate implementation.
    \item \textbf{No Probabilistic Interpretation:}  
      Hinge loss produces a margin-based decision function rather than a probability estimate, unlike cross-entropy which naturally yields class probabilities via softmax.
    \item \textbf{Potential Large Margin Tuning:}  
      Selecting the margin constraints or regularizer weights can be non-trivial, impacting the stability and performance of classification or sequence labeling tasks.
\end{itemize}

\subsubsection{Cosine Similarity Loss}
\label{loss:cosine_nlp}

\emph{Cosine Similarity Loss} \cite{hoe2021one} leverages the cosine of the angle between two vector embeddings as a measure of semantic closeness. Widely used in NLP tasks involving sentence, phrase, or word embeddings, it aims to \emph{maximize} the cosine similarity for pairs deemed “similar” (e.g., paraphrases) and \emph{minimize} it for pairs deemed “dissimilar” (e.g., unrelated texts).

\paragraph{Cosine Similarity:}
Given two non-zero vectors \(\mathbf{u}, \mathbf{v}\in \mathbb{R}^d\), the \emph{cosine similarity} is:

\begin{equation}
\label{eq:cosine_sim}
\mathrm{cos\_sim}(\mathbf{u}, \mathbf{v})
= \frac{\mathbf{u}\cdot\mathbf{v}}
       {\|\mathbf{u}\|\;\|\mathbf{v}\|},
\end{equation}

\noindent
where \(\mathbf{u}\cdot\mathbf{v}\) denotes the dot product, and \(\|\mathbf{u}\|\) is the Euclidean norm of \(\mathbf{u}\). The value of \(\mathrm{cos\_sim}\) ranges from \(-1\) to \(1\); \(1\) indicates parallel (identical direction) vectors, \(0\) indicates orthogonality, and \(-1\) indicates diametrically opposed vectors.

To embed this measure into a \emph{loss function}, one design is:

\begin{equation}
\label{eq:cosine_sim2}
\mathcal{L}_{\mathrm{cos}}(\mathbf{u},\mathbf{v})
= 1 - \mathrm{cos\_sim}(\mathbf{u},\mathbf{v}),
\end{equation}

\noindent
which is minimized when \(\mathbf{u}\) and \(\mathbf{v}\) are maximally aligned (\(\mathrm{cos\_sim}=1\)). For pairs that should be dissimilar, practitioners often define a complementary margin-based penalty to push their similarity below a certain threshold, for instance:

\begin{equation}
\mathcal{L}_{\mathrm{neg}}(\mathbf{u},\mathbf{v})
= \max\!\bigl(0,\;\mathrm{cos\_sim}(\mathbf{u},\mathbf{v}) - \mathrm{margin}\bigr),
\end{equation}
so that vectors representing dissimilar text remain sufficiently far apart in the embedding space.

\paragraph{NLP Applications:}
\begin{enumerate}
    \item \textbf{Semantic Textual Similarity (STS):}
    Models produce embeddings for sentence pairs \(\{\mathbf{u},\mathbf{v}\}\). Cosine similarity loss encourages semantically similar sentences to lie close (high cosine sim) and dissimilar ones to remain distant.
    \item \textbf{Paraphrase Identification:}
    Sentences conveying the same meaning but different wording benefit from a training objective that drives embeddings of paraphrases to have near-cosine-sim = 1.
    \item \textbf{Information Retrieval \& Ranking:}
    Documents and queries are embedded. Relevant (query, doc) pairs are forced to have higher similarity than irrelevant ones, improving retrieval performance.
    \item \textbf{Text Clustering:}
    By grouping embeddings of similar texts together and separating unrelated texts, the loss fosters coherent clusters in embedding space.
\end{enumerate}

\paragraph{Limitations:}
\begin{itemize}
    \item \textbf{Zero Similarity Ambiguity:}
      A similarity of 0 suggests orthogonality, which might arise from actual dissimilarity or from insufficient representational overlap (e.g., no shared features).
    \item \textbf{Dense Vector Requirement:}
      Typically used with continuous embeddings (word2vec, BERT, etc.). For sparse or one-hot vectors, additional normalization or transformations may be needed.
    \item \textbf{Margin Definition for Negative Pairs:}
      A naive loss like \(1-\mathrm{cos\_sim}\) only accounts for pushing similar vectors together. Handling negative (dissimilar) pairs might require margin-based or multi-pair frameworks.
\end{itemize}

\emph{Cosine Similarity Loss} directly aligns with the geometric intuition of semantic closeness, making it intuitive for tasks like \emph{STS}, \emph{paraphrase detection}, or \emph{retrieval/ranking} problems in NLP. However, care is needed to handle negative examples or define appropriate margins, especially in large-scale or multi-class settings.

\subsubsection{Marginal Ranking Loss}
\label{loss:marginal_ranking_nlp}

\emph{Marginal Ranking Loss} (also called \emph{RankNet loss} or \emph{pairwise ranking loss}) \cite{bordes2013translating,trouillon2016complex} is a margin-based objective commonly used in tasks where the model must rank items—such as text passages, documents, or sentences—by relevance. By comparing a \emph{positive} (relevant) item to a \emph{negative} (irrelevant) item, the model is encouraged to assign higher scores to relevant items, improving its ability to order results effectively.

\paragraph{Definition:}
Let \(\mathbf{q}\in\mathcal{Q}\) represent a query (or context), \(\mathbf{p}\) a positive/relevant item, and \(\mathbf{n}\) a negative/irrelevant item. Suppose the model defines a scoring function \(f(\mathbf{q}, \mathbf{z})\), producing a scalar indicating how well item \(\mathbf{z}\) matches or is relevant to \(\mathbf{q}\). The \emph{Marginal Ranking Loss} is then:

\begin{equation}
\label{eq:marg_rank}
\mathcal{L}(\mathbf{q}, \mathbf{p}, \mathbf{n})
= \max\!\Bigl(
    0,\,
    \mathrm{margin}
    - f(\mathbf{q},\mathbf{p})
    + f(\mathbf{q},\mathbf{n})
  \Bigr),
\end{equation}

\noindent
where \(\mathrm{margin}>0\) is a hyperparameter that enforces a separation between the score of positive pairs and negative pairs. Intuitively, the loss remains 0 if \(f(\mathbf{q},\mathbf{p})\) already exceeds \(f(\mathbf{q},\mathbf{n})\) by \(\mathrm{margin}\) or more, indicating sufficient discrimination between relevant and irrelevant items. Otherwise, the model incurs a penalty proportional to the shortfall.

\paragraph{Applications in NLP:}
\begin{enumerate}
    \item \textbf{Information Retrieval:}
      Given a user query \(\mathbf{q}\), documents or passages are ranked by \(f(\mathbf{q}, \mathbf{d})\). Marginal Ranking Loss compels the model to assign higher scores to relevant (\(\mathbf{p}\)) than to non-relevant (\(\mathbf{n}\)) documents, improving retrieval quality.

    \item \textbf{Question-Answering (QA):}
      For QA systems, each candidate answer can be scored relative to the question. The model thus learns to rank correct/precise answers \(\mathbf{p}\) above incorrect/less relevant answers \(\mathbf{n}\). This leads to higher-ranking correct answers at inference time.

    \item \textbf{Recommendation \& Personalization:}
      In scenarios where items (e.g., articles or products) are recommended based on user preference \(\mathbf{q}\), the ranking loss encourages relevant items to appear above irrelevant ones, tailoring suggestions to user tastes.

    \item \textbf{Semantic Similarity / Paraphrase Detection:}
      The model ranks paraphrased or near-duplicate sentences \(\mathbf{p}\) above dissimilar ones \(\mathbf{n}\) for a given reference. This fosters more consistent semantic grouping or paraphrase identification.
\end{enumerate}

\paragraph{Key Considerations:}
\begin{itemize}
    \item \textbf{Margin Tuning:}
      The choice of \(\mathrm{margin}\) is critical. If set too large, many pairs may incur non-zero loss, making optimization difficult; if too small, the model might not robustly separate relevant from irrelevant items.
    \item \textbf{Computational Cost:}
      For large-scale ranking tasks, enumerating all (\(\mathbf{p},\mathbf{n}\)) pairs can be expensive. Practitioners often use negative sampling or partial approximations to scale training.
    \item \textbf{Fine-Grained Relevance:}
      When items differ only subtly in relevance, a simple margin-based approach may insufficiently penalize smaller ranking errors. Additional constraints or label-based weighting might be needed for nuanced tasks.
    \item \textbf{Pairwise Nature:}
      Marginal Ranking Loss compares pairs of items at a time, ignoring potential broader context or synergy among multiple items. More advanced methods (e.g., listwise ranking) can capture global ordering constraints.
\end{itemize}

\emph{Marginal Ranking Loss} provides a direct mechanism for training models to \emph{rank} relevant text segments higher than irrelevant ones. By imposing a margin-based separation in scores, it is effective in retrieval, QA, recommendation, and paraphrase detection tasks. However, handling subtle relevance differences, controlling computational complexity, and capturing global ordering constraints remain challenges in applying this pairwise ranking approach at scale.

\subsection{Losses for Sequence Generation}
\label{subsec:loss_seq_gen_nlp}

Sequence generation tasks in NLP involve predicting a sequence of words or tokens given an input, such as translating a sentence from one language to another, summarizing a document, or generating a response in a conversation. The challenges associated with sequence generation include maintaining fluency and coherence and ensuring that the sequence generated accurately reflects the intended meaning or information. This section explores the loss functions commonly used to optimize models for these tasks.

\subsubsection{Connectionist Temporal Classification (CTC) Loss}
\label{loss:CTC}

\emph{Connectionist Temporal Classification (CTC)} \cite{graves2006connectionist} is designed for sequence-to-sequence learning tasks where the alignment between input and output sequences is unobserved or uncertain. Commonly used in automatic speech recognition and handwriting recognition, CTC aligns a long input sequence \(\mathbf{x}\) of length \(T\) to a shorter target sequence \(\mathbf{y}\) of length \(U\), without the need for manually specified frame-to-label correspondences.

CTC introduces a special “blank” symbol (\(\emptyset\)) to handle variable-length alignments and repeated tokens. For each time step \(t=1,\dots,T\), the model outputs a probability distribution over an extended label set \(\mathcal{L} = \{\text{vocabulary tokens}\}\cup\{\emptyset\}\). By \emph{collapsing} consecutive identical tokens and removing blanks, one obtains a final predicted sequence \(\hat{\mathbf{y}}\). The probability \(P(\mathbf{y}\mid \mathbf{x})\) is then computed by summing over the probabilities of all possible alignments \(\pi\) that collapse to \(\mathbf{y}\).

Formally, let \(\text{Align}(\mathbf{y})\) be the set of all valid alignment paths of length \(T\) that collapse to the sequence \(\mathbf{y}\). Each alignment path \(\pi = (\pi_1,\ldots,\pi_T)\) assigns a label \(\pi_t \in \mathcal{L}\) at each time \(t\). Then the total probability of \(\mathbf{y}\) is:

\begin{equation}
\label{eq:ctc}
P(\mathbf{y} \,\mid\, \mathbf{x})
= \sum_{\pi \,\in\, \mathrm{Align}(\mathbf{y})}
  P(\pi \,\mid\, \mathbf{x}),
\end{equation}

\noindent
where

\begin{equation}
\label{eq:align}
P(\pi \mid \mathbf{x})
= \prod_{t=1}^T
   P(\pi_t \,\mid\, \mathbf{x}, t),
\end{equation}

\noindent
and \(P(\pi_t \,\mid\, \mathbf{x}, t)\) is the predicted probability of label \(\pi_t\) (including blank) at time \(t\). The alignment path \(\pi\) is thus assigned a probability equal to the product of the model’s label probabilities at each frame.

\paragraph{Definition:}
The \emph{CTC Loss} is defined as the negative log-probability of the target sequence \(\mathbf{y}\):

\begin{equation}
\label{eq:ctc_loss}
\mathcal{L}_{\mathrm{CTC}}
= -\log
   P(\mathbf{y} \,\mid\, \mathbf{x}).
\end{equation}

Minimizing this objective encourages the model to increase the sum of probabilities of all valid alignment paths that produce \(\mathbf{y}\). Since \(P(\mathbf{y} \mid \mathbf{x})\) involves a summation over exponentially many alignments, efficient dynamic programming algorithms (e.g., the forward-backward algorithm specialized for CTC) are used to compute \(P(\mathbf{y}\mid \mathbf{x})\) and its gradient in polynomial time.

\paragraph{Use Cases in NLP:}
\begin{itemize}
    \item \textbf{Speech and Character-level Recognition:}
      CTC often appears in speech-to-text systems, mapping audio frames to characters or phonemes without explicit frame-to-label alignments.
    \item \textbf{Handwriting Recognition:}
      By treating the image of a handwritten text line as the input sequence, CTC can learn to align and generate text transcriptions.
    \item \textbf{Any Sequence Task Without Alignment Supervision:}
      If an NLP task has unaligned input-output pairs and a monotonic alignment assumption (i.e., tokens in \(\mathbf{y}\) must appear in chronological order in \(\mathbf{x}\)), CTC can be adapted.
\end{itemize}

\paragraph{Limitations:}
\begin{itemize}
    \item \textbf{Strict Monotonic Alignment Assumption:}
      CTC presumes that the output sequence cannot reorder input steps, which might be restrictive if the alignment is not strictly left-to-right.
    \item \textbf{Blank Symbol Overhead:}
      The blank token handles variable lengths but adds complexity to the output space, requiring careful training and hyperparameter tuning.
    \item \textbf{Complexity in Multi-sequence or Non-monotonic Tasks:}
      In general, tasks that violate the linear alignment property (e.g., reordering in machine translation) may not align well with CTC’s core assumptions.
\end{itemize}

In general, \emph{CTC Loss} provides a framework for learning sequence mappings without manual alignments, effectively summing over all plausible ways to match a shorter target sequence to a longer input. Dynamic programming algorithms enable tractable training, making CTC a powerful choice for tasks like speech recognition and character-level text generation under monotonic alignment constraints.

\subsubsection{Minimum Risk Training (MRT)}
\label{loss:mrt_nlp}

\emph{Minimum Risk Training (MRT)} \cite{shen2015minimum} is a sequence-level optimization strategy that directly aligns model training with the final evaluation metric (e.g., BLEU \cite{papineni2002bleu}, ROUGE \cite{lin2004rouge}), rather than relying on token-level losses like cross-entropy. By treating metric-based discrepancies as “risk,” MRT minimizes the expected risk under the model’s predicted distribution, thereby better capturing global sequence quality.

Many NLP tasks—such as machine translation or text summarization—are evaluated by sequence-level metrics (e.g., BLEU, ROUGE) that aggregate the coherence, coverage, or overlap of the entire output. However, standard token-level losses, like cross-entropy, do not necessarily correlate with these metrics. MRT addresses this by replacing the local objective with a global risk function tied to a chosen evaluation metric.

\paragraph{Definition:}
Let \(\mathbf{x}\) be an input, and \(\mathbf{y}\) a target reference sequence from a dataset of size \(N\). The model produces a distribution \(p_\theta(\hat{\mathbf{y}}\mid \mathbf{x})\) over possible output sequences \(\hat{\mathbf{y}}\). Define a loss function (or “risk”) \(\Delta(\hat{\mathbf{y}}, \mathbf{y})\) that measures the discrepancy between a predicted sequence \(\hat{\mathbf{y}}\) and the ground truth \(\mathbf{y}\) according to a specific metric. MRT then seeks to minimize the expected risk:

\begin{equation}
\label{eq:MRT1}
\mathcal{L}_{\mathrm{MRT}}
= \mathbb{E}_{\hat{\mathbf{y}}\sim p_\theta(\hat{\mathbf{y}}\mid \mathbf{x})}
  \bigl[\,
     \Delta(\hat{\mathbf{y}}, \mathbf{y})
  \bigr].
\end{equation}

\noindent
However, summing over all possible sequences \(\hat{\mathbf{y}}\in\mathcal{Y}\) is generally intractable. A common approach samples a subset \(\mathcal{Y}_s\subseteq \mathcal{Y}\) from \(p_\theta(\hat{\mathbf{y}}\mid\mathbf{x})\). The MRT objective becomes:

\begin{equation}
\label{eq:MRT2}
\mathcal{L}_{\mathrm{MRT}}
\,\approx\,
\sum_{\hat{\mathbf{y}}\in \mathcal{Y}_s}
  \frac{\exp\bigl(\alpha\,\log p_\theta(\hat{\mathbf{y}}\mid \mathbf{x})\bigr)}{
        \sum_{\hat{\mathbf{y}}'\in \mathcal{Y}_s}
        \exp\bigl(\alpha\,\log p_\theta(\hat{\mathbf{y}}'\mid \mathbf{x})\bigr)
        }
  \;\Delta(\hat{\mathbf{y}}, \mathbf{y}),
\end{equation}

\noindent
where \(\alpha>0\) scales or sharpens the distribution, and \(\Delta(\hat{\mathbf{y}}, \mathbf{y})\) is the sequence-level discrepancy (e.g., \(1 - \mathrm{BLEU}\)). Minimizing \(\mathcal{L}_{\mathrm{MRT}}\) pushes the model to assign higher probability to sequences with lower risk (i.e., better metric scores).

\paragraph{Applications:}
\begin{itemize}
  \item \textbf{Machine Translation}:  
    Directly optimize for BLEU or related translation metrics by defining \(\Delta(\hat{\mathbf{y}}, \mathbf{y}) = 1 - \mathrm{BLEU}(\hat{\mathbf{y}}, \mathbf{y})\). The model is incentivized to produce outputs scoring higher in BLEU, thus often correlating better with human judgments.
  \item \textbf{Text Summarization}:  
    Let \(\Delta(\hat{\mathbf{y}}, \mathbf{y}) = 1 - \mathrm{ROUGE}(\hat{\mathbf{y}}, \mathbf{y})\). Training under MRT encourages the system to maximize ROUGE overlap, leading to more coverage of essential information from the source text.
  \item \textbf{Dialog Systems or Generative QA}:  
    One can define a dialog metric \(\Delta\) capturing relevance and informativeness. MRT ensures the final outputs better align with conversation-level quality measures.
\end{itemize}

\paragraph{Advantages:}
\begin{itemize}
    \item \textbf{Closer Alignment with Evaluation Metrics:}  
    By training to minimize risk under a specific metric, MRT can yield improvements over token-level cross-entropy, especially for tasks demanding global coherence.
    \item \textbf{Sampling Complexity:}  
    Approximating the expectation requires sampling from the model distribution, which can be computationally expensive for large vocabularies or long sequences.
    \item \textbf{Choice of Scale and Loss Function:}  
    The scaling factor \(\alpha\) and the definition of \(\Delta(\hat{\mathbf{y}}, \mathbf{y})\) significantly influence stability and convergence. Too large an \(\alpha\) can cause overly peaked distributions, hampering exploration.
\end{itemize}

Overall, \emph{Minimum Risk Training} provides a powerful mechanism for sequence learning tasks where final performance is measured by specialized metrics (e.g., BLEU or ROUGE). By optimizing a direct alignment between training objectives and evaluation criteria, MRT can substantially boost the model’s end-task performance. However, it requires careful sampling strategies, hyperparameter tuning, and metric design to achieve stable, efficient training.

\subsubsection{REINFORCE Algorithm}
\label{loss:reinforce}

\emph{REINFORCE} \cite{williams1992simple} is a fundamental policy gradient method in reinforcement learning (RL) that has been adopted in sequence generation tasks within NLP to optimize \emph{non-differentiable} metrics (e.g., BLEU \cite{papineni2002bleu}, ROUGE \cite{lin2004rouge}). Instead of backpropagating through each token with a differentiable loss like cross-entropy, REINFORCE treats each generated sequence as an outcome of a \emph{stochastic policy}, using \emph{reward signals} from final outputs to guide parameter updates.

Let a model parameterized by \(\theta\) define a stochastic policy \(\pi_{\theta}\) over sequences \(\tau = (x_1, x_2, \dots, x_T)\). We interpret generating token \(x_t\) at step \(t\) as choosing an \emph{action}, and the entire generated sequence \(\tau\) then yields a scalar reward \(R(\tau)\) that reflects performance under some non-differentiable metric (e.g., BLEU for machine translation). The RL objective is to \emph{maximize} the expected reward:

\begin{equation}
\label{eq:reinforce_theta}
J(\theta)
= \mathbb{E}_{\tau \sim \pi_{\theta}} \bigl[R(\tau)\bigr].
\end{equation}

By the \emph{policy gradient theorem} \cite{sutton1999policy}, the gradient of \(J(\theta)\) w.r.t.\ \(\theta\) can be approximated by sampling trajectories \(\tau\) from \(\pi_{\theta}\) and computing:

\begin{equation}
\label{eq:reinforce_nabla}
\nabla_{\theta} J(\theta)
= \mathbb{E}_{\tau \sim \pi_{\theta}}
  \Bigl[
    R(\tau)\,\nabla_{\theta}\log\pi_{\theta}(\tau)
  \Bigr],
\end{equation}

\noindent
where \(\log \pi_{\theta}(\tau)\) is the log-probability of the sequence \(\tau\). This gradient term indicates that sequences \(\tau\) with higher rewards \(R(\tau)\) should have their log-probability increased, while lower-reward sequences should be deprioritized.

\paragraph{Algorithmic Update (REINFORCE Rule):}
Given a sample \(\tau\) from the current policy, we update parameters \(\theta\) via:

\begin{equation}
\label{eq:reinforce_delta}
\Delta \theta
= \eta\,R(\tau)\,\nabla_{\theta}\log\pi_{\theta}(\tau),
\end{equation}

\noindent
where \(\eta\) is a learning rate. Over multiple samples, the expectation in Equation~\eqref{eq:reinforce_nabla} is approximated.

\paragraph{Application to NLP Sequence Generation:}
\begin{itemize}
    \item \textbf{Machine Translation:}
      The model generates a translation \(\tau\) for an input sentence. A metric like BLEU is used as the reward \(R(\tau)\). By maximizing expected BLEU, the model aligns with how translations are ultimately scored.
    \item \textbf{Summarization:}
      Summaries are generated, then ROUGE or another measure of coverage is the reward. REINFORCE updates the model to produce more informative or coherent summaries.
    \item \textbf{Dialog Systems:}
      In conversation, custom rewards based on user satisfaction or engagement can be used. Each generated utterance leads to a final reward, guiding dialogue policy to produce relevant, context-aware responses.
\end{itemize}

\paragraph{Challenges:}
\begin{enumerate}
    \item \emph{High Variance in Gradient Estimates:}  
      Random sampling of sequences can lead to noisy gradient updates, slowing convergence. Baseline functions or variance-reduction methods (e.g., subtracting the average reward) help stabilize training.
    \item \emph{Credit Assignment:}  
      The reward is often only observed after generating an entire sequence, making it difficult to assign credit or blame to individual tokens. Techniques like reward shaping or partial intermediate rewards can alleviate this.
    \item \emph{Sample Efficiency:}  
      Large action spaces (vocabularies) and long sequences require many samples to accurately estimate expected rewards, potentially making REINFORCE computationally expensive in complex NLP tasks.
\end{enumerate}

\emph{REINFORCE} provides a principled approach for training NLP models on non-differentiable objectives by treating generation as a reinforcement learning problem. It can directly optimize final evaluation metrics, bridging the gap between local (token-level) training signals and global sequence-level quality. However, practitioners must address issues of variance, sample inefficiency, and delayed rewards to achieve stable, effective results in large-scale NLP applications.

\subsection{Performance Metrics Used in NLP}
\label{subsec:metrics_in_nlp}

Performance metrics provide quantifiable measures to assess how well a model performs a given task. In NLP, these metrics vary widely depending on the nature of the task and the type of output the model produces. For tasks involving classification, metrics like Accuracy, Precision, Recall, F1 Score, and AUC-ROC are crucial for determining the model's ability to correctly classify text into categories. While these metrics are commonly used in general classification problems, their application in NLP involves specific considerations, such as handling imbalanced datasets and the multi-label nature of some tasks.

Metrics like BLEU \cite{papineni2002bleu} and ROUGE \cite{lin2004rouge} are essential for sequence generation and text comparison tasks. These metrics evaluate the quality of generated text, such as translations or summaries, by comparing it to reference texts. They provide information on the fluency, adequacy, and informativeness of the generated content.

Other metrics, such as Perplexity \cite{jelinek1977perplexity}, are specific to language modeling and measure how well a model predicts a sequence of words. Exact Match is often used in tasks like question answering, where the precision of the model's output is critical.

In this section, we will explore these performance metrics in detail, focusing on their application in NLP tasks. We will discuss their definitions, how they are calculated, and the specific challenges and considerations when applying them to NLP models. This analysis will help in understanding the strengths and limitations of each metric and guide the selection of appropriate metrics for evaluating different NLP applications.

Table \ref{tab:guidelines-nlp-metrics} provides an overview of commonly employed metrics in NLP. It covers recommendations for use, data characteristics, advantages, and limitations.

\begin{table*}[ht!]
\caption{Guidelines for selecting a metric in NLP based on usage, data characteristics, advantages, and limitations.}
\label{tab:guidelines-nlp-metrics}
\centering
\footnotesize
\begin{tabular}{p{1.3cm}p{2.8cm}p{2.8cm}p{2.8cm}p{2.8cm}}
\toprule
Metric 
& Usage 
& Data Characteristics 
& Advantages 
& Limitations \\
\midrule

Accuracy 
& Classification tasks \newline General correctness measure 
& Suitable for balanced data \newline Discrete labels 
& Easy to interpret \newline Widely recognized 
& Misleading with imbalanced classes \newline Ignores nuances of errors \\
\midrule

Precision, Recall, F1 Score 
& Tasks requiring focus on false positives/negatives \newline Useful in NER, spam detection 
& Often imbalanced datasets \newline Varied cost of errors 
& Provide deeper insight than accuracy \newline F1 balances precision and recall 
& Hard to capture overall performance with all three \newline May ignore true negatives in some contexts \\
\midrule

AUC-ROC 
& Binary classification \newline Probabilistic outputs 
& Suited for large datasets \newline Typically two-class problems 
& Threshold-invariant \newline Summarizes TPR-FPR trade-off 
& May be less intuitive in multi-class \newline Can be over-optimistic with skewed data \\
\midrule

BLEU Score 
& Machine translation \newline Some text generation tasks 
& Requires reference translations \newline Text overlap-based 
& Widely adopted \newline Measures n-gram precision 
& Ignores synonyms \newline Overly reliant on reference variety \\
\midrule

METEOR
& Machine translation \newline Summarization \newline Text generation
& Reference-based \newline Token-level alignments \newline Uses lexical resources (synonyms, stemming)
& Considers precision and recall \newline Rewards synonyms, morphological variants \newline Often better correlation with human judgments 
& Relies on external dictionaries \newline Still mostly token-based \newline Single reference can penalize valid paraphrases \\
\midrule

ROUGE Score 
& Summarization \newline Content coverage tasks 
& Requires reference summaries \newline Focus on text overlap 
& Recall-oriented \newline Captures coverage of key content 
& Surface-level overlap \newline Limited semantic insight \\
\midrule

Perplexity 
& Language modeling \newline Text generation 
& Large, diverse corpora \newline Unconstrained vocabulary 
& Measures predictive fit \newline Standard for evaluating LM fluency 
& Depends on test corpus \newline Does not reflect semantic correctness \\
\midrule

Exact Match 
& QA, NLI \newline Tasks needing strict output 
& Typically short answers \newline Matching exact reference 
& Clear indication of full correctness \newline Binary measure 
& No partial credit \newline Fails with acceptable paraphrases \\
\midrule

WER 
& Speech recognition \newline Token-level output comparison 
& Sequence of words \newline Reference transcriptions 
& Direct measure of lexical errors \newline Intuitive fraction of incorrect words 
& Ignores semantic closeness \newline Penalizes all word errors equally \newline Highly domain-sensitive \\
\midrule

CER 
& Speech recognition \newline Character-based or script-based tasks 
& Character sequences \newline Useful for languages lacking clear word boundaries 
& Finer-grained than WER \newline Handles scripts with unclear token segmentation 
& No semantic insight \newline All character differences treated equally \newline Can over-penalize small edits \\
\bottomrule
\end{tabular}
\end{table*}

\subsubsection{Accuracy}
\label{metric:accuracy_nlp}

\emph{Accuracy} is a standard metric for classification tasks, evaluating the proportion of inputs classified correctly out of the total. Formally, for a dataset of size \(N\) with gold-standard labels \(\{y_i\}_{i=1}^N\) and model predictions \(\{\hat{y}_i\}_{i=1}^N\), the accuracy metric is:

\begin{equation}
\label{eq:acc_nlp}
\text{Accuracy}
= \frac{1}{N}
  \sum_{i=1}^N
  \mathbb{I}\bigl[\hat{y}_i = y_i\bigr],
\end{equation}

\noindent
where \(\mathbb{I}[\cdot]\) is the indicator function, returning 1 if the condition is true (the prediction matches the gold label) and 0 otherwise.

\paragraph{Usage in NLP:}
\begin{enumerate}
    \item \textbf{Text Classification:} 
    Tasks such as \emph{sentiment analysis}, \emph{spam detection}, and \emph{topic classification} involve mapping an input text to one of several discrete labels. Accuracy measures the fraction of texts for which the model assigns the correct label. For sentiment analysis, for instance, accuracy reflects how frequently the system correctly labels input as positive, negative, or neutral \cite{pang2008opinion}.

    \item \textbf{Sequence Labeling:}
    In \emph{part-of-speech tagging}, each token in a sentence receives a syntactic tag. Accuracy then represents the percentage of tokens correctly tagged. Similarly, in \emph{named entity recognition (NER)}, accuracy can measure the proportion of tokens correctly identified as person, location, organization, etc. \cite{schmid1994part,etzioni2005unsupervised}.

    \item \textbf{Language Detection:}
    For models that classify the language of a given text (e.g., English, Spanish, French), accuracy quantifies the fraction of texts for which the model identifies the correct language. 
\end{enumerate}

\paragraph{Advantages and Limitations:}
\begin{itemize}
    \item \textbf{Advantages:}
    \begin{itemize}
        \item \emph{Simplicity and Interpretability:} Accuracy is intuitive, requiring only a count of correct vs.\ incorrect predictions.
        \item \emph{Broad Applicability:} Suited for any classification or labeling scenario with discrete labels.
    \end{itemize}
    \item \textbf{Limitations:}
    \begin{itemize}
        \item \emph{Imbalanced Data}: In highly skewed class distributions, accuracy can be misleading if a model trivially predicts the majority class. 
        \item \emph{Ignores Fine-Grained Quality}: Accuracy only indicates correct vs.\ incorrect classification, not how confident or semantically close an incorrect prediction might be.  
        \item \emph{Context Sensitivity}: Accuracy fails to capture contextual or structured relationships in predictions (e.g., dependencies among tokens in sequence labeling).
    \end{itemize}
\end{itemize}

Despite its simplicity, \emph{accuracy} remains a fundamental metric for validating classification performance in many NLP tasks. Nevertheless, practitioners often complement it with other metrics (e.g., precision, recall, F1-score, or more specialized measures) to address class imbalance, structured dependencies, and nuanced performance considerations.

\subsubsection{Precision, Recall, and F1 Score}
\label{metric:precision_nlp}

In earlier sections (\S\ref{metric:precision}, \S\ref{metric:recall_tpr}, \S\ref{metric:f1}), we introduced the formal definitions of \emph{precision}, \emph{recall}, and the \emph{F1 score}, which combine both. In \emph{Natural Language Processing (NLP)}, these metrics are indispensable for tasks where correct identification or extraction of linguistic units is essential—particularly under conditions of data imbalance or different costs associated with false positives and false negatives.

\paragraph{Precision in NLP:}
Precision measures how many of the predicted positive (or relevant) instances truly belong to the positive class. It is often critical when the cost of a false positive is high. For instance:
\begin{itemize}
    \item \textbf{Named Entity Recognition (NER):}
      Precision indicates the fraction of model-identified entities that are actually correct (e.g., valid person, location, or organization). High precision ensures fewer spurious entity labels, which is essential in information extraction pipelines \cite{etzioni2005unsupervised}.
    \item \textbf{Text Classification (e.g., Spam Detection):}
      Among all emails labeled “spam,” precision reflects how many are truly spam. High precision is vital to avoid misclassifying important messages as spam.
\end{itemize}

\paragraph{Recall in NLP:}
Recall measures the proportion of true positive instances that the model successfully identifies. It is typically crucial when missing a positive instance is costly. Examples include:
\begin{itemize}
    \item \textbf{NER with High-Stakes Data:}
      Failing to detect critical entities (e.g., legal references, medical conditions) can have severe consequences. High recall helps ensure fewer missed entities.
    \item \textbf{Sentiment Analysis:}
      In some scenarios, detecting \emph{all} instances of a particular sentiment (e.g., all positive reviews) might be more important than mislabeling a few neutral items as positive \cite{pang2002thumbs}.
\end{itemize}

\paragraph{F1 Score in NLP:}
The \emph{F1 score} is the harmonic mean of precision and recall. It succinctly balances both metrics, which is valuable when class distributions are imbalanced or both false positives and false negatives incur significant costs:

\begin{equation}
\mathrm{F1} 
= 2 \times 
  \frac{(\mathrm{Precision} \times \mathrm{Recall})}{
        (\mathrm{Precision} + \mathrm{Recall})
       }.
\end{equation}

\noindent
In NLP tasks:
\begin{itemize}
    \item \textbf{Text Classification and NER:}
      F1 helps ensure that the model is not only precise but also captures the majority of positive cases. For instance, in legal or medical document classification, both incorrect inclusions and omissions carry high risk.
    \item \textbf{Information Retrieval / QA Systems:}
      Combining high precision (relevant documents or answers) and high recall (comprehensive retrieval) is often crucial. The F1 score offers a single measure balancing both.
\end{itemize}

\paragraph{Considerations:}
\begin{itemize}
    \item \textbf{Data Imbalance:}
      Precision and recall are especially important when the positive class is rare (e.g., spam vs.\ non-spam). Accuracy alone can be misleading if the negative class dominates.
    \item \textbf{Semantic Nuances:}
      In language tasks, near-miss cases (e.g., partial entity matches or closely related sentiments) can complicate strict definitions of precision and recall. Domain-specific guidelines sometimes refine how “true positives” are determined.
    \item \textbf{Task Suitability:}
      While these metrics are widely applicable to classification or sequence labeling, they may be less indicative of broader sequence coherence. Complementary metrics (e.g., BLEU, ROUGE) might be required for generative tasks.
\end{itemize}

In general, \emph{precision}, \emph{recall}, and \emph{F1 score} remain core metrics in NLP for classification- and labeling-based tasks. Their ability to tease apart the impact of false positives vs.\ false negatives—then combine these insights into a single balanced score—makes them fundamental to evaluating model performance, especially under imbalanced class distributions or cost-sensitive environments.

\subsubsection{AUC-ROC}
\label{metric:auc_roc_nlp}

In Section~\ref{metric:AUC}, we introduced \emph{Area Under the Receiver Operating Characteristic} (AUC-ROC) as a measure of a classifier’s ability to distinguish between positive and negative classes across various decision thresholds. In \emph{Natural Language Processing (NLP)}, \textbf{AUC-ROC} is especially valuable in binary classification tasks where the model outputs class probabilities (e.g., spam vs.\ non-spam). By plotting the \emph{true positive rate} (TPR) against the \emph{false positive rate} (FPR) at different threshold settings, the area under this ROC curve summarizes the model’s performance into a single scalar, effectively capturing how well the model discriminates between two classes overall.

\paragraph{Usage in Binary NLP Tasks:}
\begin{enumerate}
    \item \textbf{Spam Detection} \cite{sahami1998bayesian}:
    The model assigns a probability to each email/message indicating the likelihood of it being spam. The AUC-ROC measures how effectively the model ranks actual spam messages above legitimate ones across all possible thresholds. High AUC implies that, regardless of the specific threshold, the spam/non-spam ranking is robust.

    \item \textbf{Binary Sentiment Analysis} \cite{pang2002thumbs}:
    When classifying text into positive or negative sentiment, AUC-ROC indicates the trade-off between correctly classifying positive reviews (TPR) and misclassifying negative reviews as positive (FPR). A large area under the curve indicates a strong ability to separate positive from negative sentiments at multiple classification cutoffs.

    \item \textbf{Topic vs.\ Non-Topic Classification} \cite{joachims1998text,lewis1998naive,devlin2018bert}:
    For a scenario where the model classifies whether a document pertains to a particular topic (positive) or not (negative), AUC-ROC evaluates how well the model ranks truly relevant documents above irrelevant ones. This is central in content filtering or recommendation pipelines, where false positives can degrade user experience.
\end{enumerate}

\paragraph{Advantages and Considerations for NLP:}
\begin{itemize}
    \item \textbf{Threshold-Independent Assessment:}
      AUC-ROC evaluates classifier performance across all threshold settings, offering a more holistic assessment than accuracy at a fixed cutoff—especially relevant if one needs to calibrate the decision boundary post-training.
    \item \textbf{Robustness to Class Imbalance:}
      By examining TPR vs.\ FPR, AUC-ROC captures how the model handles skewed class distributions, often found in real-world text classification (e.g., spam detection).
    \item \textbf{Interpretation Cautions:}
      An AUC near 1.0 signifies robust discrimination. However, in highly imbalanced settings or with cost-sensitive decisions, practitioners may also consult metrics like precision-recall AUC (especially if the positive class is rare).
\end{itemize}

\emph{AUC-ROC} is a strong metric in NLP’s binary classification tasks that require probabilistic outputs (spam detection, sentiment polarity, topic vs.\ non-topic classification). Integrating performance across a continuous spectrum of decision thresholds offers a more comprehensive view than single-threshold metrics like accuracy, particularly under class imbalance or variable cost scenarios. 

\subsubsection{BLEU Score}
\label{metric:bleu_score}

\emph{BLEU (Bilingual Evaluation Understudy)} \cite{papineni2002bleu} is a widely adopted metric for evaluating machine-generated text, particularly in machine translation. It estimates how closely a candidate translation (or generated text) aligns with one or more human reference translations. BLEU focuses on matching overlapping \emph{n-grams} between candidate and references, capturing lexical and local ordering similarities.

\paragraph{Steps:}
Let the candidate output be \(\mathbf{c}\) and let \(\{\mathbf{r}^{(1)}, \mathbf{r}^{(2)}, \dots\}\) be a set of reference strings (translations). BLEU aggregates the following steps:

\begin{enumerate}
    \item \textbf{N-gram Precision Computation:}
    For \(n\)-grams of lengths \(n=1,2,\ldots,N\) (often up to 4), compute the “clipped” precision \(p_n\). That is, count how many \(n\)-grams in \(\mathbf{c}\) also appear in any reference, limiting each n-gram’s count to the maximum times it appears in any single reference. Formally:
    \begin{equation}
    p_n
    = \frac{
        \sum_{\textrm{n-gram}\in \mathbf{c}}
          \min\!\bigl[\mathrm{count}_{\mathbf{c}}(\textrm{n-gram}),
                      \max_{k}\mathrm{count}_{\mathbf{r}^{(k)}}(\textrm{n-gram}) 
                    \bigr]
      }{
        \sum_{\textrm{n-gram}\in \mathbf{c}}
          \mathrm{count}_{\mathbf{c}}(\textrm{n-gram})
      }.
    \end{equation}

    \item \textbf{Geometric Mean of N-gram Precisions:}
    Compute an average of \(\{\log p_1, \log p_2, \dots, \log p_N\}\). Typically, each \(p_n\) is equally weighted by \(\frac{1}{N}\).

    \item \textbf{Brevity Penalty (BP):}
    To penalize excessively short candidates, a factor \(\mathrm{BP}\) is applied if \(\mathbf{c}\) is shorter than the reference set. Let \(c = \lvert \mathbf{c}\rvert\) (length of the candidate) and \(r\) be the effective reference length (e.g., choosing the closest or shortest reference). The brevity penalty is:

    \begin{equation}
    \label{eq:brev_pen}
    \mathrm{BP}
    = \begin{cases}
       1, & \text{if } c > r,\\
       \exp\Bigl(1 - \frac{r}{c}\Bigr), & \text{if } c \leq r.
      \end{cases}
    \end{equation}

    \item \textbf{Final BLEU Score:}
    Combine the geometric mean of n-gram precisions with \(\mathrm{BP}\):
    \begin{equation}
    \label{eq:bleu}
    \mathrm{BLEU}
    = \mathrm{BP}\times
       \exp\Bigl(
         \sum_{n=1}^{N}
           w_n \,\log p_n
       \Bigr),
    \end{equation}
    where \(w_n\) are weights for each \(n\)-gram precision (often \(w_n = \frac{1}{N}\)).
\end{enumerate}

\paragraph{Applications in NLP:}
\begin{itemize}
    \item \textbf{Machine Translation \cite{vaswani2017attention}:}
      BLEU is a de facto standard metric, comparing candidate translations to human references. A higher BLEU indicates better lexical overlap. Although it does not measure semantic adequacy or fluency perfectly, BLEU remains a valuable baseline measure.
    \item \textbf{Text Summarization \cite{torres2014automatic}:}
      BLEU can be used to compare generated summaries against reference summaries. However, because summarization may allow paraphrasing or different sentence orderings, BLEU’s precision-based approach might underestimate the model’s actual summarization quality.
    \item \textbf{Text Generation \cite{heryanto2023evaluating,amin2022towards}:}
      In open-ended generation tasks, BLEU can help gauge how closely a generated response aligns with a known or expected sequence, though it may not fully capture creative or semantically correct variations.
\end{itemize}

\paragraph{Limitations:}
\begin{itemize}
    \item \textbf{Lexical Overlap Bias:}
      BLEU focuses on n-gram matches, ignoring semantic equivalences that use synonyms or rephrasings. Thus, perfectly valid translations may score lower if they differ lexically from references.
    \item \textbf{Reference Dependence:}
      More reference translations can capture diverse valid outputs. A single reference might penalize legitimate variations, lowering BLEU scores arbitrarily.
    \item \textbf{Precision Emphasis \& Brevity Penalty:}
      The brevity penalty deters overly short outputs, but the overall precision-based design can undervalue correct expansions or rephrasings absent in references.
\end{itemize}

Although \emph{BLEU} remains a dominant automatic metric for machine translation and related tasks, practitioners often supplement it with complementary metrics (e.g., METEOR \S\ref{metric:meteor}, ROUGE \S \ref{metric:rouge_score}) or human evaluations to obtain a fuller view of translation fluency and adequacy.

\subsubsection{METEOR}
\label{metric:meteor}

\emph{METEOR} (Metric for Evaluation of Translation with Explicit ORdering) \cite{banerjee2005meteor} is another automatic metric commonly used for evaluating machine translation outputs and other text generation tasks. Unlike BLEU (\S\ref{metric:bleu_score}), which relies on \(n\)-gram precision and brevity penalties, METEOR attempts to capture both precision and recall at the unigram level while also considering synonyms, stem matches, and phrase reordering to some extent. Consequently, it often correlates better with human judgments, particularly on more nuanced outputs.

\paragraph{Key Concepts:}
\begin{enumerate}
    \item \textbf{Token Matching:} 
    METEOR begins by aligning unigrams between a candidate (\(\mathbf{c}\)) and reference (\(\mathbf{r}\)) translation. Alignments consider exact token matches, stem or lemma matches, synonyms (using external lexical resources like WordNet), and optional paraphrase matches.
    \item \textbf{Precision and Recall:}
    Let \(m\) be the number of matched unigrams after alignments. Then, define:
    \begin{equation}
      P 
      = \frac{m}{\lvert \mathbf{c}\rvert}
      \quad\text{and}\quad
      R 
      = \frac{m}{\lvert \mathbf{r}\rvert},
    \end{equation}
    where \(\lvert \mathbf{c}\rvert\) and \(\lvert \mathbf{r}\rvert\) are the lengths (in tokens) of the candidate and reference, respectively.
    \item \textbf{Fragmentation Penalty:}
    METEOR adds a “fragmentation” or “chunking” penalty, which accounts for how well the matched tokens align in contiguous chunks. More scattered matches (i.e., many small segments) incur a higher penalty than fewer, larger segments.
\end{enumerate}

\paragraph{Scoring Formula:}
A common METEOR scoring variant uses the harmonic mean \(F_{\alpha}\) of \(P\) and \(R\), combined with a fragmentation penalty \(\mathrm{Pen}\). For example:

\begin{equation}
\label{eq:meteor_score}
\mathrm{METEOR}
= \bigl(1 - \mathrm{Pen}\bigr)\times
  \frac{(1+\alpha^2)\,P\,R}{
        \alpha^2\,P + R
       },
\end{equation}

\noindent
where \(\mathrm{Pen}\in[0,1]\) is computed from the number of matched segments vs.\ total matches, and \(\alpha\) is a parameter (often set around 0.9) controlling the relative weight of recall vs.\ precision.

\paragraph{Usage in NLP Tasks:}
\begin{itemize}
    \item \textbf{Machine Translation:}  
      METEOR aims to address some BLEU limitations by providing partial credit for synonyms or morphological variants. This can yield higher correlation with human judgments, especially in languages with rich morphology \cite{banerjee2005meteor}.
    \item \textbf{Text Summarization:}
      Like BLEU, METEOR can be employed to measure overlap between candidate summaries and reference summaries, taking into account stem/synonym matches to reward semantically equivalent expressions.
    \item \textbf{Any Generation Task with References:}
      For dialogues or paraphrase generation, METEOR offers a more flexible matching scheme than simple $n$-gram overlap, though it still may not fully capture discourse coherence or complex rephrasings.
\end{itemize}

\paragraph{Limitations:}
\begin{itemize}
    \item \textbf{Dependence on Lexical Resources:}  
      METEOR’s ability to handle synonyms or morphological variants relies on external dictionaries (e.g., WordNet), which may be domain-dependent or incomplete for certain languages.
    \item \textbf{Unigram Focus:}
      While chunk-based penalties help somewhat, METEOR mainly operates at the token level. It may still miss longer-range syntactic or semantic structures.
    \item \textbf{Reference Dependence:}
      Like BLEU, the metric is sensitive to the reference set’s coverage. A single reference might penalize valid alternate translations not reflected in the reference text.
\end{itemize}

\emph{METEOR} serves as an improvement over simpler $n$-gram matching metrics by incorporating a more flexible matching strategy and balancing precision, recall, and fragmentation considerations. While it does not capture global coherence or deep semantics fully, METEOR often correlates better with human evaluations than BLEU, making it an important complementary metric for machine translation, summarization, and other generation tasks.

\subsubsection{ROUGE Score}
\label{metric:rouge_score}

\emph{ROUGE (Recall-Oriented Understudy for Gisting Evaluation)} \cite{lin2004rouge} is a family of metrics used extensively for \emph{automatic summarization} and, to a lesser extent, \emph{machine translation} and other text generation tasks. ROUGE mainly focuses on \emph{recall} over various text overlaps (e.g., $n$-grams, sequences), indicating how much of the content in the reference text is recovered by the candidate text.

\paragraph{Core Variants:}
\begin{itemize}
    \item \textbf{ROUGE-N}: Measures the recall of $n$-gram overlaps between candidate ($\mathbf{c}$) and reference ($\mathbf{r}$). For ROUGE-1 (unigrams) or ROUGE-2 (bigrams), define
    \begin{equation}
    \label{eq:rouge-n}
    \mathrm{ROUGE\mbox{-}N}
    = \frac{
        \sum_{\text{gram}_n \in \mathbf{r}}
          \min\!\Bigl(
            \mathrm{count}_{\mathbf{c}}(\text{gram}_n),\,
            \mathrm{count}_{\mathbf{r}}(\text{gram}_n)
          \Bigr)
      }{
        \sum_{\text{gram}_n \in \mathbf{r}}
          \mathrm{count}_{\mathbf{r}}(\text{gram}_n)
      },
    \end{equation}
    where \(\mathrm{count}_{\mathbf{c}}\) and \(\mathrm{count}_{\mathbf{r}}\) are counts of \(n\)-grams in candidate and reference, respectively. Only overlapping $n$-grams that appear in both texts are credited, reflecting content recall.

    \item \textbf{ROUGE-L}: Uses the \emph{Longest Common Subsequence} (LCS) to capture the maximum sequence overlap, potentially skipping tokens in between. If $\mathrm{LCS}(\mathbf{r},\mathbf{c})$ is the length of the longest common subsequence, then:
    \begin{equation}
    \label{eq:rouge-l}
    \mathrm{ROUGE\mbox{-}L}
    = \frac{
        \mathrm{LCS}(\mathbf{r},\mathbf{c})
      }{
        \mathrm{Length}(\mathbf{r})
      }.
    \end{equation}
    This measure can detect re-ordered or spaced-out matches, better reflecting sentence-level structural overlap.

    \item \textbf{ROUGE-W}: A weighted extension of ROUGE-L emphasizing consecutive matches more heavily. Longer contiguous overlaps in word sequences yield higher scores.

    \item \textbf{ROUGE-S}: Also called \emph{skip-bigram} ROUGE, it counts pairs of words in the reference (in order) that appear in the candidate, allowing gaps. This addresses non-consecutive word matches.

\end{itemize}

\paragraph{Usage in NLP Tasks:}
\begin{enumerate}
    \item \textbf{Text Summarization:}
      ROUGE is the \emph{de facto} standard for measuring how much important content from the source text is retained in the candidate summary. High recall of key phrases or word sequences is crucial \cite{lin2004rouge}.
    \item \textbf{Machine Translation} \cite{lewis2019bart}:
      Although BLEU (\S\ref{metric:bleu_score}) is more common, ROUGE can provide additional recall-based insight, especially if capturing coverage of reference tokens is essential.
    \item \textbf{General Text Generation} \cite{wang2023automated}:
      ROUGE can help evaluate how comprehensive or on-topic the generated text is compared to a gold-standard reference, such as in question-answering or dialogue systems.
\end{enumerate}

As an example, \emph{ROUGE-N} for $n$-gram-based recall is:
\begin{equation}
\mathrm{ROUGE\mbox{-}N}
= \frac{\sum_{\text{gram}_n \in \mathbf{r}}
         \min\Bigl(\mathrm{count}_{\mathbf{c}}(\text{gram}_n),
                   \mathrm{count}_{\mathbf{r}}(\text{gram}_n)\Bigr)}
       {\sum_{\text{gram}_n \in \mathbf{r}}
         \mathrm{count}_{\mathbf{r}}(\text{gram}_n)},
\end{equation}
where the numerator sums clipped counts of $n$-grams that appear in both candidate and reference, reflecting how much of the reference $n$-gram space is covered by the candidate.

\paragraph{Considerations and Limitations:}
\begin{itemize}
    \item \textbf{Surface-Level Overlaps:}
      Like BLEU, ROUGE’s $n$-gram comparisons may undervalue semantically correct paraphrases or synonyms that differ lexically.
    \item \textbf{Dependency on Reference Diversity:}
      Multiple, varied references can improve the metric’s fairness by accommodating valid expression alternatives. A single reference might penalize correct solutions not reflected in its wording.
    \item \textbf{Emphasis on Recall:}
      ROUGE is intentionally recall-oriented, which can over-reward longer candidates that contain extraneous material. Combining with precision-oriented measures or brevity controls might be necessary.
\end{itemize}

\emph{ROUGE} has become a mainstay for automatic summarization evaluation and is also applicable to other generation tasks needing content-coverage assessments. Although it provides a systematic approach to measuring $n$-gram and subsequence recall, users should be aware of its limited semantic sensitivity and reliance on reference coverage.

\subsubsection{Perplexity}
\label{metric:perplexity}

\emph{Perplexity} \cite{jelinek1977perplexity} is a standard metric for evaluating the predictive power of language models, reflecting how “surprised” a model is by observed data. Formally, it measures the geometric mean of the inverse probabilities that a model assigns to each token in a sequence. Lower perplexity indicates greater confidence and accuracy in the model’s token predictions, suggesting it captures language structure more effectively.

\paragraph{Definition:}
Let \(\mathbf{w} = (w_1, w_2, \dots, w_T)\) be a sequence of \(T\) words. A language model defines a conditional probability \(p(w_t \mid w_{1:t-1})\) for each word \(w_t\). The \emph{perplexity} of this model on the sequence \(\mathbf{w}\) is:

\begin{equation}
\label{eq:perplex}
\mathrm{Perplexity}(\mathbf{w})
= \exp\!\Bigl(
   -\frac{1}{T}
    \sum_{t=1}^T
      \log p\bigl(w_t \mid w_{1:t-1}\bigr)
  \Bigr).
\end{equation}

\noindent
Equivalently, perplexity can be viewed as \(\exp(H)\), where \(H\) is the cross-entropy of the model distribution relative to the empirical distribution of the words in \(\mathbf{w}\).

\paragraph{Usage in NLP:}
\begin{enumerate}
    \item \textbf{Language Modeling:} 
    Perplexity is often the principal metric for \emph{n-gram models} \cite{masataki1997task,li2022n}, \emph{RNN-based} \cite{sundermeyer2015feedforward}, or \emph{transformer-based} language models \cite{radford2019language,brown2020language}. A lower perplexity on a test set indicates that the model better predicts held-out text and thus better captures the syntactic/semantic structure of the language.

    \item \textbf{Text Generation:}
    In chatbots and other generative systems, perplexity reflects how fluently or coherently the model can produce text. A model with lower perplexity generally generates fewer implausible token sequences and exhibits more context consistency.

    \item \textbf{Embedded LM in Other Tasks:}
    Perplexity is also used in \emph{speech recognition} or \emph{machine translation} systems to evaluate the quality of the underlying language model. Better perplexity often correlates with improved final task performance, though it is not always strictly deterministic.
\end{enumerate}

\paragraph{Advantages and Limitations:}
\begin{itemize}
    \item \textbf{Advantages:}
    \begin{itemize}
      \item \emph{Direct Measure of Predictive Quality}: 
        Because perplexity is derived from probabilities assigned to the observed tokens, it succinctly captures how well the model “fits” the data.
      \item \emph{Broad Applicability}: 
        Perplexity is easy to compute for any language model that provides next-token probabilities.
    \end{itemize}

    \item \textbf{Limitations:}
    \begin{itemize}
      \item \emph{Corpus/Domain Sensitivity}: 
        Perplexity scores can vary significantly across corpora with different vocabularies or styles. Comparisons are meaningful only if done on the same test set or similar domains.
      \item \emph{Vocabulary Size Effects}:
        Increasing the vocabulary might inflate perplexity, as the model spreads probability mass over more tokens. Normalization or consistent vocabulary usage is essential for fair evaluation.
      \item \emph{Calibration Assumption}:
        Perplexity presupposes well-calibrated probabilities. Overconfident or poorly calibrated models may yield deceptively low (or high) perplexities that do not translate to better generative performance.
    \end{itemize}
\end{itemize}

\emph{Perplexity} remains the cornerstone metric for language modeling, measuring the model’s uncertainty about token sequences. Low perplexity typically indicates strong predictive capacity and alignment with linguistic regularities. Nevertheless, for certain tasks requiring sequence-level metrics (e.g., BLEU or ROUGE), perplexity should be complemented by evaluations more focused on semantic or global coherence aspects of text.

\subsubsection{Exact Match (EM)}
\label{metric:exact_match_nlp}

\emph{Exact Match (EM)} \cite{rajpurkar2016squad} is a binary evaluation metric particularly useful when the \emph{entire} predicted output must match the reference output verbatim. It is widely adopted in tasks where minor deviations (such as rewording, word order, or formatting differences) invalidate the answer or reduce its utility.

\paragraph{Definition:}
Let there be \(N\) examples, each with a ground-truth output \(y_i\) and a predicted output \(\hat{y}_i\). The EM metric assigns 1 if \(\hat{y}_i\) is \emph{exactly} equal to \(y_i\) (token-for-token, character-for-character) and 0 otherwise:

\begin{equation}
\label{eq:ex_match}
\text{Exact Match}
= \frac{1}{N}
  \sum_{i=1}^{N}
  \mathbb{I}\bigl(y_i = \hat{y}_i\bigr),
\end{equation}

\noindent
where \(\mathbb{I}(\cdot)\) is an indicator function returning 1 if the condition is true and 0 otherwise.

\paragraph{Usage in NLP:}
\begin{enumerate}
    \item \textbf{Extractive Question Answering (QA)} \cite{rajpurkar2016squad}:  
    In tasks where the answer must be an exact substring (span) of a reference context, EM measures how frequently the predicted span perfectly matches the gold span. Even slight deviations—such as an omitted word—yield no credit.

    \item \textbf{Natural Language Inference (NLI)}:  
    When the model outputs one of a discrete set of labels (e.g., “entailment,” “contradiction,” “neutral”), EM checks if the predicted label precisely matches the correct label, making it a straightforward accuracy measure in a 3-class setting.

    \item \textbf{Strictly Constrained Generation}:  
    In specialized machine translation or summarization tasks (e.g., legal text or heavily format-dependent outputs), EM ensures the model’s output reproduces an expected sequence exactly, leaving no room for paraphrase. Similarly, code generation or formula generation can rely on EM to verify syntactic correctness.

    \item \textbf{Label Prediction with Immutable Formats}:  
    If a classification label or keyword-based annotation must match a reference’s exact format (e.g., a specific category name, diagnostic code), EM confirms total match.
\end{enumerate}

\paragraph{Limitations:}
\begin{itemize}
    \item \textbf{Absolute Rigidity}:  
    EM provides a score of 0 unless the prediction is an exact surface-form match, which can be overly strict in tasks where synonyms or slight re-phrasings are acceptable.
    \item \textbf{No Partial Credit}:  
    The metric does not differentiate between predictions that are slightly off and those that are completely incorrect. Thus, it fails to capture near-correct attempts.
    \item \textbf{Semantic Insensitivity}:  
    EM checks for literal equivalence and cannot assess deeper meaning. Tasks emphasizing semantic correctness (rather than exact wording) may find EM too unforgiving.
\end{itemize}

\emph{Exact Match} is a reliable, straightforward metric in high-stakes or format-specific scenarios, such as extractive QA or code generation, where any discrepancy from the reference invalidates the model output. However, due to its binary nature and strict string matching, EM often requires supplementation with other metrics or human evaluation for tasks permitting equivalent, paraphrased, or partially correct responses.

\subsubsection{Word Error Rate (WER)}
\label{metric:wer}

\emph{Word Error Rate (WER)} is a standard metric for evaluating speech recognition or any system that produces a tokenized output (e.g., words) against a reference transcription. It quantifies the total number of substitutions, deletions, and insertions needed to transform the predicted word sequence into the reference sequence.

\paragraph{Definition:}
Let the reference be \(\mathbf{r} = (r_1, r_2, \dots, r_M)\) of length \(M\), and the hypothesis (model output) be \(\mathbf{h} = (h_1, h_2, \dots, h_N)\) of length \(N\). WER is computed as:
\begin{equation}
\mathrm{WER}
= \frac{S + D + I}{M},
\end{equation}
where 
\begin{itemize}
  \item \(S\) is the number of substitutions (words in \(\mathbf{h}\) that differ from reference words at matching positions),
  \item \(D\) is the number of deletions (words in \(\mathbf{r}\) that have no corresponding word in \(\mathbf{h}\)),
  \item \(I\) is the number of insertions (extra words in \(\mathbf{h}\) not found in \(\mathbf{r}\)).
\end{itemize}
Finding \(S\), \(D\), and \(I\) can be done via the Levenshtein (edit) distance algorithm, aligning words between the hypothesis and reference.

\paragraph{Usage in NLP:}
\begin{itemize}
  \item \textbf{Speech Recognition:} 
  The de facto metric for comparing automatically recognized transcripts to their ground-truth references. A lower WER indicates fewer word-level errors.
  \newline
  \item \textbf{Spoken Language Translation or Dictation Systems:} 
  Evaluating if the transcribed text in a target language or domain closely matches a reference script or user’s intended words.
\end{itemize}

\paragraph{Advantages:}
\begin{itemize}
  \item \emph{Easy Interpretation}: WER directly corresponds to a fraction of incorrect words, making it intuitive for end-users.
  \newline
  \item \emph{Complete Overlap Check}: Captures the count of exact lexical mismatches (substitutions, insertions, deletions).
\end{itemize}

\paragraph{Limitations:}
\begin{itemize}
  \item \emph{No Semantic Equivalence}: Fails to give partial credit if a semantically correct but lexically different word is used.
  \newline
  \item \emph{Same Penalty for All Errors}: Substitutions vs.\ insertions/deletions are all included. This may not reflect user perceptions (some errors might be more severe).
  \newline
  \item \emph{Vocabulary and Domain Specificity}: Uncommon words or domain-specific terms can inflate WER if the model struggles with them.
\end{itemize}

WER is widely used to benchmark ASR (Automatic Speech Recognition) systems, but can be complemented with other measures like \(\mathrm{CER}\) (Character Error Rate) or domain-specific metrics to capture subtleties of language use and user tolerance for certain error types.

\subsubsection{Character Error Rate (CER)}
\label{metric:cer}

\emph{Character Error Rate (CER)} parallels WER but measures errors at the \emph{character} (or sometimes subword) level rather than words. This can be especially relevant for languages with agglutinative morphology or for tasks involving scripts where word segmentation is less clear.

\paragraph{Definition:}
Given a reference string \(\mathbf{r} = (r_1, r_2, \dots, r_P)\) of length \(P\) in characters and a hypothesis string \(\mathbf{h} = (h_1, h_2, \dots, h_Q)\) of length \(Q\), the CER is:

\begin{equation}
\mathrm{CER}
= \frac{S + D + I}{P},
\end{equation}
where 
\begin{itemize}
  \item \(S\) is the number of substituted characters,
  \item \(D\) is the number of deleted characters,
  \item \(I\) is the number of inserted characters.
\end{itemize}
This again corresponds to the edit distance in characters, normalized by the reference length \(P\).

\paragraph{Usage in NLP:}
\begin{itemize}
  \item \textbf{Speech Recognition for Character-Based Languages:} 
  When the system outputs characters directly (e.g., in Chinese or Japanese), CER can be more precise than WER because word boundaries might be ambiguous.
  \newline
  \item \textbf{ASR in Low-Resource Scripts or Morphologies:} 
  For languages where consistent word segmentation is not well-defined, character-level evaluation can be more stable.
  \newline
  \item \textbf{Handwriting Recognition:} 
  CER is also commonly used if the system outputs textual characters from an image-based input.
\end{itemize}

\paragraph{Advantages:}
\begin{itemize}
  \item \emph{More Granular Than WER}: 
  Reflects partial correctness in words if only some characters are off, offering finer error insight.
  \newline
  \item \emph{Language Agnostic}: 
  Avoids complexities of word segmentation, especially valuable in languages with no whitespace or morphological complexity.
\end{itemize}

\paragraph{Limitations:}
\begin{itemize}
  \item \emph{No Semantic Interpretation}: 
  As with WER, CER penalizes any character difference equally, ignoring synonyms or small morphological variants with the same meaning.
  \newline
  \item \emph{Not Always Reflecting Human-Readable Errors}: 
  A few character changes can drastically alter a word’s meaning, but CER does not weigh severity of those changes differently.
\end{itemize}

CER is essential in scripts or domains where word-based segmentation is artificial or prone to errors. Together with WER, it provides complementary perspectives on the accuracy of textual outputs at different granularities.

\section{Retrieval-Augmented Generation (RAG)}
\label{sec:RAG}

Retrieval-Augmented Generation (RAG) \cite{lewis2020retrieval} combines retrieval-based and generation-based NLP models to enrich tasks requiring substantial knowledge. By merging large language models (LLMs) with external knowledge sources, RAG enhances the accuracy, factuality, and relevance of generated content.

\paragraph{RAG Components:}
\begin{enumerate}
    \item \textbf{Retrieval:}
    A dedicated neural retriever and dense index query external databases or knowledge bases. This step fetches relevant passages or records, allowing the system to incorporate up-to-date information. The retrieval mechanism is crucial for tasks like open-domain QA or domain-specific applications, where static model knowledge may be insufficient \cite{lewis2020retrieval}.

    \item \textbf{Generation:}
    A pre-trained language model synthesizes coherent, context-aware text. It leverages both its internal learned representations and the retrieved external data to produce responses that are contextually aligned and factually supported.

    \item \textbf{Augmentation:}
    The retrieved content is combined or appended to the model’s input representation, enriching the model’s generative process. This reduces hallucinations and outdated knowledge, since the LLM can refer to the retrieved material for accurate details and domain-specific terminologies.
\end{enumerate}

RAG excels in tasks demanding broad or specialized knowledge—such as answering open-domain questions, drafting legal documents, or specialized advisory systems (e.g., TaxTajweez \cite{chouhan2024lexdrafter,habib2024taxtajweez}). By integrating external databases, RAG improves the precision and reliability of model outputs, crucial for applications where correctness is paramount. In addition, RAG can dynamically adapt to evolving knowledge, seamlessly incorporating new data over time \cite{gao2023retrieval}.

\subsection{Loss Functions Used in RAG}
\label{loss:RAG}

Although \emph{Retrieval-Augmented Generation (RAG)} systems can achieve strong performance without explicit fine-tuning—often by simply prompting a pre-trained large language model (LLM) to condition on retrieved information—certain applications demand deeper domain adaptation or more reliable retrieval. In such cases, fine-tuning allows the system to better integrate external knowledge and produce outputs aligned with specific domains, tasks, or user requirements. Below, we discuss common loss functions used \emph{if} RAG requires fine-tuning, but many practical scenarios rely on carefully constructed prompts alone.

\begin{enumerate}
    \item \textbf{Cross-Entropy Loss:} 
    Typical for sequence-to-sequence training, improving the generation module’s ability to produce correct target tokens. During fine-tuning, the model adjusts its vocabulary-level probabilitieto handle domain-specific terms betterms and reduce off-topic generation. Refer to \S\ref{loss:t-cce}.

    \item \textbf{Contrastive Loss:}
    Applied when refining the retrieval module to differentiate relevant documents from irrelevant ones. By embedding queries and documents in a shared space, contrastive loss ensures higher similarity for relevant query-document pairs than for negative pairs. Refer to \S\ref{loss:contrastive}

    \item \textbf{Marginal Ranking Loss:}
    Another retrieval-oriented objective that enforces a margin by which relevant documents must outrank irrelevant ones. This is valuable in large corpora scenarios, where fine-tuned ranking can significantly improve retrieved passages. Refer to \S\ref{loss:marginal_ranking_nlp}.

    \item \textbf{Negative Log-Likelihood (NLL):}
    Often used interchangeably with cross-entropy in sequence-to-sequence contexts. Fine-tuning on NLL encourages the model to assign higher probability mass to correct token sequences, beneficial for tasks requiring precise or domain-aware generation.

    \item \textbf{KL Divergence Loss:}
    Employed in teacher-student or multi-component RAG settings to align probability distributions among sub-models. For instance, a teacher model’s output can guide a student model to produce similar distributions for relevant passages or token probabilities. Refer to \S\ref{loss:kl-loss}.
\end{enumerate}

Although many RAG systems work effectively via \emph{prompting only}, the option of fine-tuning remains critical for specialized domains or high-stakes use cases. By applying these losses—targeting either retrieval or generation—the fine-tuned RAG system learns to retrieve highly pertinent information and produce domain-specific text with enhanced accuracy and relevance.

\subsection{Performance Metrics Used in RAG}
\label{subsec:metrics:RAG}

Evaluating Retrieval-Augmented Generation (RAG) systems poses unique challenges due to their \emph{hybrid} structure—integrating both a retrieval mechanism (to fetch relevant external context) and a generation model (to produce final responses). This dual component design requires evaluating:

\begin{enumerate}
    \item \textbf{Retrieval Quality:} How accurately and precisely the system fetches relevant information from external knowledge sources.
    \item \textbf{Generation Quality:} How the model uses the retrieved content to formulate coherent, contextually grounded, and factually correct outputs.
\end{enumerate}

\noindent
Conventional metrics (e.g., BLEU, ROUGE) often assume ground-truth references exist, but in many real-world RAG scenarios, explicit references or labeled answers may be unavailable \cite{salemi2024evaluating,rackauckas2024evaluating}. Evaluators thus turn to frameworks like RAGAS \cite{es2023ragas} and ARES \cite{saad2023ares}, which propose reference-free or partially reference-based metrics to handle diverse RAG use cases (e.g., question answering, knowledge-grounded dialogue).

\paragraph{Evaluation Considerations:}
\begin{itemize}
    \item \textbf{Absence of Strict Reference Answers:} 
    RAG tasks often address open-domain or knowledge-intensive queries for which no official ground truth is provided. Metrics must allow partial or reference-free evaluation and handle issues like model hallucination \cite{salemi2024evaluating}.
    \item \textbf{Faithfulness to Retrieved Context:} 
    Ensuring that generated responses align with retrieved passages (i.e., do not hallucinate beyond the evidence) demands specialized metrics that cross-check model statements against the retrieved text \cite{es2023ragas}.
    \item \textbf{Relevance and Focus in Retrieval:} 
    The system should extract passages that are on-topic and concise, avoiding extraneous or off-topic data. Evaluating retrieval precision is essential to ensure the final generation is not contaminated by irrelevant or noisy content.
\end{itemize}

Table \ref{tab:guidelines-rag-metrics} summarizes the metrics employed to assess RAG systems, highlighting their benefits and drawbacks.

\begin{table*}[ht!]
\caption{Guidelines for selecting a performance metric in RAG-based systems, based on usage, data characteristics, advantages, and limitations.}
\label{tab:guidelines-rag-metrics}
\centering
\footnotesize
\begin{tabular}{p{1.3cm}p{2.8cm}p{2.8cm}p{2.8cm}p{2.8cm}}
\toprule
Function 
& Usage 
& Data Characteristics 
& Advantages 
& Limitations \\
\midrule

answer semantic similarity 
& Checks semantic closeness to reference
& Embedding or cross-encoder\newline needs ground-truth 
& Captures meaning not just words
& No guarantee of factual accuracy\newline depends on embedding quality \\
\midrule

answer correctness 
& Combines token F1 and semantic similarity
& Requires reference answer\newline merges lexical + semantic checks 
& Penalizes missing tokens\newline rewards accurate rephrasing
& Sensitive to small token differences\newline weighting factor needed \\
\midrule

answer relevance 
& Rates topical alignment with the question
& Embedding-based “reverse queries”\newline does not test correctness 
& Detects off-topic or redundant text
& Relevant yet possibly incorrect\newline relies on LLM to generate reverse queries \\
\midrule

context precision 
& Emphasizes high-rank relevance
& Binary relevant/irrelevant labeling\newline top K chunks 
& Ensures key evidence is top-ranked
& Ignores recall\newline no coverage guarantee \\
\midrule

context recall 
& Checks if retrieval covers correct answer
& Matches ground-truth statements\newline ratio of “supported” content 
& Measures completeness of retrieval
& No direct link to generation\newline partial coverage untracked \\
\midrule

faithfulness 
& Ensures generated text is grounded in context
& Splits output into claims\newline verifies each 
& Reduces hallucination\newline crucial for factual tasks
& Needs reliable claim-checking\newline does not measure relevance \\
\midrule

summarization score 
& Balances coverage vs.\ brevity
& QA on key points + length ratio
& Encourages concise, accurate summaries
& Relies on question generation\newline synonyms might lower QA score \\
\midrule

context entities recall 
& Ensures key named entities are preserved
& Requires entity detection in both context and output
& Good for fact-focused tasks
& Fails on synonyms\newline depends on NER accuracy \\
\midrule

aspect critique 
& Binary checks for custom aspects (e.g., harmlessness)
& Repeated LLM prompts for yes/no verdict 
& Fast content moderation\newline flexible aspects
& Subject to LLM bias\newline yes/no may be simplistic \\
\midrule

context relevance (ARES) 
& Classifies passage-query pairs for alignment
& Contrastive training with positives/negatives
& Improves retrieval targeting\newline uses confidence intervals
& Discrete notion of relevance\newline depends on negative sample quality \\
\midrule

answer faithfulness (ARES) 
& Tests if answer is faithful to retrieved passages
& Synthetic faithful/unfaithful sets\newline contrastive approach
& Reduces hallucination\newline fosters context-based correctness
& Needs good example generation\newline subtle contradictions may slip \\
\midrule

answer relevance (ARES) 
& Rates if response addresses the query
& Pos/neg samples\newline another LM judge
& Ensures query alignment\newline complements correctness
& May miss factual errors\newline irrelevant details can be true \\
\bottomrule
\end{tabular}
\end{table*}

In the following, we delve into the metrics provided by RAGAS and ARES.

\subsubsection{Answer Semantic Similarity}
\label{metric:ragas_seman_simil}

\emph{Answer Semantic Similarity} measures how closely the model’s generated answer aligns semantically with a reference or “ground-truth” answer. RAGAS \cite{es2023ragas} computes this score via a cross-encoder or embedding-based approach, typically using a cosine similarity:

\begin{equation}
\label{eq:sem_sim}
\mathrm{SemanticSimilarity}
= \cos\bigl(\mathbf{E}_g,\; \mathbf{E}_t\bigr),
\end{equation}

\noindent
where
\begin{itemize}
    \item \(\mathbf{E}_g\) is the embedding (vector) of the generated answer,
    \item \(\mathbf{E}_t\) is the embedding of the target or reference answer,
    \item \(\cos(\cdot)\) is the cosine similarity function, yielding values in \([-1,1]\).
\end{itemize}

In practice, the score is often scaled or clipped to lie in \([0,1]\). Higher values indicate closer semantic overlap, even if the surface wording differs. This metric helps reveal whether the model captures the essential meaning of an answer, rather than just lexical matches.

\subsubsection{Answer Correctness}
\label{metric:ragas_ans_correct}

\emph{Answer Correctness} assesses both the factual accuracy and semantic coherence of the response, combining a token-level F1 measure with the previously described semantic similarity. Formally, RAGAS \cite{es2023ragas,ragas_answer_correctness} defines:

\begin{equation}
\label{eq:ans_corr}
\mathrm{AnswerCorrectness}
= \alpha \,\times \,\mathrm{F1}
\;+\;
(1 - \alpha)\,\times \,\mathrm{SS},
\end{equation}

\noindent
where:
\begin{itemize}
    \item \(\mathrm{SS}\) is the semantic similarity from Equation~\eqref{eq:sem_sim},
    \item \(\mathrm{F1}\) is a token-level F1 score, capturing overlap in terms of lexical correctness,
    \item \(\alpha\in [0,1]\) is a weighting factor controlling the relative importance of F1 vs.\ semantic similarity.
\end{itemize}

The token-level F1 score \(\mathrm{F1}\) is computed as:

\begin{equation}
\label{eq:f1_ans_corr}
\mathrm{F1}
= \frac{2 \,\times\, \mathrm{Precision}\,\times\,\mathrm{Recall}}
        {\mathrm{Precision} \;+\;\mathrm{Recall}},
\end{equation}

\noindent
where 
\(\mathrm{Precision} = \frac{\mathrm{TP}}{\mathrm{TP}+\mathrm{FP}}\)
and
\(\mathrm{Recall}   = \frac{\mathrm{TP}}{\mathrm{TP}+\mathrm{FN}}\),
with:
\begin{itemize}
    \item \(\mathrm{TP}\) (true positives): tokens in the generated answer matching those in the ground truth,
    \item \(\mathrm{FP}\) (false positives): extra or incorrect tokens in the generated answer absent in the ground truth,
    \item \(\mathrm{FN}\) (false negatives): tokens in the ground truth that the model’s answer omitted.
\end{itemize}

By combining a token-level overlap check (\(\mathrm{F1}\)) with an embedding-based similarity check (\(\mathrm{SS}\)), the \emph{Answer Correctness} metric aims to penalize missing or incorrect tokens while rewarding semantically accurate rephrasings. A higher score signals a more comprehensive and faithful response to the question or query.

\subsubsection{Answer Relevance}
\label{metric:ragas_ans_rel}

\emph{Answer Relevance} \cite{ragas_answer_relevance} quantifies how well the generated answer addresses the posed question in terms of topicality and completeness, independently of factual accuracy. A relevant answer should not be redundant or off-topic, and it should align with the question’s intent. 

\paragraph{Definition:}
Let \(Q\) be the original question, and let \(\mathbf{A}\) be the model’s generated answer. Using a language model, we generate \(N\) “reverse-engineered” queries \(\{Q_{g_1}, Q_{g_2}, \ldots, Q_{g_N}\}\) by prompting the model to produce plausible questions that could lead to \(\mathbf{A}\). Each \(Q_{g_i}\) is then embedded into a vector space (e.g., via a sentence encoder) to obtain \(\mathbf{E}_{g_i}\). The original question \(Q\) is similarly embedded as \(\mathbf{E}_o\). The \emph{Answer Relevance} is computed by averaging the cosine similarities:

\begin{equation}
\label{eq:ans_rel}
\text{AnswerRelevance}
= \frac{1}{N}
  \sum_{i=1}^{N}
    \cos\!\bigl(\mathbf{E}_{g_i}, \mathbf{E}_o\bigr),
\end{equation}

\noindent
where \(\cos(\cdot,\cdot)\) is the cosine similarity function. A higher value indicates that, on average, the “reverse-engineered” queries resemble the original question in embedding space, suggesting the generated answer remains close to the question’s topic.

\paragraph{Interpretation:}
\begin{itemize}
    \item \emph{Completeness vs.\ Redundancy:}
    Answers failing to address the question or containing extraneous text typically yield lower cosine similarity to the original query.
    \item \emph{No Accuracy Guarantee:}
    Since this metric does not incorporate factual correctness, an answer can be “relevant” yet factually incorrect. 
\end{itemize}

\noindent
Thus, \emph{Answer Relevance} captures how directly the system’s output responds to the question’s core intent, leaving exact correctness to other metrics (\S\ref{metric:ragas_ans_correct}, etc.).

\subsubsection{Context Precision}
\label{metric:ragas_context_prec}

\emph{Context Precision} \cite{es2023ragas,ragas_context_precision} evaluates how well a system ranks relevant context items at the top among all retrieved passages or chunks. The rationale is that truly pertinent evidence should appear in the highest ranks, maximizing the precision in early positions. 

\paragraph{Score Formulation:}
Consider a set of $K$ retrieved context chunks $\{c_1, \dots, c_K\}$ sorted by relevance score. Let $v_k$ be 1 if $c_k$ is relevant to the question and 0 otherwise. Define $\mathrm{Precision@}k$ as:

\begin{equation}
\label{eq:cont_prec2}
\mathrm{Precision@}k 
= \frac{\mathrm{TP@}k}{\mathrm{TP@}k + \mathrm{FP@}k},
\end{equation}

\noindent
where $\mathrm{TP@}k$ is the count of true positives in the top $k$ ranks (i.e., relevant chunks among the first $k$), and $\mathrm{FP@}k$ is the count of false positives. The \emph{Context Precision} at top $K$ chunks, $\mathrm{CP@K}$, is then:

\begin{equation}
\label{eq:cont_prec}
\mathrm{CP@K}
= \frac{\sum_{k=1}^{K}
   \bigl(\mathrm{Precision@}k \times v_k\bigr)}
  {\text{Total relevant items in top }K}.
\end{equation}

A higher $\mathrm{CP@K}$ indicates that relevant chunks consistently appear at the top. Unlike simple precision, $\mathrm{CP@K}$ weights each rank $k$ by whether the chunk at that rank is truly relevant. 

\paragraph{Interpretation:}
\begin{itemize}
    \item \emph{Emphasis on Early Ranks:}
    Critical in RAG systems where only top $K$ chunks feed into the generation step. Poorly ranked relevant chunks effectively vanish if not included in the final context.
    \item \emph{No Recall Component:}
    A system can have high context precision but fail to retrieve all relevant chunks if those items are missing from the top $K$ set altogether.
\end{itemize}

\noindent
\emph{Context Precision} thus provides a lens on \emph{how well} the retrieval module prioritizes relevant evidence in the top positions crucial for the subsequent generation.

\subsubsection{Context Recall}
\label{metric:ragas_context_rec}

\emph{Context Recall} \cite{es2023ragas,ragas_context_recall} measures the extent to which the retrieved context supports the correct (ground truth) answer. Intuitively, all statements within the reference answer should be attributable to at least some portion of the retrieved text. The higher the ratio of “attributable” statements, the better the context coverage. Mathematically:

\begin{equation}
\label{eq:cont_recall}
\text{Context Recall}
= \frac{\lvert \mathrm{GT}_{\mathrm{sentences\,attributed}} \rvert}
        {\lvert \mathrm{GT}_{\mathrm{total\,sentences}} \rvert},
\end{equation}

\noindent
where 
\(\mathrm{GT}_{\mathrm{sentences\,attributed}}\) are the ground-truth answer’s sentences that can be linked to (or found in) the retrieved context, and 
\(\mathrm{GT}_{\mathrm{total\,sentences}}\) is the total number of sentences in the ground-truth answer. A higher Context Recall means the system retrieved enough relevant passages so that each statement in the correct answer is well-supported by the retrieved material.

\paragraph{Interpretation:}
\begin{itemize}
    \item \emph{Completeness of Retrieval:}
    A perfect score (1.0) indicates that every statement from the ground truth has corresponding evidence in the retrieved text.
    \item \emph{Ignoring Generation Quality:}
    Context Recall only evaluates retrieval completeness, not whether the final generated answer is accurate. A system may retrieve relevant data yet generate an incorrect response (or vice versa).
\end{itemize}

\subsubsection{Faithfulness}
\label{metric:ragas_faithfulness}

\emph{Faithfulness} \cite{es2023ragas,ragas_faithfulness} evaluates whether all statements in the generated response follow logically from the given context. A higher faithfulness score implies that the model does not fabricate or “hallucinate” details absent in the provided context. Each statement in the generated output is checked against the retrieval context for verifiability.

\paragraph{Score Definition:}
Let the generated response contain \(C_t\) total claims (statements). Out of these, \(C_s\) are directly supported by the retrieved context. Then:

\begin{equation}
\label{eq:faith_score}
\mathrm{Faithfulness\ Score}
= \frac{C_s}{C_t},
\end{equation}

\noindent
where
\(\mathrm{Faithfulness\ Score}\in [0,1]\). A score of 1.0 indicates every statement in the response is context-grounded.

\paragraph{Interpretation:}
\begin{itemize}
    \item \emph{Focus on Source-Consistency:}
    If a statement is unsupported or contradicts the context, it lowers \(C_s\). This metric is crucial for applications requiring factual accuracy tied to explicit references (e.g., legal or medical).
    \item \emph{Domain Constraints:}
    In specialized fields, verifying faithfulness may require deeper semantic checks or external knowledge that goes beyond raw text matching.
\end{itemize}

\subsubsection{Summarization Score}
\label{metric:ragas_sum_score}

\emph{Summarization Score} \cite{es2023ragas,ragas_summarization_score} determines how effectively a generated summary captures the essential information from the original text while encouraging brevity. It operates by generating questions from key phrases in the context and then verifying if the summary answers these questions correctly.

\paragraph{Two Key Components:}
\begin{enumerate}
    \item \textbf{QA Score:}
    Let the system produce \(\mathrm{N}\) questions based on important points in the context. For each question, the summary either answers it correctly (1) or incorrectly (0). Then,

    \begin{equation}
    \label{eq:qa_score}
    \mathrm{QA\ Score}
    = \frac{\text{Correctly Answered Questions}}{\text{Total Questions}}.
    \end{equation}

    \item \textbf{Conciseness Score:}
    To prevent trivial success by copying large swaths of text, a conciseness factor penalizes excessively long summaries:

    \begin{equation}
    \label{eq:cons_score}
    \mathrm{Conciseness\ Score}
    = \frac{\mathrm{Length(Summary)}}{\mathrm{Length(Context)}}.
    \end{equation}

    Summaries that replicate the entire context might answer all questions but yield a high ratio in Equation~\eqref{eq:cons_score}, which indicates poor conciseness.
\end{enumerate}

\paragraph{Final Summarization Score:}
\begin{equation}
\label{eq:sum_score}
\mathrm{Sum\ Score}
= \frac{
    \mathrm{QA\ Score}
    + \mathrm{Conciseness\ Score}
  }{2}.
\end{equation}

\paragraph{Interpretation:}
\begin{itemize}
    \item \emph{Balancing Completeness and Brevity:}
    This metric rewards precise, minimal summaries that still answer key questions from the context. 
    \item \emph{Limitations:}
    Summaries may be semantically correct but use synonyms that do not directly answer enumerated questions, or the question-generation module might produce suboptimal queries. Nonetheless, this approach ensures some measure of coverage \emph{and} conciseness.
\end{itemize}

\subsubsection{Context Entities Recall}
\label{metric:ragas_context_ent_recall}

\emph{Context Entities Recall} \cite{ragas_context_entities_recall} quantifies how effectively a system retrieves or maintains the entities from the original context. It is particularly relevant in fact-intensive tasks (e.g., tourism info, historical QA), where omitting certain named entities (cities, dates, persons) can degrade the overall usefulness of the generated response.

\paragraph{Definition:}
Let \(\mathrm{Entities}_\mathrm{context}\) be the set of named entities (or key domain-specific entities) identified in the reference context, and let \(\mathrm{Entities}_\mathrm{generated}\) be the entities identified in the system’s generated text. The metric:

\begin{equation}
\label{eq:cont_ent_rec}
\mathrm{CER}
= \frac{
    \lvert
      \mathrm{Entities}_\mathrm{context}
      \,\cap\,
      \mathrm{Entities}_\mathrm{generated}
    \rvert
  }{
    \lvert
      \mathrm{Entities}_\mathrm{context}
    \rvert
  },
\end{equation}

\noindent
yields a ratio in \([0,1]\). A higher value means more contextually significant entities from the original text reappear in the generated output.

\paragraph{Interpretation and Use:}
\begin{itemize}
    \item \emph{Entity-Focused Scenarios:}
    Ideal for verifying that essential names, places, or terms from the context remain present in the final generation. 
    \item \emph{Factual Adequacy:}
    While not a direct measure of correctness, a low CER might hint that crucial references were lost or omitted.
\end{itemize}

\paragraph{Limitations:}
\begin{itemize}
    \item \emph{Entity Identification Errors:}
    Dependence on NER or entity extraction might produce inaccuracies if either the reference or generation’s entity recognizer is imperfect.
    \item \emph{No Semantic Variation:}
    If the system uses synonyms or paraphrases for certain terms (instead of direct mentions), such entities might not match exactly.
\end{itemize}

\subsubsection{Aspect Critique}
\label{metric:ragas_aspect_critique}

\emph{Aspect Critique} is designed to evaluate submissions (model outputs) against a set of predefined or custom aspects (e.g., correctness, harmfulness) \cite{es2023ragas}. Each aspect yields a binary decision indicating whether the submission meets (Yes) or violates (No) the criterion.

\paragraph{Procedure:}
\begin{enumerate}
    \item \textbf{Aspect Definition:}
    Users specify the aspect or dimension of interest (e.g., “harmlessness,” “correctness,” or any custom criterion). 

    \item \textbf{LLM-based Verification:}
    The system issues multiple prompts (often 2 or 3) to a Large Language Model (LLM), asking if the submission meets the aspect requirement. For instance, a \emph{harmfulness} aspect might ask: “Does the submission cause or risk harm to individuals, groups, or society?”

    \item \textbf{Majority Verdict:}
    If the LLM returns “No” (harmless) in 2 out of 3 calls, the final aspect critique is \emph{No Harm}. If it returns “Yes” in the majority, the verdict is “Harmful.” A \emph{strictness} parameter (2 to 4) can regulate the level of confidence or self-consistency the model must display to finalize the verdict.
\end{enumerate}

\paragraph{Interpretation:}
\begin{itemize}
    \item \emph{Binary Feedback:}
    The output is simply \texttt{Yes} or \texttt{No} for each aspect, offering a direct measure of compliance.
    \item \emph{Modular Aspects:}
    Ragas Critiques or custom scripts can add or remove aspects (such as bias, factual correctness, etc.), tailoring the critique to domain requirements.
\end{itemize}

\paragraph{Limitations:}
\begin{itemize}
    \item \emph{Reliance on LLM Consistency:}
    Because the method re-prompts an LLM to judge the submission, it is subject to the model’s own biases and variability.
    \item \emph{Subjective or Complex Aspects:}
    Certain aspects (like “correctness” in specialized domains) may require deeper validation than a short textual query can offer. 
\end{itemize}

\paragraph{Use Cases:}
\begin{itemize}
    \item \emph{Content Moderation:}
    The “harmfulness” aspect can identify harmful, offensive, or risky content automatically.
    \item \emph{Regulatory or Ethical Checklists:}
    In finance or healthcare domains, aspects such as “compliance with policy” can be verified via aspect critiques before final deployment.
\end{itemize}

ARES aims to improve the precision and accuracy of evaluating RAG systems by using a combination of lightweight language model (LM) judges and prediction-powered inference (PPI) techniques. The framework generates its own synthetic data and uses them to fine-tune models that can assess the quality of individual RAG components. This approach allows ARES to offer automated, efficient, and domain-adaptive evaluation methods, which require minimal human annotations. ARES can be found at \url{https://github.com/stanford-futuredata/ARES}.

The three metrics provided by ARES are explained below.

\subsubsection{Context Relevance}
\label{metric:ares_context_rel}

Context Relevance assesses whether the information retrieved by the system is pertinent to the given query. In the ARES framework, context relevance is determined by fine-tuning a lightweight language model to classify query-passage pairs as either relevant or irrelevant. The process involves generating synthetic datasets of question-answer pairs from the corpus passages, which include both positive examples (relevant passages) and negative examples (irrelevant passages). The model is trained using a contrastive learning objective, which helps it distinguish between relevant and irrelevant contexts. The effectiveness of the context relevance metric is enhanced by using prediction-powered inference (PPI) to provide statistical confidence intervals, ensuring the accuracy of the evaluation \cite{saad2023ares}.

The process to compute Context Relevance is the following:

\begin{enumerate}
    \item Synthetic Dataset Generation: The process begins with the generation of synthetic datasets. A large language model (LLM) is used to create synthetic question-answer pairs from the corpus passages. This involves generating both positive examples (where the passage is relevant to the query) and negative examples (where the passage is irrelevant).
    \item Fine-Tuning LLM Judges: A lightweight language model, specifically DeBERTa-v3-Large \cite{he2020deberta}, is fine-tuned to serve as a judge for context relevance. This model is trained using a contrastive learning objective to classify query-passage pairs as relevant or irrelevant. The training leverages the synthetic dataset, which includes both positive and negative examples.
    \item Negative Example Generation: To enhance the training process, two strategies are employed to generate negative examples.
   Weak Negatives are randomly sampled passages that are unrelated to the synthetic query. Strong Negatives are passages from the same document as the relevant passage or similar passages retrieved using BM25, but they do not directly answer the query.
    \item Evaluation and Scoring: The fine-tuned LLM judge evaluates the context relevance of query-passage-answer triples from various RAG systems. The model assigns scores based on its classification of each triple as relevant or irrelevant.
    \item Prediction-Powered Inference (PPI): To improve the accuracy of the evaluation and provide statistical confidence intervals, ARES uses PPI. This involves using a small set of human-annotated datapoints to adjust the predictions made by the LLM judge. PPI helps in generating confidence intervals for the context relevance scores, ensuring that the evaluations are robust and reliable.
\end{enumerate}

\subsubsection{Answer faithfulness}
\label{metric:ares_answer_faith}

Answer faithfulness measures whether the generated response is grounded in the retrieved context, without introducing hallucinated or extrapolated information. This metric is useful to ensure that the answers provided by the RAG system are accurate and reliable. In ARES, answer faithfulness is evaluated by fine-tuning a separate LM judge to classify query-passage-answer triples as either faithful or unfaithful. Training involves creating synthetic datasets with both faithful answers (correctly grounded in the retrieved passages) and unfaithful answers (containing hallucinated or contradictory information). The model learns to identify discrepancies between the generated answer and the retrieved context. As with context relevance, PPI is used to enhance the accuracy of the faithfulness evaluation by leveraging a small set of human-annotated datapoints to generate confidence intervals.

The process to compute Answer faithfulness follows the same strategy as Context Relevance. Through the following steps, ARES measures answer faithfulness, ensuring that the generated answers are accurately grounded in the retrieved context.

\begin{enumerate}
    \item Synthetic Dataset Creation: The process begins with generating synthetic datasets that include both faithful and unfaithful examples. This involves using a large language model (LLM) to generate question-answer pairs based on passages from a corpus. The synthetic dataset includes positive examples where the answer is faithful to the retrieved passage and negative examples where the answer is not faithful.
    \item Fine-Tuning LLM Judges: A lightweight language model, specifically DeBERTa-v3-Large \cite{he2020deberta}, is fine-tuned to serve as a judge for answer faithfulness. The model is trained to classify query-passage-answer triples as either faithful or unfaithful using a contrastive learning objective. This training helps the model learn to identify when a generated answer is grounded in the retrieved passage without introducing hallucinated or contradictory information.
    \item Negative Example Generation: To enhance the training process, two strategies are used to generate negative examples. Weak Negatives are randomly sampled answers from other passages that are not related to the given query-passage pair and Strong Negatives are contradictory answers generated by prompting the LLM with few-shot examples to produce answers that conflict with the information in the passage.
    \item Evaluation and Scoring: The fine-tuned LLM judge evaluates the answer faithfulness of query-passage-answer triples produced by various RAG systems. The model assigns scores based on its classification of each triple as faithful or unfaithful.
    \item Prediction-Powered Inference (PPI): Same as with Context Relevance. This step is used to improve the accuracy of the evaluation and provide statistical confidence intervals, ARES uses PPI. This involves using a small set of human-annotated datapoints to adjust the predictions made by the LLM judge. 
\end{enumerate}

\subsubsection{Answer Relevance}
\label{metric:ares_answ_rel}

Answer relevance assesses whether the generated response is relevant to the original query, considering both the retrieved context and the query itself. This metric ensures that the answer not only aligns with the retrieved information but also addresses the query effectively. In the ARES framework, answer relevance is evaluated by fine-tuning another LM judge to classify query-passage-answer triples as either relevant or irrelevant. The training process involves generating synthetic datasets with examples of both relevant and irrelevant answers, allowing the model to learn the nuances of relevance in the context of the query and the retrieved passage. PPI is again employed to provide confidence intervals, improving the reliability of the relevance evaluation by combining the predictions from the LM judge with a small set of human annotations.

The process to measure answer relevance is similar to the previous ARES metrics. Through these steps, ARES ensures that the generated answers not only align with the retrieved information but also effectively address the original query.

\begin{enumerate}
    \item Synthetic Dataset Creation: Similar to the other metrics, the process begins with generating synthetic datasets. A large language model (LLM) is used to create synthetic question-answer pairs from the corpus passages. This dataset includes both positive examples (where the answer is relevant to the query and retrieved passage) and negative examples (where the answer is irrelevant).
    \item Fine-Tuning LLM Judges: A lightweight language model, such as DeBERTa-v3-Large \cite{he2020deberta}, is fine-tuned to act as a judge for answer relevance. The model is trained to classify query-passage-answer triples as either relevant or irrelevant using a contrastive learning objective. This training helps the model learn to identify when a generated answer appropriately addresses the query using information from the retrieved passage.
    \item Negative Example Generation: To enhance the training process, two types of negative examples are generated. Weak Negatives are randomly sampled answers from other passages that are not related to the given query-passage pair. Strong Negatives are answers generated by prompting the LLM to produce responses that do not address the query or are contradictory to the retrieved passage.
    \item Evaluation and Scoring: The fine-tuned LLM judge evaluates the answer relevance of query-passage-answer triples produced by various RAG systems. The model assigns scores based on its classification of each triple as relevant or irrelevant.
    \item Prediction-Powered Inference (PPI): To improve the accuracy of the evaluation and provide statistical confidence intervals, ARES uses PPI using a small set of human-annotated datapoints to adjust the predictions made by the LLM judge. PPI helps in generating confidence intervals for the answer relevance scores, ensuring that the evaluations are robust and reliable.
\end{enumerate}

\section{Combining Multiple Loss Functions}
\label{sec:combining_loss}

A practical strategy in deep learning is the use of \emph{multi-loss setups}, wherein two or more loss functions are combined---often in a weighted sum---to capture different facets of a complex task. This approach allows a model to optimize multiple objectives simultaneously, often improving final performance \cite{xu2018pad,kendall2018multi}. Below are several exemplary cases:

\begin{itemize}
    \item \textbf{Localization vs.\ Classification in Object Detection.} In object detection architectures such as YOLO~\cite{redmon2016you} or Faster R-CNN~\cite{ren2015faster}, a regression loss (e.g., Smooth \(\ell_1\)~\cite{girshick2015fast}) is used for bounding-box coordinates, while a classification loss (e.g., Cross-Entropy) handles object labels. Training on both ensures accurate box localization \emph{and} robust classification, thereby balancing spatial accuracy with semantic correctness.

    \item \textbf{Generative Adversarial Networks (GANs).} GANs typically rely on an \emph{adversarial} loss (standard~\cite{goodfellow2014} or Wasserstein~\cite{arjovsky2017wasserstein}) to push generated samples to be indistinguishable from real data. However, to control additional properties (e.g., content consistency, style, or sharper details), auxiliary terms such as pixel-level or perceptual losses~\cite{gatys2015neural,ledig2017} may be introduced. This can help avoid mode collapse and improve the visual fidelity of generated images, as each component of the total loss addresses distinct aspects of image realism.

    \item \textbf{Language Modeling with Additional Constraints.} In summarization or machine translation, a standard token-level Cross-Entropy loss may be paired with a sequence-level loss (e.g., Minimum Risk Training~\cite{shen2015minimum} or REINFORCE~\cite{williams1992simple}) that directly optimizes global quality metrics such as BLEU~\cite{papineni2002bleu} or ROUGE~\cite{lin2004rouge}. Balancing token-by-token accuracy with entire-sequence objectives often yields more coherent and context-aware outputs, reducing exposure bias~\cite{ranzato2015sequence}.

    \item \textbf{Perceptual + Reconstruction Loss in Image-to-Image Tasks.} In tasks like super-resolution or style transfer, a reconstruction loss (e.g., pixel-level MSE) may be combined with a perceptual loss computed via intermediate features of a pretrained CNN~\cite{johnson2016perceptual}. This ensures both low-level fidelity (sharp edges, stable colors) and high-level feature preservation (texture consistency, semantic alignment), ultimately leading to more realistic results.
\end{itemize}

A common method to combine losses is to form a weighted sum:
\begin{equation}
    L_{\text{total}} \;=\; 
    \lambda_{1}\,L_{1} \;+\; \lambda_{2}\,L_{2} \;+\;\cdots\;+\; \lambda_{k}\,L_{k},
    \label{eq:multi_loss}
\end{equation}
where \( \lambda_{i} \) are nonnegative weighting factors for each objective \( L_{i} \). The assignment of these weights remains an important challenge. On the one hand, \emph{i)} if an auxiliary loss is overemphasized, it can overshadow the primary objective. On the other hand, \emph{ii)} if underemphasized, it offers little benefit~\cite{kendall2018multi,chen2018gradnorm}. Researchers often tune these hyperparameters empirically or employ heuristic methods such as gradient normalization (GradNorm) \cite{chen2018gradnorm} or uncertainty weighting \cite{kendall2018multi} to manage them automatically.

\paragraph{Synergistic Effects of Multi-Loss Learning.}
In many cases, combining carefully selected loss terms can produce synergy by tackling complementary aspects of a task. For example, in image segmentation, cross-entropy might be integrated with a loss based on Dice or IoU \cite{milletari2016v,rahman2016optimizing} to handle class imbalance and emphasize overlapping regions. Similarly, in retrieval-augmented generation (RAG) \cite{lewis2020retrieval}, pairing a contrastive or ranking loss for the retrieval component with a sequence modeling loss for text generation enables the system to more effectively leverage external knowledge and produce high-fidelity responses.

\paragraph{Challenges and Future Directions.}
While multi-loss sets often improve performance, they can also introduce training instability if the losses have conflicting gradients \cite{yu2020gradient}. In addition, too many objectives can lead to loss competition, where improvements in one objective degrade another. To address these issues, recent works study adaptive reweighting strategies (e.g., dynamic uncertainty weighting \cite{kendall2018multi}, GradNorm \cite{chen2018gradnorm}) or multi-objective optimization techniques~\cite{sener2018multi}, allowing the model to balance objectives in a principled way. Looking forward, automated methods to discover or tune weights for multi-loss training are a promising avenue \cite{zhang2021ada,groenendijk2021multi}.

Multi-loss frameworks are now ubiquitous across a variety of deep learning applications. As tasks become more complex and multifaceted, designing or automating optimal loss combinations will likely remain a key area of research. Incorporating task-specific constraints, domain knowledge, and trade-offs among different objectives will be essential to push the boundaries of performance and reliability in deep learning systems.

\section{Challenges and Trends}
\label{sec:challenges_and_trends}

In the past decade, deep learning has significantly advanced fields like computer vision \cite{krizhevsky2012,he2016,lin2017focal}, natural language processing \cite{devlin2018bert,brown2020language,radford2019language}, reinforcement learning \cite{mnih2015human}, and generative modeling \cite{goodfellow2014,kingma2013auto}. These advancements pose various challenges for researchers and practitioners. This subsection outlines these challenges, how current loss functions and metrics address them, and the trends influencing deep learning's future.

\subsection{Challenges in Deep Learning}

\subsubsection*{Data Limitations and Imbalances}
Large amounts of high-quality labeled data are necessary for training deep learning models. In many practical scenarios, data are expensive to annotate or the classes are highly imbalanced. Traditional loss functions such as cross-entropy (\S\ref{loss:BCE}) or MSE (\S\ref{loss:MSE}) can struggle in these scenarios, because they treat every instance equally, potentially causing overfitting on the majority classes \cite{lin2017focal}. Methods like Weighted Binary Cross-Entropy (\S\ref{loss:weighted_cross_entropy}) and Focal Loss (\S\ref{loss:focal}) specifically address these issues by assigning higher weight to harder or minority samples, improving performance on imbalanced datasets.

\subsubsection*{Sensitivity to Outliers and Noise}
Standard losses such as (MSE) are known to be sensitive to outliers or noisy data, disproportionately penalizing large errors. This could lead to compromised performance on tasks with heavy-tailed error distributions, such as time series forecasting \cite{zhao2012review_energy_consumption} or in real-time sensor data analyses. Robust alternatives, such as Huber Loss (\S\ref{loss:Huber}) or Smooth L1 (\S\ref{loss:smooth_l1}), can mitigate these issues by transitioning between the squared and absolute error regimes based on a threshold.

\subsubsection*{Exposure Bias and Sequence-Level Learning}
In natural language generation tasks, models trained with next-token prediction suffer from exposure bias since, at inference time, they generate tokens sequentially but were trained to predict one token at a time using the ground-truth prefix \cite{ranzato2015sequence}. This mismatch can lead to the accumulation of errors and degradation of sequence-level performance. Methods like Minimum Risk Training (\S\ref{loss:mrt_nlp}) and REINFORCE (\S\ref{loss:reinforce}) address this issue by optimizing metrics such as BLEU (\S\ref{metric:bleu_score}) or ROUGE (\S\ref{metric:rouge_score}) directly, bridging the gap between training and inference conditions.

\subsubsection*{Complex Evaluation Requirements}
As tasks become more sophisticated (e.g. multi-object detection \cite{redmon2016you,ren2015faster}, panoptic segmentation \cite{kirillov2019panoptic}, retrieval-augmented generation \cite{lewis2020retrieval}), a single metric or loss measure is insufficient. Models that are state-of-the-art in one metric might fail in others. This complexity demands task-specific or multi-faceted metrics that capture different quality dimensions (e.g., faithfulness (\S\ref{metric:ragas_faithfulness}) vs.\ relevance (\S\ref{metric:ragas_ans_rel}), bounding-box precision (\S\ref{loss:IoUloss_obj_det}) vs.\ recall \S\ref{metric:recall_tpr}).

\subsection{Addressing  Challenges}

\subsubsection*{Addressing Class Imbalance and Data Scarcity}
Loss functions like \emph{Focal Loss} (\S\ref{loss:focal}) and \emph{Weighted Cross-Entropy} (\S\ref{loss:weighted_cross_entropy}) emphasize minority classes by assigning higher loss to misclassified examples or underrepresented classes. Similarly, in retrieval-augmented generation, contrastive (\S\ref{loss:contrastive}) or ranking-based losses (\S\ref{loss:marginal_ranking_nlp}) guide the system to focus on genuinely relevant documents from vast corpora.

\subsubsection*{Robustness to Outliers}
When dealing with outliers in applications like healthcare \cite{pham2000current}, robust losses such as Huber (\S\ref{loss:Huber}), Smooth L1 (\S\ref{loss:smooth_l1}), or Log-Cosh (\S\ref{loss:logcos}) are essential. These losses are less sensitive to extreme values, helping maintain stability and reducing the tendency to overfit.

\subsubsection*{Sequence-Level Training and Global Metrics}
Optimizing only at the token level often fails to capture global sequence properties. Metrics like BLEU (\S\ref{metric:bleu_score}), ROUGE (\S\ref{metric:rouge_score}), and specialized ones like METEOR (\S\ref{metric:meteor}) can be integrated via Minimum Risk Training (\S\ref{loss:mrt_nlp}) or Reinforcement Learning \cite{williams1992simple,ranzato2015sequence}, driving improvements in fluency and semantic alignment.

\subsubsection*{Multi-Task and Multi-Loss Objectives}
Modern deep learning tasks increasingly require balancing multiple objectives \cite{kendall2018multi}. For instance, in object detection (bounding-box regression + classification), models combine smooth L1 or IoU-based losses for localization \cite{rezatofighi2019generalized} with cross-entropy losses for classification. Meanwhile, generative models can integrate \emph{adversarial} losses \cite{goodfellow2014} alongside reconstruction or perceptual losses \cite{gatys2015neural,ledig2017} to achieve realistic and content-preserving outputs.

\subsection{Trends and Future Directions}

\subsubsection*{Automated Loss Function Search}
Researchers have begun exploring methods to automatically discover or design loss functions (``Loss Function Search''), aiming to reduce the reliance on trial and error. Using gradient-based meta-learning or evolutionary algorithms, these approaches can search a space of differentiable loss operators \cite{liu2021loss,li2022autoloss} to find the most effective for a given dataset or task.

\subsubsection*{Task-Adaptive and Data-Adaptive Metrics}
Generic metrics such as MSE or Cross-Entropy can be suboptimal when they do not align with real-world goals. Emerging methods adapt losses and evaluation metrics to specific objectives: for example, survival analysis in healthcare requires specialized metrics for censored data \cite{efron1977efficiency}, while retrieval-augmented generation employs metrics like \emph{Answer Faithfulness} (\S\ref{metric:ragas_faithfulness}) or \emph{Context Relevance} (\S\ref{metric:ares_context_rel}) for real-world utility and factual correctness.

\subsubsection*{Continual and Online Learning Scenarios}
In many dynamic settings (e.g., streaming data \cite{hasan2016}, or rapidly changing domains), metrics and losses must adapt to evolving data distributions \cite{hoi2021online}. Online surrogates of cross-entropy or adversarial objectives have been proposed, while incremental evaluation protocols address shifts in data distribution without complete retraining.

\subsubsection*{Robustness and Safety in Deployment}
As deep models are deployed in safety-critical areas like medical imaging \cite{pham2000current}, autonomous driving \cite{bojarski2016}, or legal advisory \cite{chouhan2024lexdrafter}, new evaluation metrics (fairness, interpretability, reliability) and novel losses that penalize harmful behaviors are gaining prominence \cite{rajpurkar2018deep,agarwal2021deep}. Tailored metrics like \emph{Answer Correctness} or \emph{Answer Faithfulness} \cite{es2023ragas,saad2023ares} are already emerging in retrieval-augmented systems.

\subsubsection*{Enhanced Interpretability and Explainability}
Future loss functions and metrics may incorporate interpretability constraints, ensuring that model predictions can be more transparent \cite{rudin2019stop}. This is particularly relevant in retrieval-augmented generation \cite{lewis2020retrieval}, where the chain-of-thought or evidence path can be traced back to external documents.

\subsubsection*{Large Language Models and Prompt-Engineering Paradigms}
As large-scale language models \cite{brown2020language,radford2019language} become prevalent, new metrics are needed to evaluate context usage, reasoning chains, or prompt understanding. We foresee research into specialized loss functions that penalize factual inconsistencies or hallucinations, a rising concern in modern LLM-based applications \cite{es2023ragas,saad2023ares}.

\bigskip
\noindent
Deep learning loss functions and metrics are evolving to address challenges like imbalanced data, outliers, sequence-level training, and multi-objective optimization. Trends focus on robust, context-aware, application-specific objectives, highlighting the field's diversity and complexity. Innovations in task-adaptive losses and advanced evaluation frameworks offer reliable, interpretable, domain-specific solutions for future applications.

\section{Conclusion}
\label{sec:conclusion}

Loss functions and performance metrics lie at the core of effective deep learning. As this review demonstrates, no single approach applies universally across tasks: regression, classification, computer vision (for instance, object detection, face recognition, depth estimation, image generation), and NLP all require careful tailoring of the loss function and evaluation metric. The varying data characteristics, particularly in scenarios with class imbalance or noisy labels, emphasize the critical role of selecting or designing losses that align with the problem at hand.

Beyond conventional choices such as cross-entropy or MSE, specialized losses (e.g., focal loss, smooth L1, adversarial objectives) and advanced evaluation metrics (e.g., AP/AR, IoU-based measures, BLEU, ROUGE) have emerged to address domain-specific challenges. Moreover, multi-loss setups, in which two or more objectives are combined, often produce richer learning signals as long as weighting factors are tuned carefully or adapted to balance competing goals.

Looking ahead, several directions can strengthen our use of loss functions and metrics in deep learning. One key area is the automation of loss and metric selection through search algorithms or meta-learning techniques, thereby reducing human trial and error. Another involves designing robust and interpretable objectives that remain stable in the presence of noise or domain shifts while offering more transparent optimization signals. There is also growing interest in task-adaptive metrics that align more closely with practical objectives, such as fairness, explainability, or retrieval faithfulness, rather than relying exclusively on generic measures. Finally, hybrid approaches will need to be extended for emerging tasks such as retrieval-augmented generation, where both retrieval accuracy and generative fidelity are crucial.

By advancing these avenues, researchers and practitioners will obtain stronger, more reliable deep learning models. In particular, better comprehension of losses and metrics will help overcome challenges associated with complex tasks, ultimately pushing modern AI systems to remain high-performing and robust in a rapidly evolving field.

\section{Acknowledgments}
We thank the National Council for Science and Technology (CONACYT) for its support through the National Research System (SNI).

\section*{Declaration of generative AI and AI-assisted technologies in the writing process}

We acknowledge the use of three AI tools: Grammarly Assistant to improve the grammar, clarity, and overall readability of the manuscript, Perplexity for finding relevant academic works, and GPT-4o to help with the wording and proofreading of the manuscript.

\bibliographystyle{ieeetr}
\bibliography{references}  

\end{document}